\documentclass[10pt,twocolumn,letterpaper]{article}
\usepackage[algorithms]{wacv}
\usepackage{times}
\usepackage{epsfig}
\usepackage{graphicx}
\usepackage{amsmath}
\usepackage{amssymb}
\usepackage[accsupp]{axessibility}
\usepackage{caption}
\usepackage{xcolor}
\usepackage{makecell}
\usepackage{multirow}
\usepackage{booktabs}
\usepackage{tabularx}
\usepackage{paralist}
\usepackage[numbers,sort,compress]{natbib}
\usepackage[pagebackref=true,breaklinks=true,colorlinks,bookmarks=false]{hyperref}
\usepackage{float}

\def\wacvPaperID{2587}
\def\confName{WACV}
\def\confYear{2024}

\usepackage[capitalize]{cleveref}
\crefname{section}{Sec.}{Secs.}
\Crefname{section}{Section}{Sections}
\Crefname{table}{Table}{Tables}
\crefname{table}{Tab.}{Tabs.}

\newcommand{\xhdr}[1]{\vspace{2pt}\noindent\textbf{#1}}

\newcommand{\textmr}[2]{\multirow{#2}{*}{#1}}

\definecolor{darkgreen}{rgb}{0,0.4,0}
\definecolor{groundtruth}{rgb}{0,0.63,0.38}
\definecolor{partialmatch1}{rgb}{1,0.4,0}
\definecolor{partialmatch2}{rgb}{1,0.65,0}
\definecolor{text}{rgb}{0.44,0.68,0.32}
\definecolor{image}{rgb}{0.22,0.45,0.75}
\definecolor{voxel}{rgb}{0.95,0.47,0.23}
\definecolor{imagevoxel}{rgb}{0.58,0.40,0.73}

\newcommand{\tv}{\ensuremath{\text{t}\rightarrow \text{v}}} 
\newcommand{\vt}{\ensuremath{\text{v}\rightarrow \text{t}}}
\newcommand{\incorrect}[1]{\underline{#1}}

\newcommand{\bimodi}{Bi(\textbf{I})\xspace} 
\newcommand{\trimodi}{Tri(\textbf{I})\xspace}
\newcommand{\bimodv}{Bi(\textbf{V})\xspace}
\newcommand{\trimodv}{Tri(\textbf{V})\xspace}
\newcommand{\trimodiv}{Tri(\textbf{I}+\textbf{V})\xspace}

\newcommand{\TITLE}{TriCoLo: Trimodal Contrastive Loss for Text to Shape Retrieval}

\newcolumntype{Y}{>{\centering\arraybackslash}X}

\newcommand{\denselist}{\itemsep 0pt\parsep=0pt\partopsep 0pt\vspace{-\topsep}}

\setdefaultleftmargin{1.5em}{}{}{}{.5em}{.5em}

\begin{document}

\title{\TITLE}

\author{Yue Ruan$^1$\thanks{indicates equal contribution.}~ \qquad Han-Hung Lee$^1$\footnotemark[1]~ \qquad Yiming Zhang$^1$ \qquad Ke Zhang$^1$ \qquad Angel X. Chang$^{1,2}$\\
$^1$Simon Fraser University \qquad $^2$Alberta Machine Intelligence Institute (Amii)
\\
{\tt\small \{yuer, hla300, yza440, ke\_zhang\_4, angelx\}@sfu.ca} \\
\footnotesize{\url{https://3dlg-hcvc.github.io/tricolo/}}
}

\twocolumn[{%
	\renewcommand\twocolumn[1][]{#1}%
    \maketitle
	\thispagestyle{empty}
	\begin{center}
	    \vspace{-0.5cm}
		\includegraphics[width=\textwidth]{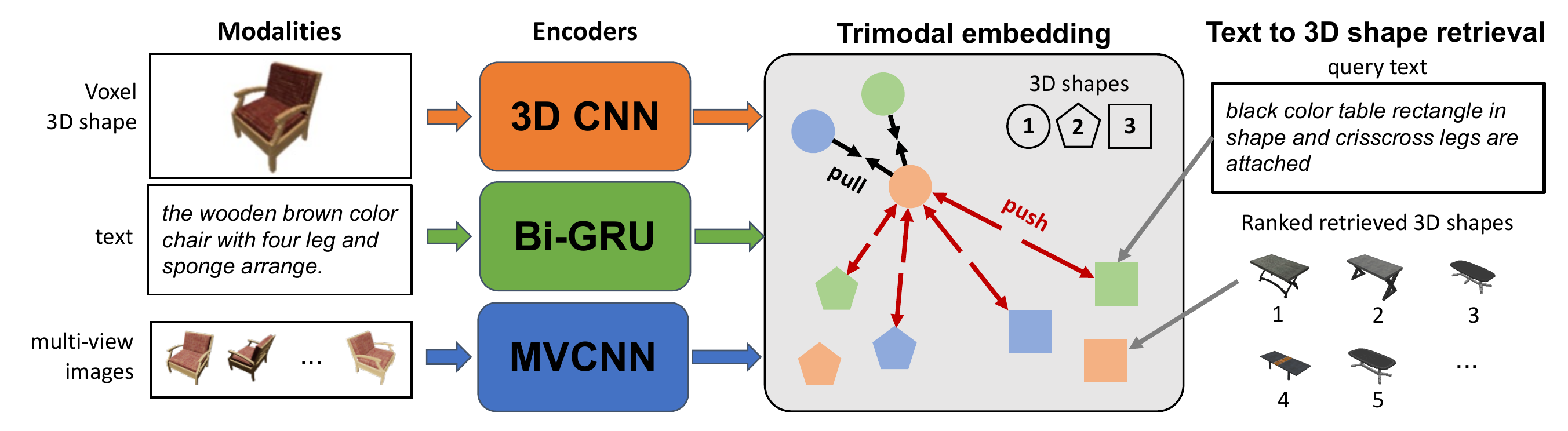}
		\captionof{figure}{
		We introduce \textbf{TriCoLo}, a \textbf{tri}modal \textbf{co}ntrastive \textbf{lo}ss for text to 3D shape retrieval. We take objects represented by 3D colored voxels, text descriptions, and multi-view images and jointly use these three modalities to train a trimodal embedding space. This trimodal embedding allows us to perform fine-grained text to shape retrieval.
		}
		\label{fig:teaser}
	\end{center}
}]
\begin{abstract}
Text-to-shape retrieval is an increasingly relevant problem with the growth of 3D shape data.
Recent work on contrastive losses for learning joint embeddings over multimodal data~\cite{radford2021learning} has been successful at tasks such as retrieval and classification.
Thus far, work on joint representation learning for 3D shapes and text has focused on improving embeddings through modeling of complex attention between representations~\cite{tang2023parts2words}, or multi-task learning~\cite{han2019y2seq2seq}.
We propose a trimodal learning scheme over text, multi-view images and 3D shape voxels, and show that with large batch contrastive learning we achieve good performance on text-to-shape retrieval without complex attention mechanisms or losses.
Our experiments serve as a foundation for follow-up work on building trimodal embeddings for text-image-shape.
\end{abstract}

\section{Introduction}
\label{sec:intro}

There has been a dramatic increase in the availability of 3D content in recent years.
Improved scanning hardware and reconstruction algorithms are democratizing 3D content creation.
The growth in virtual and augmented reality applications has also driven demand for more synthetic (i.e. human-designed) 3D content.
It is no wonder that operating systems now natively support viewing and editing 3D content (e.g., iOS/MacOS and Windows).
In addition to curated 3D object datasets for research~\cite{wu20153d,chang2015shapenet,fu20203dfuture,collins2021abo,reizenstein2021common}, large repositories of 3D shapes provide both synthetic~\cite{sketchup3dw,turbosquid,sketchfab} %
and scanned objects~\cite{downs2022google,polycam}.

As 3D assets become more pervasive, we need techniques that allow users to easily and rapidly search through large 3D collections.
In recent years, text to image search has seen renewed interest due to improved architectures~\cite{lu2019vilbert,li2020oscar,chen2020uniter,radford2021learning} and objectives~\cite{kiros2014unifying,faghri2017vse++,zhang2021contrastive,radford2021learning} for joint representation learning.
In contrast, there has been little research on text-driven 3D content search.

Early work by \citet{min2004comparison} compared the text query with text associated with the shape (essentially text-text retrieval).
\citet{chen2018text2shape} were the first to jointly embed text and 3D shapes for text-to-shape retrieval.
They learned the embedding space using triplet loss combined with learning by association~\cite{haeusser2017learning}.
Leveraging the `chairs and tables' dataset~\cite{chen2018text2shape}, followup work investigated improved methods for text-to-shape retrieval~\cite{han2019y2seq2seq,tang2023parts2words}.

So far, prior work on text-to-shape retrieval has not provided a systematic investigation of: 1) whether 3D information is necessary for text-to-shape retrieval (or whether single view images to represent a shape are sufficient); 2) whether there are benefits to incorporating information across three modalities; and 3) what contrastive learning setup and loss should be used for constructing joint text-shape embeddings.  In our work, we present a systematic study of what is important for improved text-to-shape retrieval.  We conduct experiments to examine the effect of input representation (single-view vs multi-view vs 3D voxels), loss function, batch size, and resolution.
We show that recent contrastive learning algorithms~\cite{zhang2021contrastive} are sufficient to achieve good performance while avoiding more complex mechanisms, such as combining metric learning with learning by association~\cite{chen2018text2shape} or training using part-based segmentation of the 3D shapes~\cite{tang2023parts2words}.%

In addition, we propose a joint embedding that leverages the multiple modalities offered by 3D data.  Specifically, we learn a joint embedding in a trimodal setting: voxel, multi-view images and text.
Prior work on text-to-shape retrieval either learns a joint representation with voxels and text, or multi-view images and text, both of which are bimodal settings.
We use all three modalities to learn the joint embedding space in an end-to-end fashion and show that trimodal works better than bimodal embedding for text-to-shape retrieval.
In summary, our contributions are:
\begin{itemize}\denselist
    \item We introduce a trimodal training scheme with contrastive loss that jointly embeds multi-view images, voxels, and language.  We show the trimodal embedding is effective for text to 3D shape retrieval. 
    \item We release ShapeNet c13, a dataset of paired shapes and captions for 13 object categories from ShapeNet~\cite{chang2015shapenet}.
    \item We present extensive experiments and analysis to provide guidelines on effective settings for applying contrastive loss for text-to-shape retrieval.
    \item We establish a high-performing baseline for text-to-shape retrieval.  Our simple but effective approach outperforms more complex techniques from prior work.  Since we introduced TriCoLo in 2022, several follow-up works~\cite{wang2023mxm,tang2023parts2words} have used it as a comparison baseline.
\end{itemize}

\section{Related work}
\label{sec:rel}

There has been growing interest in connecting language to 3D representations for several tasks: identifying 3D objects in scenes~\cite{chen2020scanrefer,achlioptas2020referit3d,huang2021text,yuan2021instancerefer,zhao20213dvg,roh2021languagerefer}, describing 3D objects~\cite{han2020shapecaptioner,chen2021scan2cap}, using 3D augmentation in caption-driven image retrieval~\cite{wu2021towers},  generating~\cite{chen2018text2shape} and disambiguating~\cite{achlioptas2019shapeglot,thomason2021language} 3D shapes using natural language.

\xhdr{3D shape retrieval.}
\citet{min2004comparison} was one of the first to address text to 3D shape retrieval by comparing the text query with textual information associated with the shape.
Their approach was based purely on text, and relied on each shape having an associated description.
\citet{chen2018text2shape} was the first work to create a joint embedding of text and 3D shapes and use that for text-to-shape retrieval.
The joint embedding was constructed using a CNN encoder on voxels and GRU encoders on text, with a combined triplet loss~\cite{song2016deep} and learning by association~\cite{haeusser2017learning} to align the embedded representations.
To improve retrieval, \citet{han2019y2seq2seq} used a GRU to encode image features from multiple views to represent the shape, and use reconstruction losses (both intra and inter modalities) in addition to triplet loss and classification loss to train the joint embedding. In contrast, we use multi-view and voxel representation for the shape and do not rely on reconstruction losses.
\citet{tang2023parts2words} incorporated part-level information, and used point cloud representations for the shapes.  In their work, semantic part data was used to compute attention with words to model 3D part relationship with the descriptions. However, obtaining semantic part information can be difficult. %
Following our work, \citet{wang2023mxm} shows improved performance for text-to-shape retrieval by better selection of positive and negative pairs for contrastive learning, even with just bimodal embeddings of text and multi-image views.

\xhdr{3D object disambiguation through language.}
The task of object disambiguation through language (also known as a reference game) is related to our text-to-shape retrieval.
The main difference between the two tasks is a matter of scale.
In shape retrieval, we retrieve all objects that match a textual query from a large set of candidate objects.
In contrast, in 3D object disambiguation, there is a small set of objects (typically just two or three) from which we select the one that best matches the description.
Reference games involving images and language have a long history~\cite{clark1986referring,golland2010game,frank2012predicting,kazemzadeh2014referitgame,monroe2017colors}, but there is significantly less work that takes advantage of the 3D nature of objects.
\citet{achlioptas2019shapeglot} used a speaker-listener model for selecting the correct object based on the text description from among three objects.
They showed that combining 3D features (from point clouds) with 2D features (from images) is better than just using 3D or 2D features.
More recently, \citet{thomason2021language} showed that using multi-view images can improve the disambiguation power of a model.  %
Unlike this line of prior work, we focus on text-to-3D shape retrieval and examine the benefit of combining multi-view images and colored 3D voxel representations.
Note that the text-to-shape retrieval problem has different characteristics and challenges as it requires selecting from a large number of instances at inference time (vs disambiguating two or three).

\xhdr{Joint embedding.}
Joint embedding spaces for text and images~\cite{weston2011wsabie,frome2013devise,kiros2014unifying,faghri2017vse++,zhang2021contrastive,radford2021learning} have enabled retrieval and generation between text and 2D images.
Most joint embedding approaches use contrastive learning, focusing on one modality such as images~\cite{hadsell2006dimensionality,schroff2015facenet,sohn2016improved,chen2020simple}, or two modalities~\cite{radford2021learning}.
Recently, an increasing number of works explore combinations of more modalities~\cite{alayrac2020self,akbari2021vatt,mai2021hybrid,liu2021contrastive}, with prior work (typically in combining vision, audio, and language) showing that multiple modalities can improve performance~\cite{alayrac2020self,akbari2021vatt,mai2021hybrid}.
\citet{liu2021contrastive} introduce a data augmentation technique where modalities are disturbed to generate negative samples.
These lines of prior work are orthogonal to our work as we investigate the use of trimodal contrastive loss on creating a joint embedding with 3D shape, language, and multiview images for text-to-shape retrieval.
Since our work was introduced in 2022, other works started to investigate trimodal embeddings of text-image-shape~\cite{xue2023ulip,liu2023openshape,zhao2023michelangelo}, showing it is useful to train 3D encoders to align 3D embeddings against frozen CLIP text and vision embeddings.
ULIP~\cite{xue2023ulip} and OpenShape~\cite{liu2023openshape} demonstrated that such aligned point cloud embeddings are useful by quantitatively evaluating on classification, and \citet{zhao2023michelangelo} showed that using an aligned space results in more faithful text-to-shape generation.
These works did not focus on evaluating how well the aligned space works for more fine-grained text-to-shape retrieval or comparing different encoders.

\section{Problem statement}
\label{sec:task}

We tackle the problem of 3D shape retrieval given an input query sentence $x_t$.
We use the Text2Shape~\cite{chen2018text2shape} dataset which contains tables and chairs from ShapeNet~\cite{chang2015shapenet} paired with several text descriptions for each object.  
The text descriptions provide fine-grained information about the appearance of the objects such as color, texture, shape, and whether the object has a certain part (e.g. armrest or a circular base).
Accurate retrieval requires learning a good similarity measure between text description and 3D shape.
To this end, we learn a shared latent space to facilitate the process of text-shape alignment.

\section{Approach}
\label{sec:app}

\begin{figure}
\includegraphics[width=\linewidth]{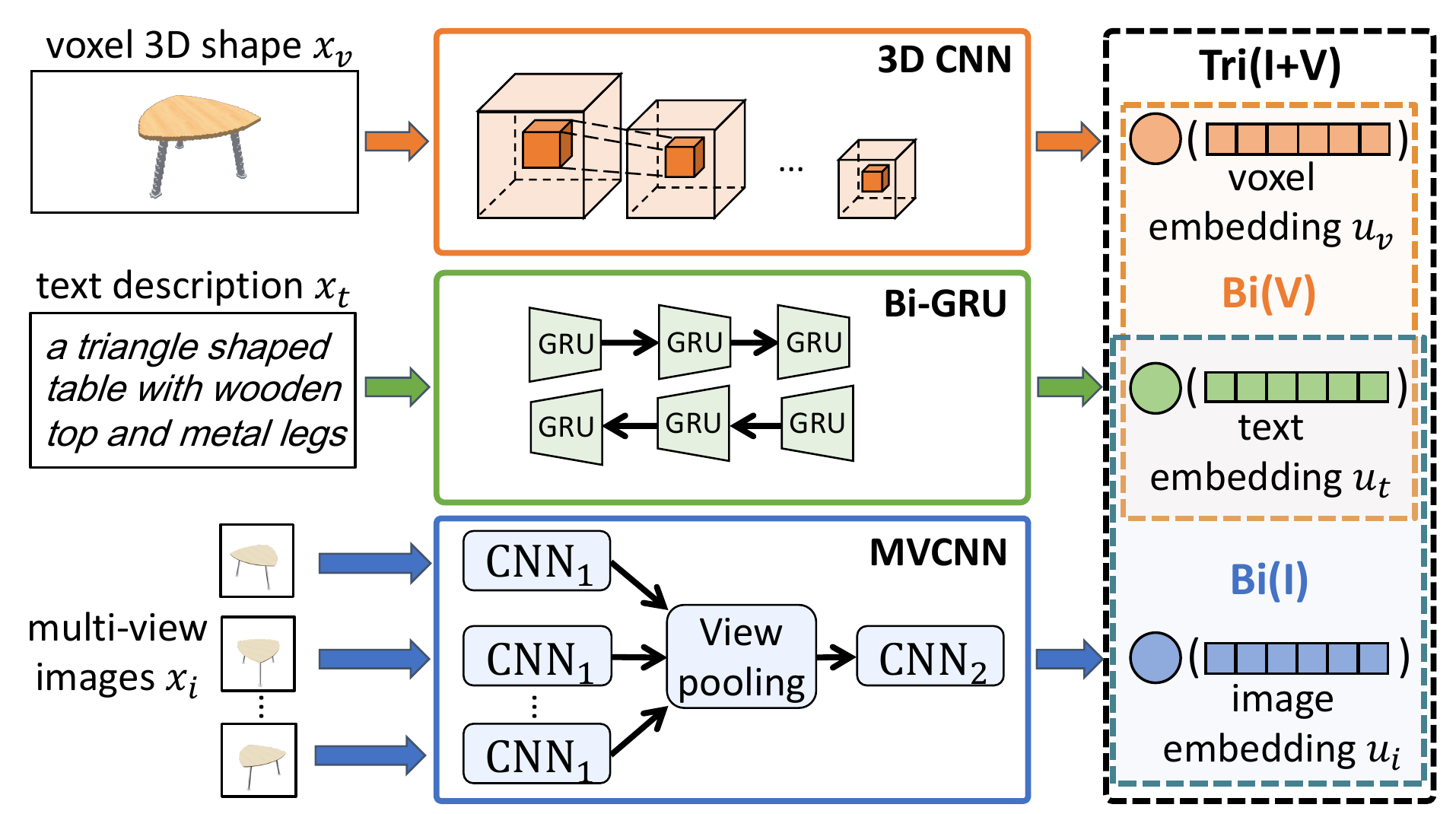}
\vspace{-15pt}
\caption{Given the voxel shapes $x_v$, input text description $x_t$ and rendered images $x_i$, 3D CNN, Bi-GRU and MVCNN transform them to feature vectors $u_v$, $u_t$ and $u_i$. We then minimize a bidirectional contrastive loss to learn effective shape, text, and image representations that are close to each other if they are from the same object.}
\label{fig:tri-modality-pipeline}
\end{figure}

Inspired by recent developments in multimodal contrastive learning~\cite{alayrac2020self,akbari2021vatt,mai2021hybrid,liu2021contrastive}, we use 3D voxels, multi-view images and language to learn a shared embedding using contrastive learning.
\cref{fig:tri-modality-pipeline} shows how we encode different modalities with per-modality architectures.
Embeddings of the same object are pulled closer, while those of different objects are pushed apart using contrastive loss.

\xhdr{Encoder models.}
We represent the input 3D voxels, text description and multi-view images as $x_v, x_t$ and $x_i$ respectively. For each modality $m \in (v,i,t)$, an encoder $f_m$ takes the input $x_m$ and outputs an encoding $u_m \in \mathbb{R}^d$.  The text encoder $f_t$ is a Bi-directional Gate Recurrent Unit (Bi-GRU)~\cite{cho2014learning} which takes a text description $x_t\in R^{L\times e_t}$ and outputs the embedding $u_t\in R^{d}$, where $L$ and $e_t$ are the sentence and word embedding lengths respectively.
For voxels, a 3D CNN model $f_v$ takes a 3D input of $x_v\in \mathbb{R}^{r_v\times r_v\times r_v\times4}$ and outputs $u_v\in \mathbb{R}^{d}$ where $r_v$ is the voxel resolution.
Finally, the image encoder takes $M$ views of the object $x_i\in R^{M\times r_i\times r_i\times3}$ through an MVCNN~\cite{su2015multiview} architecture with pretrained ResNet18~\cite{he2016deep} backbone $f_i$ to obtain the representation $u_i\in \mathbb{R}^{d}$ where $r_i$ is the image resolution.

\xhdr{Loss function.}
We adopt the bimodal loss from ConVIRT \cite{zhang2021contrastive}.
Specifically for two modalities $m_1, m_2 \in (v,i,t)$ so that $m_1\neq m_2$ and a batch size of $N$ we construct $N$ positive pairs $( u_{m_1j},u_{m_2j})$ for embeddings of the same object and $N^2-N$ negative pairs $( u_{m_1j},u_{m_2k})_{j \neq k}$ for different objects. We then apply the symmetric NT-Xent contrastive loss from ConVIRT\cite{zhang2021contrastive} and popularized by CLIP~\cite{radford2021learning}:
\begin{equation}
l_j^{\vt}=- \log \frac{\exp(\langle u_{v_j}, u_{t_j}\rangle/\tau)}{\sum_{k=1}^N \exp(\langle u_{v_j}, u_{t_k}\rangle / \tau)},
\end{equation}
\begin{equation}
l_j^{\tv}=- \log \frac{\exp(\langle u_{t_j}, u_{v_j}\rangle /\tau)}{\sum_{k=1}^N \exp(\langle u_{t_j}, u_{v_k}\rangle /\tau)} 
\end{equation}%
where $\tau \in \mathbb{R}^+$ is a temperature parameter that controls the concentration of the distribution and smoothness of softmax, and $\langle,\rangle$ is the cosine similarity.
Finally we calculate a weighted sum of $l_j^{\vt}$ and $l_j^{\tv}$ and average over the minibatch:
$L(v, t) = \frac{1}{N} \sum_{j=1}^{N}(\alpha l_j^{\vt} + (1 - \alpha) l_j^{\tv})$,
where $\alpha \in [0,1]$
\vspace{4pt}
\noindent\emph{Trimodal loss:}
To extend the loss to three modalities we simply calculate the contrastive loss over all pair possibilities for the text, voxel and image representations.
This gives the final loss:
$
L_{\text{tri}}=L(v, i)+L(v, t)+L(i, t)
$.

\xhdr{Retrieval.}
For the retrieval task, we are given an input text description and we have to return the matching object.
To do this, we leverage the shared embedding space we have built between three modalities.
We consider three strategies for matching the text and shape by calculating the cosine similarity between: 1) text and voxel embeddings, 2) text and image embeddings, and 3) text and sum of image and voxel embeddings.

\section{Experiments}
\label{sec:exp}

In the main paper, we present experiments on text-to-shape retrieval on the `chair and tables' dataset from Text2Shape~\cite{chen2018text2shape}. The `chairs and tables' dataset consists of solid colored voxels of $6521$ chairs and $8378$ tables from ShapeNet~\cite{chang2015shapenet}, and text descriptions collected from humans ($\approx5$ per shape).  We follow the train/val/test split by \citet{chen2018text2shape} which ensures that shapes do not occur in the same split.
We present additional results on shape-to-text retrieval, and results on the primitives dataset in the supplement. 
To show our method works beyond `chairs and tables', we conduct text-to-shape retrieval on an extended set of 13 object categories from ShapeNet~\cite{chang2015shapenet} (see supplement).
Results across experiments consistently show that our trimodal model outperforms bimodal models.

\subsection{Metrics}

We follow prior work on text-to-shape retrieval~\cite{chen2018text2shape,han2019y2seq2seq,tang2023parts2words} and use standard metrics of Recall Rate (RR@k) and Normalized Discounted Cumulative Gain (NDCG)~\cite{jarvelin2002cumulated}.
RR@k deems a retrieval successful if the ground truth (GT) appears in the top $k$ candidates (we set $k$ to $1$ and $5$).
NDCG compares the ranked retrieval results with optimal ranking.
However, since we assume there is only one relevant shape for each query, we do not take full advantage of the NDCG metric.
We also evaluate using Mean Reciprocal Rank (MRR), the average of the inverse of the rank of the GT.

We note that there are often multiple shapes that can match the text description.
Since the text description can be underspecified, we also measure the similarity of the top $k$ retrieved shapes to the GT shape.
Following work in shape retrieval~\cite{kuo2020mask2cad}, we use a point-wise $F1^{\tau}$ with $\tau=0.1$ to calculate shape similarity.
$F1^{\tau}$ is the harmonic mean of the fraction of points from retrieved shapes within $\tau$ of a point from GT (point-wise precision), and the fraction of points from GT within $\tau$ of a point from retrieved shapes (point-wise recall).
To compute $F1^{\tau}$, we sample 10K points uniformly on the mesh surface of GT and retrieved shapes.
See the supplement for more details. %

\subsection{Implementation details}
We use a one-layer bi-directional GRU~\cite{cho2014learning} for the text encoder, and a 3D CNN architecture for the voxel encoder.
We use the pretokenized and lemmatized text from \citet{chen2018text2shape}, with a vocabulary consisting of 3587 unique words and 1 pad token. 
For the Bi-GRU, we use word embedding size of 256, and a hidden state size of 128.
Word embeddings are randomly initialized from a standard Normal distribution.
For the 3D CNN, we use 5 Conv3D layers of sparse convolutions from the \texttt{spconv} library.\footnote{\href{https://github.com/traveller59/spconv}{https://github.com/traveller59/spconv}}
For multi-view images we use the MVCNN~\cite{su2015multiview} architecture with pretrained ResNet18~\cite{he2016deep} backbone.
A fully-connected layer is added to ensure the output dimension for all encoders is 512.
Unless otherwise specified, training uses batch size $128$, voxel resolution $64^3$, image resolution $128^2$ and 6 images for the MVCNN.
In preprocessing, we normalize image and voxel values from 0-255 to 0-1.
We implement our models using PyTorch~\cite{pytorch} and train with the Adam optimizer~\cite{kingma2014adam}.
We use a learning rate of $0.00035$ for batch size $128$, and adopt the linear scaling rule~\cite{goyal2017accurate} to scale the learning rate for other batch sizes.
We train for up to 20 epochs until convergence, and select the checkpoint with the best performance on the val set.
All models are trained on an A40 GPU with each experiment taking about 1 hour.
Our models are memory efficient with most models requiring less than 12GB (see supplement).
We render the multiview images with Pyrender\footnote{\href{https://github.com/mmatl/pyrender}{https://github.com/mmatl/pyrender}}
from 12 camera positions elevated slightly above the object, pointing towards the object, and separated by 30 degrees.
For multiview experiments using fewer images, we subsample so images are evenly spaced.

\subsection{Models}

\xhdr{Baselines.}
We compare to Text2Shape~\cite{chen2018text2shape}, Y2Seq2Seq \cite{han2019y2seq2seq} and Parts2Words~\cite{tang2023parts2words}.
Text2Shape~\cite{chen2018text2shape} uses a triplet loss~\cite{song2016deep} combined with learning by association~\cite{haeusser2017learning}.
Y2Seq2Seq~\cite{han2019y2seq2seq} uses a view-based model and a triplet constraint.  For Parts2Words, we report results for a global model (no part modeling) and their full model that uses part information. 
Parts2Words~\cite{tang2023parts2words} uses point clouds as input instead of voxels.
The global model uses  PointNet~\cite{qi2017pointnet} as the feature encoder and a Bi-GRU as the text encoder and aggregates the point and text features.  
Parts2Words jointly embeds point clouds and text by aligning parts from shapes and words from sentences.
Both the global and the part-based models use a semi-hard negative mining triplet ranking loss.  We note that these methods use either more complex losses or require additional labelled data compared to our model.
In addition to baselines from prior work, we use two random baselines: one computes the expected metric mathematically, and the other uses our architecture with random weights.

\xhdr{Our models.}
We train variants of our model with just two modalities (Bi) or all three modalities (Tri).
For the bimodal models, we only consider text and image (\textbf{I}), or text and voxels (\textbf{V}).
During retrieval, we compute the similarity of the text with image (\textbf{I}), or text with voxels (\textbf{V}). In the case of trimodal embedding, we also use a combination of the image and voxel when computing the similarity, with (\textbf{I}+\textbf{V}) denoting that the retrieval was done by calculating similarity with text and the sum of image and voxel representations.

\begin{figure*}
\setkeys{Gin}{width=\linewidth}
\begin{tabularx}{\textwidth}{@{}p{0.2cm}p{6cm}YYYYY@{}}
\toprule
 & & top1 & top2 & top3 & top4 & top5\\
\midrule
1&\footnotesize{an L-shaped dark brown colored wooden table.}&
\includegraphics[trim=50 120 50 270,clip]{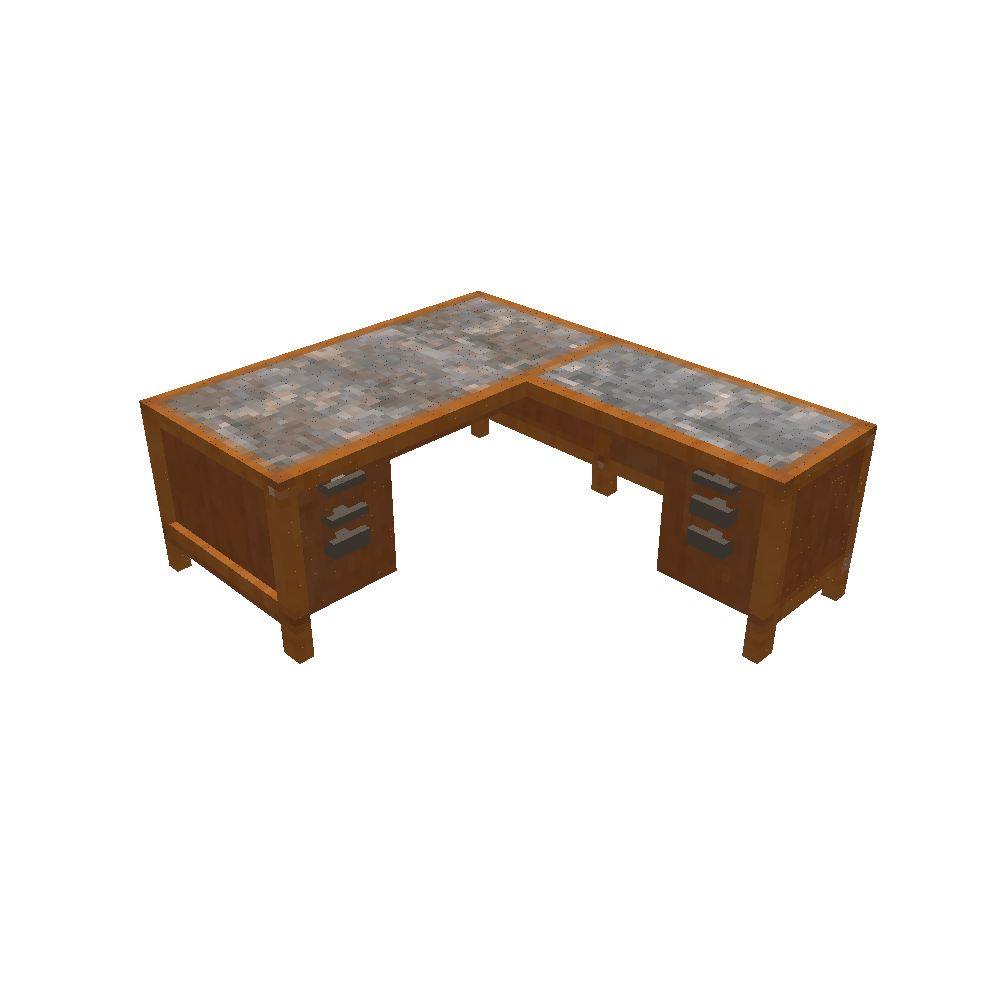}& 
\includegraphics[trim=50 120 50 270,clip]{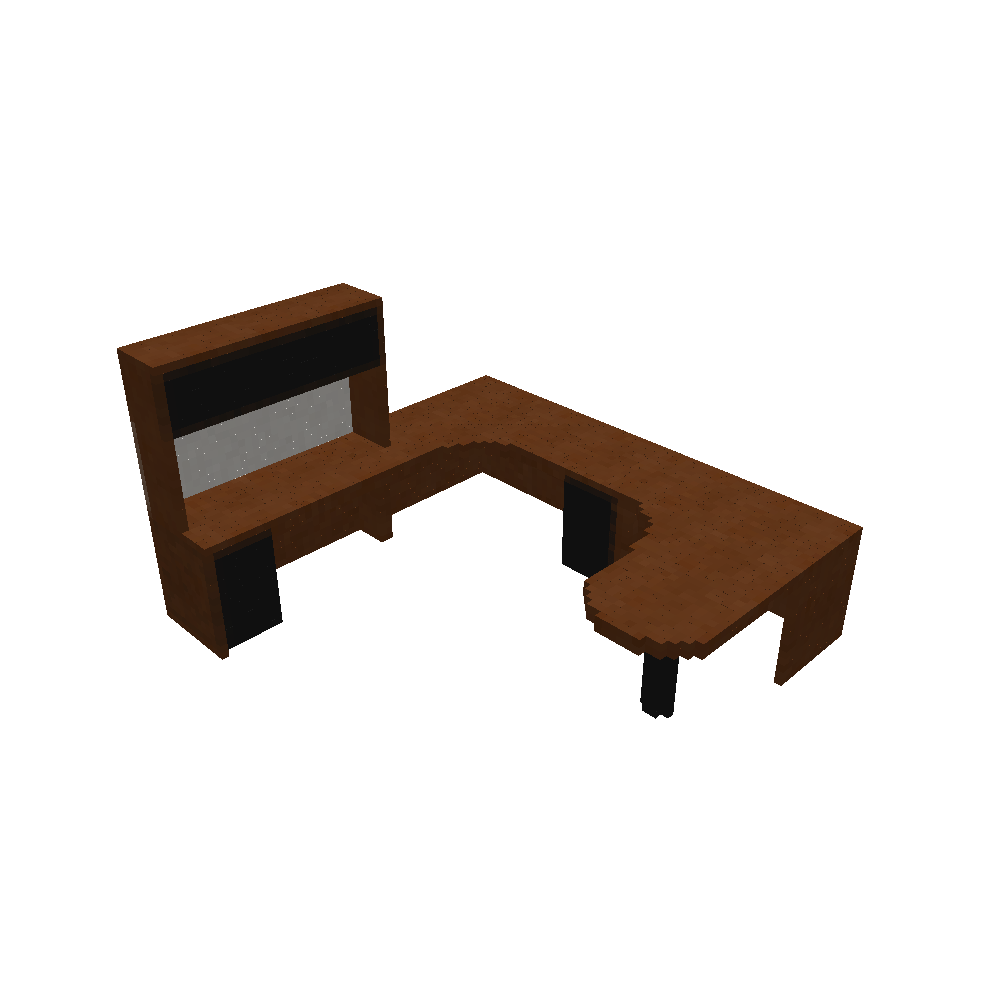}&
\includegraphics[trim=50 120 50 270,clip]{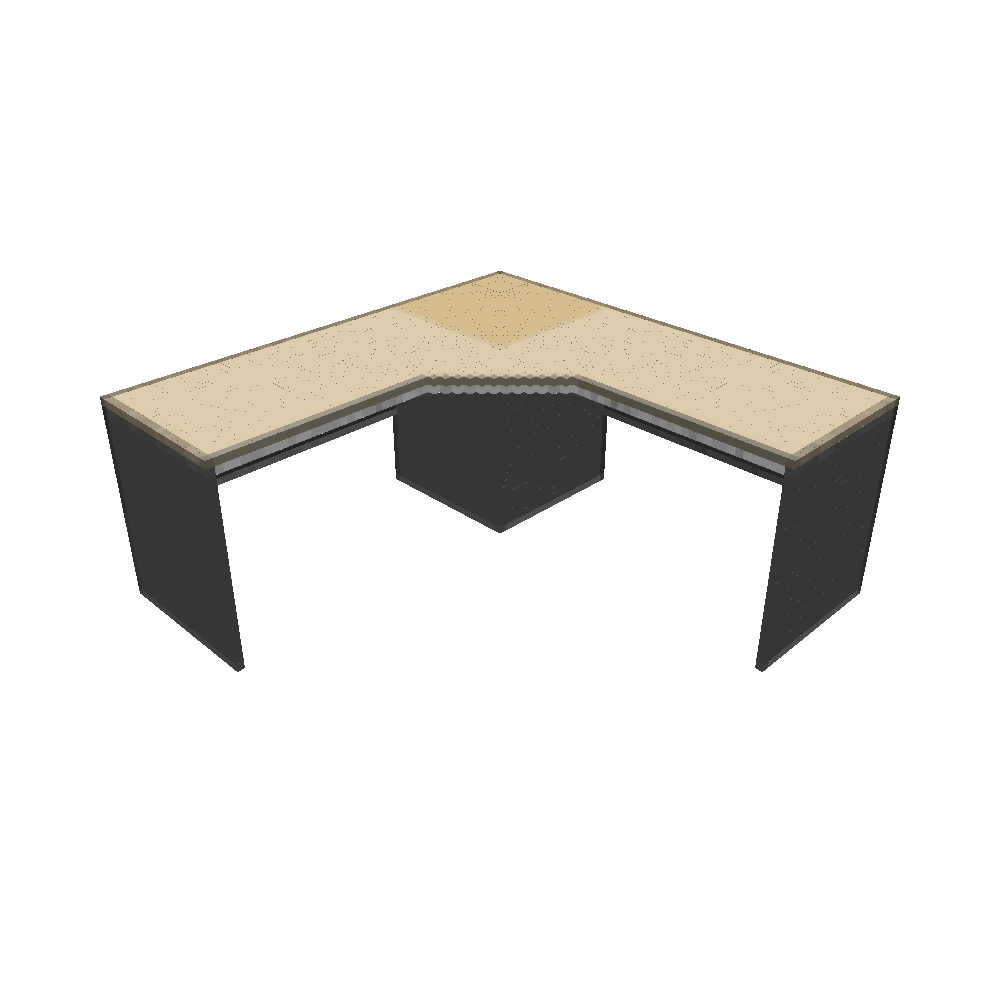}&
\includegraphics[trim=50 120 50 270,clip]{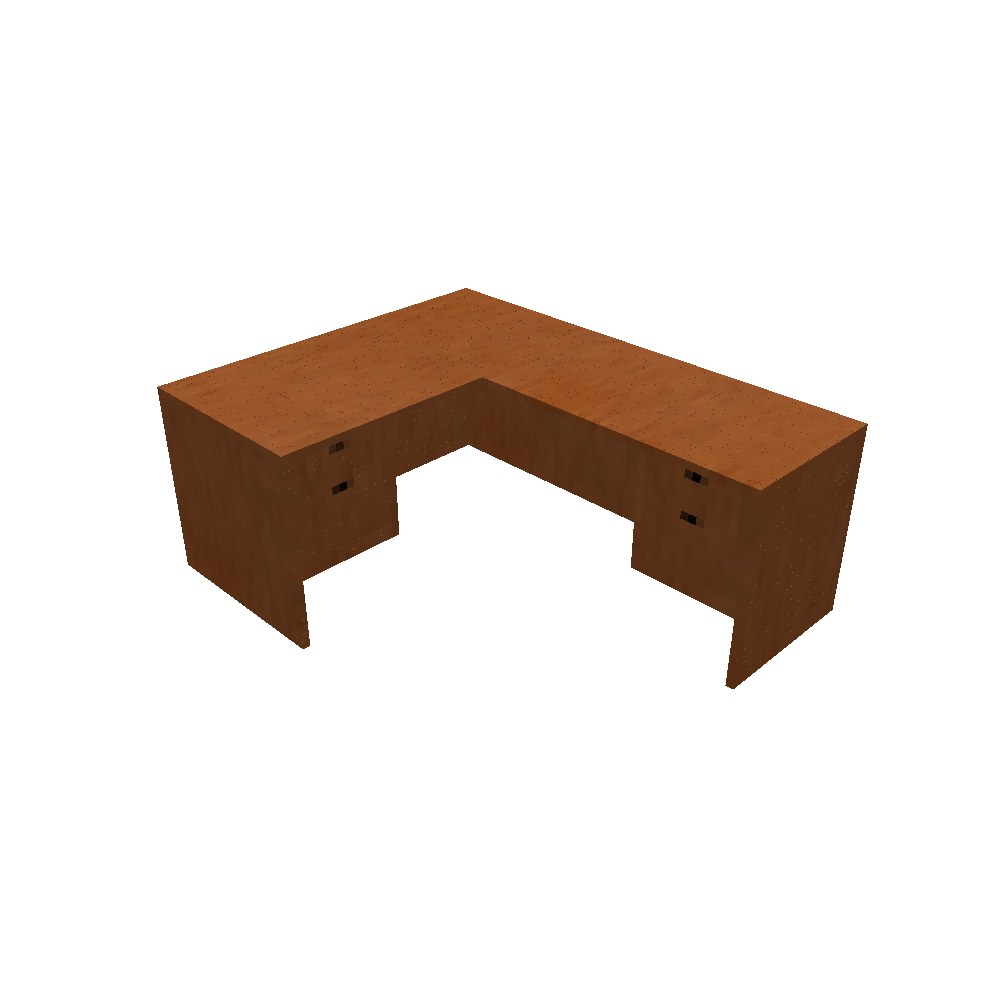}& 
\includegraphics[trim=50 120 50 270,clip]{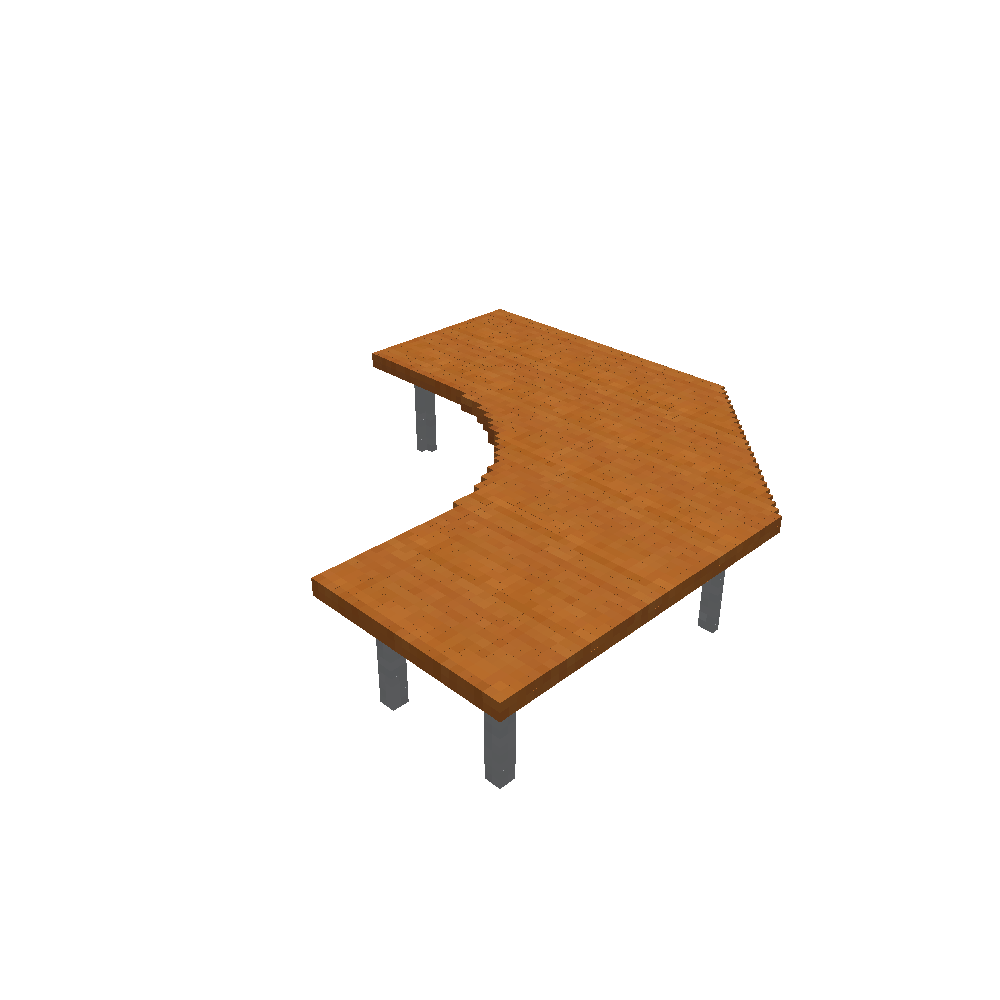}\\
[-0.2cm]
& & 17.31 & 6.60 & {\color{partialmatch1}20.34} & {\color{groundtruth}GT} & 2.19 \\
\midrule

2&\footnotesize{a luxurious gray leather modern concept plush chair with stainless steel frame foots}&
\includegraphics[trim=50 40 50 160,clip]{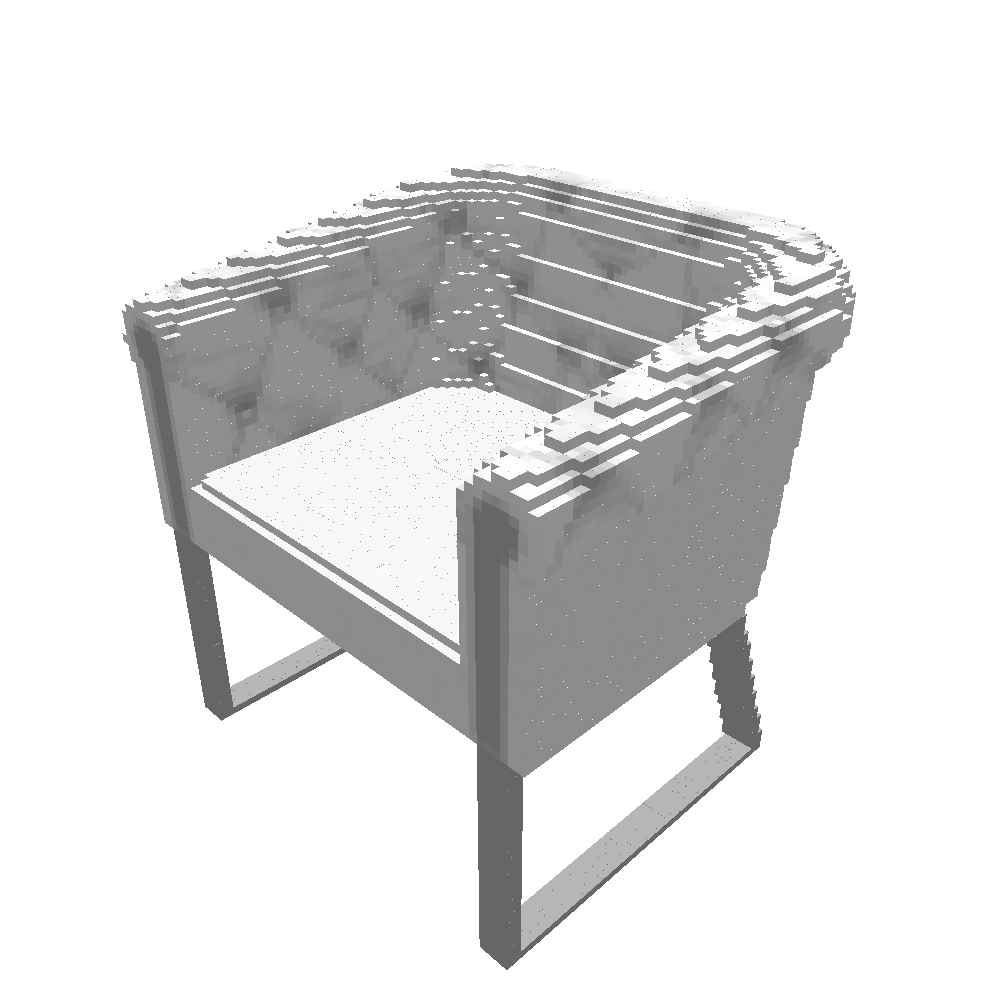}&
\includegraphics[trim=50 40 50 160,clip]{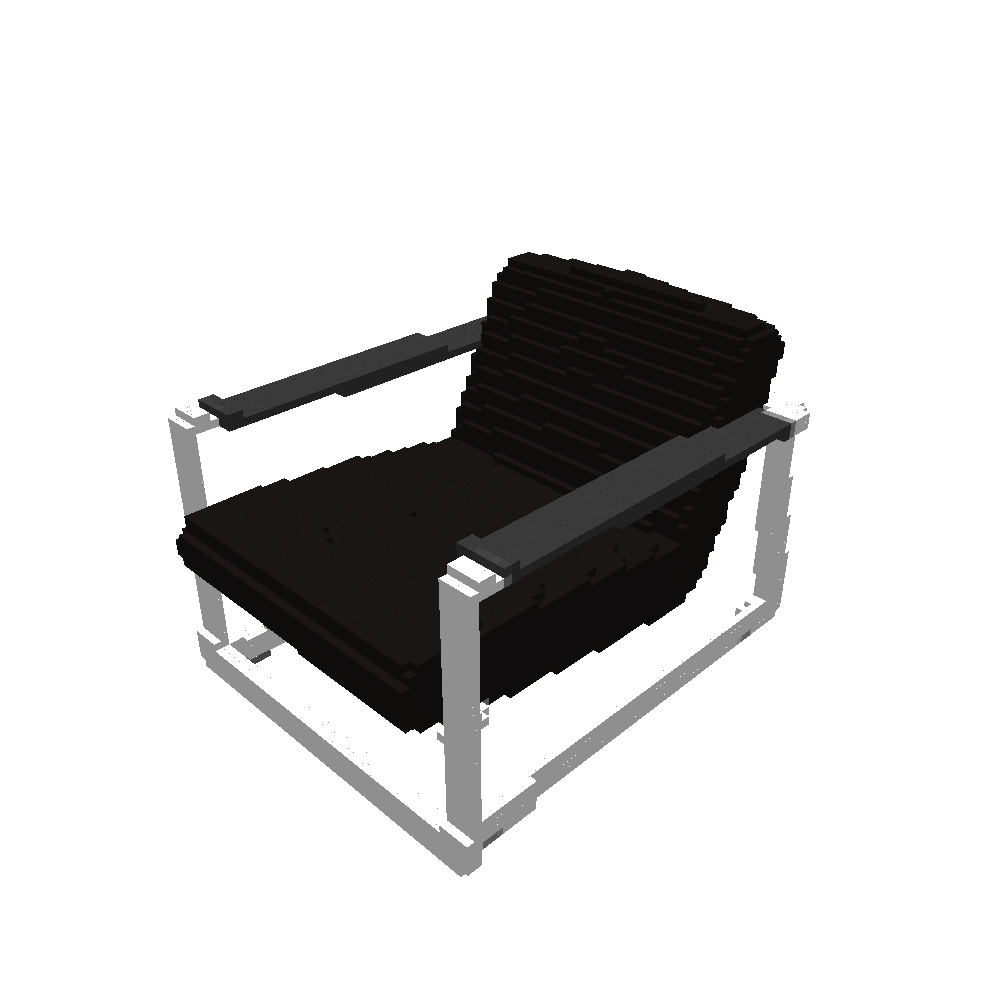}&
\includegraphics[trim=50 40 50 160,clip]{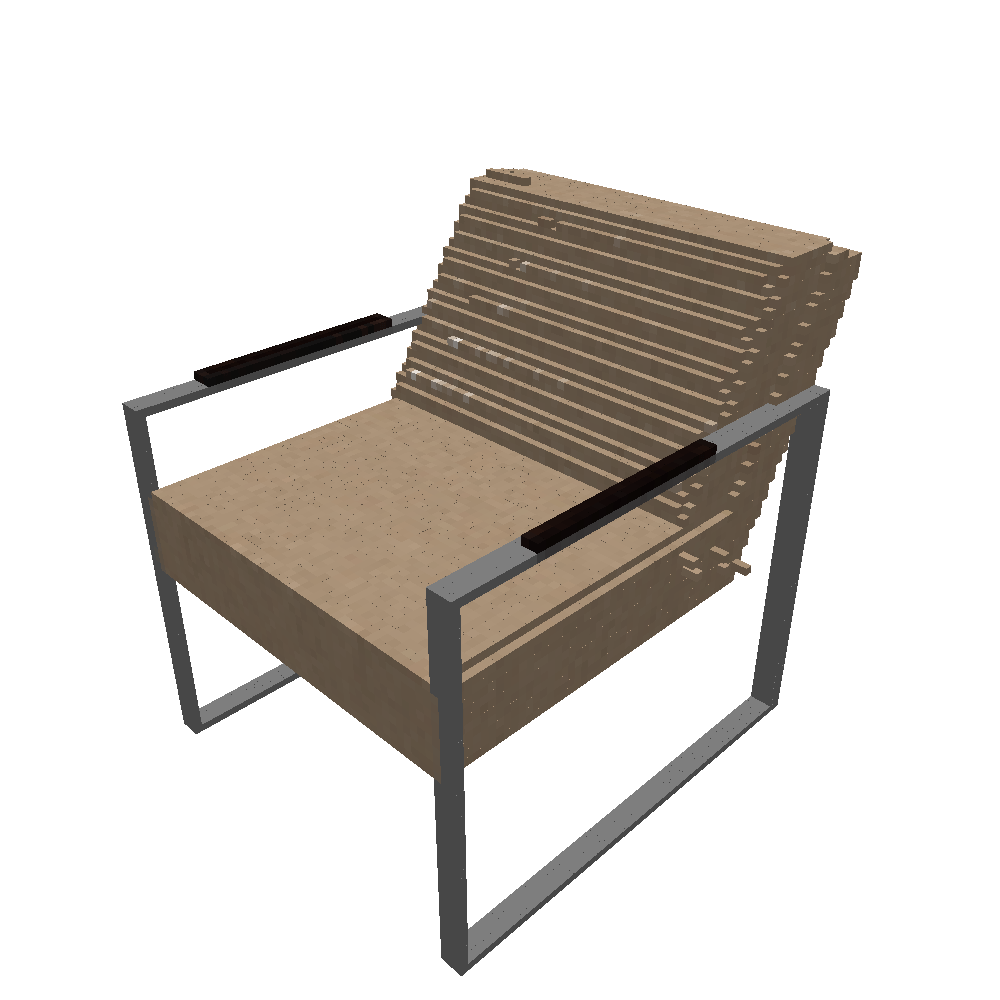}&
\includegraphics[trim=50 40 50 160,clip]{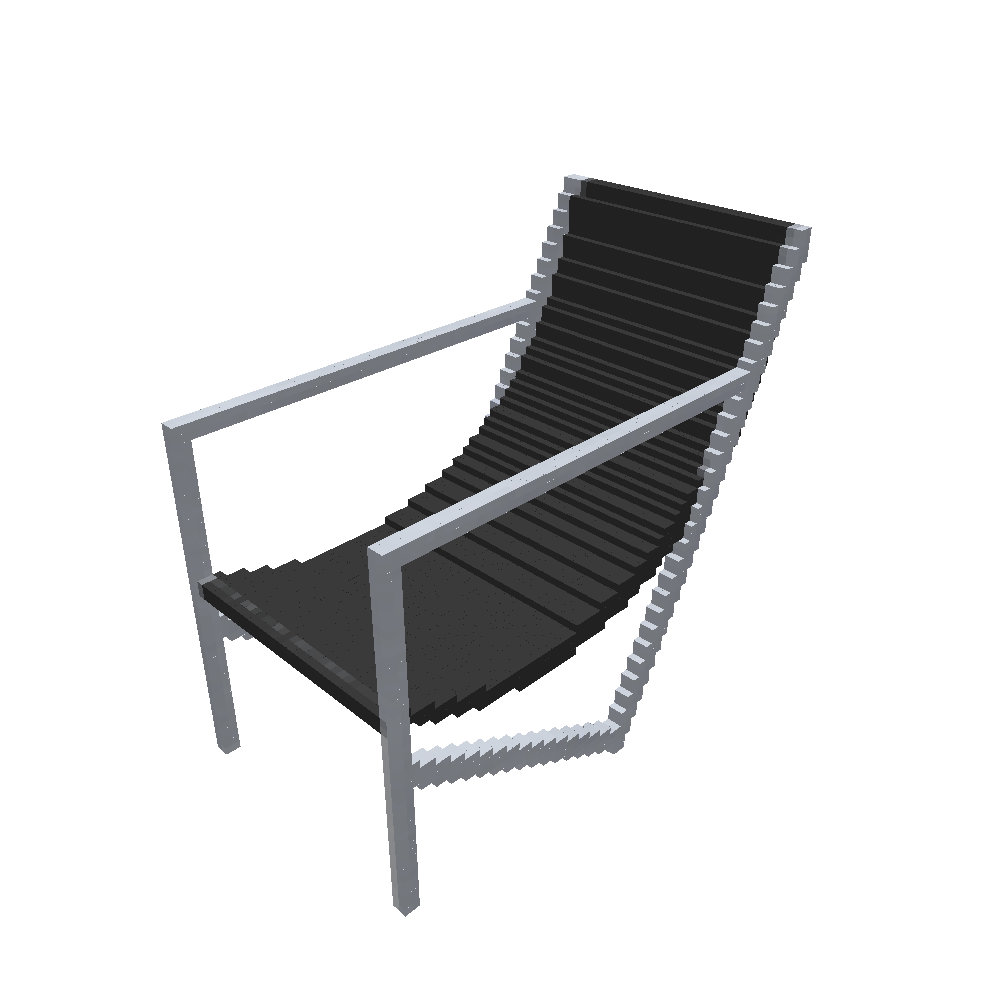}&
\includegraphics[trim=50 40 50 160,clip]{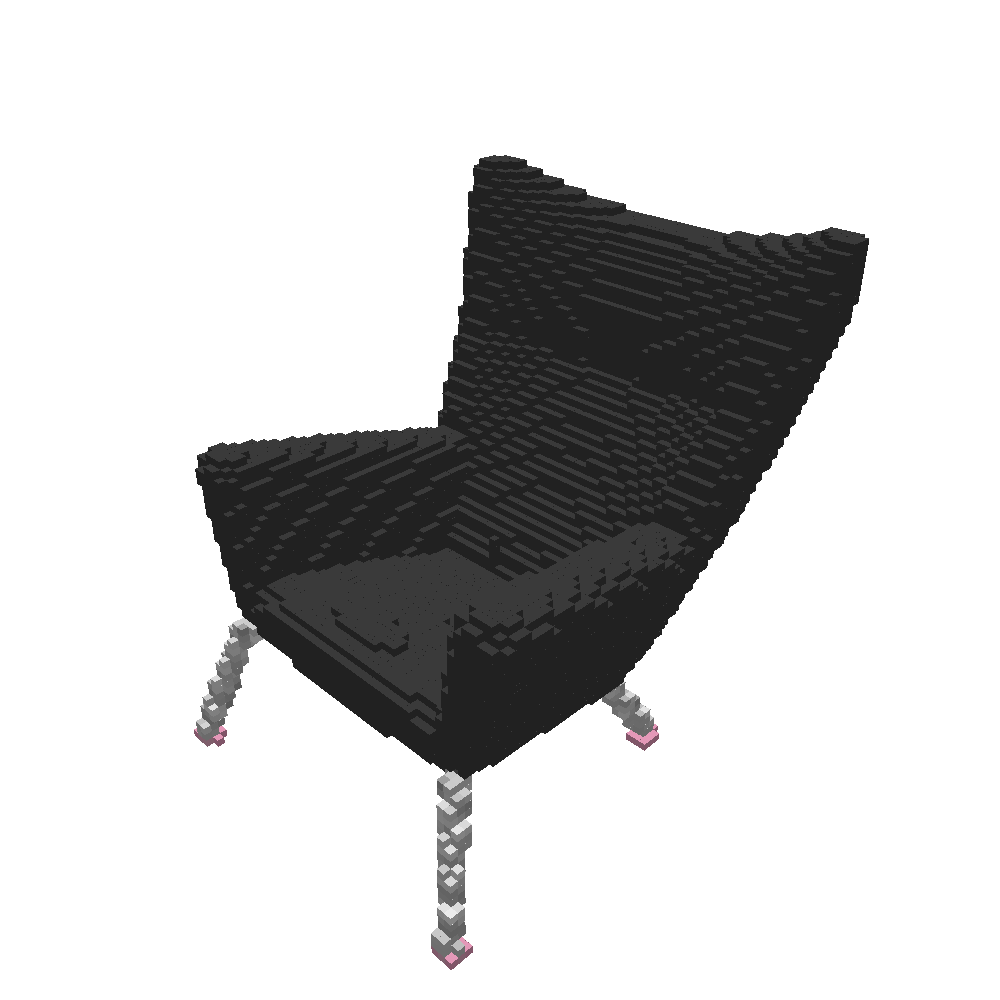}\\
& & {\color{groundtruth}GT} & {\color{partialmatch1}2.89} & {\color{partialmatch1}9.50} & {\color{partialmatch1}1.78} & {\color{partialmatch1}3.42}\\
\midrule

3&\footnotesize{simple circular table with no leg and only one circular base.}&
\includegraphics[trim=50 100 50 170,clip]{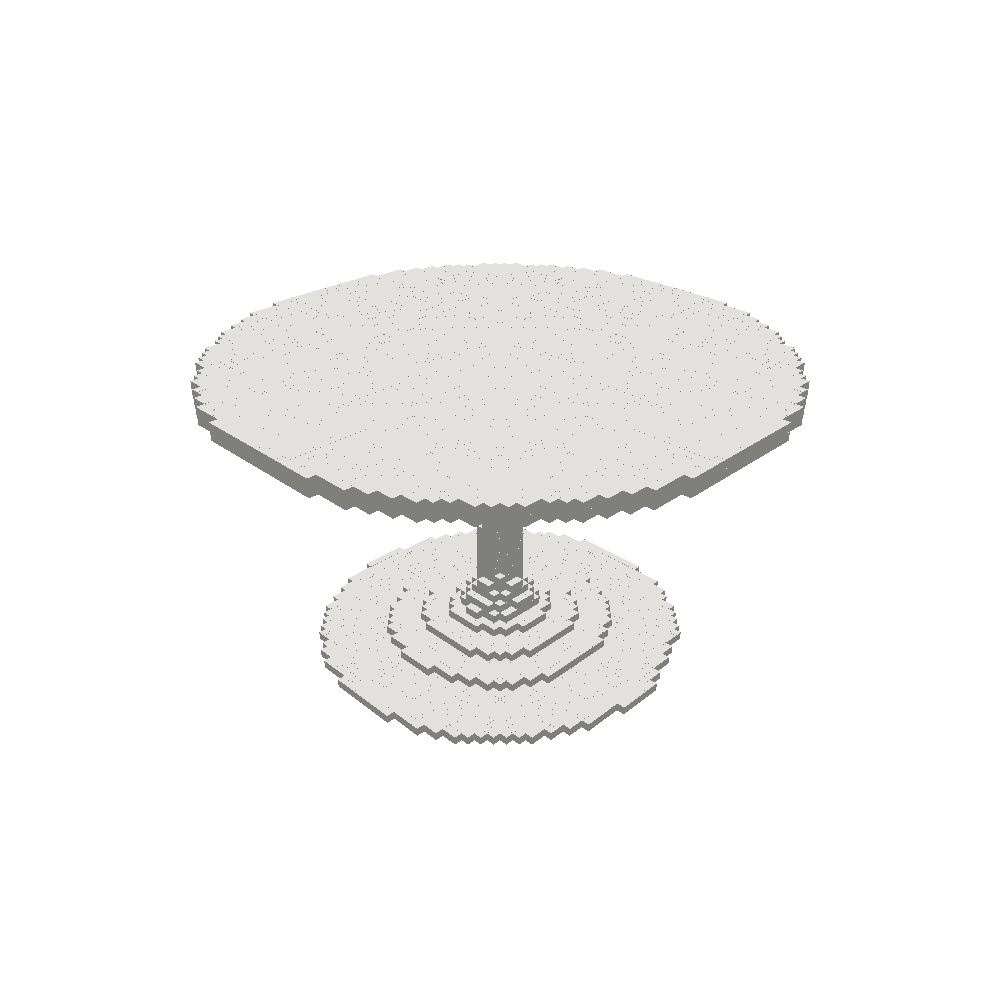}&
\includegraphics[trim=50 100 50 170,clip]{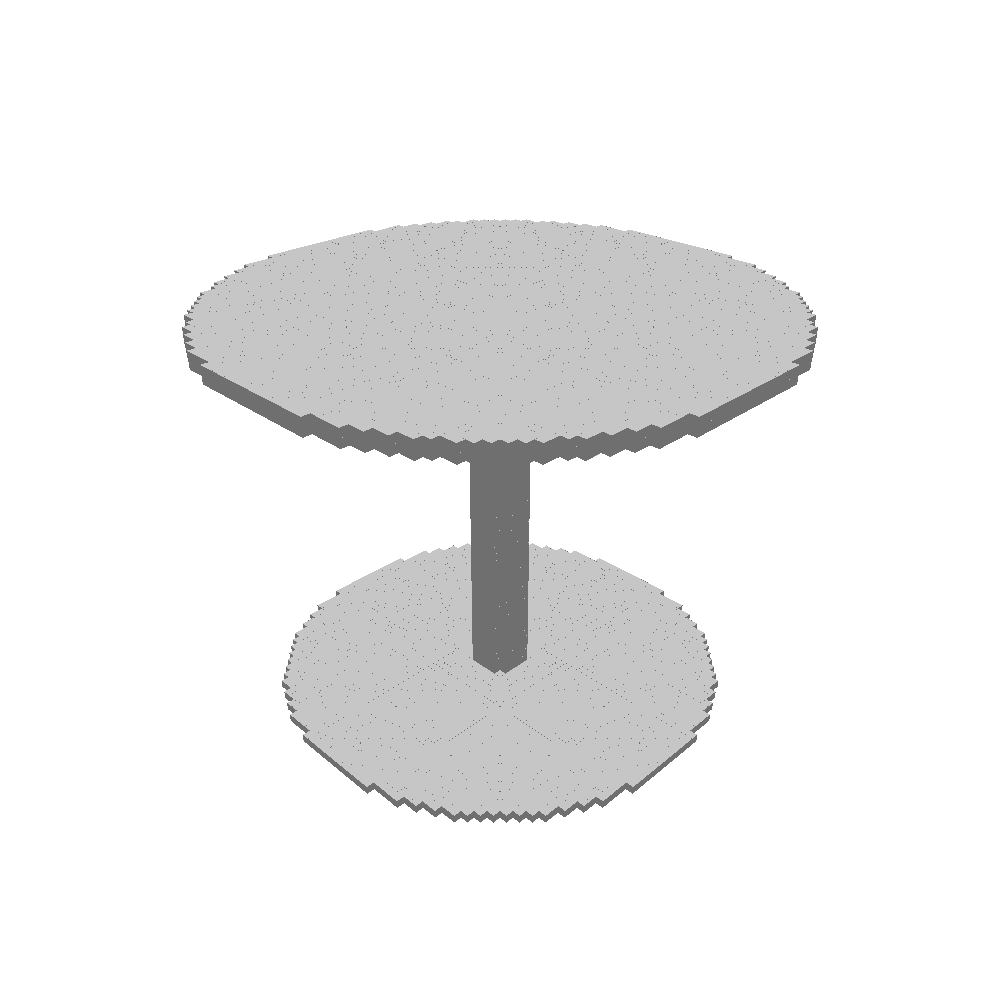}&
\includegraphics[trim=50 100 50 170,clip]{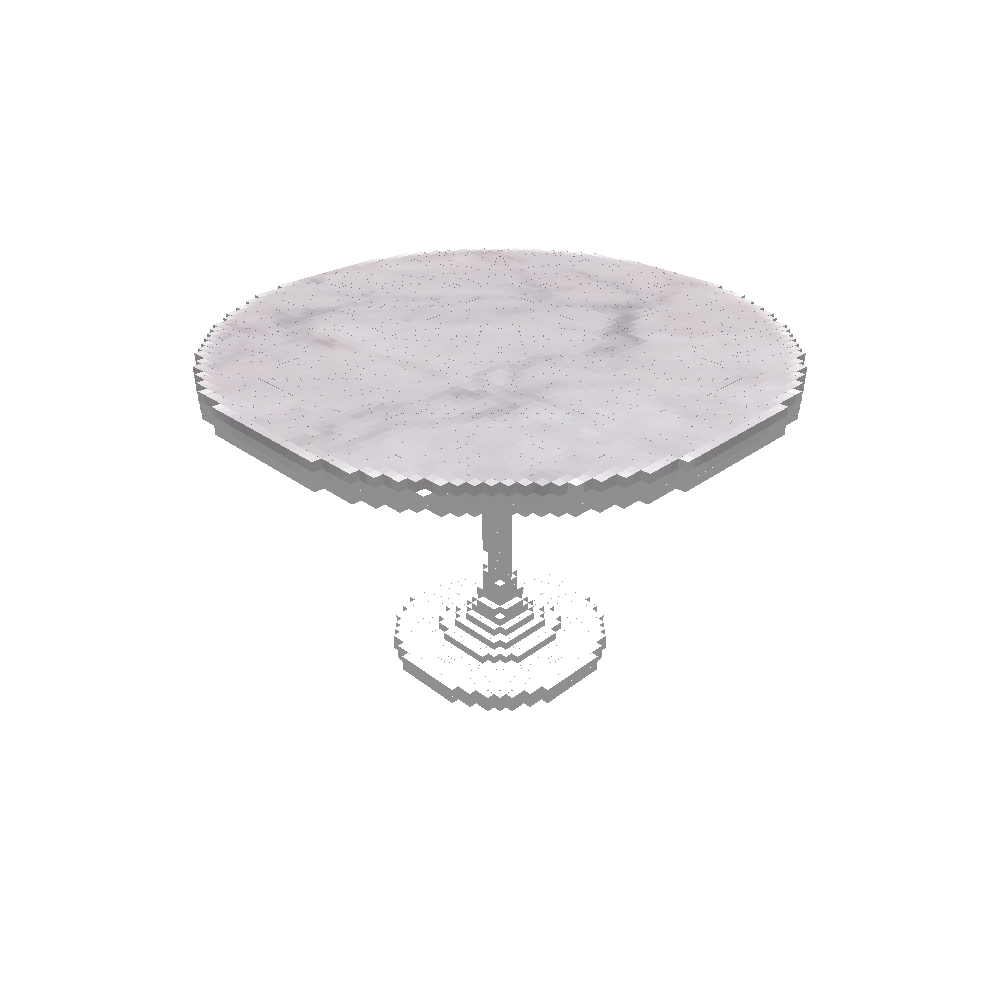}&
\includegraphics[trim=50 100 50 170,clip]{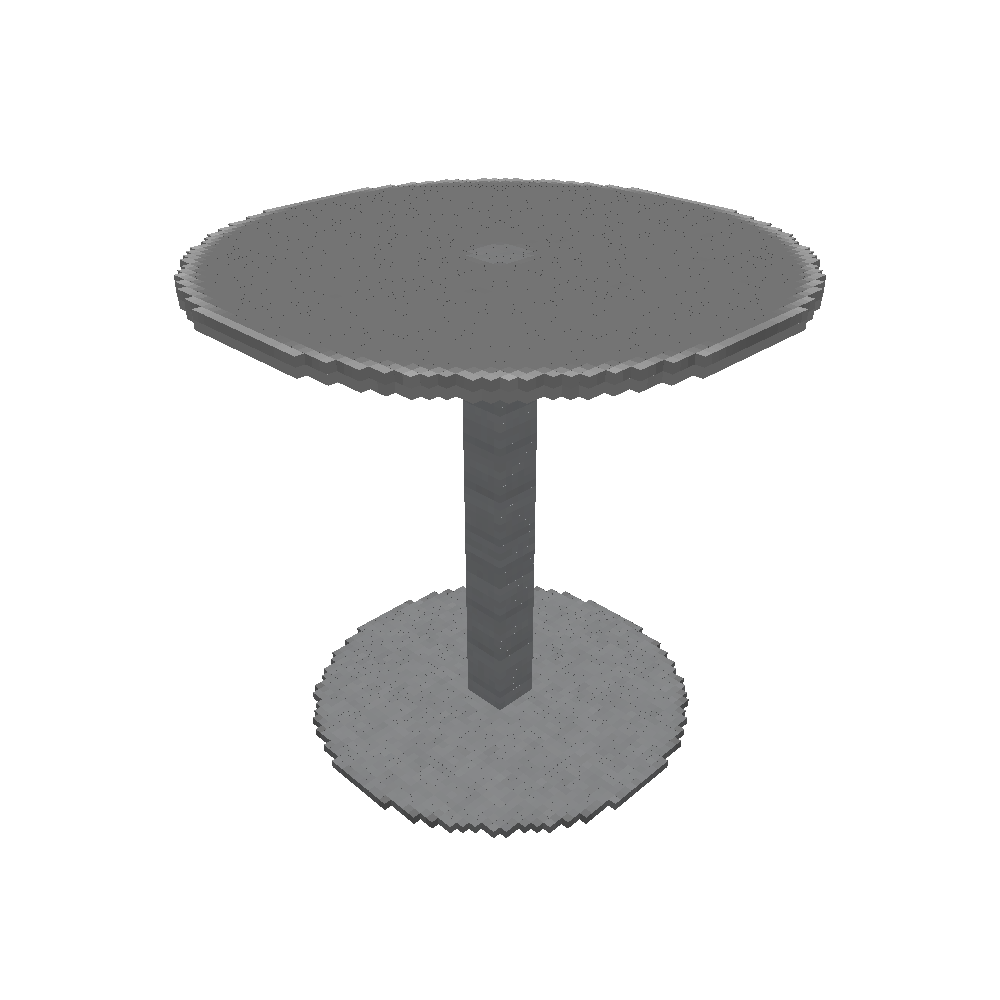}&
\includegraphics[trim=50 100 50 170,clip]{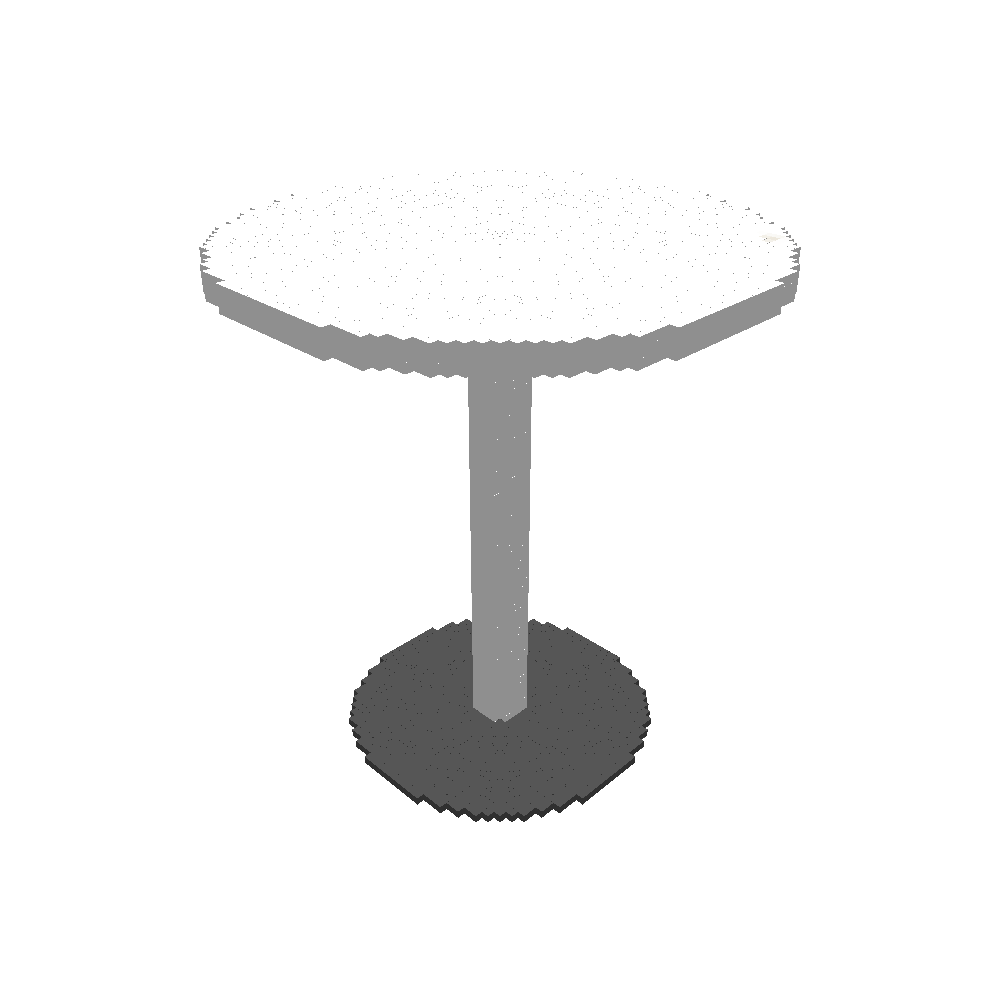}\\
[-0.1cm]
& & 0.79 & 5.77 & 0.59 & {\color{groundtruth}GT} & 6.32\\
\midrule

4&\footnotesize{This is greenish top wooden billiards table.}&
\includegraphics[trim=50 100 50 240,clip]{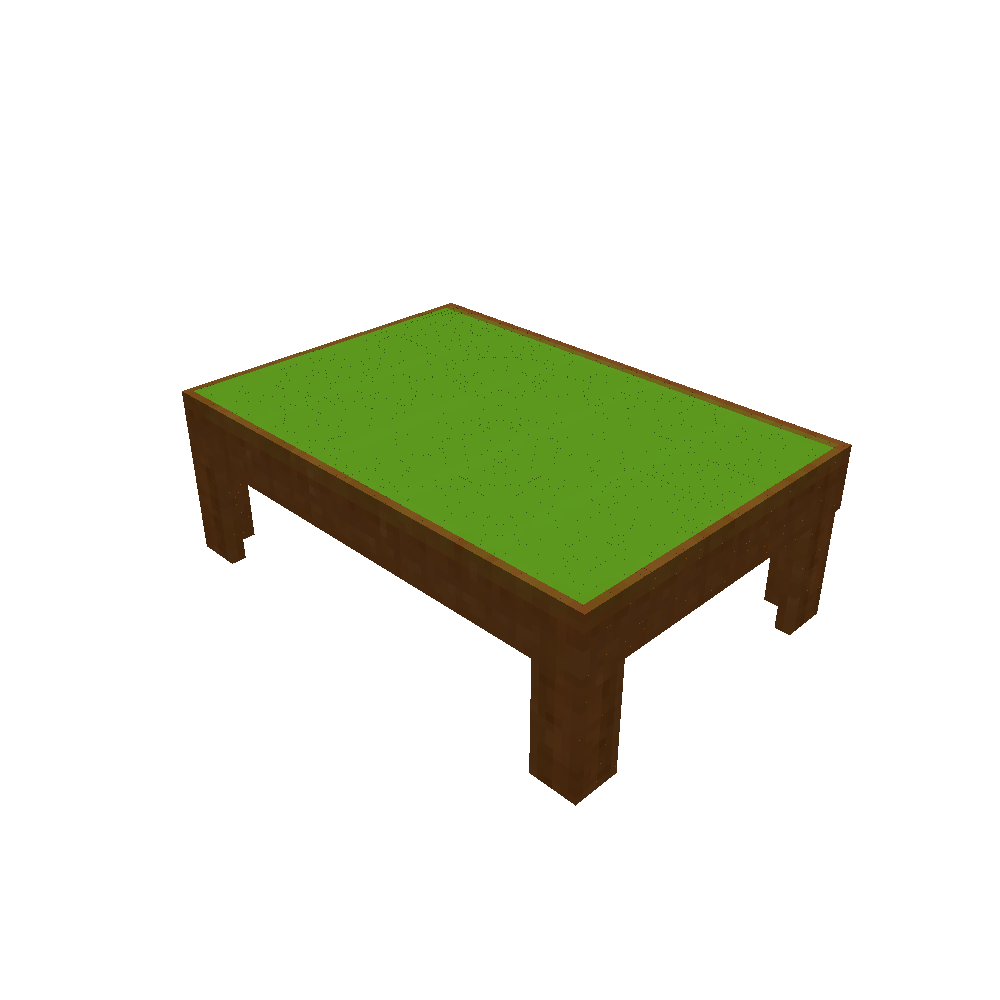}&
\includegraphics[trim=50 100 50 240,clip]{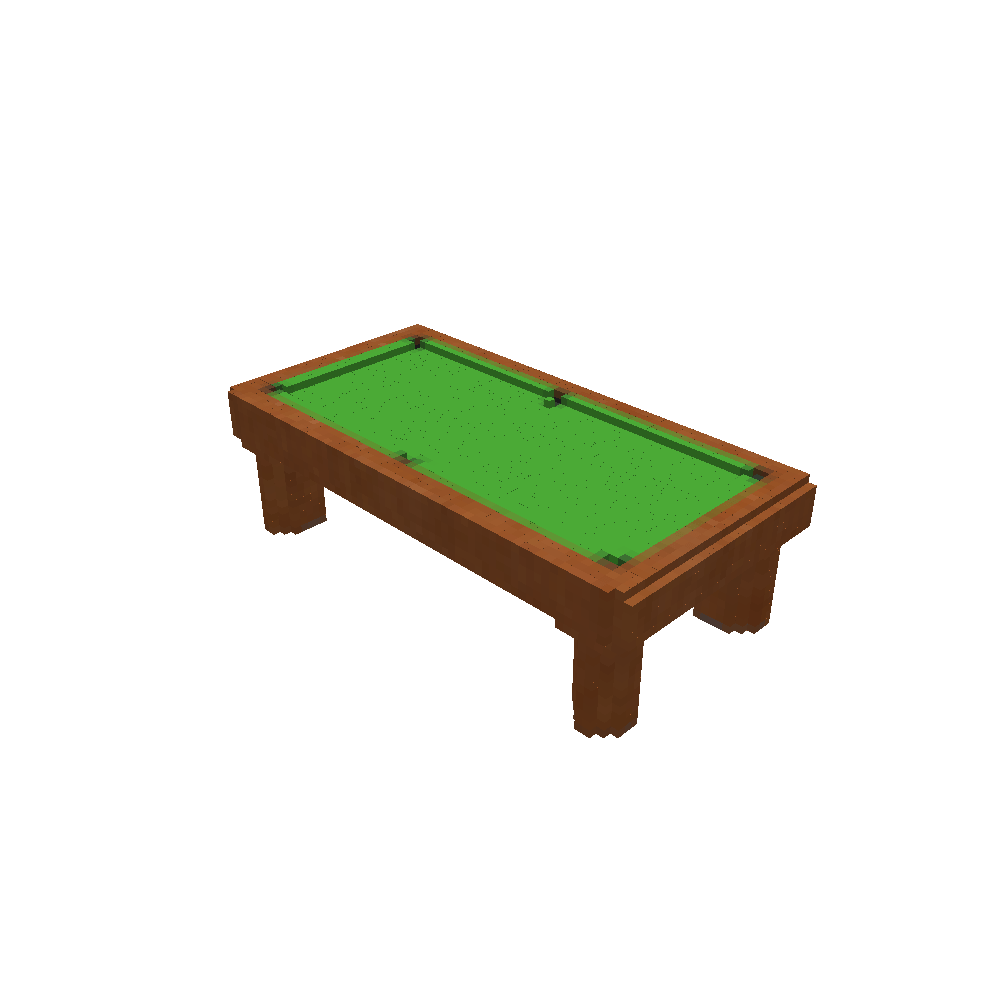}&
\includegraphics[trim=50 100 50 240,clip]{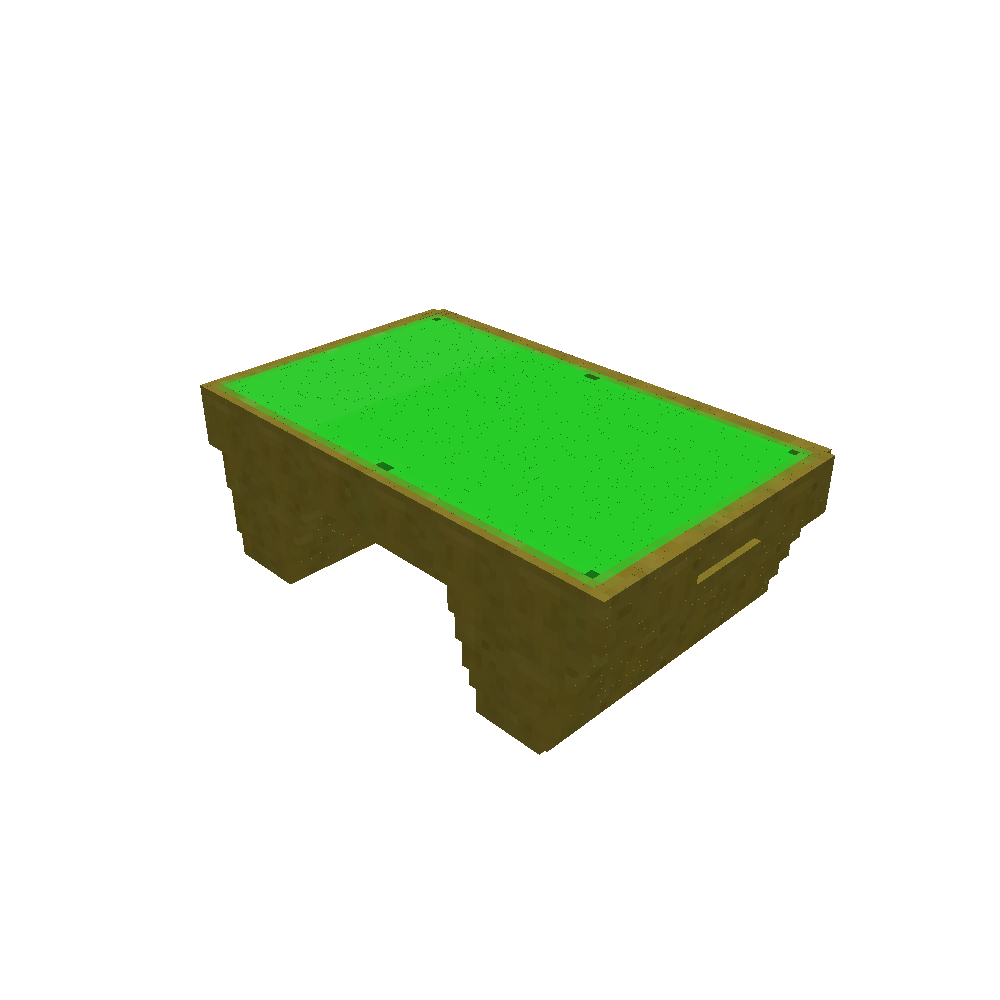}&
\includegraphics[trim=50 100 50 240,clip]{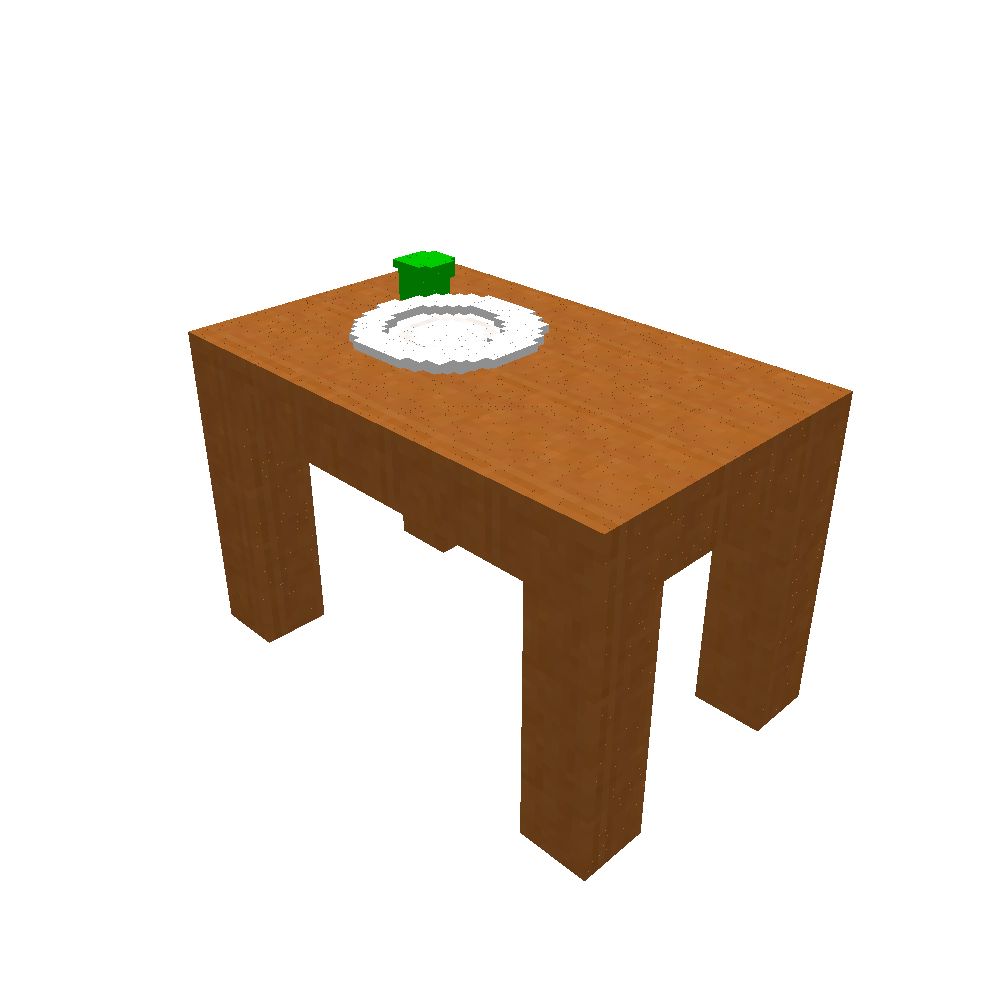}&
\includegraphics[trim=50 100 50 240,clip]{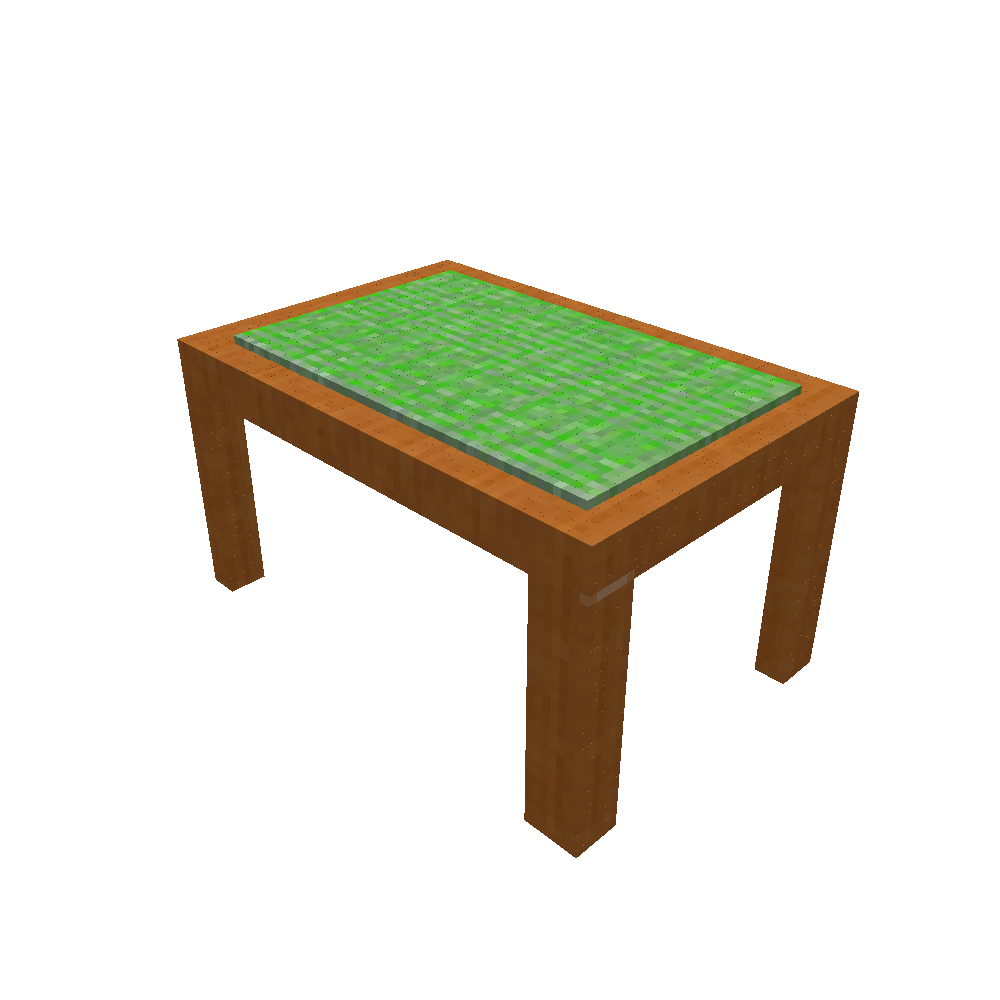}\\
[-0.1cm]
& & 15.79 & {\color{groundtruth}GT} & 8.04 & {\color{partialmatch1}19.59} & 3.18\\
\midrule

5&\footnotesize{this is a boxy look gray chair. It appears to be made out of granite and is gray with 4 short legs and a high, arched back.}&
\includegraphics[trim=50 110 50 160,clip]{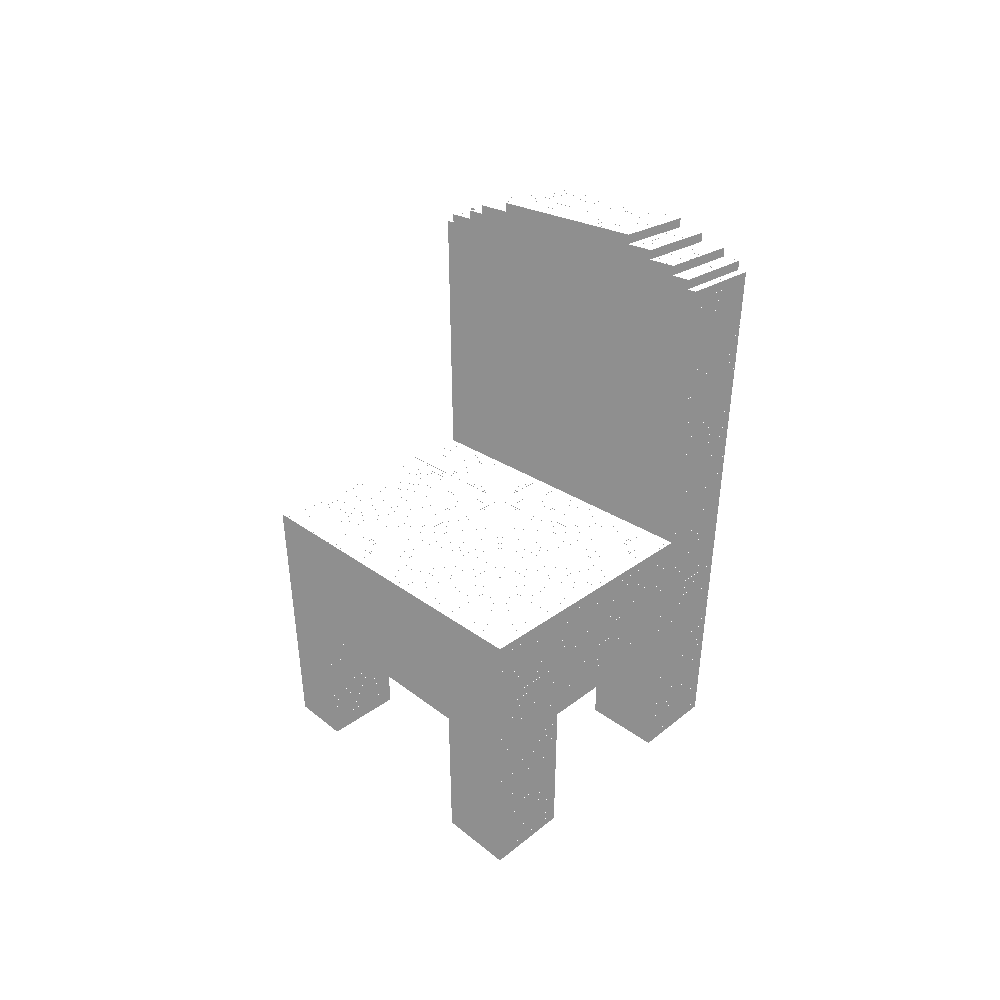}&
\includegraphics[trim=50 110 50 160,clip]{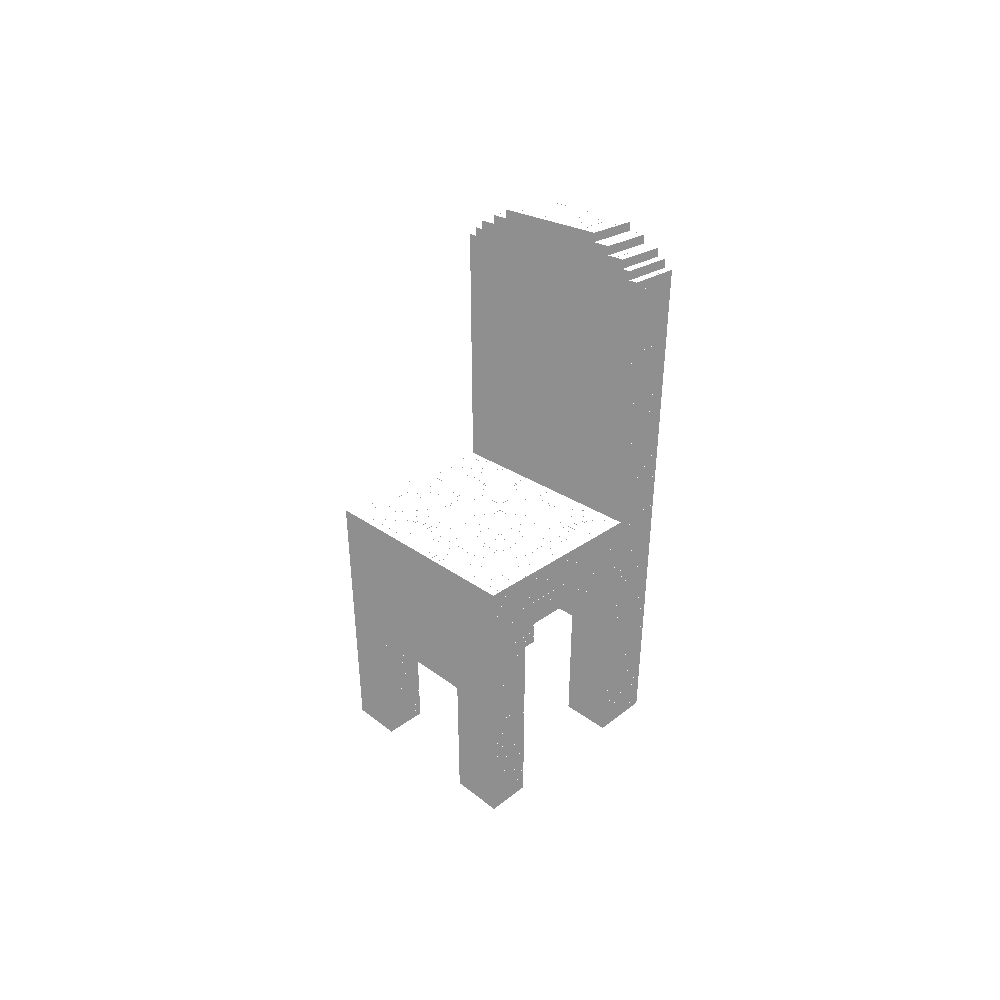}&
\includegraphics[trim=50 110 50 160,clip]{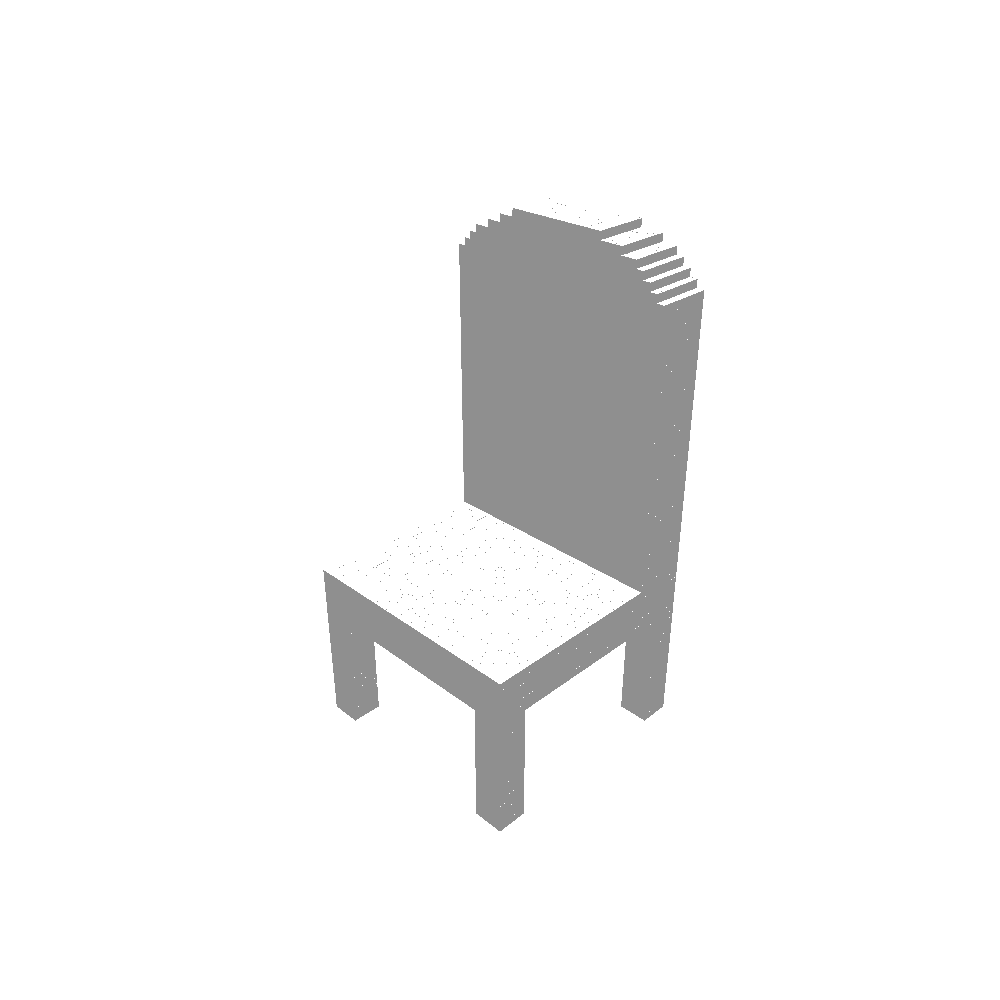}&
\includegraphics[trim=50 110 50 160,clip]{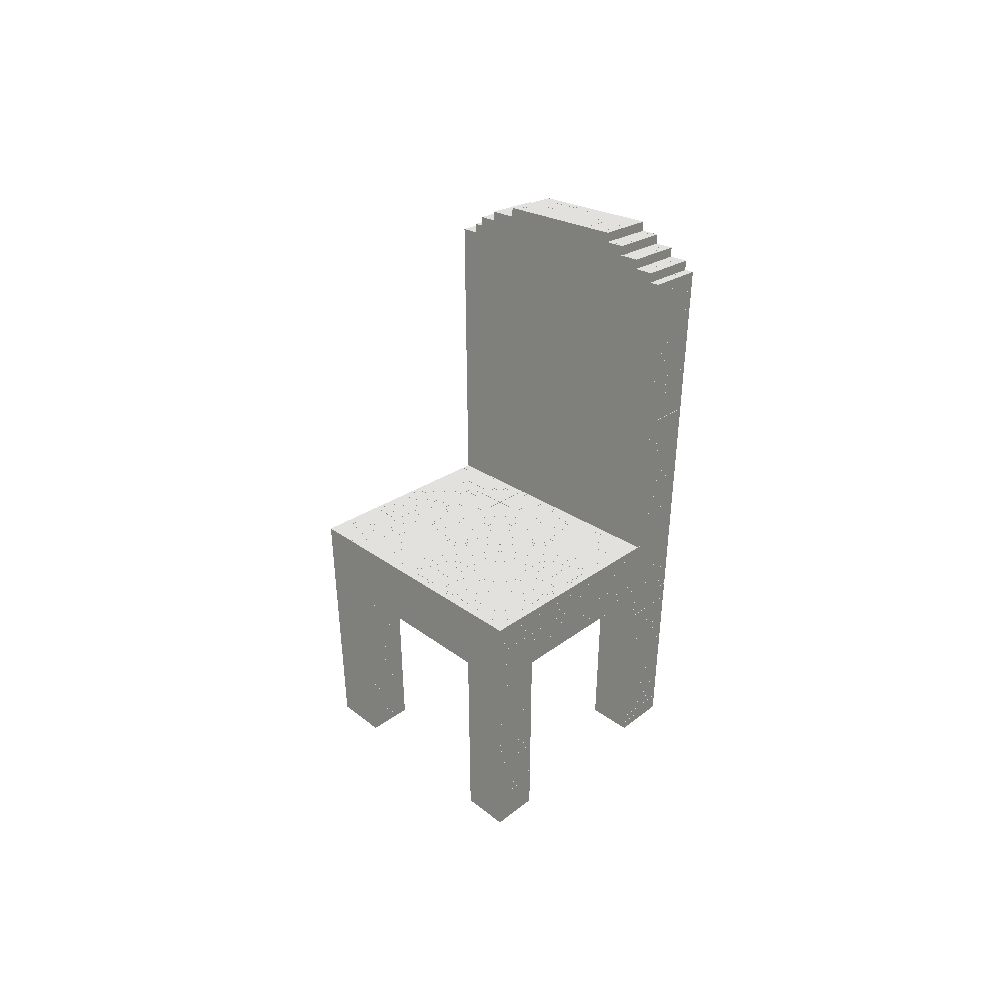}&
\includegraphics[trim=50 110 50 160,clip]{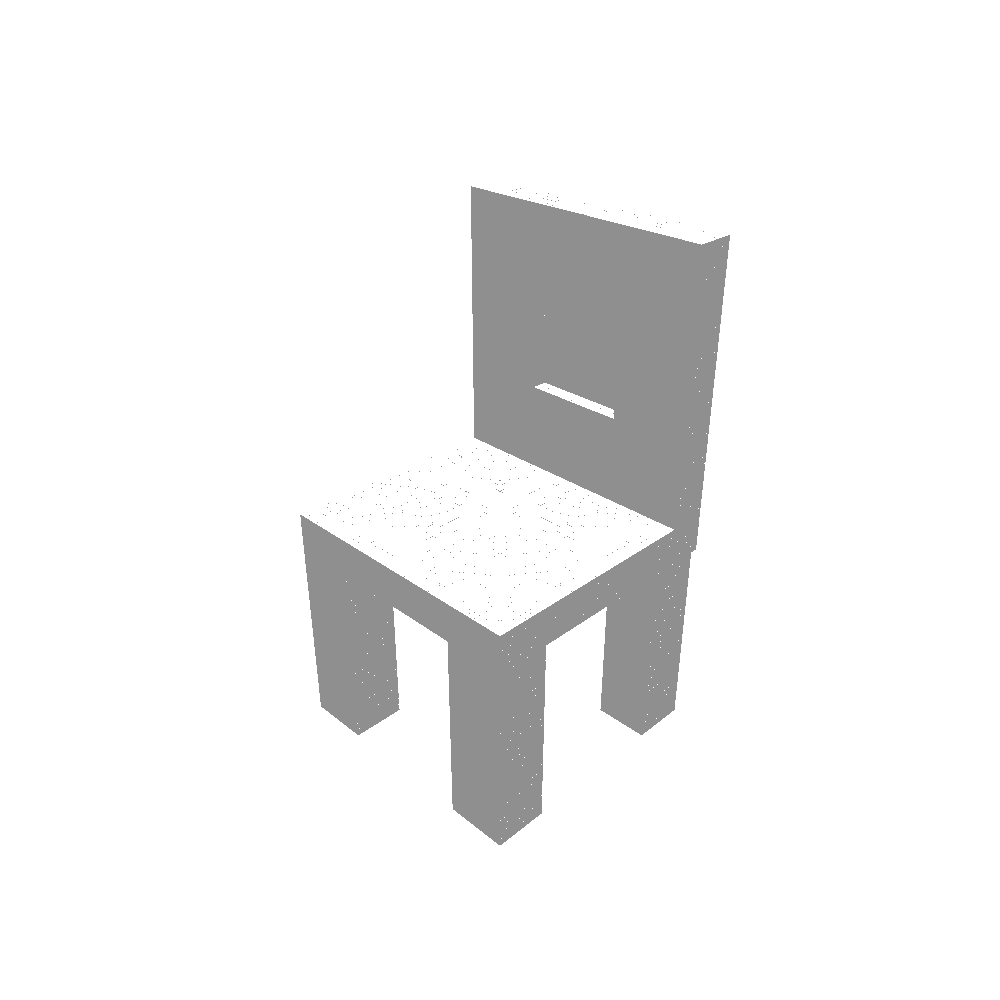}\\
[-0.1cm]
& & {\color{groundtruth}GT} & 4.64 & 22.32 & 12.97 & 11.42\\
\midrule

6&\footnotesize{wooden armless dining room chair with open nine-square back.}&
\includegraphics[trim=50 110 50 170,clip]{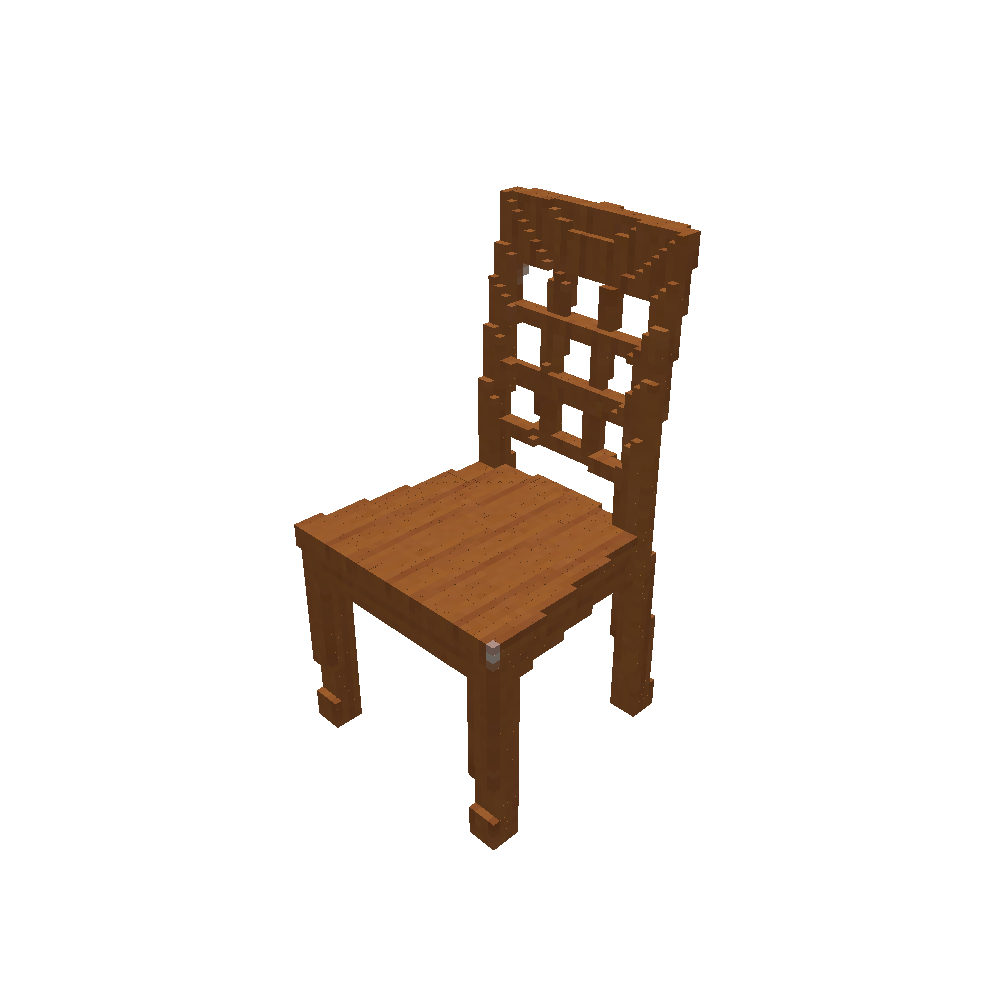}&
\includegraphics[trim=50 110 50 170,clip]{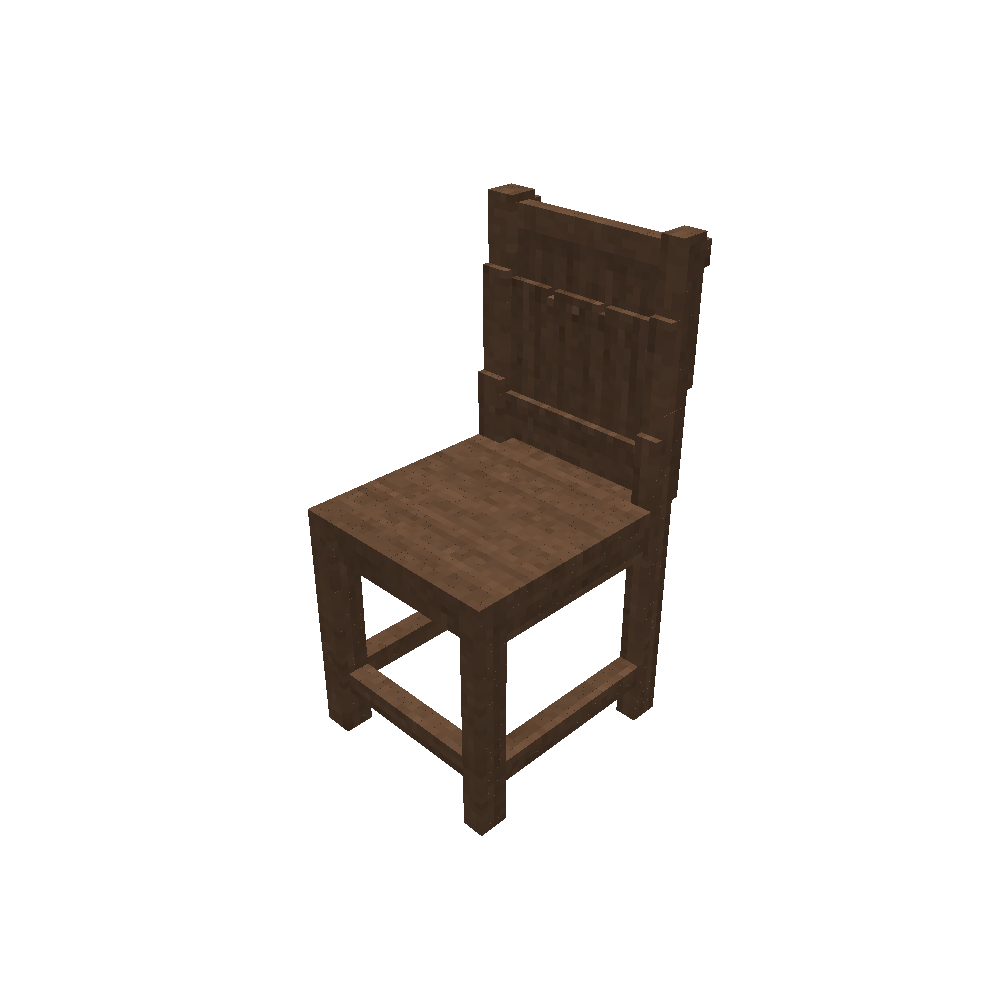}&
\includegraphics[trim=50 110 50 170,clip]{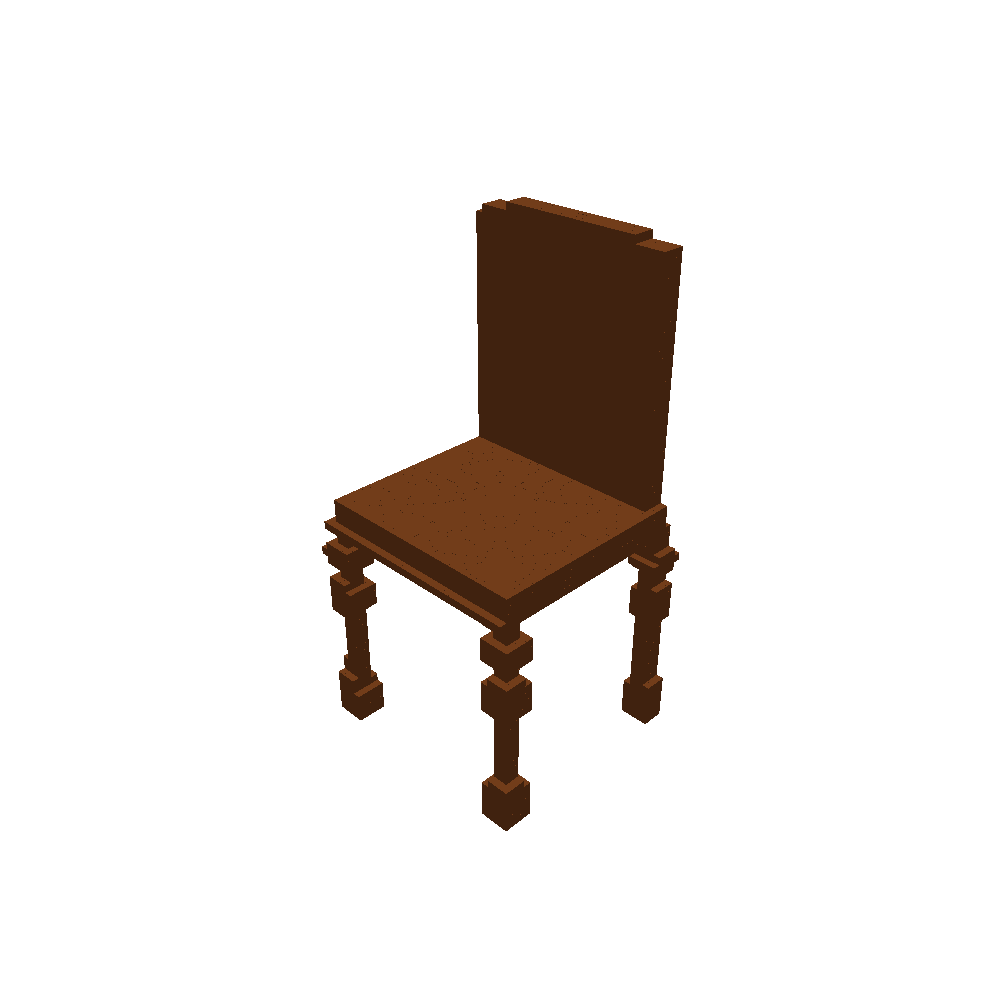}&
\includegraphics[trim=50 110 50 170,clip]{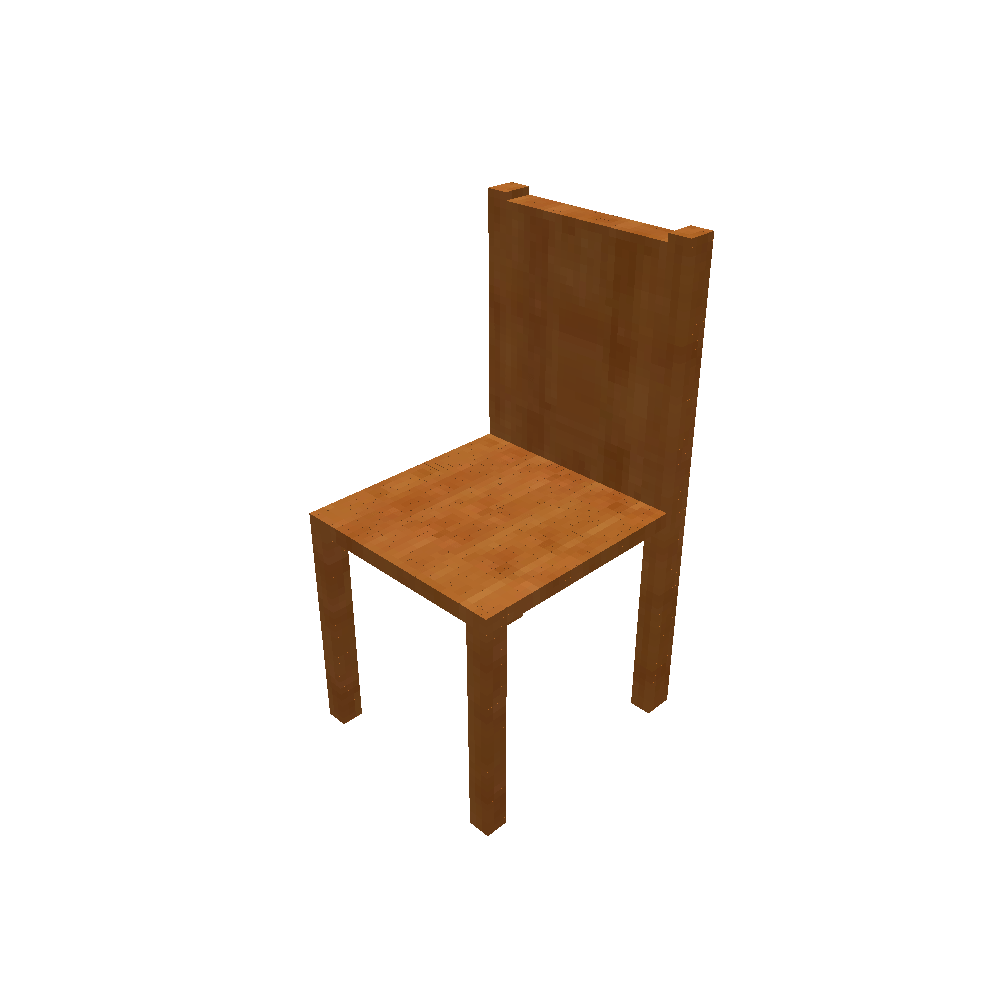}&
\includegraphics[trim=50 110 50 170,clip]{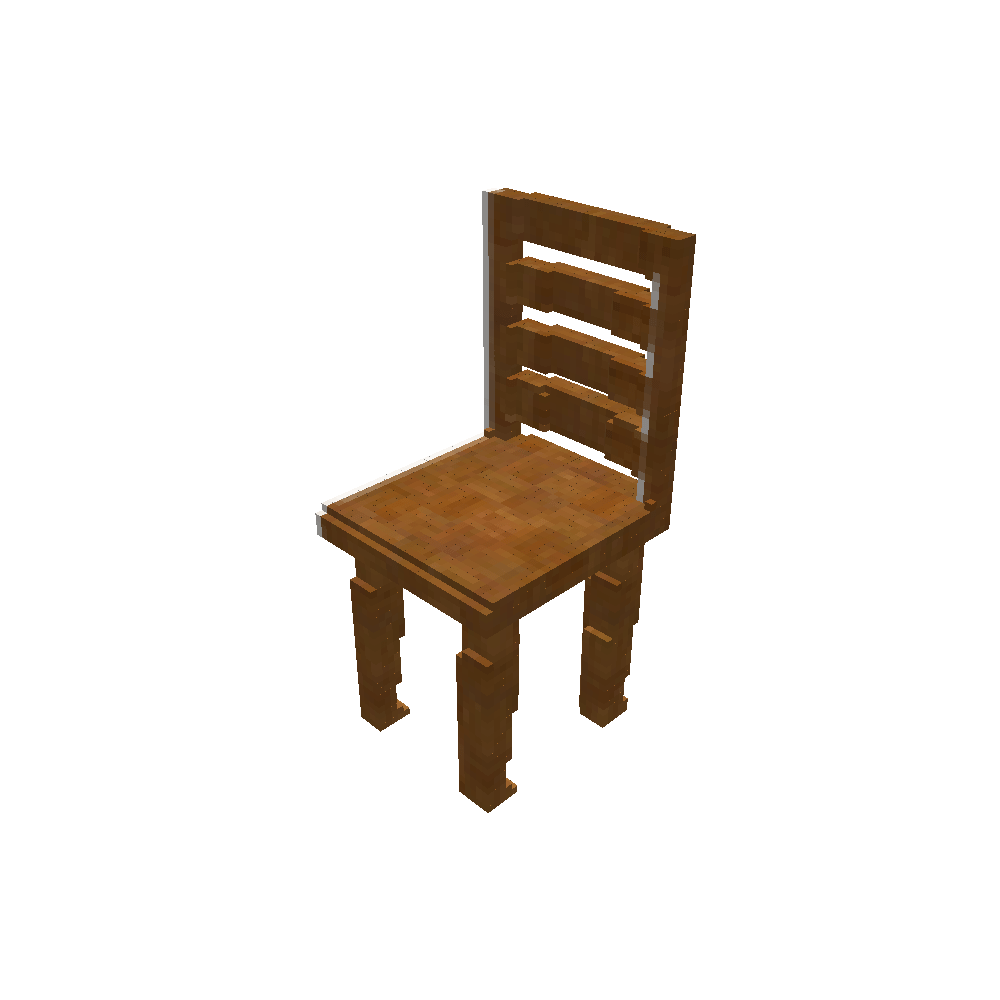}\\
[-0.1cm]
& & {\color{groundtruth}GT} & {\color{partialmatch2}19.36} & {\color{partialmatch2}13.31} & {\color{partialmatch2}18.78} & {\color{partialmatch2}12.38}\\
\bottomrule
\end{tabularx}
\caption{Retrieval results on the \emph{test} set with \trimodiv. For each description, we use \trimodiv to retrieve the top-5 shapes.  We show the $\text{F}1^{0.1}$ score (as a percentage) for each retrieved shape and mark the ground-truth shape (indicated by {\color{groundtruth}green} GT). The expected F1 score for GT is 100. Shapes that are not a perfect match to the description are marked in {\color{partialmatch1}dark orange} (color mismatch), and {\color{partialmatch2}gold} (shape detail mismatch). Results show that our network has good language grounding ability overall. It can retrieve shapes that match \textit{L-shaped} (row 1), \textit{stainless steel frame foots} (row 2), \textit{circular table} (row 3), \textit{no leg} (row 3), \textit{circular base} (row 3), \textit{greenish top}  (row 4), \textit{wooden}  (row 4), \textit{boxy look}  (row 5), \textit{gray}  (row 5), \textit{armless}  (row 6) and \textit{nine-square back} (row 6). Though we assume one ground-truth shape, multiple shapes can match the query description (row 1, 3-5).}
\label{fig:retrieval-visualization}
\vspace{-1em}
\end{figure*}

\subsection{Results}

We present qualitative retrieval examples and quantitative evaluations comparing our method to prior work. We examine the choice of different loss functions and hyperparameters  (see supplement for experiments on image and voxel resolution, and experiments with different backbones).
We train models with different seeds and report mean and standard error across $3$ runs.   

\xhdr{Example retrievals}
\cref{fig:retrieval-visualization} shows successful retrievals of shapes using \trimodiv, our best-performing model (see supplement for more examples).
Our model successfully grounds language describing shape (\textit{L-shaped}, \textit{boxy}), color (\textit{brown}, \textit{greenish}), and texture (\textit{wooden}).
It can also handle negation (\textit{armless}).
Note that many shapes match the description despite not being the ground-truth shape, indicating that there are indeed many matching shapes for a given description.
For example, in row 3 
the retrieved shapes all match the description, but the four of the five shapes would be negatives in our training and retrieval metrics.

\begin{table}
\centering
{
\begin{tabular}{@{}rrrr@{}}
\toprule
  & RR@1 & RR@5 & NDCG@5\\
\midrule
Random (expected) & 0.06 & 0.30 & 0.20 \\
Random (weights)  & 0.08 & 0.32 & 0.20 \\
Text2Shape~\cite{chen2018text2shape}  & 0.40 & 2.37 & 1.35 \\
Y2Seq2Seq~\cite{han2019y2seq2seq} & 2.93 & 9.23 & 6.05 \\
Parts2Words (global) ~\cite{tang2023parts2words} & 8.60 & 24.82 & 16.83 \\
Parts2Words (full)~\cite{tang2023parts2words} & 12.72 & 32.98 & 23.13 \\
\bimodi (our)  & 11.09 & 30.78 & 21.10 \\
\bimodv (our)  & 8.93 & 26.71 & 17.98 \\
\trimodiv (our)  & 12.22 & 32.23 & 22.46 \\

\bottomrule
\end{tabular}
}
\vspace{-8pt}
\caption{Text to shape retrieval comparison against prior work on the \emph{test} set. We report the recall rate (RR@1, RR@5) and NDCG@5 as percentages. We train with a batch size of $128$, $64^3$ voxels, and $6$ multi-view images at a resolution of $128^2$ each.
Our bimodal joint embedding (\bimodi, \bimodv) trained using the NT-XEnt loss outperforms prior work with global matching.
Our trimodal embedding (\trimodiv) further improves performance, and is close to Parts2Words~\cite{tang2023parts2words} which uses part annotations for training.}
\label{tab:retrieval-shapenet}
\end{table}

\xhdr{Comparison with prior work.}
We report the text-to-shape retrieval results in \cref{tab:retrieval-shapenet}.
The full Parts2Words~\cite{tang2023parts2words} model assumes prior part segmentation knowledge to compute attention with the word embeddings and trains using the triplet loss with negative sampling.
In contrast, we do not leverage any part prior knowledge, or attention mechanisms.  For comparison, we include Parts2Words (global) with average pooling, which does not use any part information. 
\cref{tab:retrieval-shapenet} shows that our method performs better on all retrieval metrics, and we can achieve slightly better performance than even the full Parts2Words model.
Note that there are several differences in the prior work compared to our own: the network architectures and specifics of the loss functions, as well as different input representations.
\citet{chen2018text2shape} used $32^3$ colored voxels, while Y2Seq2Seq~\cite{han2019y2seq2seq} used multi-view images, and Parts2Words~\cite{tang2023parts2words} used colored point clouds.

\begin{table}
\centering
\resizebox{\linewidth}{!}
{
\begin{tabular}{@{}lrrrrr@{}}
\toprule
  & RR@1($\uparrow$) & RR@5($\uparrow$) & NDCG@5($\uparrow$) &  MRR($\uparrow$) &  $F1^{0.1}$($\uparrow$) \\
\midrule

\bimodi & 11.61 $\pm$ 0.20 & 30.65 $\pm$ 0.19 & 21.36 $\pm$ 0.23 & 21.46 $\pm$ 0.25 & 16.69 $\pm$ 0.50 \\
\trimodi & 12.19 $\pm$ 0.45 & 32.33 $\pm$ 0.60 & 22.54 $\pm$ 0.54 & 22.62 $\pm$ 0.49 & 17.39 $\pm$ 0.41 \\
\bimodv & 9.59 $\pm$ 0.27 & 27.14 $\pm$ 0.48 & 18.54 $\pm$ 0.13 & 19.03 $\pm$ 0.08 & 14.96 $\pm$ 0.20 \\
\trimodv & 9.83 $\pm$ 0.21 & 27.75 $\pm$ 0.35 & 18.97 $\pm$ 0.21 & 19.32 $\pm$ 0.20 & 15.08 $\pm$ 0.23 \\
\trimodiv & \textbf{12.52 $\pm$ 0.28} & \textbf{32.67 $\pm$ 0.61} & \textbf{22.87 $\pm$ 0.46} & \textbf{22.68 $\pm$ 0.32} & \textbf{17.45 $\pm$ 0.30} \\

\bottomrule
\end{tabular}
}
\vspace{-8pt}
\caption{Comparison of bimodal and trimodal models for text-to-shape retrieval on the \emph{val} set.
Trimodal embeddings (\trimodi,\trimodv) give better performance than bimodal embeddings (\bimodi,\bimodv).
By summing the image and voxel embeddings from the trimodal model (\trimodiv), we further improve retrieval performance.}
\label{tab:retrieval-modality-shapenet}
\vspace{-14pt}
\end{table}

\begin{figure}
\centering
\setkeys{Gin}{width=\linewidth}
\footnotesize
\begin{tabularx}{\linewidth}{@{}lYYYYY@{}}
\toprule
\multicolumn{5}{p{5.5cm}}{\vspace{-1.5em}GT: It looks like one you would use at a picnic. It is wooden and has bench seating.} &
\includegraphics[trim=50 200 50 250,clip]{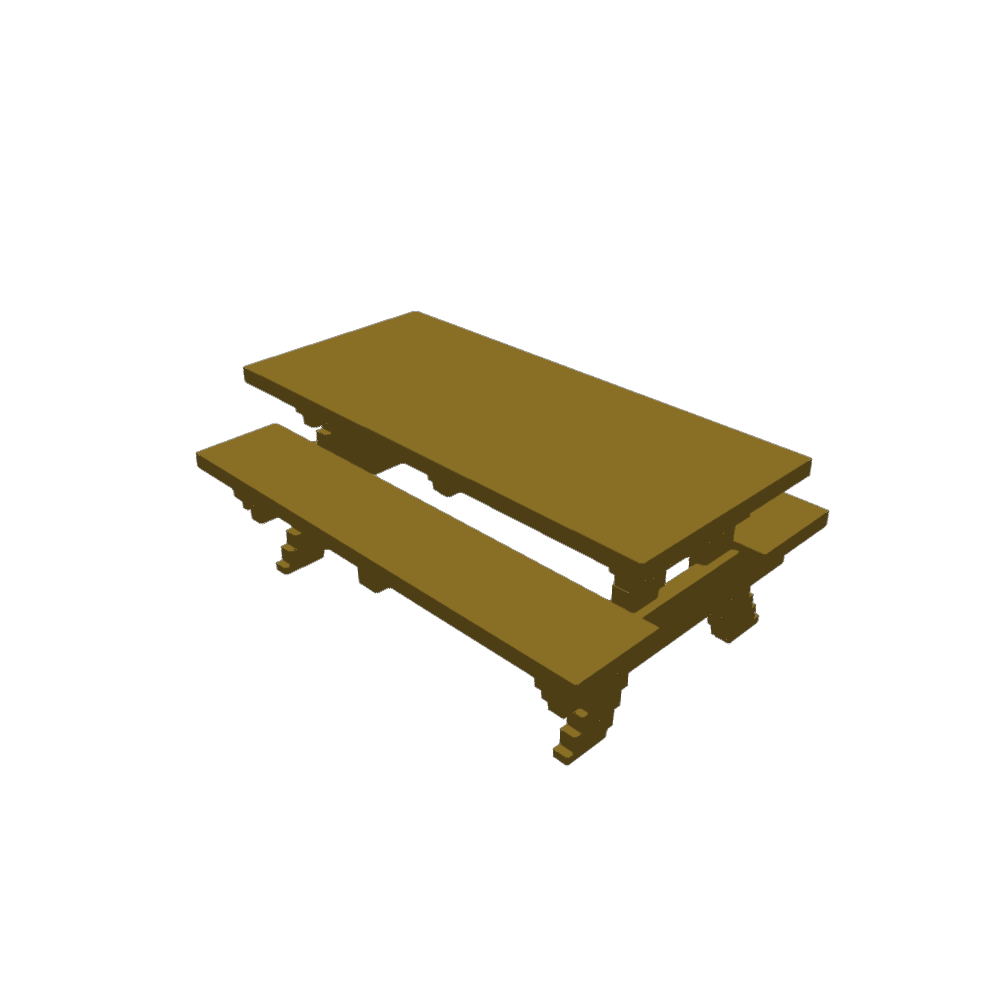} \\
\midrule
\bimodi & 
\includegraphics[trim=50 40 50 150,clip]{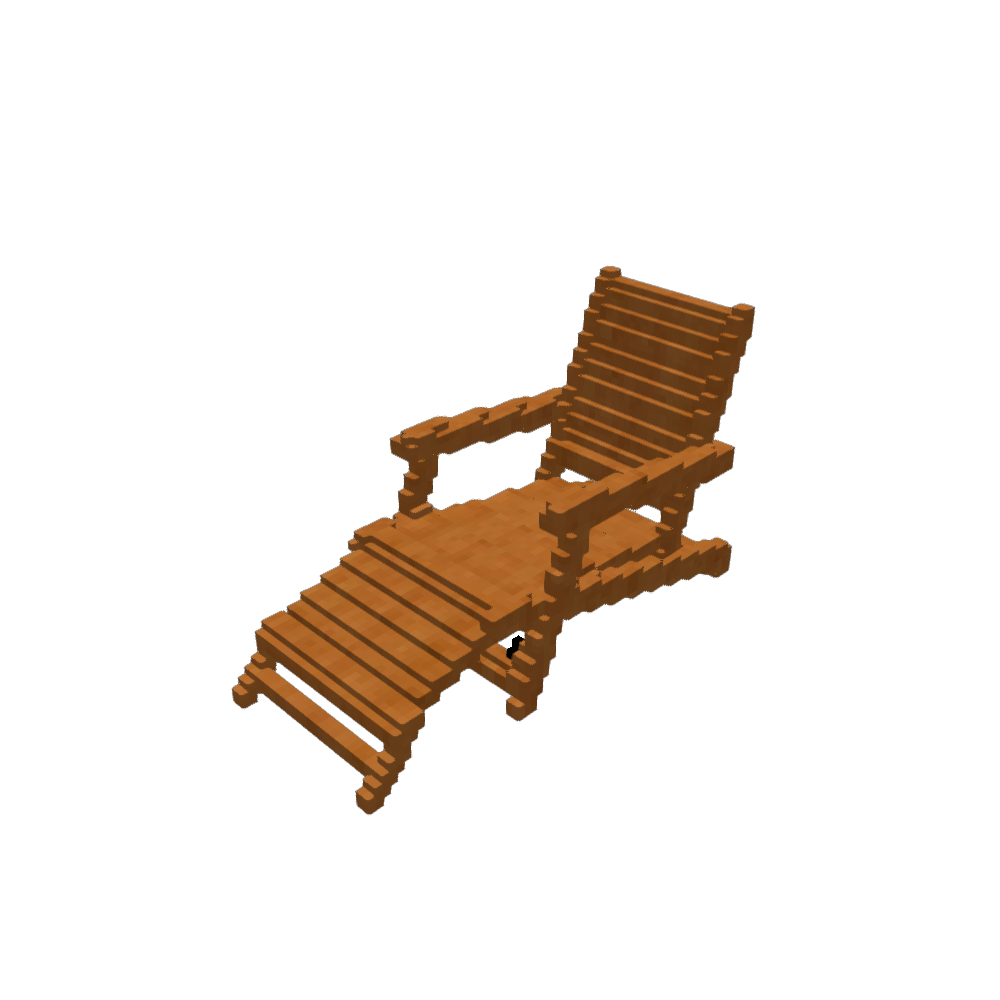} & 
\includegraphics[trim=50 40 50 150,clip]{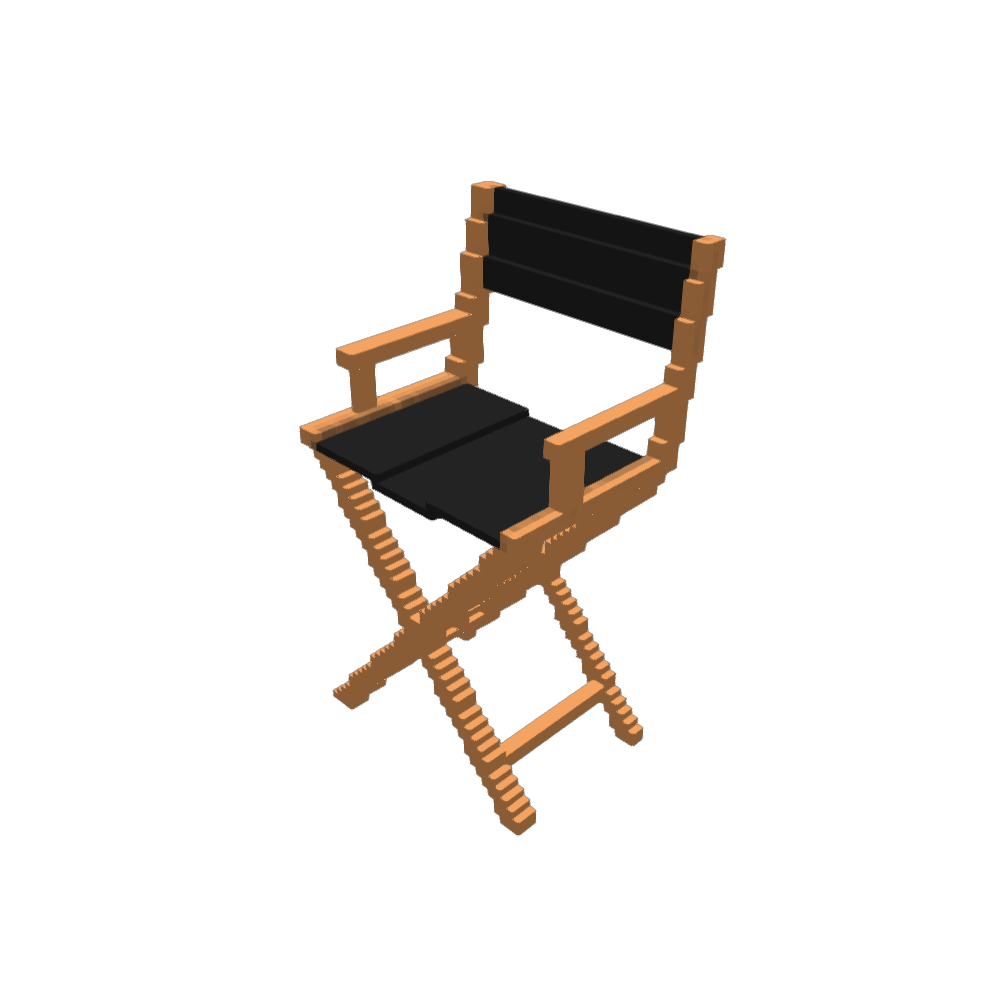} &
\includegraphics[trim=50 40 50 150,clip]{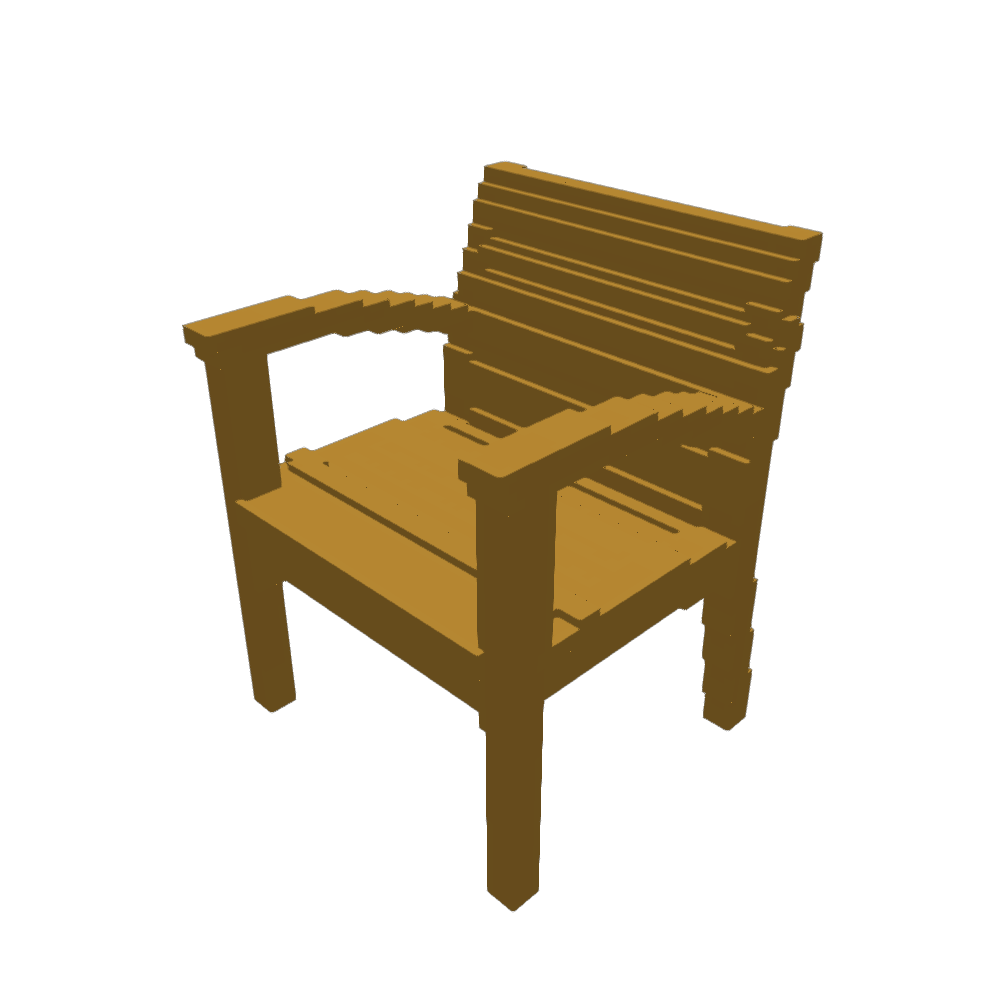} & 
\includegraphics[trim=50 40 50 150,clip]{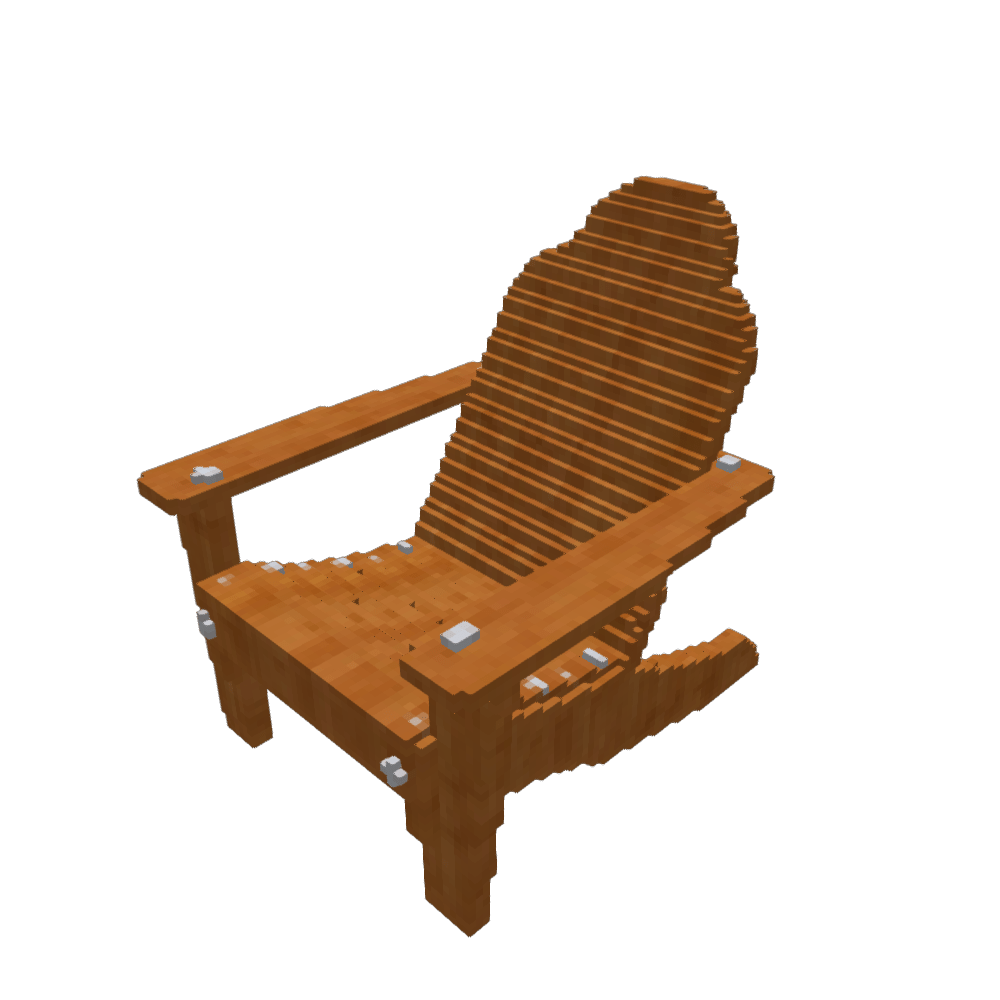} & 
\includegraphics[trim=50 40 50 150,clip]{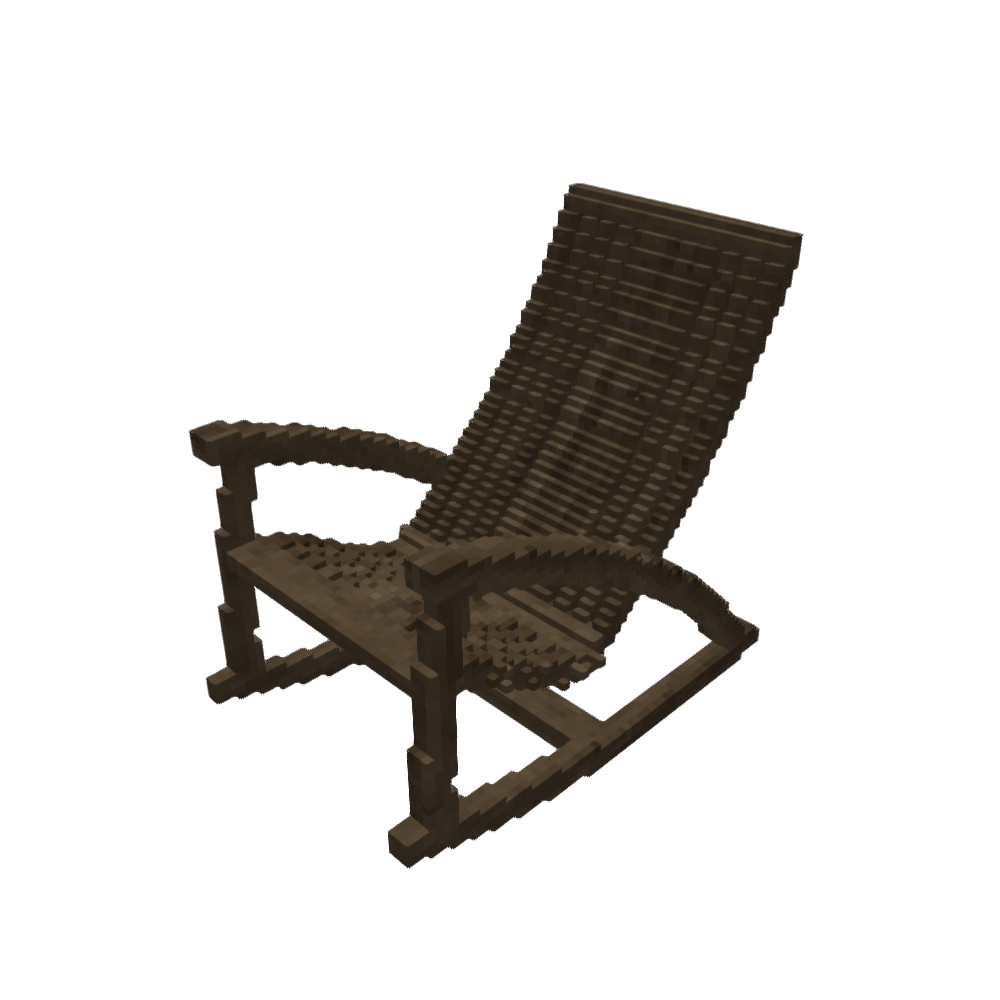}  \\
\bimodv & 
\includegraphics[trim=50 200 50 250,clip]{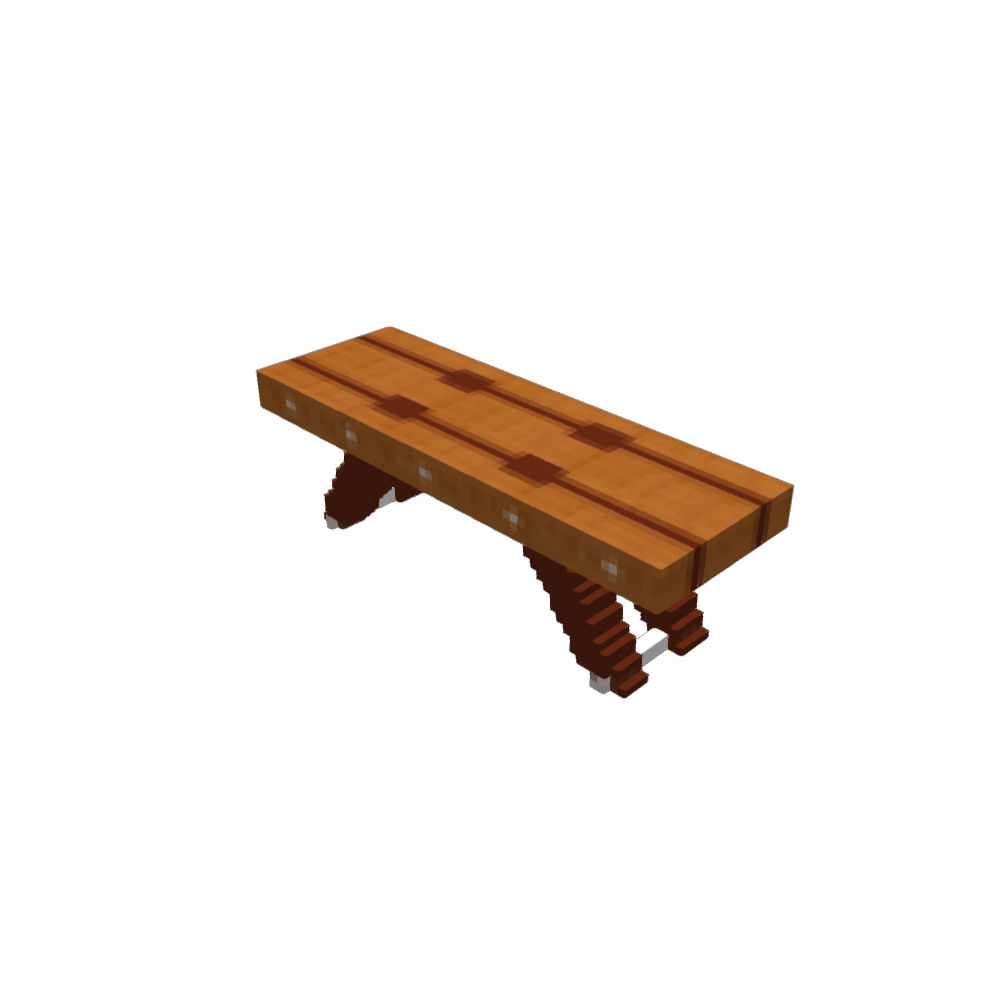} & 
\includegraphics[trim=50 200 50 250,clip]{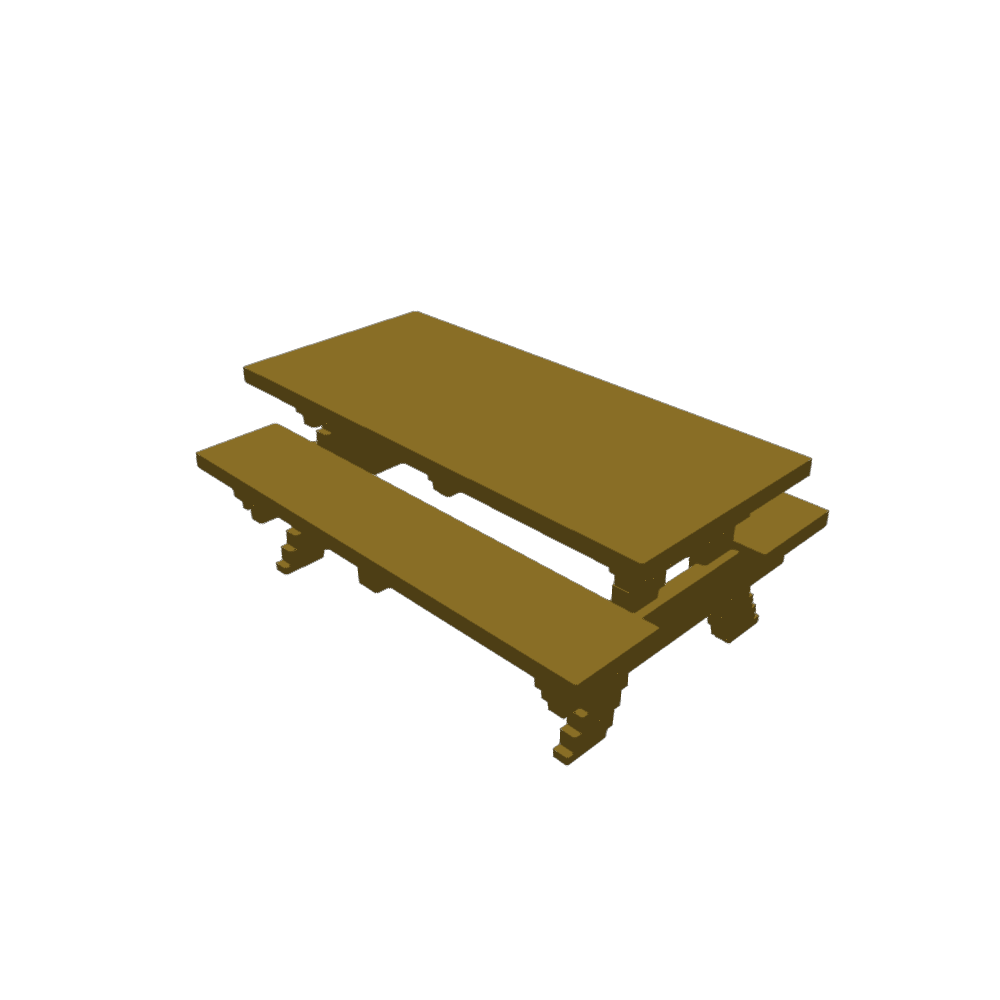} &
\includegraphics[trim=50 200 50 250,clip]{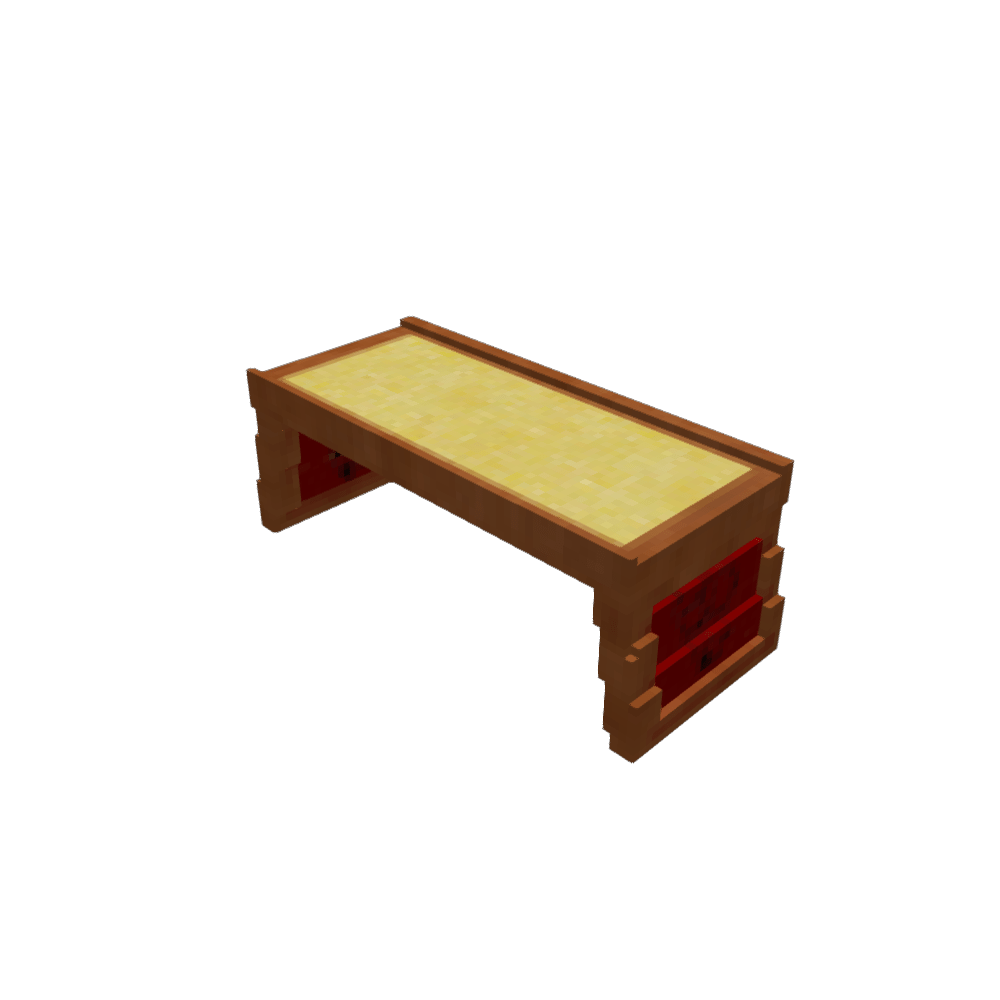} & 
\includegraphics[trim=50 200 50 250,clip]{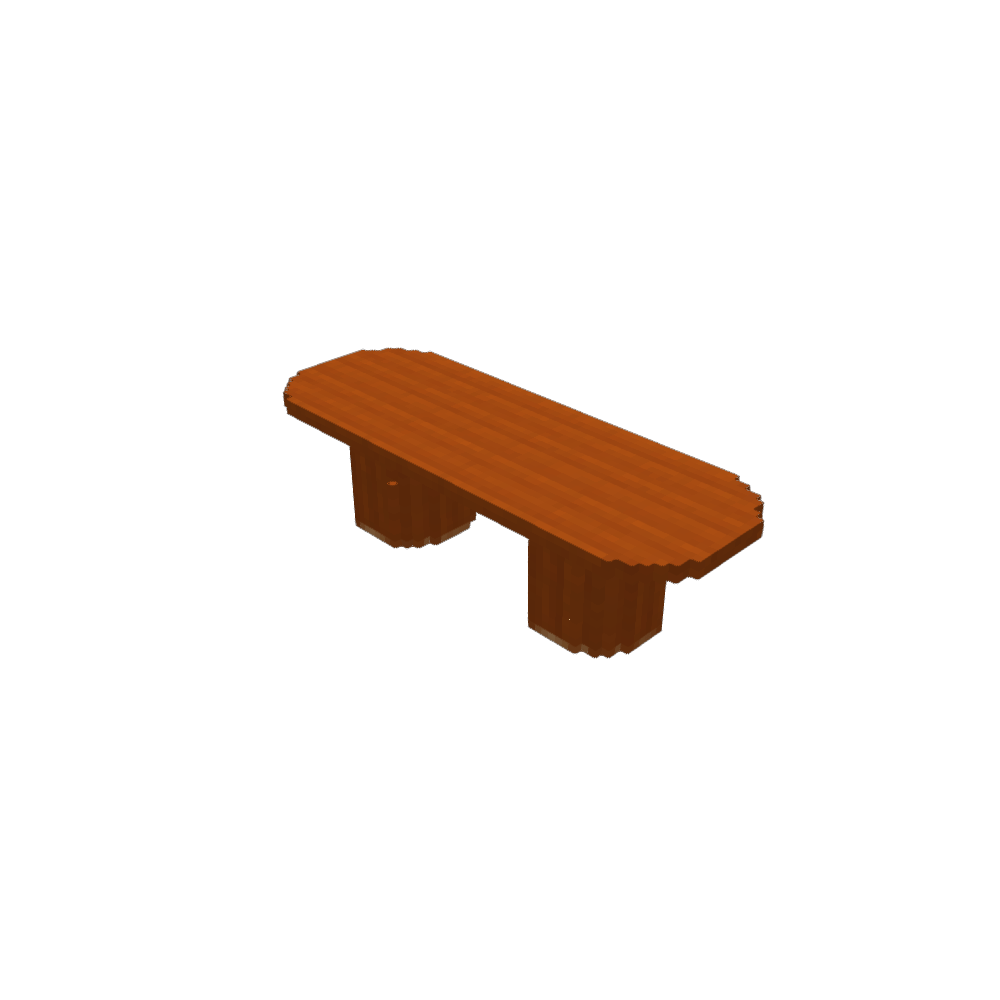} &
\includegraphics[trim=50 200 50 250,clip]{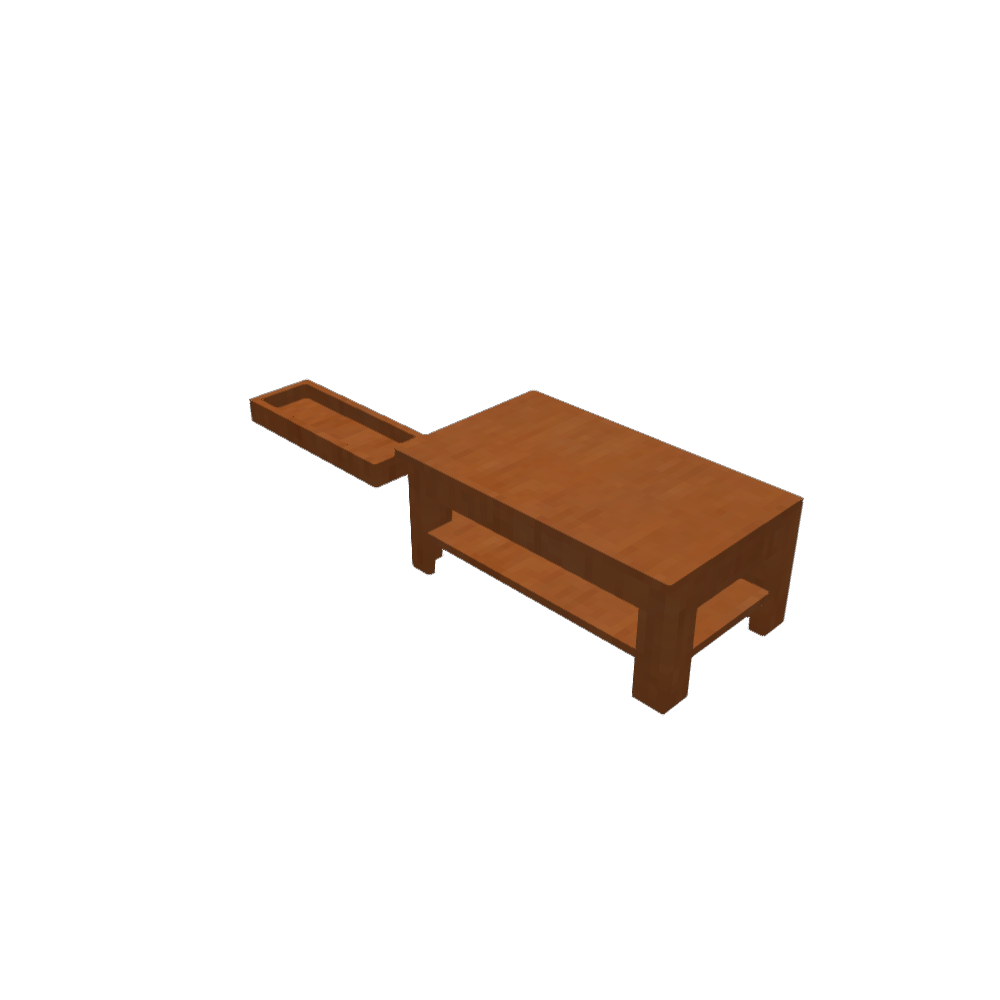} \\
\trimodiv & 
\includegraphics[trim=50 200 50 250,clip]{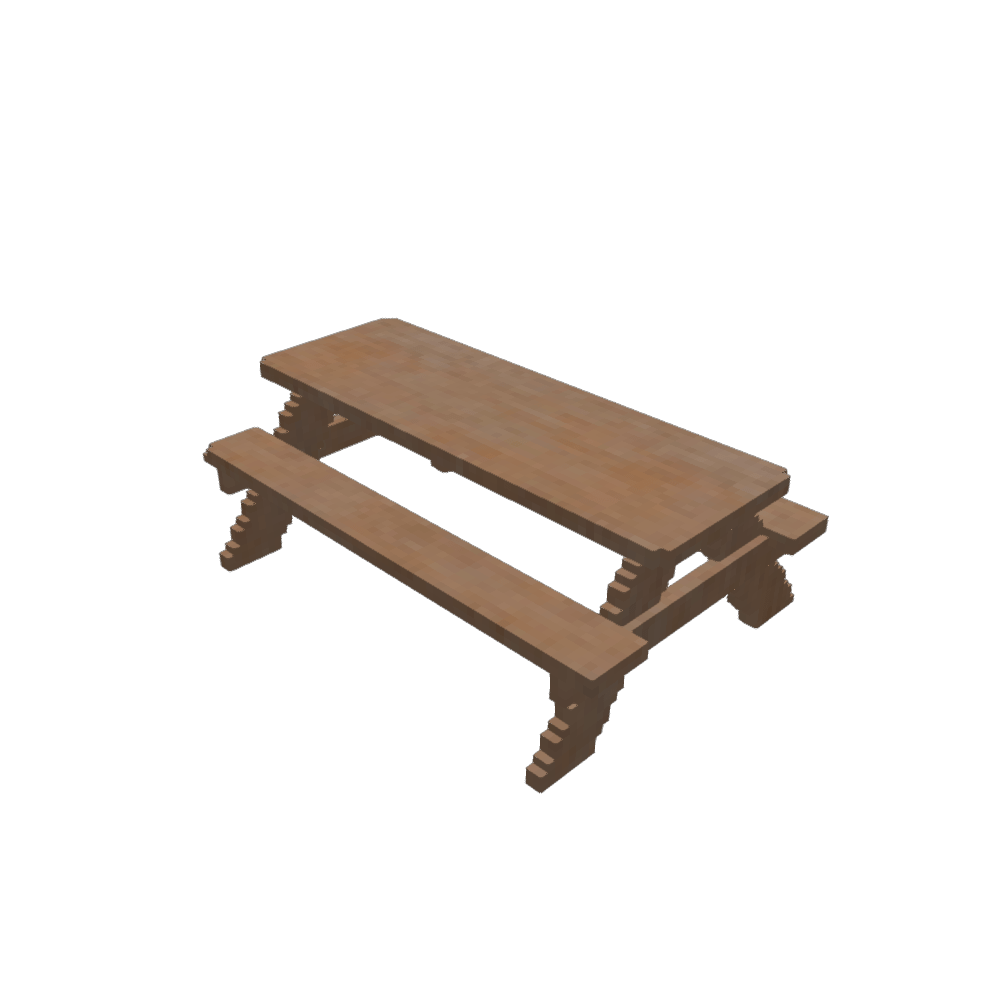} & 
\includegraphics[trim=50 200 50 250,clip]{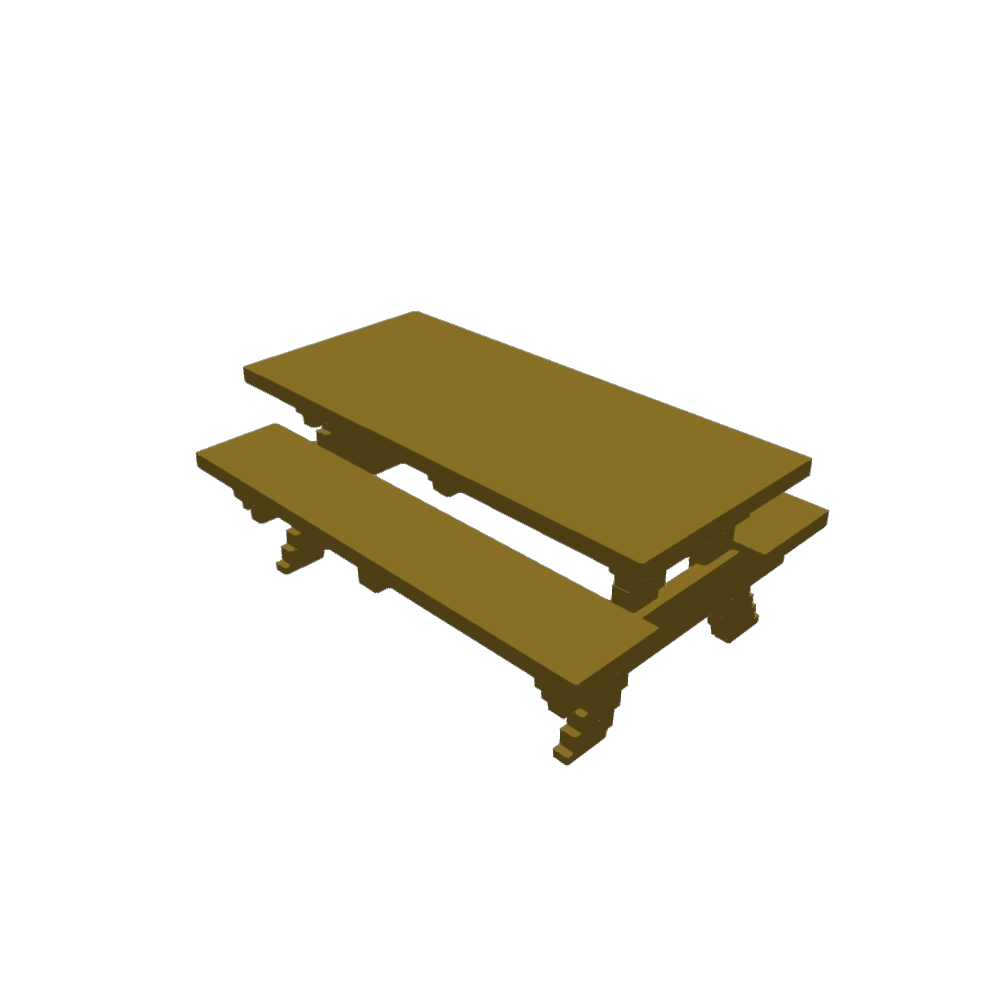} & 
\includegraphics[trim=50 200 50 250,clip]{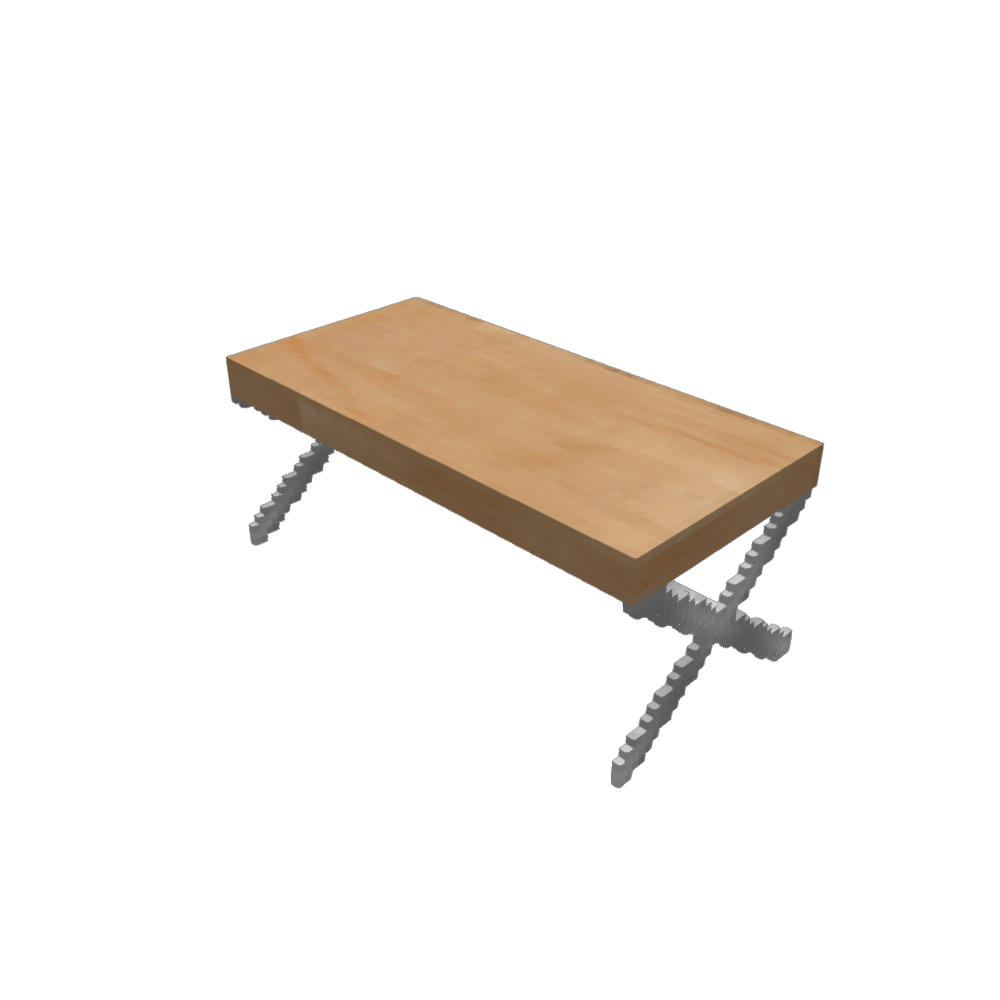} & 
\includegraphics[trim=50 200 50 250,clip]{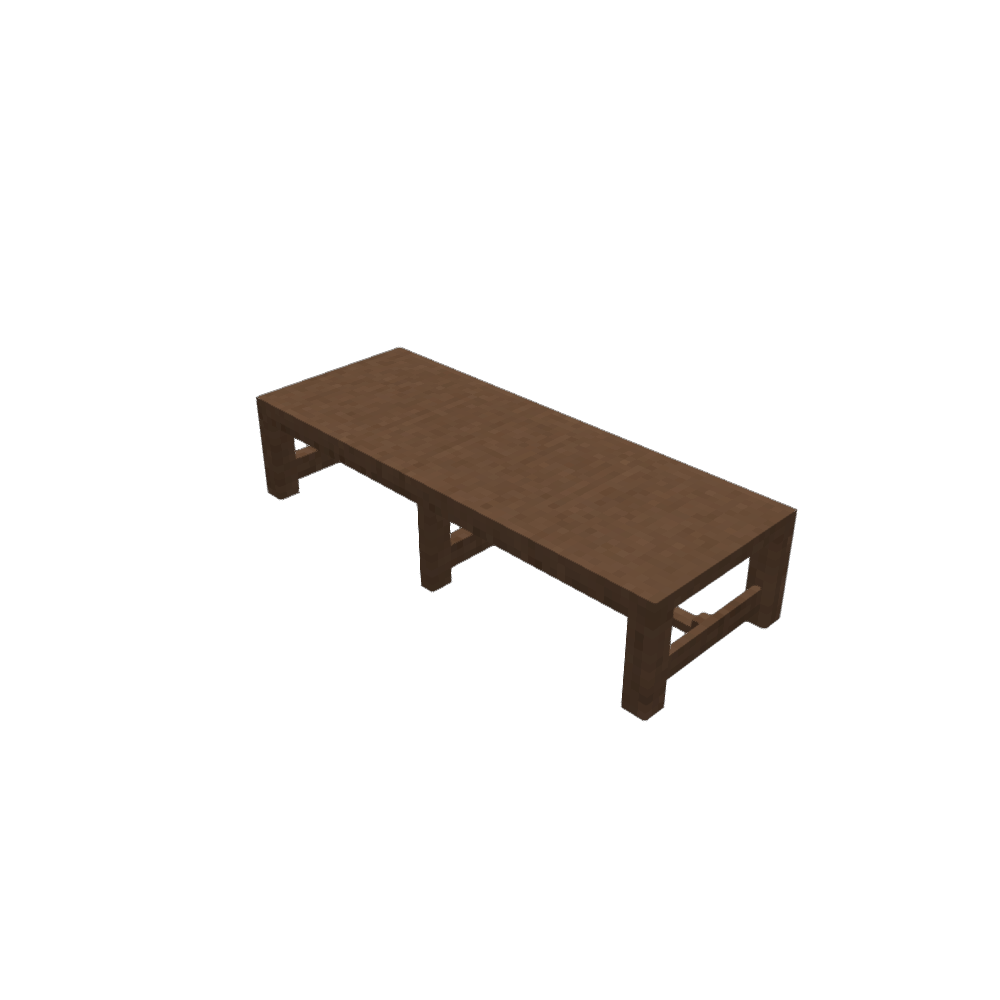} &
\includegraphics[trim=50 200 50 250,clip]{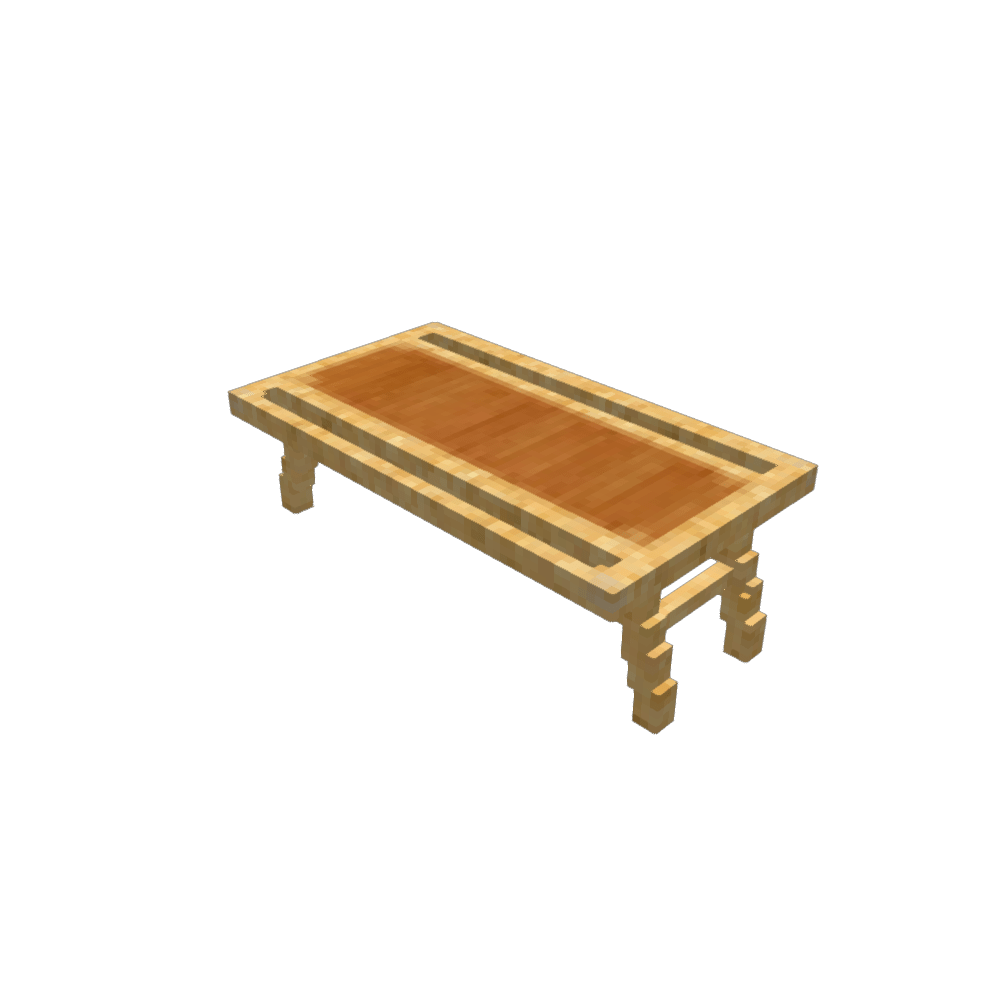} \\
\bottomrule
\end{tabularx}
\vspace{-8pt}
\caption{Top 5 retrieved shapes from the \emph{val} set using \bimodi, \bimodv, and \trimodiv. We see that \bimodi understands abstract concepts such as \textit{picnic} poorly. \trimodiv retrieves the most shapes consistent with the description.
}
\label{fig:retrieval-comparison-visualization-main}
\vspace{-8pt}
\end{figure}

\xhdr{Bimodal vs Trimodal.} %
We compare the trimodal joint embedding with bimodal ones (see \cref{tab:retrieval-modality-shapenet}).
The modalities in the parentheses indicate which representation was used to retrieve the 3D shapes with respect to the text embeddings.
We see that the trimodal embedding improves retrieval performance across all metrics when retrieving by both images and voxels.
We obtain the best result when we sum the image and voxel embeddings, indicating that the information in the voxels is complementary to the multi-view images.
\cref{fig:retrieval-comparison-visualization-main} shows a comparison of retrieved shapes for an example description from the validation set.
The retrieved shapes using \trimodiv conform to the description more closely than \bimodi or \bimodv shapes.
See supplement for more examples.

\begin{table}
\centering
\resizebox{\linewidth}{!}
{
\begin{tabular}{@{}rrrrrrr@{}}
\toprule
& Text & Image & Voxels & RR@1 & RR@5 & NDCG@5\\
\midrule
ZS$^*$ & CLIP & CLIP & - & 5.43 & 16.57 & 11.27\\
\midrule
\bimodi & CLIP & MVCNN & - & 5.79 & 16.90 & 11.49 \\
\bimodi & GRU & CLIP & - & 7.29 & 22.57 & 14.85 \\
\bimodi & CLIP & CLIP & - & 5.76 & 19.13 & 12.41  \\
\trimodiv & CLIP & CLIP & 3D-CNN & 6.72 & 21.95 & 14.52 \\
\midrule
\bimodi & GRU & MVCNN & - & 11.43 & 30.07 & 20.92 \\
\bimodv & GRU & - & 3D-CNN  & 8.98 & 26.76 & 17.99 \\
\trimodiv & GRU & MVCNN & 3D-CNN & \textbf{12.11} & \textbf{32.39} & \textbf{22.42} \\
\bottomrule
\end{tabular}
}
\vspace{-8pt}
\caption{Comparison of text to shape retrieval performance using CLIP-based models on the \emph{val} set. We report the recall rate (RR@1, RR@5) and NDCG@5 as percentages. It can be seen that zero-shot CLIP~\cite{radford2021learning} has relatively good performance considering that it has not been trained on the Text2Shape~\cite{chen2018text2shape} `chairs and tables' dataset.  Training an MLP to project the CLIP embeddings (CLIP-MLP) drastically improves the retrieval performance, but still underperforms our \trimodiv model.}
\vspace{-10pt}
\label{tab:retrieval-clip}
\end{table}

\xhdr{Can pretrained CLIP encoders help?}
We investigate whether using pretrained vision-language encoders can help improve performance.
Specifically, we experiment with CLIP~\cite{radford2021learning}, a popular text-image embedding that was trained using contrastive learning with the NT-Xent loss on a large corpus of 400M image-text pairs.
We took the pretrained CLIP with ViT-L/14~\cite{dosovitskiy2021image} backbone, and aggregated the embeddings from 6 multi-view images.

We compared the performance of using CLIP in a zero-shot (ZS) manner (without training any weights) and using the CLIP image and text encoders in our models.
For zero-shot retrieval, we use the average of the multi-view image embeddings as our overall image embedding, and match that against the text embedding by taking the dot product.
To incorporate CLIP into our models, we project frozen CLIP embeddings using a two-layer MLP (see supplement).
We present results for zero-shot CLIP and variations of using the text or image CLIP encoders in \Cref{tab:retrieval-clip}.
We find that zero-shot CLIP does not perform well, likely due to the domain gap between the rendered images and the CLIP training data.  Nevertheless, it can beat the baseline method from the original Text2Shape~\cite{chen2018text2shape} without being trained on the dataset.
Incorporating CLIP into our models and training the MLP results in higher performance, showing the value of training a task-specific MLP on the Text2Shape~\cite{chen2018text2shape} `chairs and tables' data.
We find that our models, which are trained from scratch, are able to outperform the CLIP variants.
We also experiment on the ShapeNet c13 data (see supplement), where the CLIP-based models perform much better.
We believe that is likely due to less training data for the other categories.

\begin{table}
\centering
\resizebox{\linewidth}{!}
{%
\begin{tabular}{@{}l cccc @{}}
\toprule
 & RR@1($\uparrow$) & RR@5($\uparrow$) & NDCG@5($\uparrow$) & MRR($\uparrow$)\\
\midrule
\bimodi & 6.44 $\pm$ 0.36 & 21.6 $\pm$ 0.61 & 14.08 $\pm$ 0.51 & 14.99 $\pm$ 0.44 \\
\bimodv & 6.19 $\pm$ 0.14 & 20.85 $\pm$ 1.02 & 13.58 $\pm$ 0.46 & 14.51 $\pm$ 0.30 \\
\trimodiv & \textbf{8.12 $\pm$ 0.20} & \textbf{26.39 $\pm$ 0.58} & \textbf{17.96 $\pm$ 0.55} & \textbf{18.55 $\pm$ 0.42} \\
\bottomrule
\end{tabular}
}
\vspace{-8pt}
\caption{Text-to-shape retrieval on the \emph{val} set using triplet loss with semi-hard negative mining. Performance is lower than NT-Xent (\cref{tab:retrieval-modality-shapenet}). }
\label{tab:retrieval-triplet}
\vspace{-14pt}
\end{table}

\xhdr{NT-Xent vs triplet loss.}
To validate the choice of NT-Xent as our loss function, we compare the performance of our model using a hinge-based triplet loss~\cite{schroff2015facenet} instead of NT-Xent.  We use semi-hard negative mining with a margin of $0.025$.
Semi-hard negatives have been shown to improve performance for contrastive losses~\cite{chen2020simple}.
Specifically \citet{tang2023parts2words} showed it worked better than either triplet-loss by itself or hard negatives for retrieval with the Text2Shape dataset.
\cref{tab:retrieval-triplet} shows that the retrieval performance with triplet loss is significantly lower than with NT-Xent.
Overall, our findings are consistent with prior work~\cite{chen2018learning}.
Note that our model outperforms Y2Seq2Seq~\cite{han2019y2seq2seq} even with just triplet loss.
We find that with NT-Xent loss, our bimodal models surpass the performance of Parts2Words~\cite{tang2023parts2words}.

\begin{table}
\centering
\resizebox{\linewidth}{!}
{
\begin{tabular}{@{}lrrrrr@{}}
\toprule
& \# of images & RR@1($\uparrow$) & RR@5($\uparrow$) & NDCG@5($\uparrow$) & MRR($\uparrow$) \\
\midrule

\textmr{\bimodi}{4} & 1  & 9.15 $\pm$ 0.11 & 26.34 $\pm$ 0.32 & 17.84 $\pm$ 0.20 & 18.36 $\pm$ 0.19 \\
& 3  & 11.19 $\pm$ 0.19 & 30.22 $\pm$ 0.25 & 20.97 $\pm$ 0.05 & 21.23 $\pm$ 0.12 \\
& 6  & \textbf{11.61 $\pm$ 0.20} & 30.65 $\pm$ 0.19 & 21.36 $\pm$ 0.23 & 21.46 $\pm$ 0.25 \\
& 12 & 11.23 $\pm$ 0.20 & \textbf{31.13 $\pm$ 0.10} & \textbf{21.43 $\pm$ 0.09} & \textbf{21.50 $\pm$ 0.15} \\

\bottomrule
\end{tabular}
}
\vspace{-8pt}
\caption{Comparison of number of images on shape retrieval for \bimodi on the \emph{val} set.  We find that having multiple views is important for improved performance, but increasing the number of images beyond 6 causes a slight decrease in RR@1.  We believe that 6 views is sufficient to capture the necessary information, and increasing it further increases the number of parameters and requires more compute.
}
\label{tab:retrieval-num-images}
\end{table}

\begin{table}
\centering
\resizebox{\linewidth}{!}
{
\begin{tabular}{@{}lrrrrr@{}}
\toprule
& batch size & RR@1($\uparrow$) & RR@5($\uparrow$) & NDCG@5($\uparrow$) & MRR($\uparrow$) \\
\midrule
\textmr{\trimodiv}{3} & 32  & 10.62 $\pm$ 0.27 & 30.19 $\pm$ 0.61 & 20.60 $\pm$ 0.16 & 20.86 $\pm$ 0.11  \\
& 64  & 11.48 $\pm$ 0.34 & 31.40 $\pm$ 0.55 & 21.67 $\pm$ 0.37 & 21.80 $\pm$ 0.32 \\
&  128  & \textbf{12.52 $\pm$ 0.28} & \textbf{32.67 $\pm$ 0.61} & \textbf{22.87 $\pm$ 0.46} & \textbf{22.68 $\pm$ 0.32} \\
& 256 & 12.43 $\pm$ 0.35 & 32.25 $\pm$ 0.58 & 22.53 $\pm$ 0.49 & 22.65 $\pm$ 0.40 \\
\bottomrule
\end{tabular}
}
\vspace{-8pt}
\caption{
Comparison of batch-size on shape retrieval for \trimodiv on the \emph{val} set. We find that increasing the batch size increases the performance.  However, the performance decreased for the largest batch size we tried ($256$).  This could be due to overfitting on the limited amount of negatives, or the presence of more noisy negatives in the large batch.}
\label{tab:retrieval-bs}
\vspace{-8pt}
\end{table}

\subsection{Hyper-parameter analysis}
We compare the performance of bimodal models on the validation set with different numbers of input images and batch sizes.  
We use the bimodal models as they are faster to train and require less memory than the trimodal model.

\xhdr{Do we need multi-view images?}
For \bimodi, we experiment with number of images ranging from $1$ to $12$ and find that performance increases as we increase the number of images to $6$, after which there are diminishing returns and even a small drop in performance (see \cref{tab:retrieval-num-images}).
The results indicate that multi-view images provide a benefit over a single view.

\xhdr{Does larger batch size always help?}
We also compare batch sizes of $32,64,128$ for
\trimodiv
and find that performance increases with increasing batch size from $32$ to $128$ (see \cref{tab:retrieval-bs}).
This is consistent with findings from prior work on contrastive learning~\cite{chen2020simple,oord2018representation}.
However, the performance drops when the batch size increases to $256$ for \bimodi.
For \bimodv, increasing the batch size to $256$ makes little difference.
We hypothesize this is due to more false negatives in the batch since the text description may apply to multiple shapes.
Another reason may be that since our dataset size is small compared to image datasets used in prior work~\cite{radford2021learning,chen2020simple}, having a big batch size might overfit our model.
We also note that variance is quite high between runs, which we again attribute to false negatives in the batch and randomness introduced when sampling batches.
However, more investigation is warranted.

\xhdr{Impact of other parameters.}
We also conduct additional experiments (see supplement) examining the effect of different resolutions for image and voxels (higher resolution is better), sparse convolutions (similar performance with less memory, but more time to train), image backbone (smaller backbone works better), zero-shot performance of CLIP (works but not as good as model trained on the data) as well as CLIP embeddings projected using a trained MLP (similar performance as \bimodi on `chairs and tables' but slightly better for ShapeNet c13).

\begin{figure*}
\centering
\setkeys{Gin}{width=\linewidth}
\begin{tabularx}{\textwidth}{@{}p{4cm}Y|YYYYY@{}}
\toprule
 & GT & top1 & top2 & top3 & top4 & top5 \\
\midrule

\multirow{2}{4cm}{round surface with interconnected leg}  & 
\includegraphics[trim=50 120 50 150,clip]{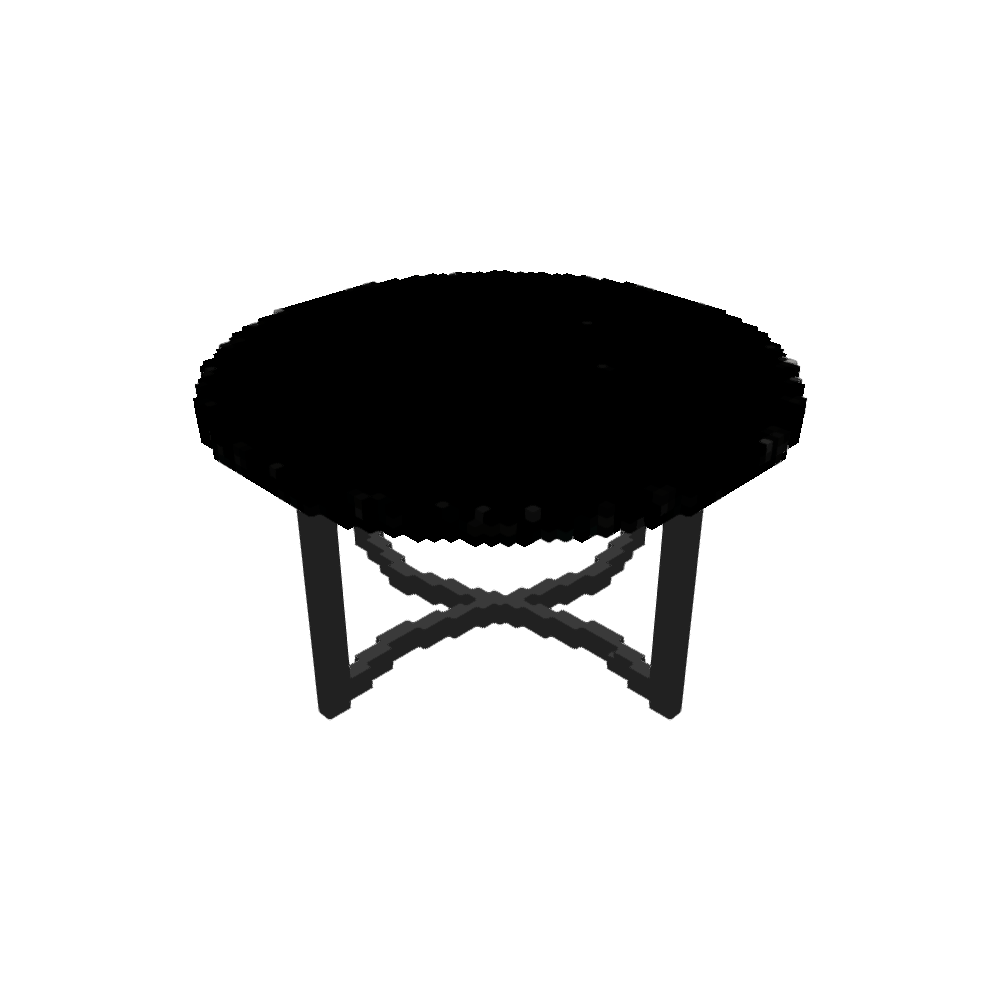} &
\includegraphics[trim=50 120 50 150,clip]{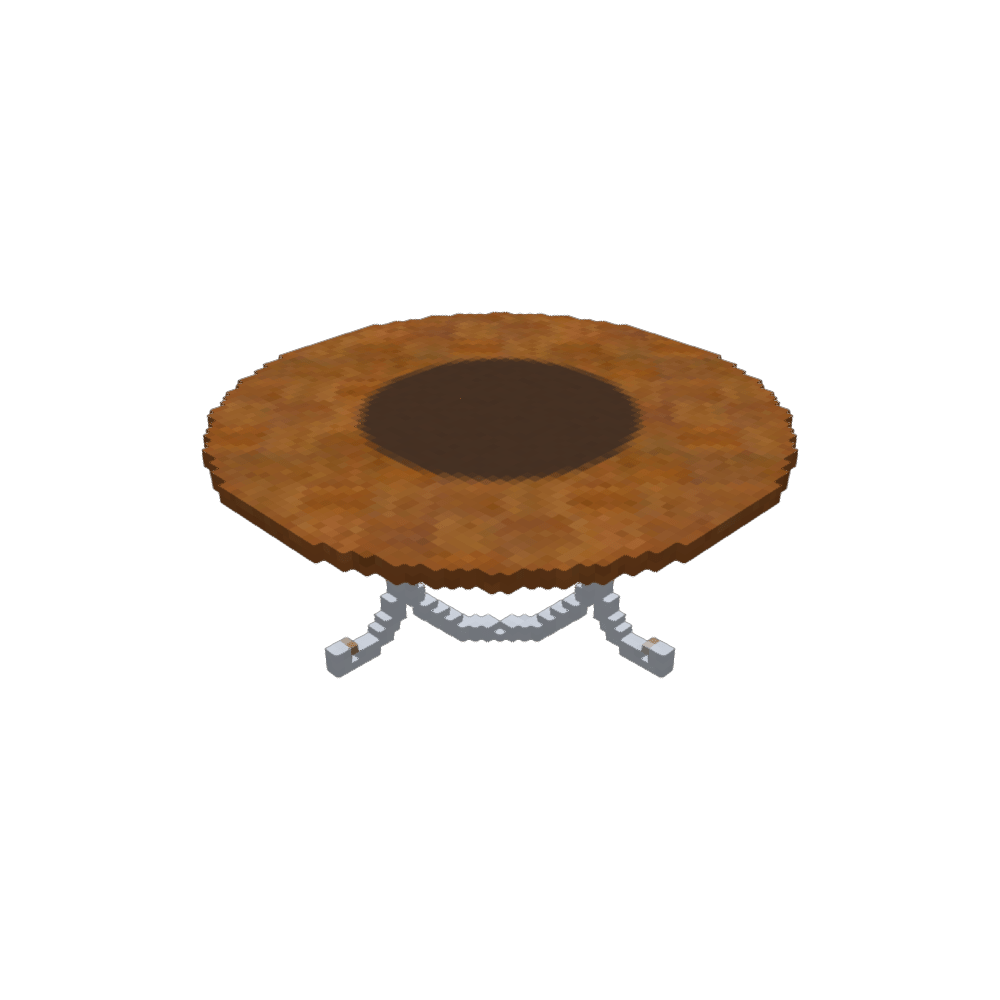} &
\includegraphics[trim=50 120 50 150,clip]{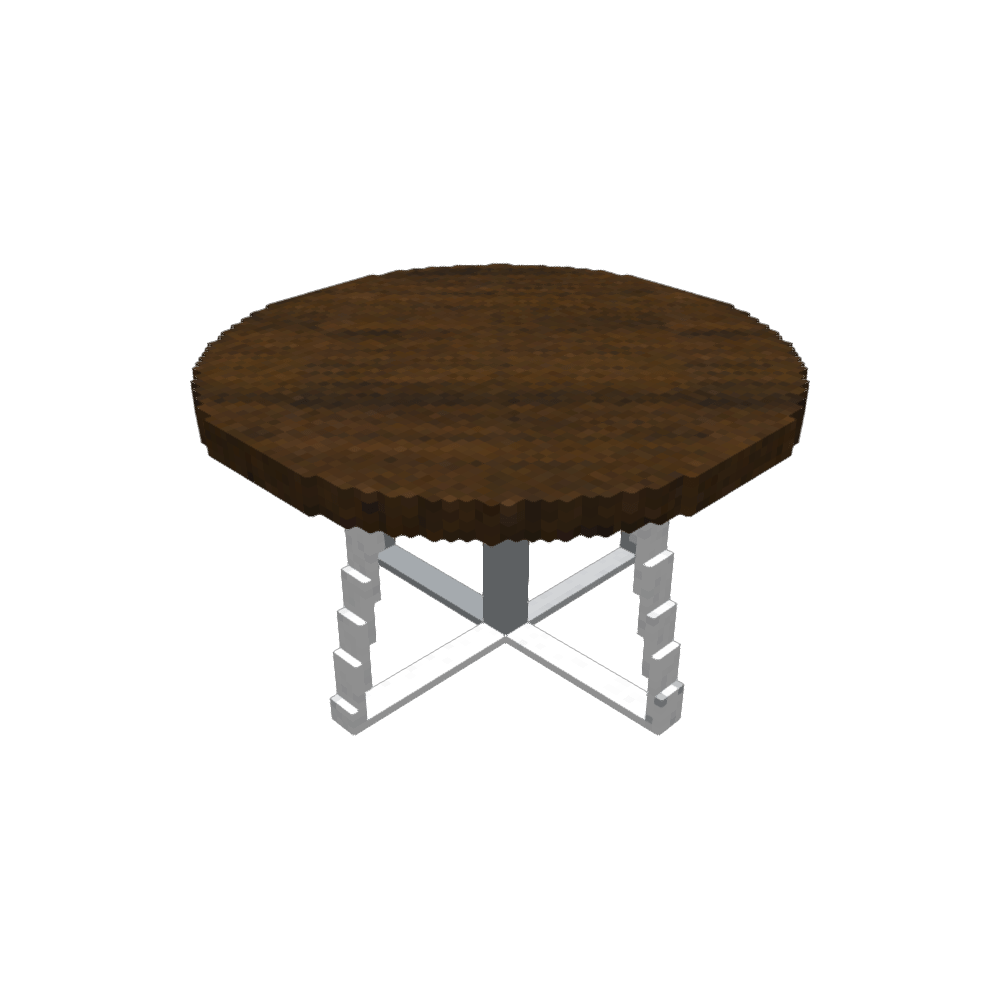} &
\includegraphics[trim=50 120 50 150,clip]{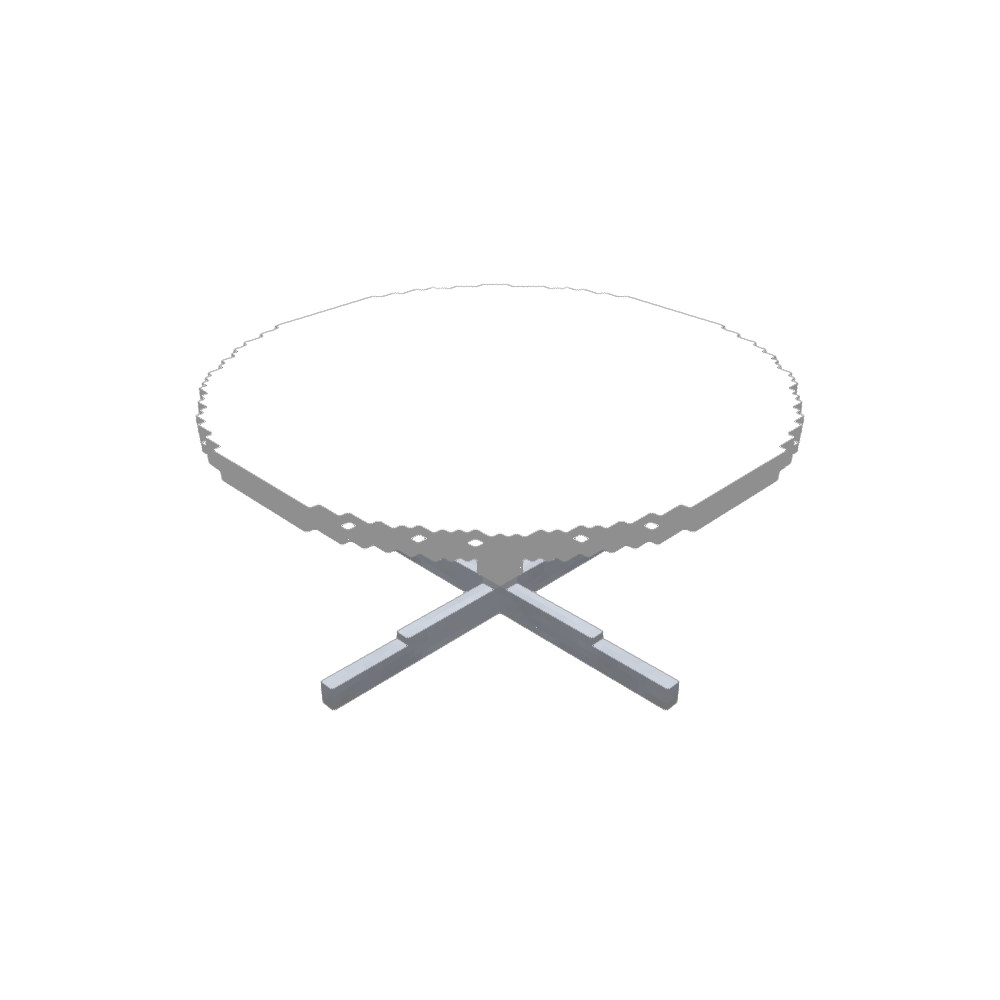} &
\includegraphics[trim=50 120 50 150,clip]{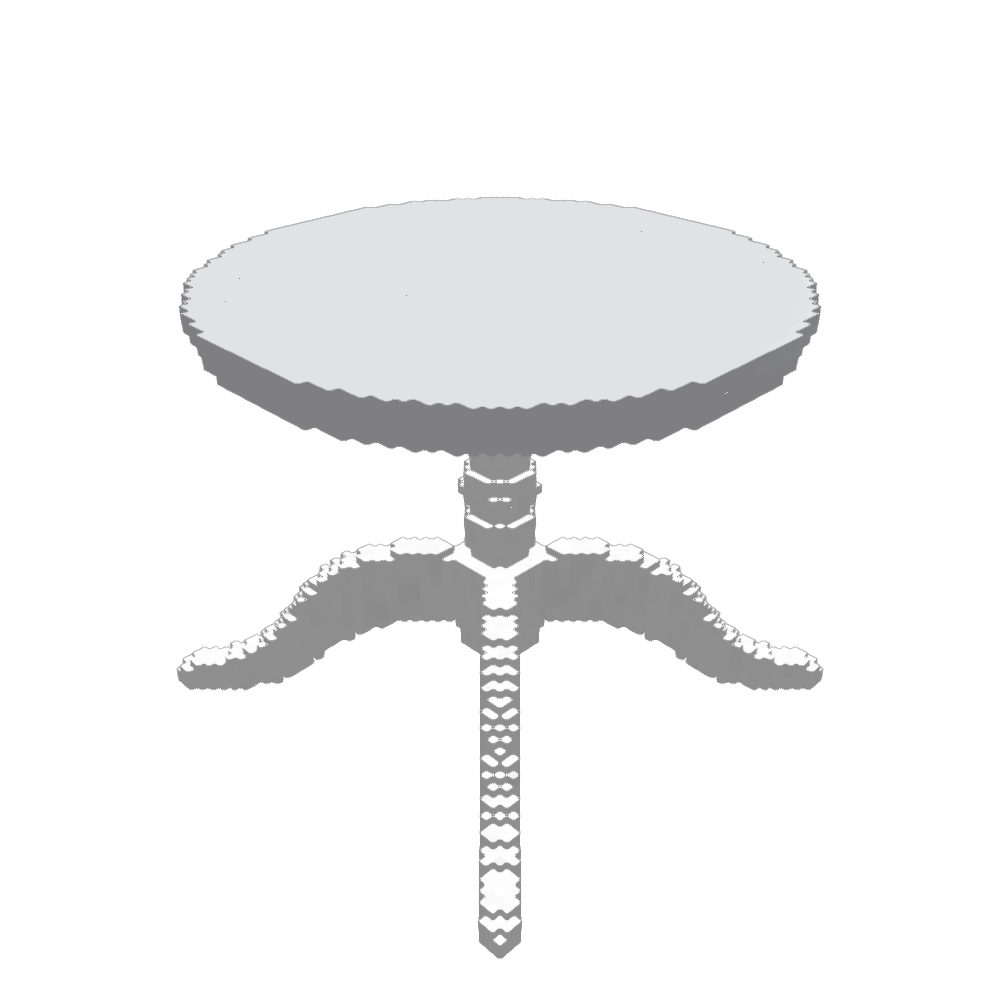} &
\includegraphics[trim=50 120 50 150,clip]{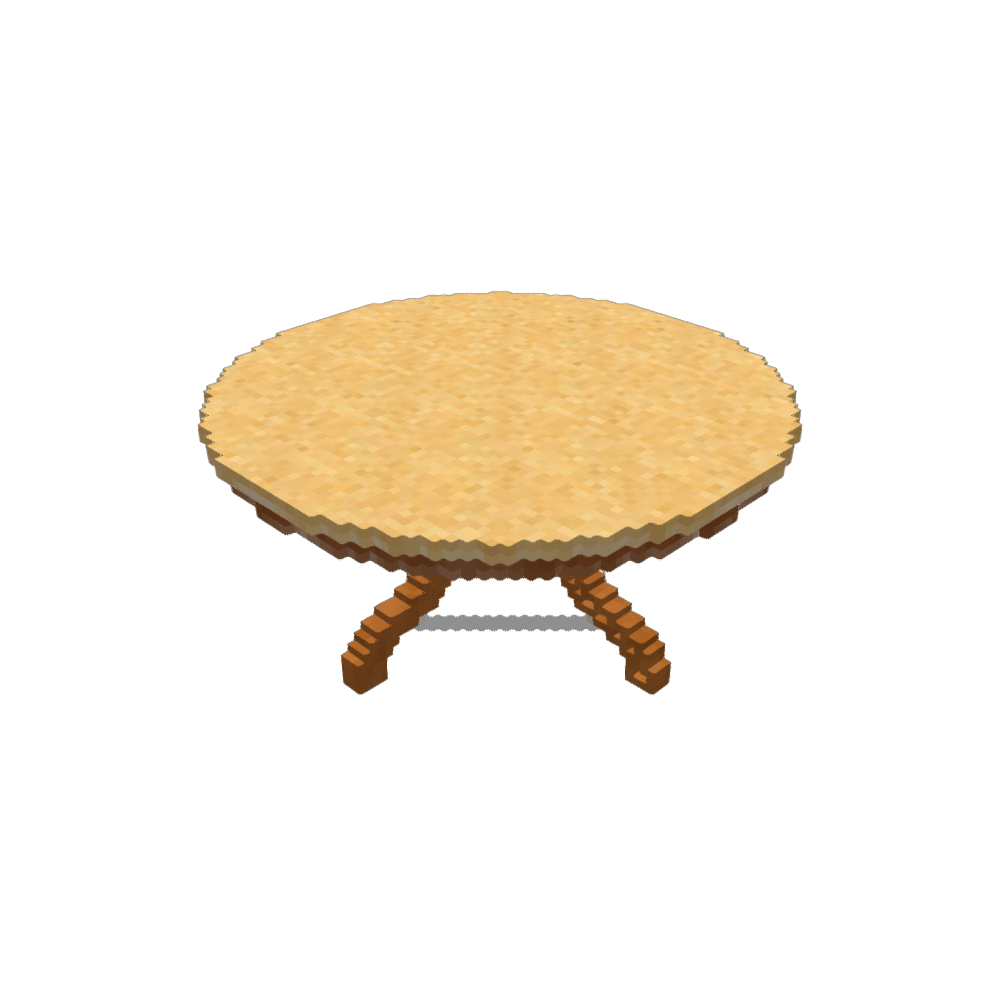} \\
	 &  & 0.62 & 25.19 & 5.06 & 0.73 & 22.28\\
\midrule
 
\multirow{2}{4cm}[1.25em]{taupe one seater sofa . it has a light brown wooden frame with four leg support . it has two \incorrect{wide} arm rest}  & 
\includegraphics[trim=50 120 50 150,clip]{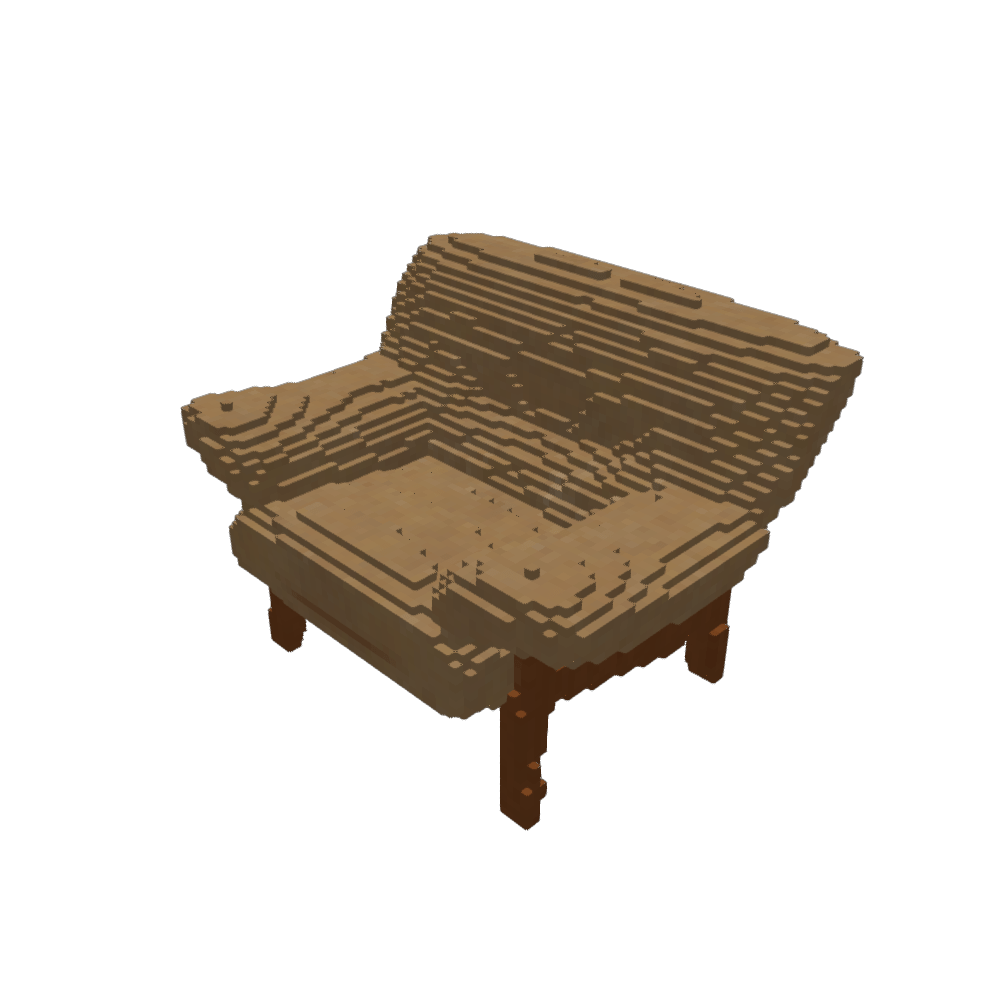} &
\includegraphics[trim=50 120 50 150,clip]{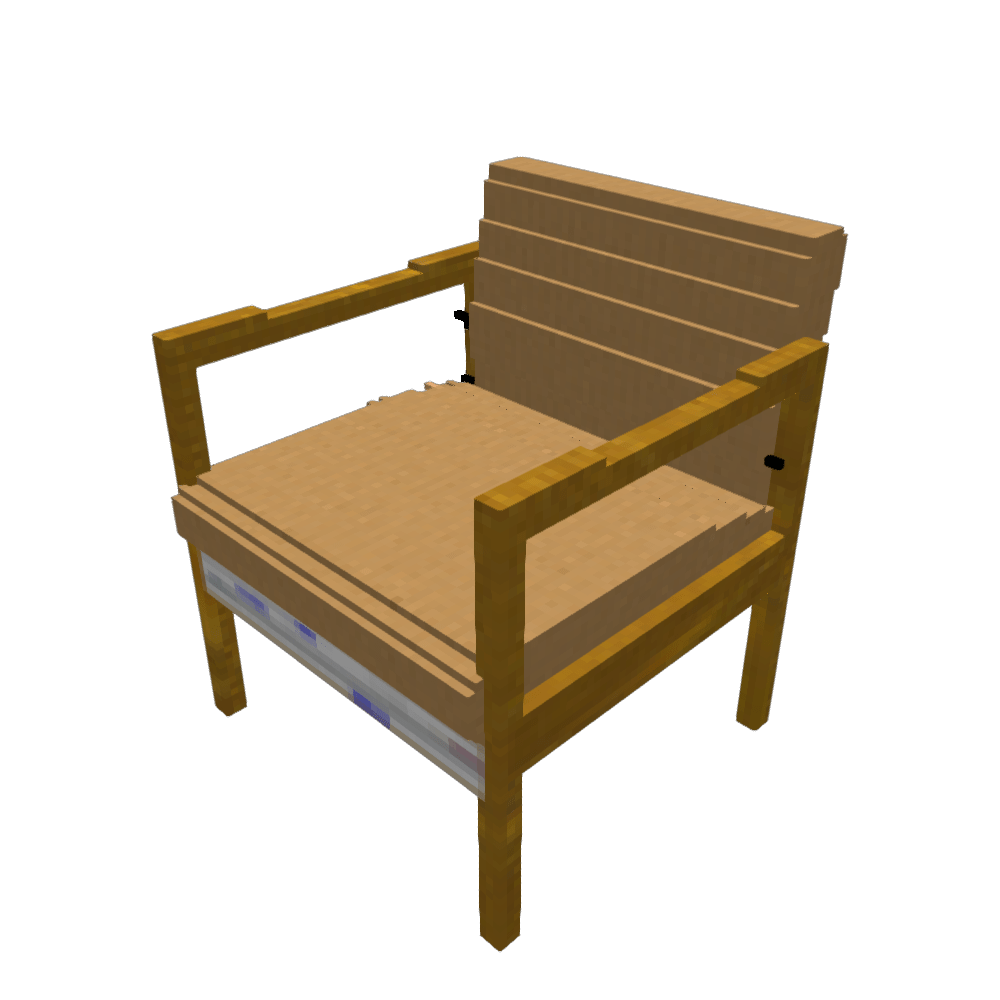} &
\includegraphics[trim=50 120 50 150,clip]{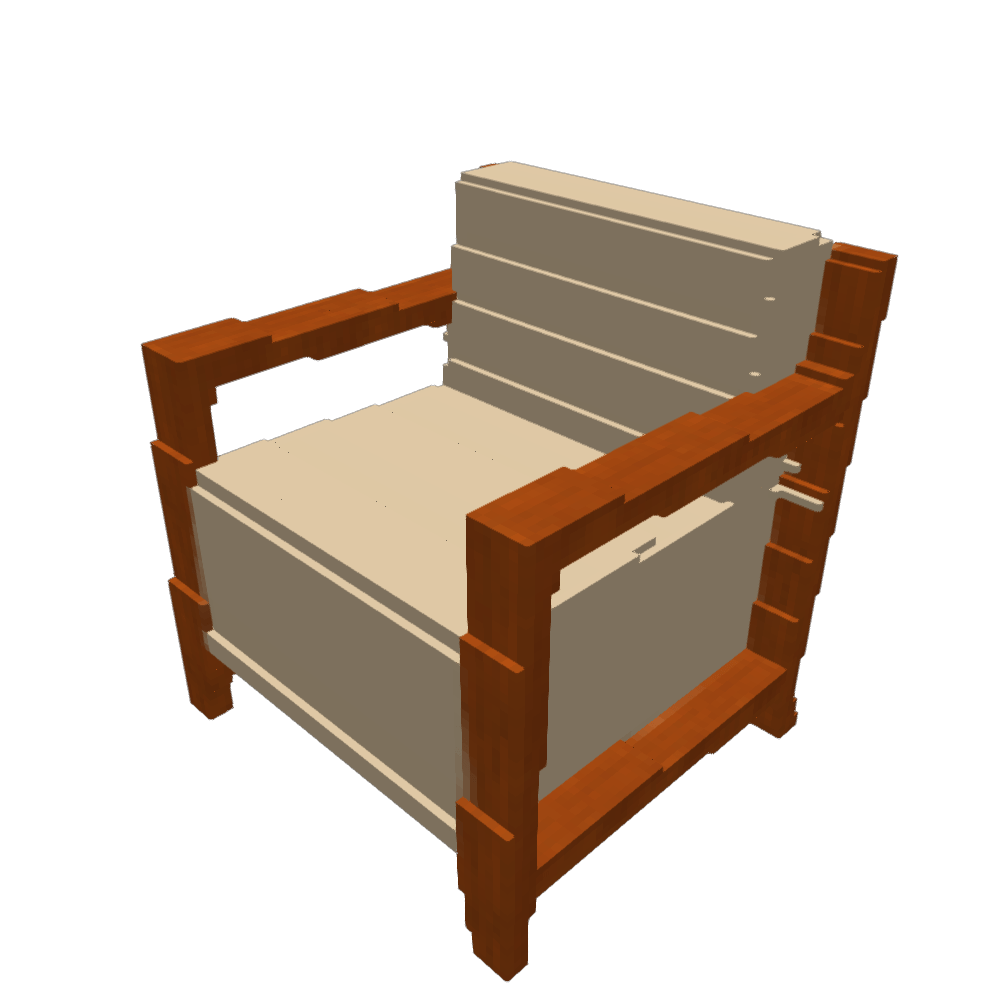} &
\includegraphics[trim=50 120 50 150,clip]{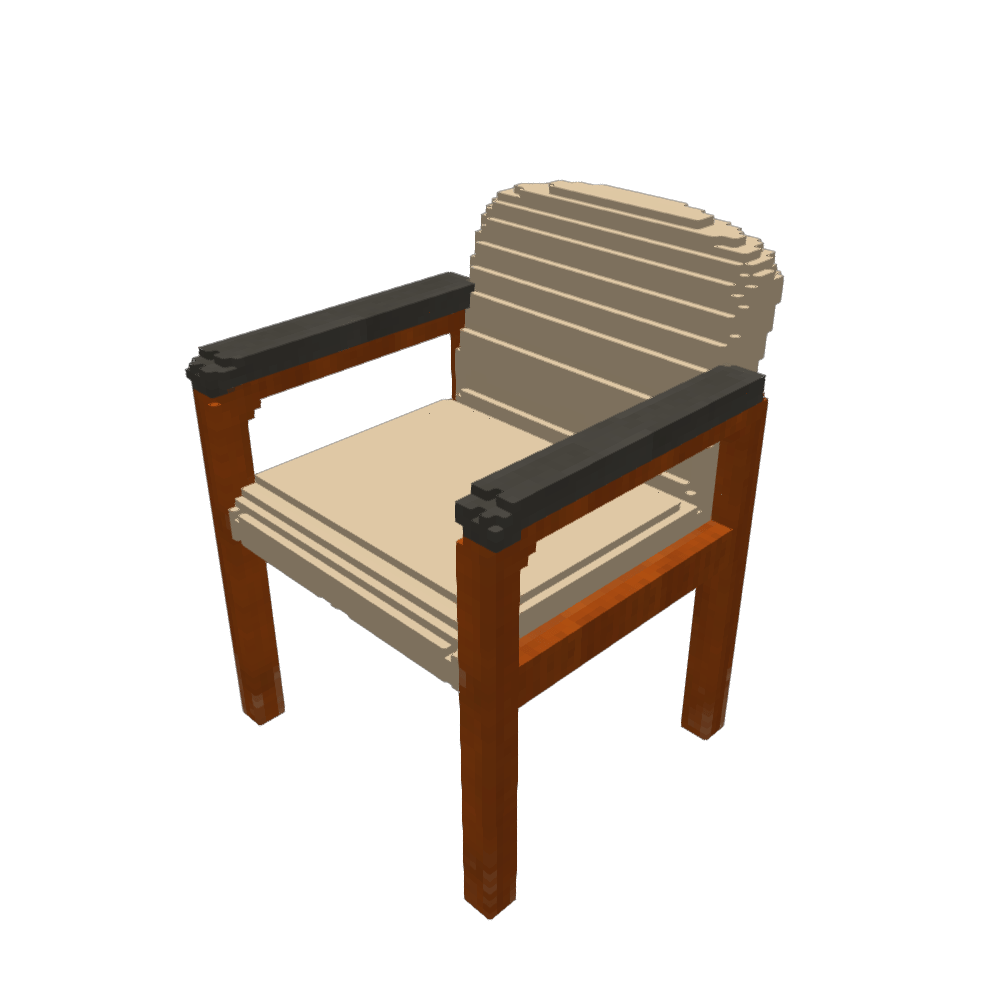} &
\includegraphics[trim=50 120 50 150,clip]{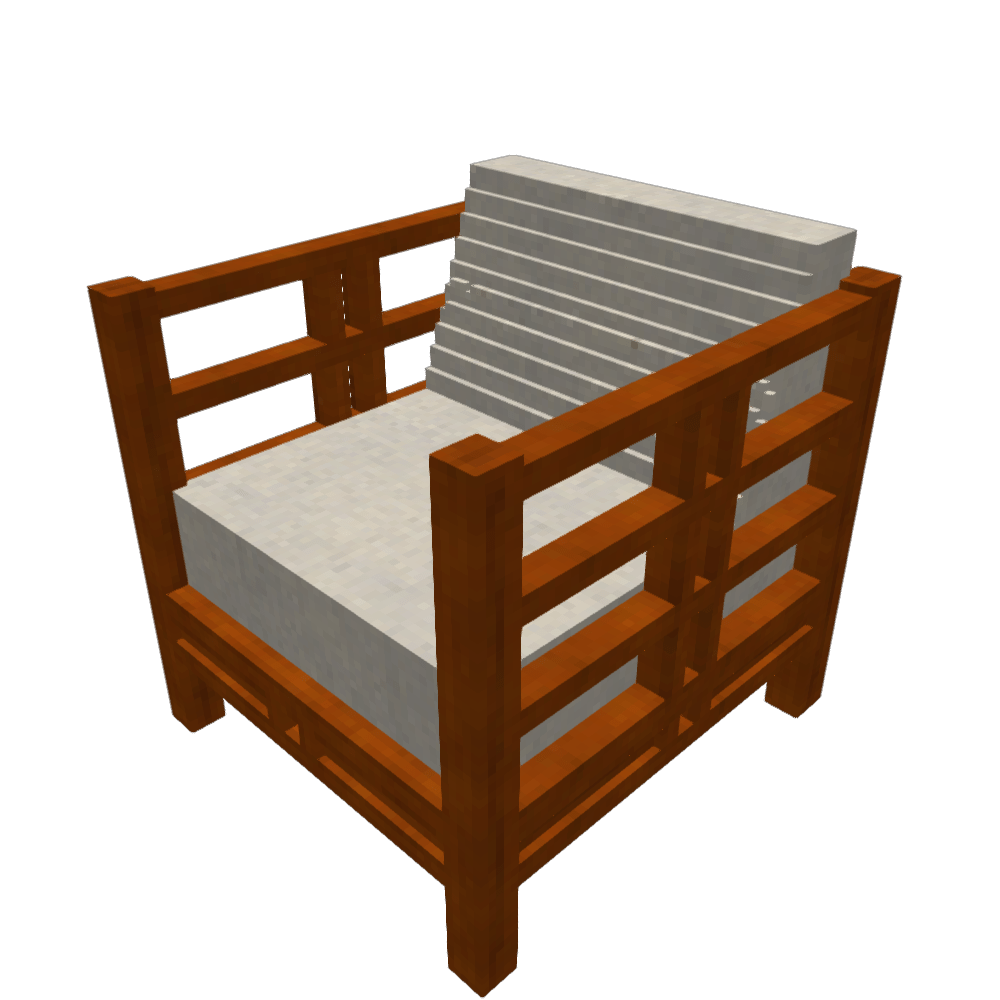} &
\includegraphics[trim=50 120 50 150,clip]{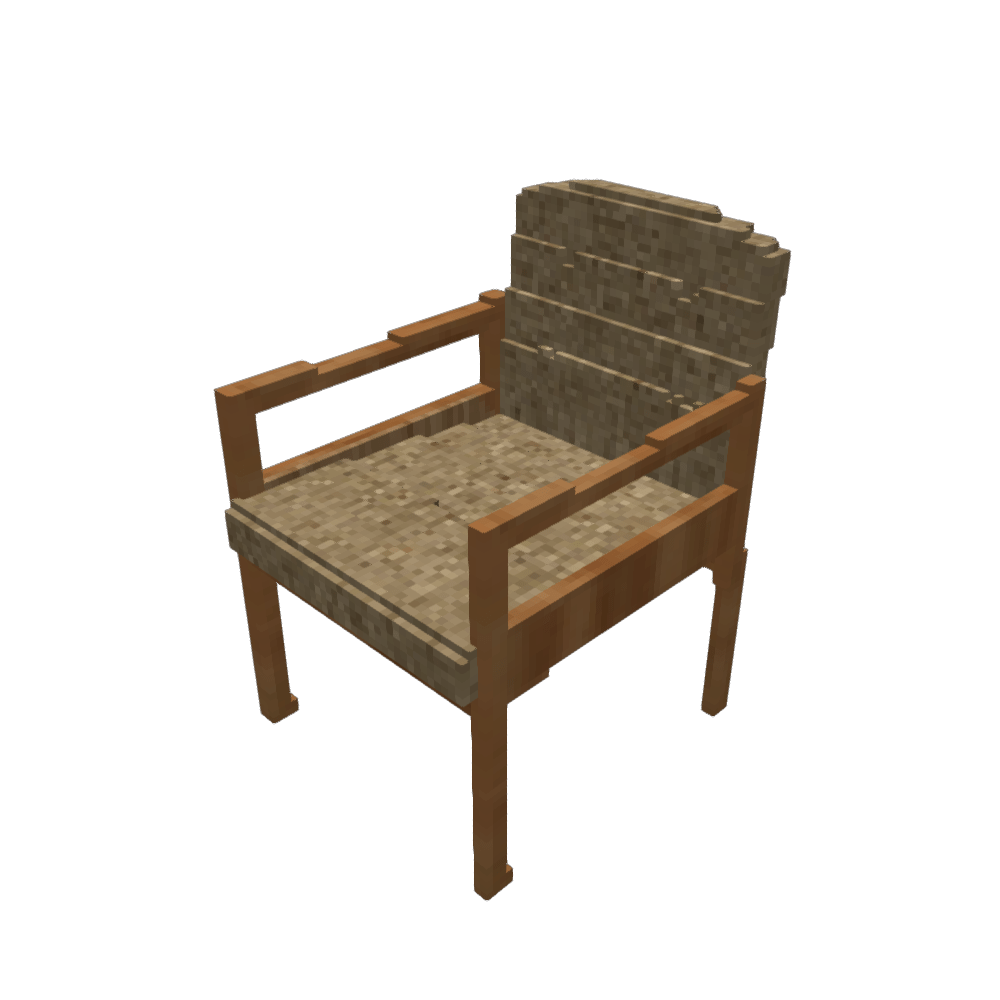}
 \\
	 & & 5.53  & 7.09  & 8.88  & 6.19  & 0.47  \\
\bottomrule
\end{tabularx}
\vspace{-8pt}
\caption{Failed retrievals on the \emph{val} set with \trimodiv (unmatched text is underlined). The ground truth (GT) is shown in the first column, followed by the retrieved results with the ${F1}^{0.1}$ score for each. 
We see that some descriptions are general and our retrieved shapes match the description but is not GT shape (first row), and retrieval of shapes with part details (\emph{wide arm rest}) is hard (second row). 
}
\label{fig:retrieval-failed-visualization}
\end{figure*}

\subsection{Error analysis}

\xhdr{Manual analysis.}
We conduct a manual analysis of the top 5 results returned for 50 text queries from the validation set for \bimodi, \bimodv, and \trimodiv.  We count the number of query results (shapes) that match the description exactly, and categorize the error into color mismatch, large shape mismatch, shape detail mismatch, and missing part (see supplement for examples and \cref{tab:error-analysis} for analysis summary).
As expected from the quantitative results, \trimodiv has the most shapes that match the text.
With the limited number of queries we examined, all models have similar performance on color and missing parts.
The \bimodi model had difficulty getting small shape details correct, and \trimodiv obtained the best performance on matching the overall shape.

\xhdr{Failure cases.}
\cref{fig:retrieval-failed-visualization} shows failure cases for the \trimodiv model.
For the top row, while our model did not retrieve the ground-truth (GT) shape in the top 5, all top 5 retrieved shapes match the input description (28/100 samples that we manually inspected belonged to this error case).
This illustrates that the evaluation data and protocol should be improved beyond what was established in Text2Shape~\cite{chen2018text2shape}.
Despite this, our manual analysis indicates the metrics do a good job of ranking the models.
The second row shows the challenge of retrieving shapes with fine detail.
While our model can retrieve \textit{taupe} colored armchairs, the retrieved objects did not have \textit{wide armrest}.
Our model is good at the overall shape and color, but can miss fine details and infrequent terms such as \textit{foldable}.
More data and modeling part-to-text correspondence (as in Parts2Words~\cite{tang2023parts2words}) can help to reduce these types of failures.

\begin{table}
\centering
\resizebox{\linewidth}{!}
{
\begin{tabular}{@{}lc|cccc@{}}
\toprule
 & match & \makecell{color\\mismatch} & \makecell{big\\shape error} & \makecell{small\\shape error} & \makecell{missing\\part}\\
\midrule
\bimodi & 106 & 65 & 22  & 85 & 5\\
\bimodv & 103 & 67 & 26  & 76 & 5\\
\trimodiv & 113 & 64 & 17 & 74 & 5\\
\bottomrule
\end{tabular}
}
\vspace{-8pt}
\caption{
Manual analysis of the top 5 results returned for 50 text queries.  We group the results into whether they perfectly match the description, or whether there is a mismatch in color or shape.  We confirm that \trimodiv has the best overall performance with the most perfect matches and the least number of shape mismatches.}
\label{tab:error-analysis}
\vspace{-16pt}
\end{table}

\subsection{Limitations}

We investigated a trimodal loss for text-to-shape retrieval and found that with careful tuning we outperformed the SoTA as of early 2022, when this work was initially performed.
We restricted our study to voxel-based 3D representations which often do not capture geometric details and fine-grained surface textures.
It would be interesting to consider other modalities such as point clouds, depth images, and textured 3D polygonal meshes which may alleviate these limitations.
One big challenge of incorporating additional modalities is the memory cost.
In addition, we focused on a specific type of contrastive loss.
Other contrastive losses, data augmentation, as well as other loss terms such as captioning loss and reconstruction loss are promising directions for further improvement.
Our dataset is limited in the style of the text and the coverage of shapes.
The evaluation also assumes that there is only one correct shape but as we have noted, multiple shapes can match a description.
Thus, a significant challenge is to handle false negative pairs in a mini-batch due to the descriptions being ambiguous.
These limitations suggest opportunities for future work.
We believe our work can serve as a good foundation for follow-up work in text-to-shape retrieval.

\section{Conclusion}
\label{sec:conc}

We carry out a systematic study of contrastive losses for text-to-shape retrieval.
We show that using simple contrastive losses can achieve comparable results to text-to-shape retrieval methods relying on extra annotation and complex losses.
Our experiments show that incorporating 3D information either via voxels or multi-view images is helpful for the task.
We identify important challenges to solve for the development of useful text-to-retrieval models.
In addition, we propose a trimodal contrastive loss which further improves performance by considering both 2D and 3D representations. 
We hope our systematic study will serve as a foundation encouraging more work on text-to-shape retrieval, which is an increasingly important task as there are more and more 3D data repositories.

\xhdr{Acknowledgements.}
This work is funded by the Canada CIFAR AI Chair program, an NSERC Discovery Grant, and a TUM-IAS Hans Fischer Fellowship (Focus Group Visual Computing).
This research was enabled in part by support provided by \href{www.westgrid.ca}{WestGrid} and \href{www.computecanada.ca}{Compute Canada}.
We thank Dave Zhenyu Chen for collecting the text descriptions for ShapeNet c13.
We also thank the anonymous reviewers for their feedback, and Manolis Savva for proofreading and editing suggestions.

\clearpage
\newpage
{\small
\bibliographystyle{plainnat}
\setlength{\bibsep}{0pt}
\bibliography{main}
}

\clearpage
\section*{Appendix}
\appendix
\label{sec:supp}
In this supplement to the main paper, we provide details about the statistics of the `chairs and tables' dataset(\cref{sec:supp:statistics}), our rendering process (\cref{sec:supp:render}) and our model implementation (\cref{sec:supp:model-details}), as well as additional evaluation experiments (\cref{sec:supp:exp}).  In \cref{sec:supp:qual}, we provide visualizations of our trimodal embedding space (\cref{sec:supp:tsne}), additional qualitative results (\cref{sec:supp:result-visuals}),  and brief discussion of shape similarity metrics (\cref{sec:supp:shape-sim-discussion}).

\section{Data statistics}
\label{sec:supp:statistics}
\begin{table}[h]
\centering
{
{\footnotesize
\begin{tabular}{@{}l ccc ccc@{}}
\toprule
& \multicolumn{3}{c}{Text} & \multicolumn{3}{c}{Shape} \\ \cmidrule{2-4} \cmidrule{5-7}
Category & Train & Val & Test & Train & Val & Test \\ \midrule
Chair & 26257 & 3313 & 3206 & 5221 & 659 & 641 \\
Table & 33520 & 4122 & 4246 & 6700 & 827 & 851 \\ 
\midrule
Total & 59777 & 7435 & 7452 & 11921 & 1486 & 1492 \\
\bottomrule
\end{tabular}
}

}
\vspace{-6pt}
\caption{`Chairs and tables' statistics~\cite{chen2018text2shape}.}
\label{tab:data-split}
\end{table}

\begin{table}[h]
\centering
{
{\footnotesize
\begin{tabular}{@{}l ccc ccc@{}}
\toprule
& \multicolumn{3}{c}{Text} & \multicolumn{3}{c}{Shape} \\ \cmidrule{2-4} \cmidrule{5-7}
Category & Train & Val & Test & Train & Val & Test \\ \midrule
Table & 33547 & 4167 & 3815 & 6773 & 838 & 765 \\
Chair & 25843 & 3413 & 3584 & 5181 & 687 & 719 \\
Sofa & 11036 & 1487 & 1440 & 2543 & 321 & 307\\
Lamp & 8078 & 1087 & 1120 & 1849 & 229 & 239 \\
Loudspeaker & 5425 & 721 & 757 & 1276 & 158 & 160 \\
Bookshelf & 1560 & 231 & 131 & 366 & 48 & 28 \\
Display & 3474 & 502 & 567 & 851 & 110 & 125 \\
Clock & 2437 & 298 & 292 & 526 & 62 & 61 \\
Trash bin  & 1296 & 129 & 143 & 286 & 28 & 29 \\
Cabinet & 5313 & 657 & 782 & 1255 & 141 & 173 \\
Bathtub & 2770 & 413 & 406 & 678 & 89 & 89 \\
File cabinet & 1022 & 108 & 174 & 230 & 23 & 37\\
Bed & 864 & 85 & 89 & 184 & 17 & 18 \\
\midrule
Total & 102665 & 13298 & 13300 & 21998 & 2751 & 2750 \\
\bottomrule
\end{tabular}
}
}
\vspace{-6pt}
\caption{`C13' statistics.}
\label{tab:data-split-c13}
\end{table}

\section{Render settings}
\label{sec:supp:render}
For the rendering setup, we use Pyrender\footnote{ https://github.com/mmatl/pyrender}. The object is placed at the center (0, 0, 0). The camera is placed at (0, 1, 0.6) with the focal length set to 35mm and the sensor width to 32mm while being pointed towards the center (0, 0, 0).
We render 12 images by rotating the camera 30 degrees per render with render resolution set to $224$.

\section{Model details}
\label{sec:supp:model-details}

\subsection{Voxel encoder details}
\label{sec:supp:voxel-encoder}
\begin{table}
\centering
{
\begin{tabular}{@{}lrrrrrr@{}}
\toprule
Layer & Kernel & Stride & Channels & BN & LR\\
\midrule
conv1 & 3 & 1 & 32 & Y & Y \\
max\_pool1 & 2 & 2 & - & - & - \\
conv2 & 3 & 1 & 64 & Y & Y \\
max\_pool2 & 2 & 2 & - & - & - \\
conv3 & 3 & 1 & 128 & Y & Y \\
max\_pool3 & 2 & 2 & - & - & - \\
conv4 & 3 & 1 & 256 & Y & Y \\
max\_pool4 & 2 & 2 & - & - & - \\
conv5 & 3 & 1 & 512 & Y & Y \\
max\_pool5 & 2 & 2 & - & - & - \\
fc6   & - & - & 512 & N & N \\
\bottomrule
\end{tabular}
}
\vspace{-6pt}
\caption{Voxel encoder for resolution $64^3$}
\label{tab:model-architecture}
\end{table}

For the voxel encoder, we use a 5-layer sparse 3D CNN architecture with input resolution $64^3$.
\cref{tab:model-architecture} shows the architectural details.
Here BN stands for Batch Normalization and LR stands for Leaky ReLU, each layer of convolution is followed by normalization then activation.

\subsection{Incorporating CLIP}
\label{sec:supp:clip-model}

To use CLIP~\cite{radford2021learning} for our retrieval task, we feed 6 multi-view images of an object into the image encoder for CLIP separately then average the vectors to get the image embedding.  Specifically, we use the ViT-L/14 pretrained model from CLIP.
For retrieval, we encode the text using the pretrained transformer-based CLIP text encoder and then retrieve relevant shapes by taking the dot product of the text and shape embeddings.  We compared the zero-shot performance of CLIP embedding to the embedding from CLIP with additional MLPs (which is equivalent to taking the cosine similarity) and found that the CLIP+MLP embeddings work better. For incorporating CLIP into our model, we take the CLIP embeddings of each image, and project the embeddings using a two-layer MLP.  The projected embeddings of the 6 multi-view images are averaged, and the weights of the MLP are trained using a Cross-Entropy loss. We train using the Adam~\cite{kingma2014adam} optimizer until the validation performance drops, with a learning rate of 0.00035 on A40.  For fast training, we preprocess the data and store frozen clip embeddings in a cache.

\subsection{Triplet loss}
\label{sec:supp:triplet-loss}

Given an anchor text embedding $\mu_{t_j}$ and its positive shape embedding $\mu_{s_j}$, we sample the semi-hard examples $\mu_{s_k}$ such that $\langle\mu_{t_j}, \mu_{s_j}\rangle < \langle \mu_{t_j},\mu_{s_k}\rangle < \langle \mu_{t_j},\mu_{s_j}\rangle+\alpha$ is satisfied, where $\alpha$ is the margin. The semi-hard example is sampled online by selecting from the mini-batch as in \citet{schroff2015facenet}. Semi-hard sampling has been shown to work better than naive sampling or hard negative sampling for triplet loss and is also used in \citet{tang2023parts2words}. We then apply the triplet loss using the triplets $(\mu_{t_j}, \mu_{s_j}, \mu_{s_k})$, here $\mu_s$ can be the voxel or image embeddings:
\begin{equation}
l_j = \sum_{\text{all satisfied k}}\max(\langle\mu_{t_j}, \mu_{s_j}\rangle - \langle \mu_{t_j},\mu_{s_k}\rangle + \alpha, 0)
\end{equation}

\subsection{Model size}
\label{sec:supp:model-size}

\begin{table}
\centering
{
\begin{tabular}{@{}lrrrr@{}}
\toprule
Model & \#Params & Resolution & BS & Memory \\
\midrule
\multirow{5}{*}{\bimodv} & \multirow{5}{*}{6.6M} & $32^3$ & $128$ & 2.6 GB \\
\cmidrule{3-5}
 &  & \multirow{4}{*}{$64^3$} & $32$ & 3.3 GB\\
& & & $64$ & 5.1 GB\\
& & & $128$ & 10.1 GB\\
& & & $256$ & 18.3 GB\\
\cmidrule{1-5}
\multirow{6}{*}{\bimodi} & \multirow{6}{*}{13.3 M}  & $64^2$ & \multirow{3}{*}{$128$} & 2.7 GB \\
& & $128^2$ & & 9.9 GB \\
& & $224^2$ & & 30.4 GB \\
\cmidrule{3-5}
& & \multirow{4}{*}{$128^2$} & $32$ & 2.6 GB \\
& & & $64$ & 5.2 GB \\
& & & $128$ & 9.9 GB \\
& & & $256$ & 20.8 GB \\
\cmidrule{1-5}
\trimodiv & 20.4 M & v$64^3$i$128^2$ & $128$ & 11.8 GB \\
\bottomrule
\end{tabular}
}
\vspace{-6pt}
\caption{Memory usage and number of parameters for the \bimodi, \bimodv and \trimodiv models.}
\label{tab:model-memory}
\end{table}

We show the number of parameters and memory for our models with different hyperparameters in \cref{tab:model-memory}.
Here BS stands for batch size.
\bimodi is trained with 6 multi-view images and a ResNet-18 backbone.
\trimodiv is trained with voxel resolution $64^3$ and image resolution $128^2$.
Here, the memory usage shown is calculated using {\footnotesize\texttt{torch.cuda.max\_memory\_reserved}} during training.

\section{Additional experiments}
\label{sec:supp:exp}

We conduct additional experiments to investigate the impact of image and voxel resolution (\cref{sec:supp:exp-resolution}) and image backbone architecture (\cref{sec:supp:exp-image-backbone}) on text-to-shape retrieval.
We present the performance of our model on an extended set of 13 object categories from ShapeNet~\cite{chang2015shapenet} (\cref{sec:supp:exp-c13}).

\subsection{Resolution experiments}
\label{sec:supp:exp-resolution}

\begin{table}
\centering
\resizebox{\linewidth}{!}
{
\begin{tabular}{@{}lrrrrr@{}}
\toprule
& resolution & RR@1($\uparrow$) & RR@5($\uparrow$) & NDCG@5($\uparrow$) & MRR($\uparrow$) \\
\midrule
\textmr{\bimodi}{3} & 64 & 9.98 $\pm$ 0.28 & 28.60 $\pm$ 0.19 & 19.56 $\pm$ 0.20 & 19.88 $\pm$ 0.24 \\
& 128 & \textbf{11.61 $\pm$ 0.20} & 30.65 $\pm$ 0.19 & 21.36 $\pm$ 0.23 & 21.46 $\pm$ 0.25 \\
& 224 & 11.45 $\pm$ 0.18 & \textbf{32.06 $\pm$ 0.22} & \textbf{22.01 $\pm$ 0.26} & \textbf{21.86 $\pm$ 0.27} \\
\midrule
\textmr{\bimodv}{3} & 32 & 8.45 $\pm$ 0.32 & 26.00 $\pm$ 0.35 & 17.32 $\pm$ 0.20 & 17.72 $\pm$ 0.19 \\
& 64 & \textbf{9.59 $\pm$ 0.27} & \textbf{27.14 $\pm$ 0.48} & \textbf{18.54 $\pm$ 0.13} & \textbf{19.03 $\pm$ 0.08} \\
\bottomrule
\end{tabular}
}
\vspace{-6pt}
\caption{Comparison of resolution settings on shape retrieval for \bimodi and \bimodv on the validation set. We find that increasing the resolution increases the performance.}
\label{tab:retrieval-resolution}
\end{table}

We conduct experiments for different resolutions of images ($64^2$, $128^2$ and $224^2$) and voxels ($32^3$ and $64^3$).
In \cref{tab:retrieval-resolution} we see that the performance increases with higher resolutions.

\subsection{Image backbone experiments}
\label{sec:supp:exp-image-backbone}

\begin{table}
\centering
\resizebox{\linewidth}{!}
{
\begin{tabular}{@{}lrrrrr@{}}
\toprule
& Model & RR@1($\uparrow$) & RR@5($\uparrow$) & NDCG@5($\uparrow$) & \#Params\\
\midrule
\textmr{\bimodi}{3} & EfficientNet-B0 & \textbf{12.52 $\pm$ 0.22} & \textbf{32.53 $\pm$ 0.28} & \textbf{22.79 $\pm$ 0.31} & 7.2M \\
& ResNet-18 & 11.61 $\pm$ 0.20 & 30.65 $\pm$ 0.19 & 21.36 $\pm$ 0.23 & 13.3M \\
& ResNet-34 & 11.41 $\pm$ 0.21 & 30.68 $\pm$ 0.17 & 21.21 $\pm$ 0.25 & 23.4M \\
\bottomrule
\end{tabular}
}
\vspace{-6pt}
\caption{Comparison of different architectures on shape retrieval for \bimodi on the validation set. We find that increasing the model parameters actually decreases the performance from ResNet-18 to ResNet-34. Using a more powerful model with fewer parameters (i.e. EfficientNet), we see that the performance improves.}
\label{tab:retrieval-architecture}
\end{table}

We conduct experiments with different image backbones for \bimodi to see how different model sizes will impact the retrieval performance.
The results are shown in \cref{tab:retrieval-architecture}.
When we increase model parameter size from ResNet-18~\cite{he2016deep} to ResNet-34 it can be seen that the performance drops, this is in conflict with prior work \cite{chen2020simple, radford2021learning} on contrastive learning that sees performance gains when using larger encoder models.
This could support our hypothesis that our dataset is relatively small, and using bigger models will result in overfitting.
To verify this we conduct another experiment using a model that uses fewer parameters but has comparable performance to ResNet-152, namely EfficientNet-B0~\cite{Tan2019EfficientNetRM}.
We see that EfficientNet-B0 performs better than ResNet-18 and ResNet-34.
For our use case, it may be more desirable to use models that are more lightweight, but still offer good performance.
Note that we don't use EfficientNet-B0 in our main model because it actually uses more than 2x memory in training compared to ResNet-18 due to the operations in EfficientNet requiring more memory for backpropagation.

\begin{figure*}
\centering 
\includegraphics[width=\textwidth]{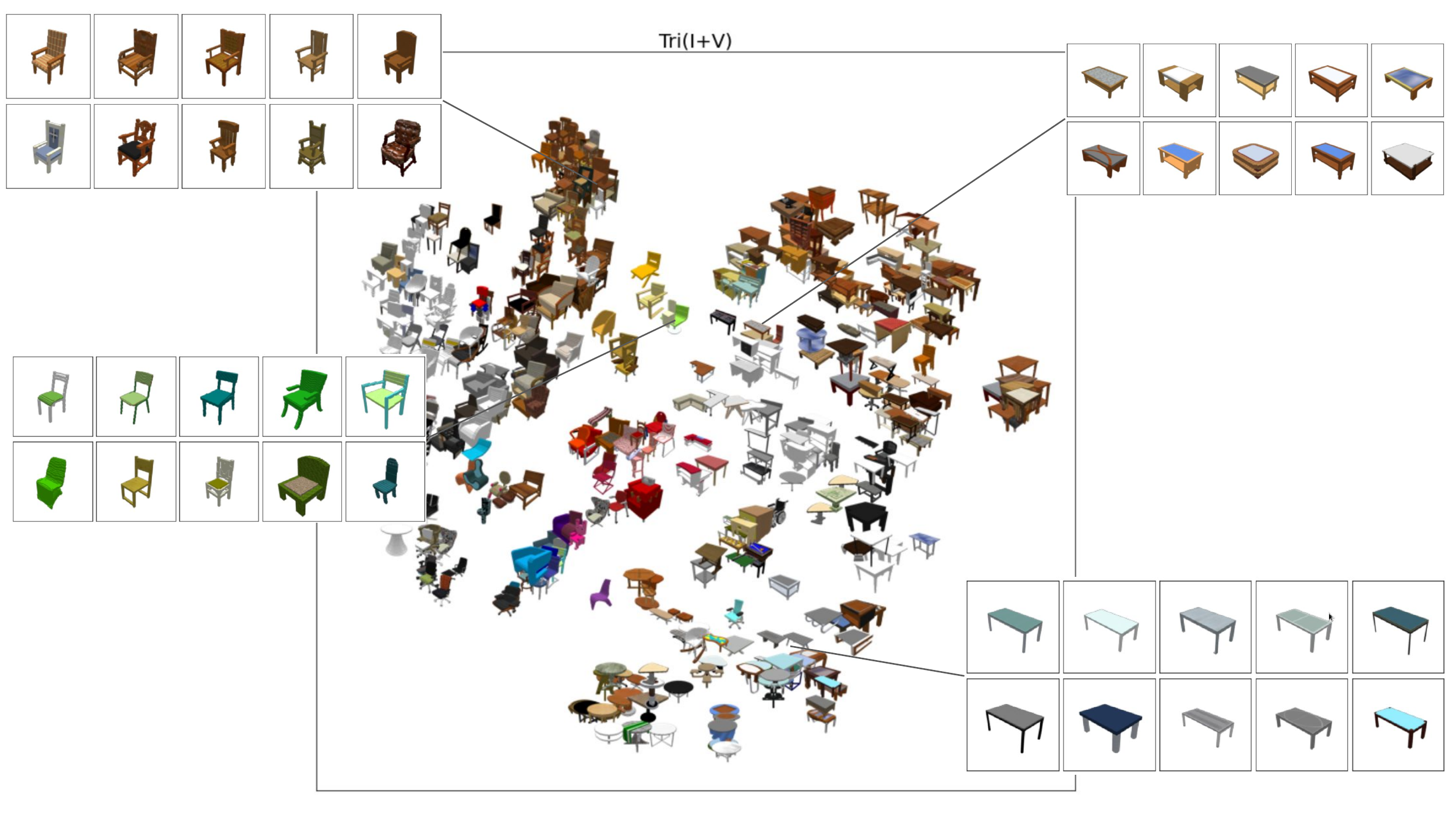}
\caption{Detailed joint embedding space for \trimodiv with t-SNE. We also zoom in on four regions of the joint embedding space. It is clear that similar shapes are close to each other in the embedding space. }
\label{fig:tsne-with-images} 
\end{figure*}
\begin{figure}
\centering
\setkeys{Gin}{width=\linewidth}
\small
\setlength\tabcolsep{1.5pt}
\begin{tabularx}{\linewidth}{@{}Y|YY@{}}
\toprule
\multirow{2}{\hsize}{a black desk chair with two legs made of metal piping which join under the back of the chair along with arm rests and a split back rest} & color mismatch & big shape error\\
 & \includegraphics[trim=0 40 0 40,clip]{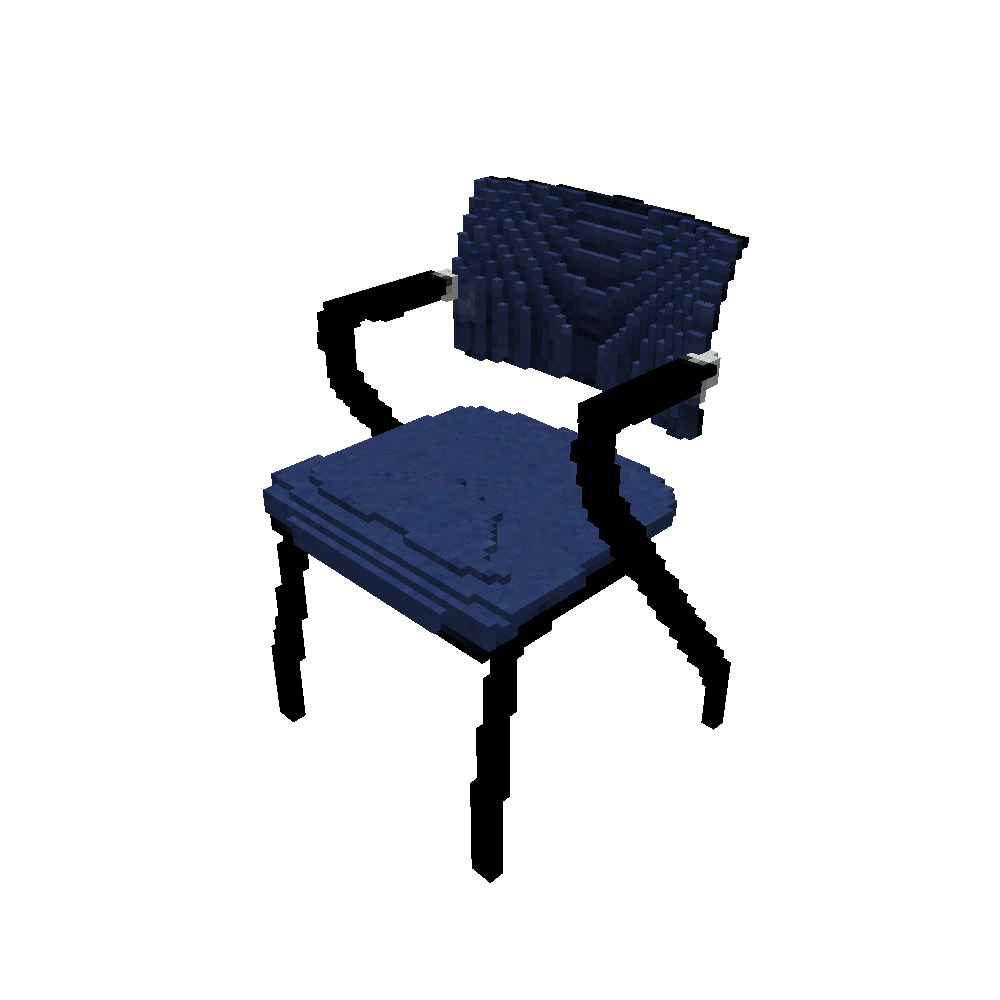} & \includegraphics[trim=0 40 0 40,clip]{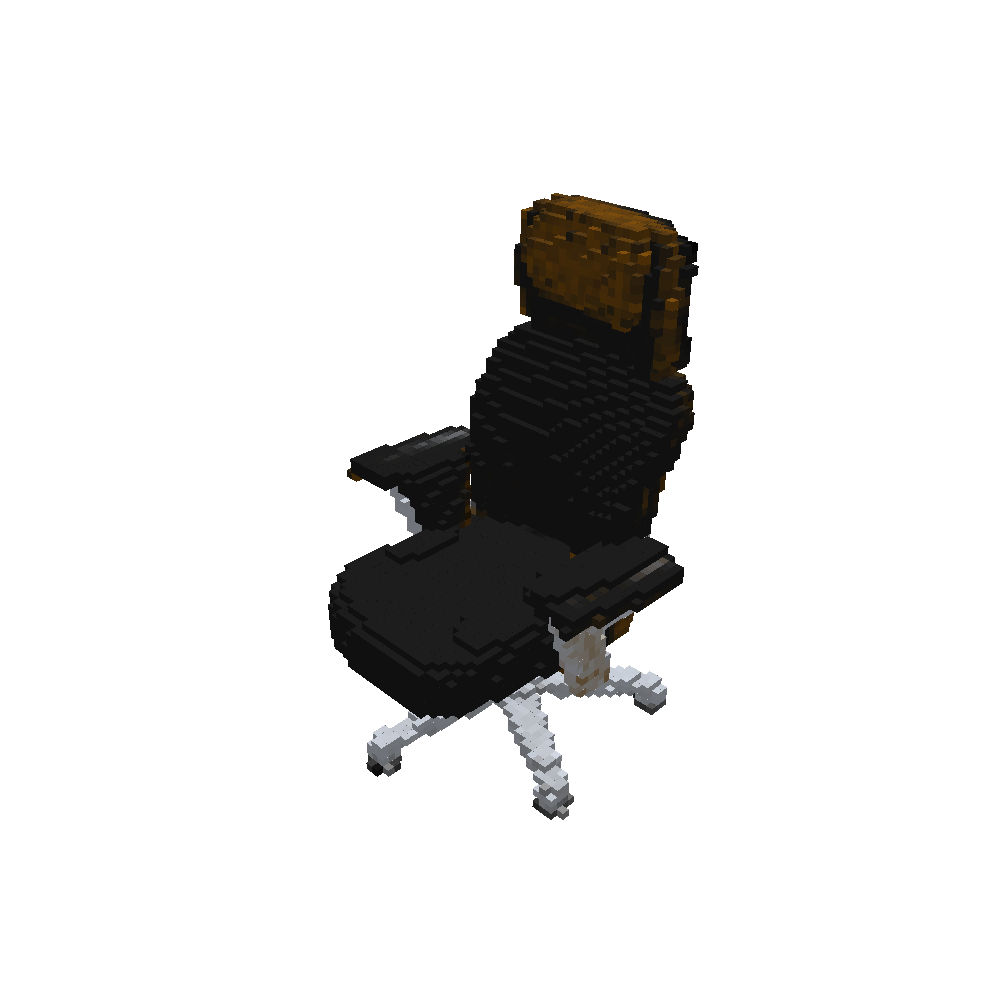} \\
GT & small shape error & missing part\\
\includegraphics[trim=0 80 0 80,clip]{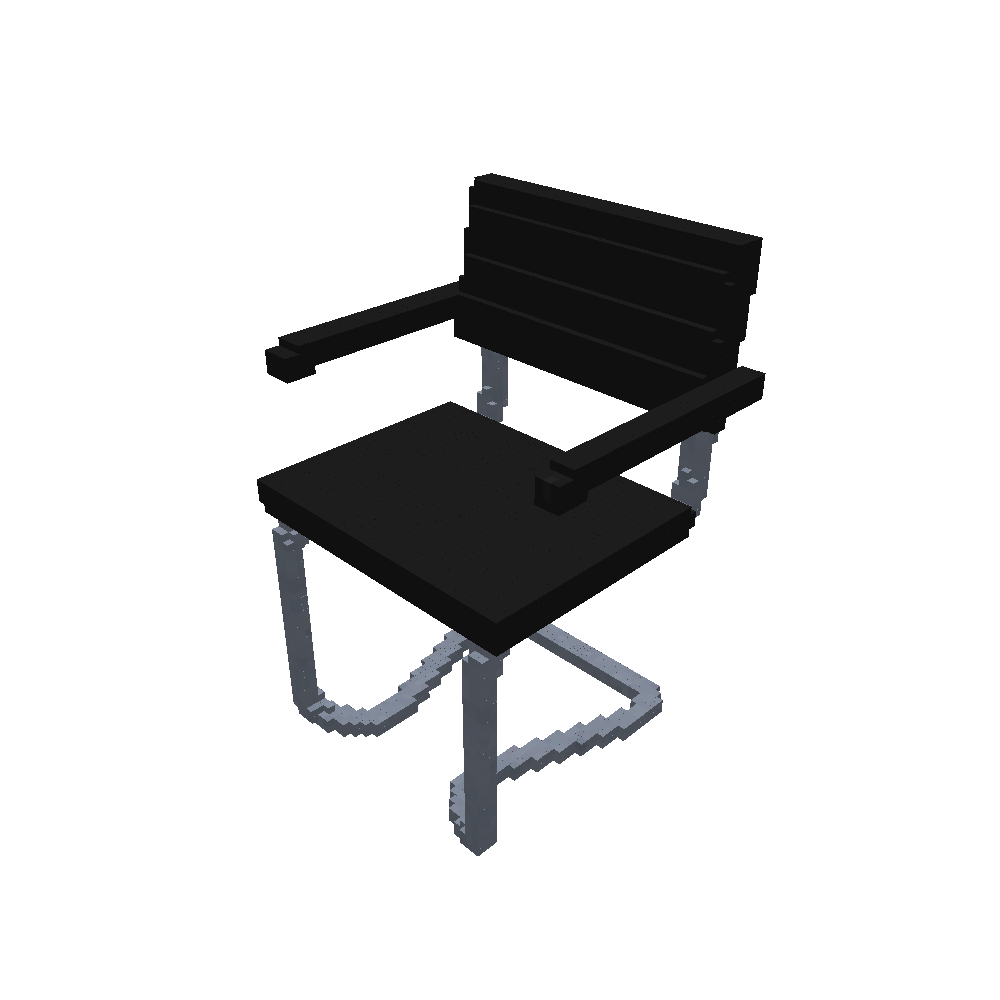} & \includegraphics[trim=0 80 0 80,clip]{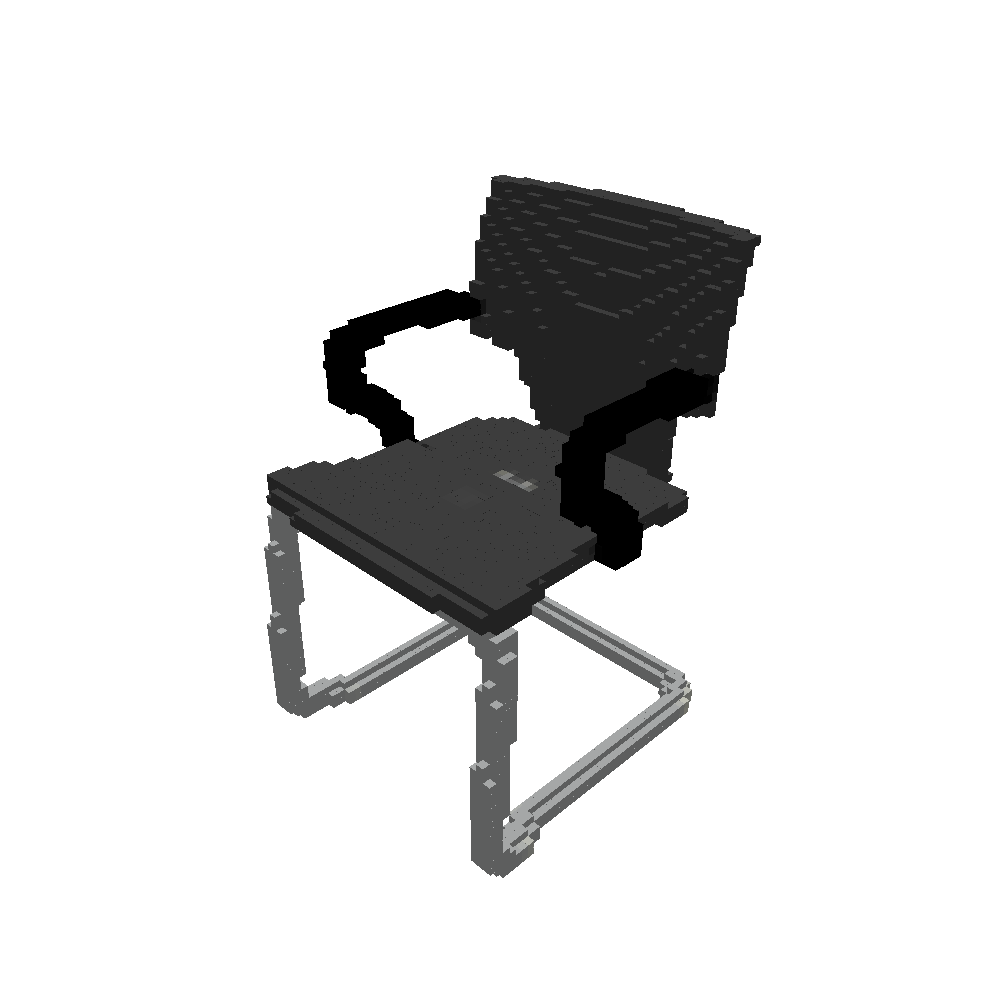} & \includegraphics[trim=0 80 0 80,clip]{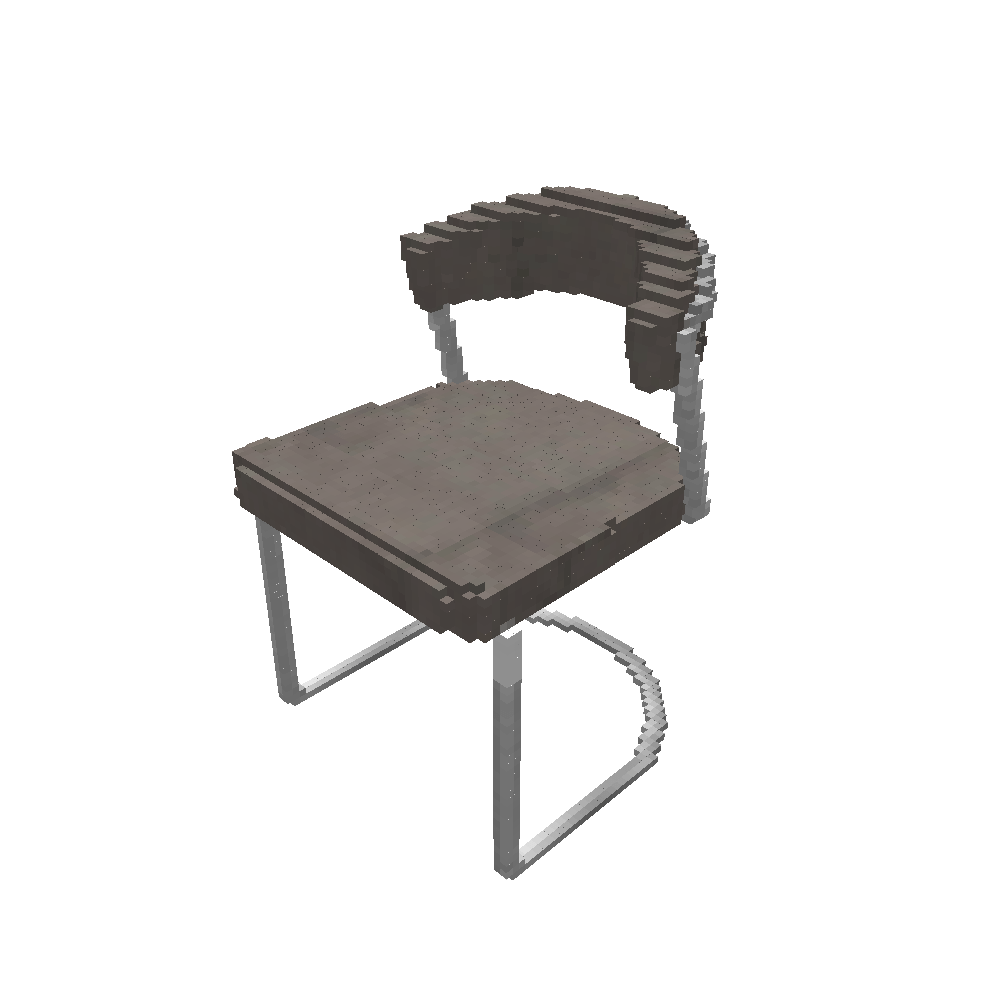} \\
\bottomrule
\end{tabularx}
\caption{Examples of the four types of error we analyzed in our manual analysis.}
\label{fig:retrieval-error-example}
\end{figure}

\begin{figure}
\centering
\scriptsize
\setkeys{Gin}{width=\linewidth}
\begin{tabularx}{\linewidth}{@{} p{2cm}YYY} %
\toprule
 & \bimodi & \bimodv & \trimodiv \\ %
\midrule
\vspace{-1.5em}a dark brown colored wooden table &
\includegraphics[trim=100 150 100 150,clip]{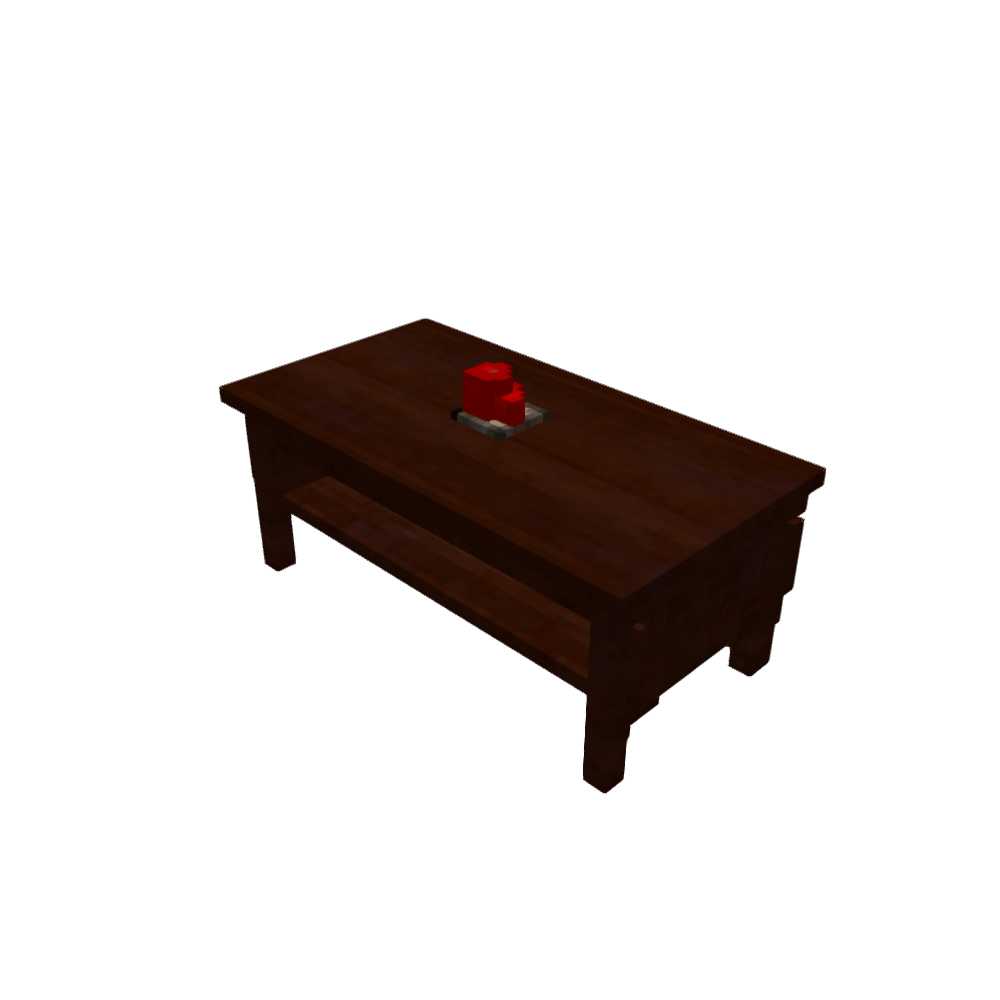} & 
\includegraphics[trim=100 150 100 150,clip]{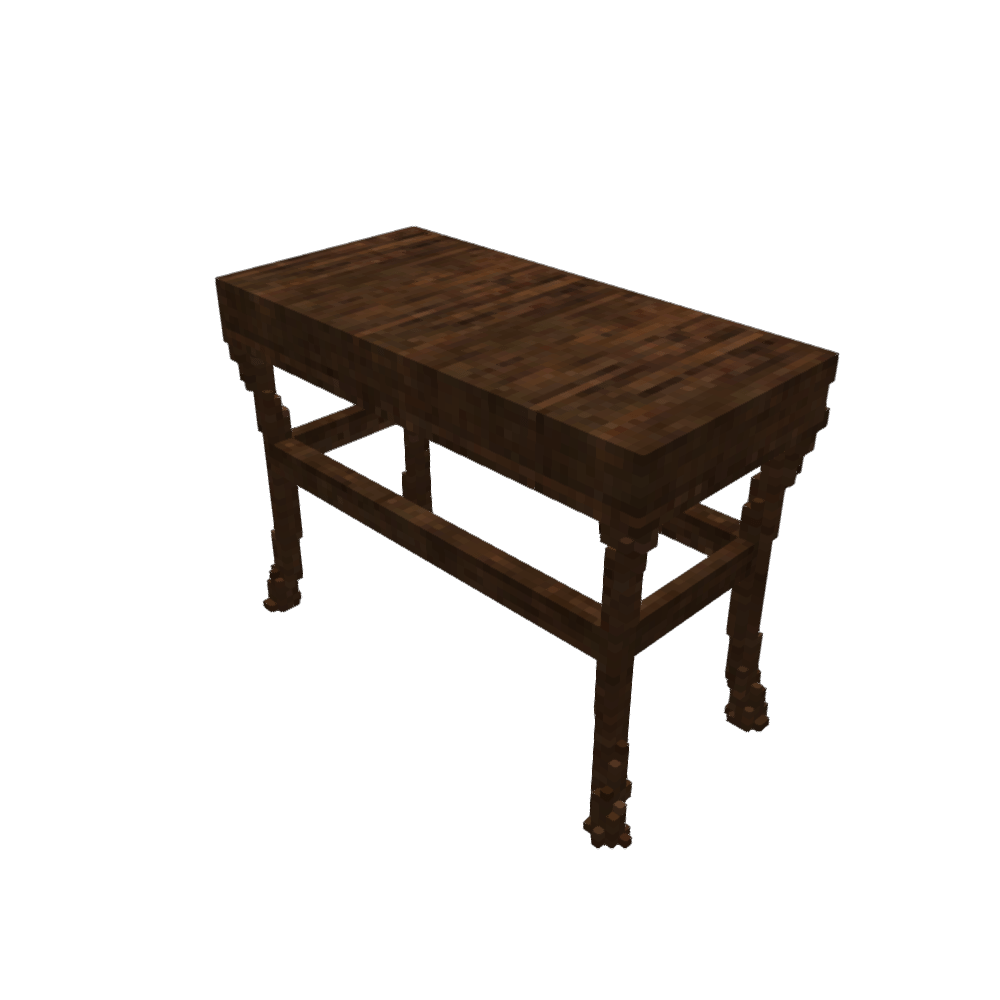} &
\includegraphics[trim=100 150 100 150,clip]{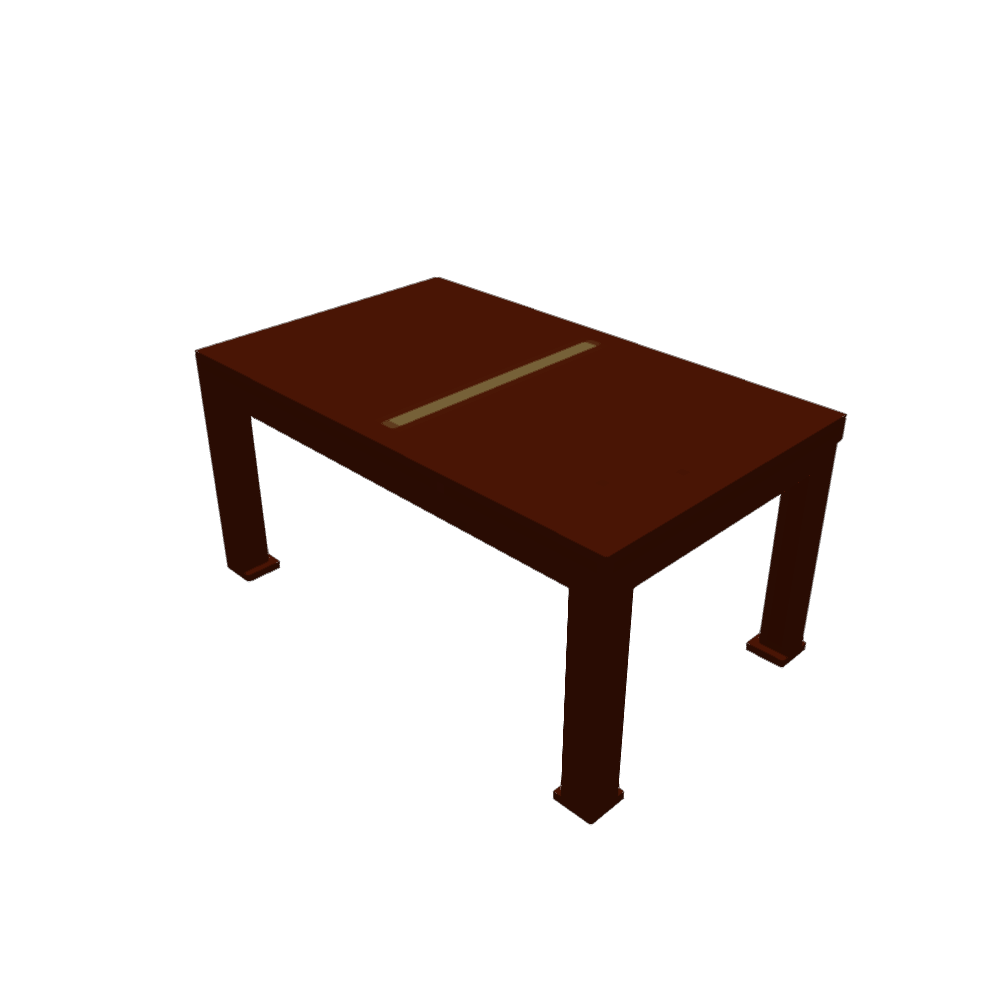} \\ %
\vspace{-1.5em}circular table with glass &
\includegraphics[trim=100 150 100 150,clip]{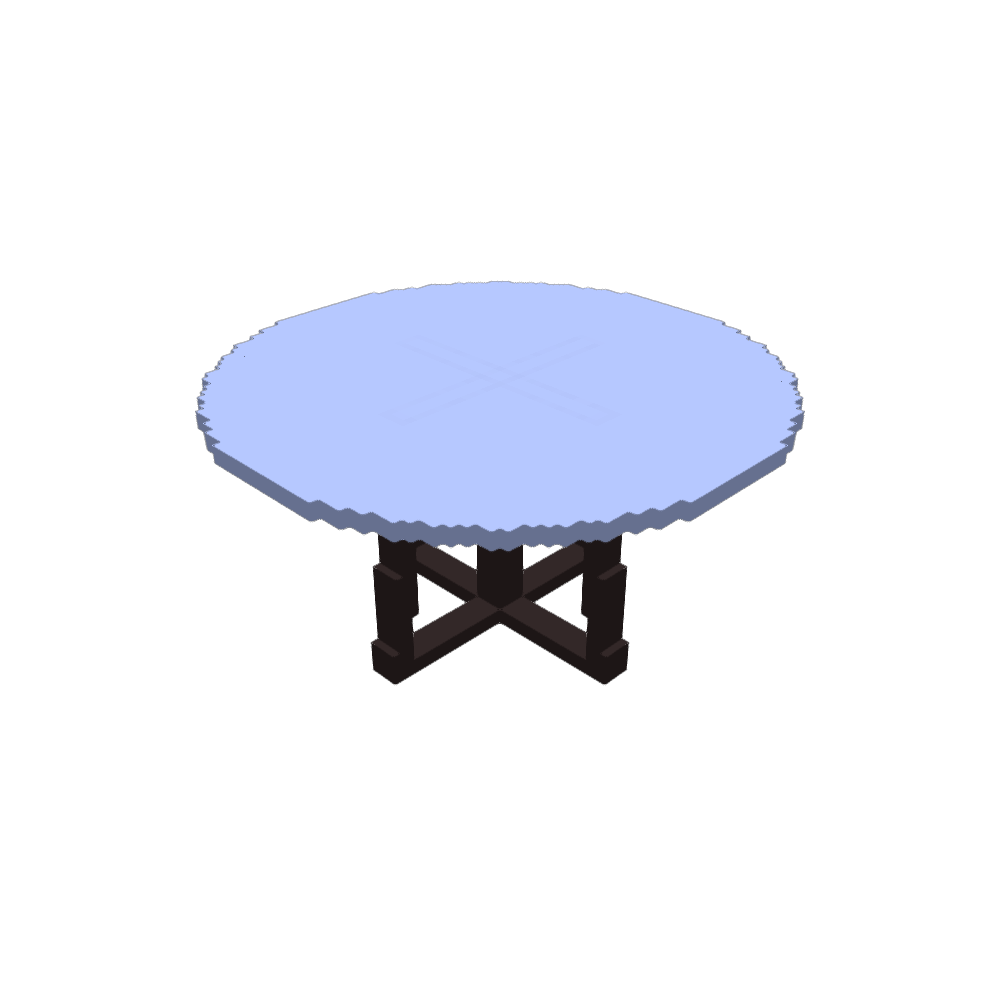} &
\includegraphics[trim=100 150 100 150,clip]{figures/fake/c3c467718eb9b2a313f96345312df593.png} &
\includegraphics[trim=100 150 100 150,clip]{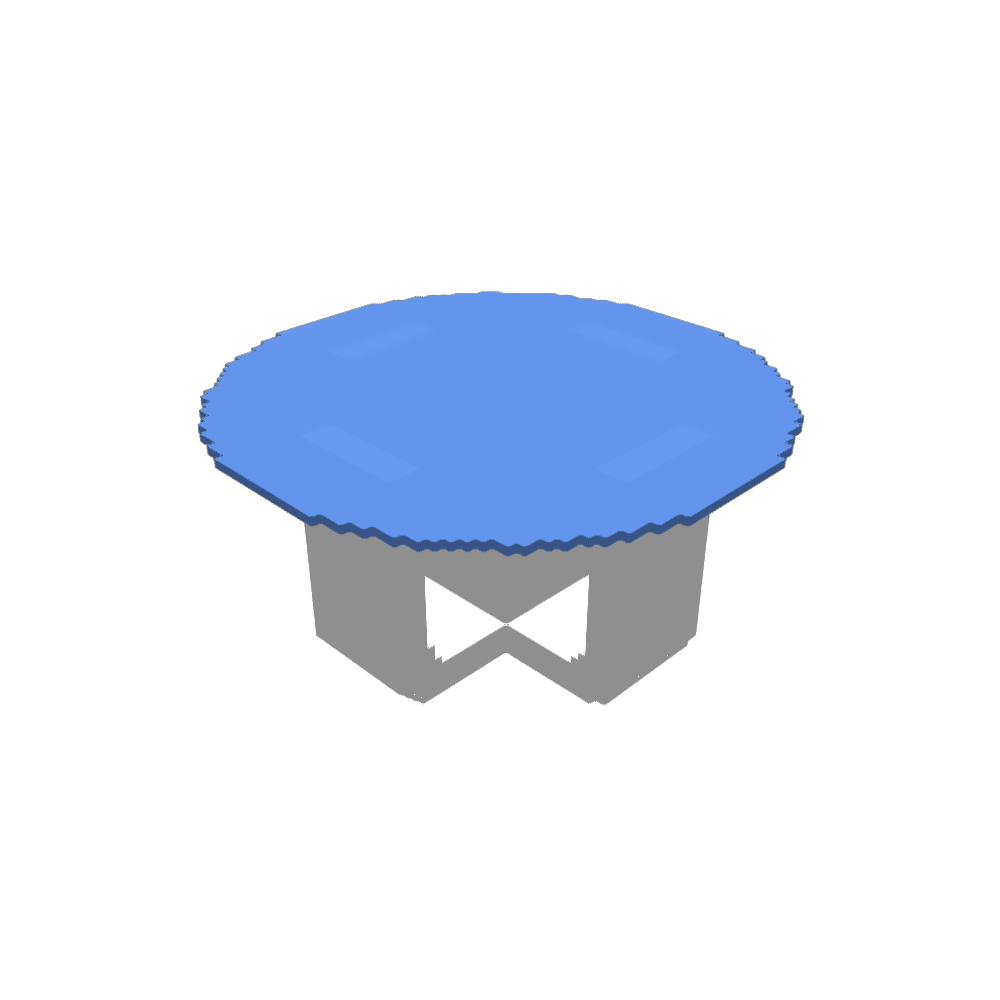} \\ %
\vspace{-1.5em}chair with arm &
\includegraphics[trim=50 100 50 100,clip]{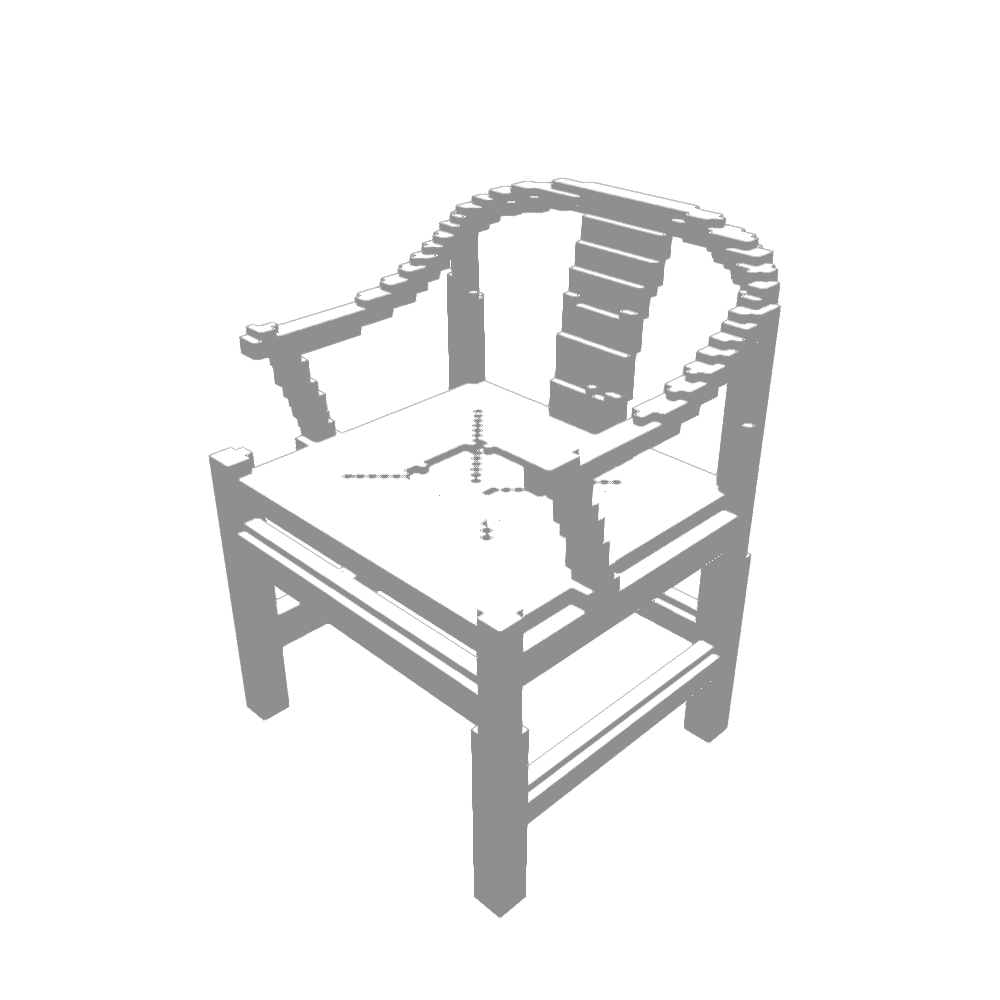} &
\includegraphics[trim=50 100 50 100,clip]{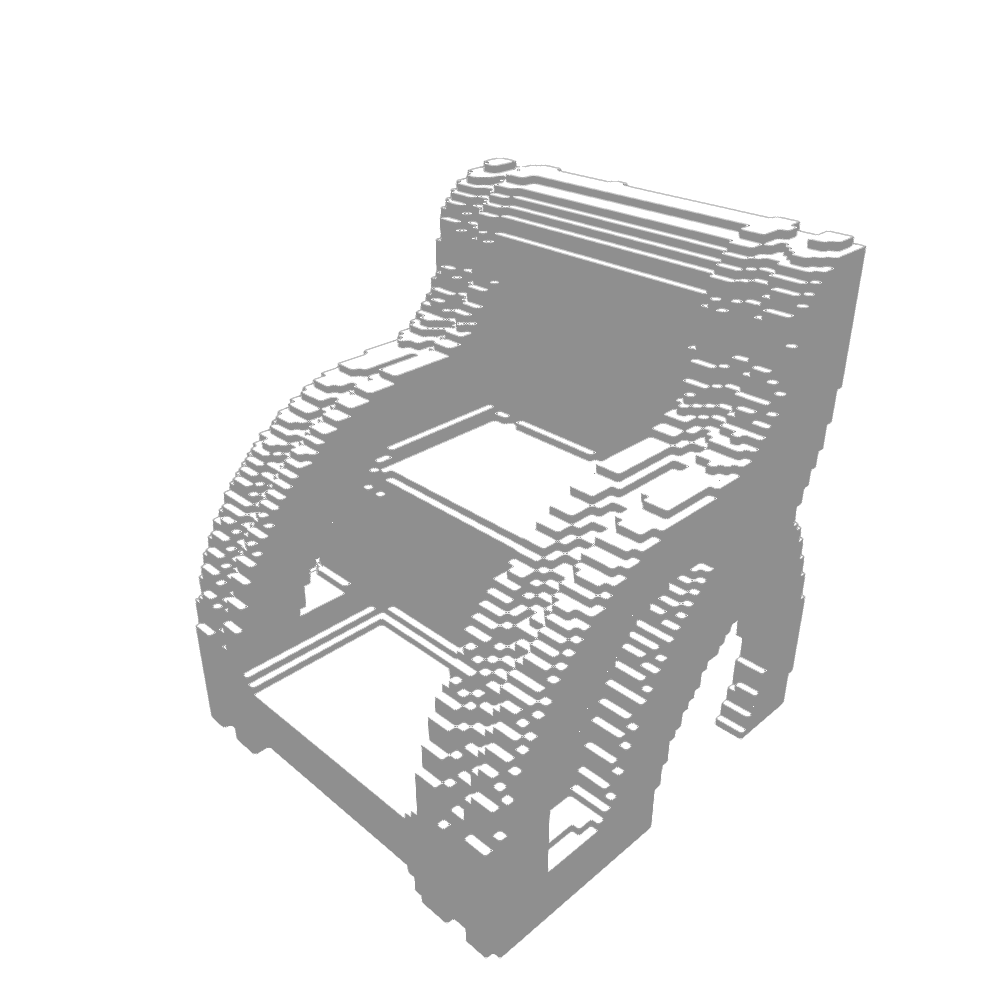} &
\includegraphics[trim=50 100 50 100,clip]{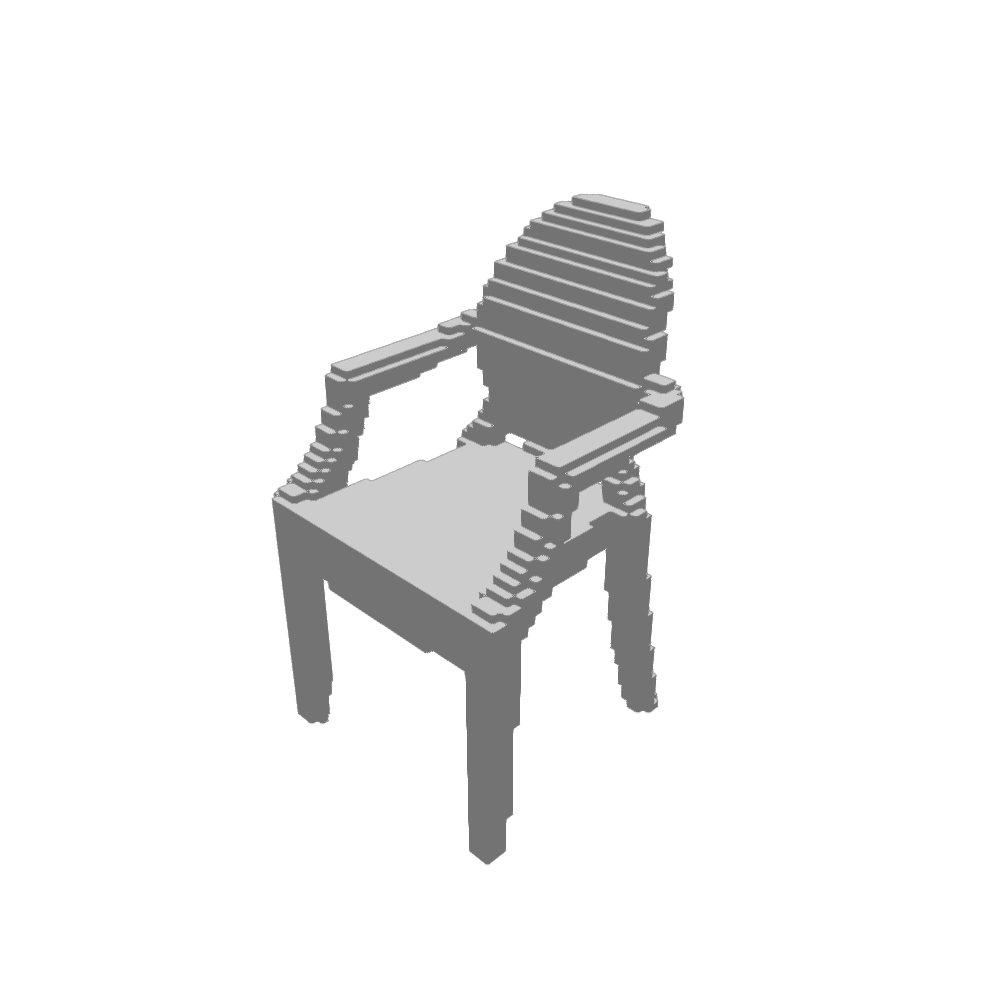} \\
\vspace{-1.5em}chair without arm &
\includegraphics[trim=50 100 50 100,clip]{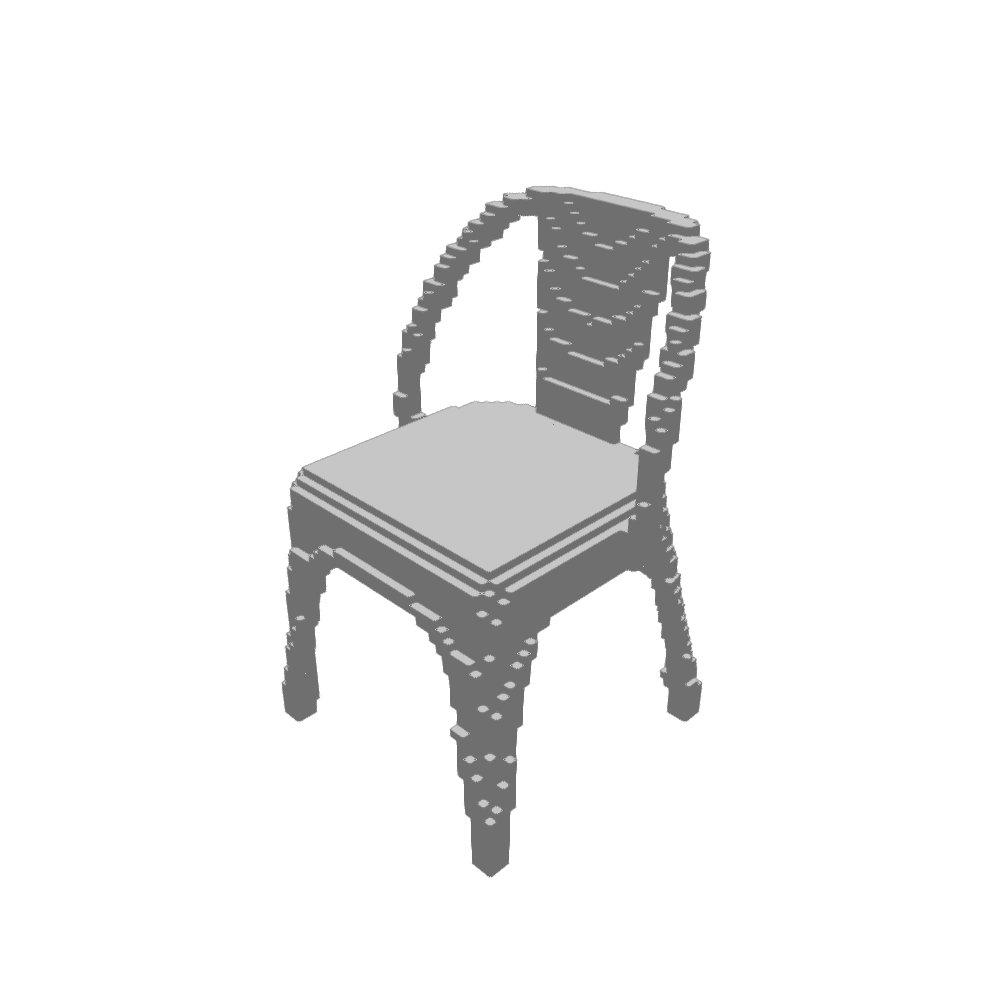} &
\includegraphics[trim=50 100 50 100,clip]{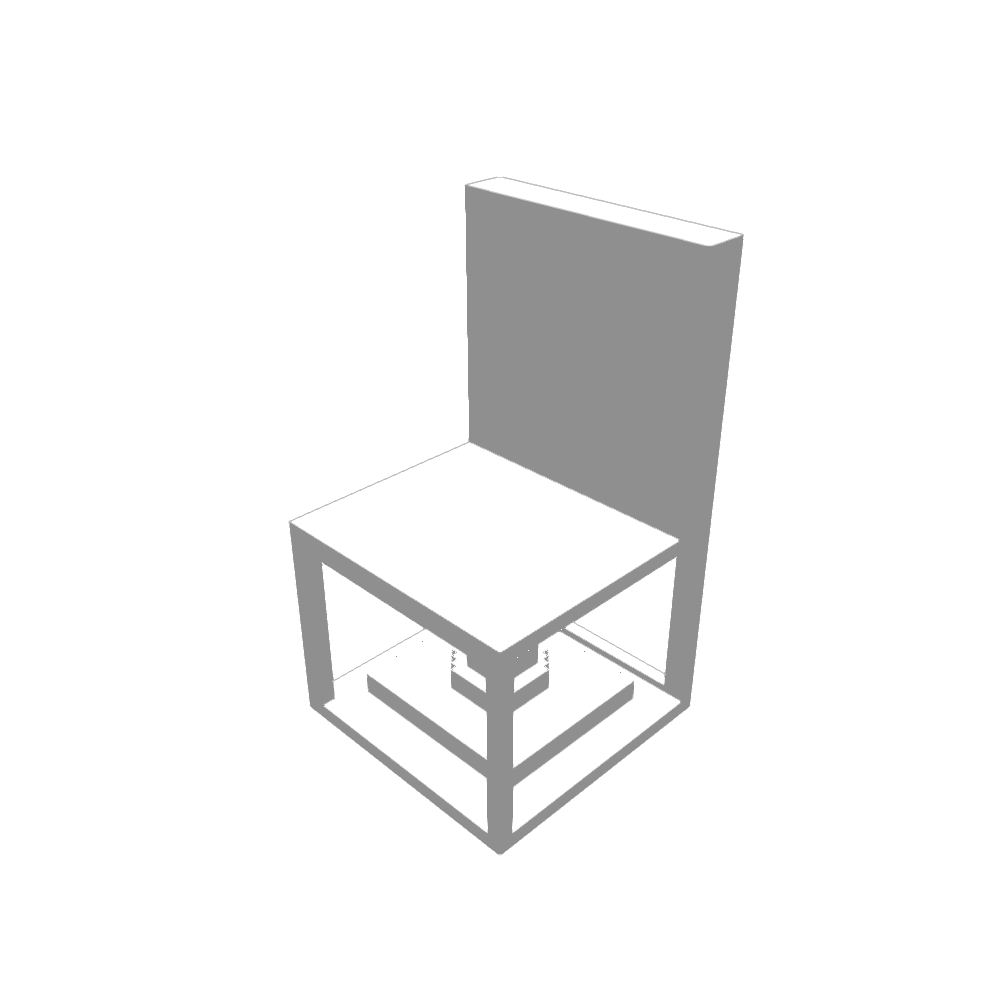} &
\includegraphics[trim=50 100 50 100,clip]{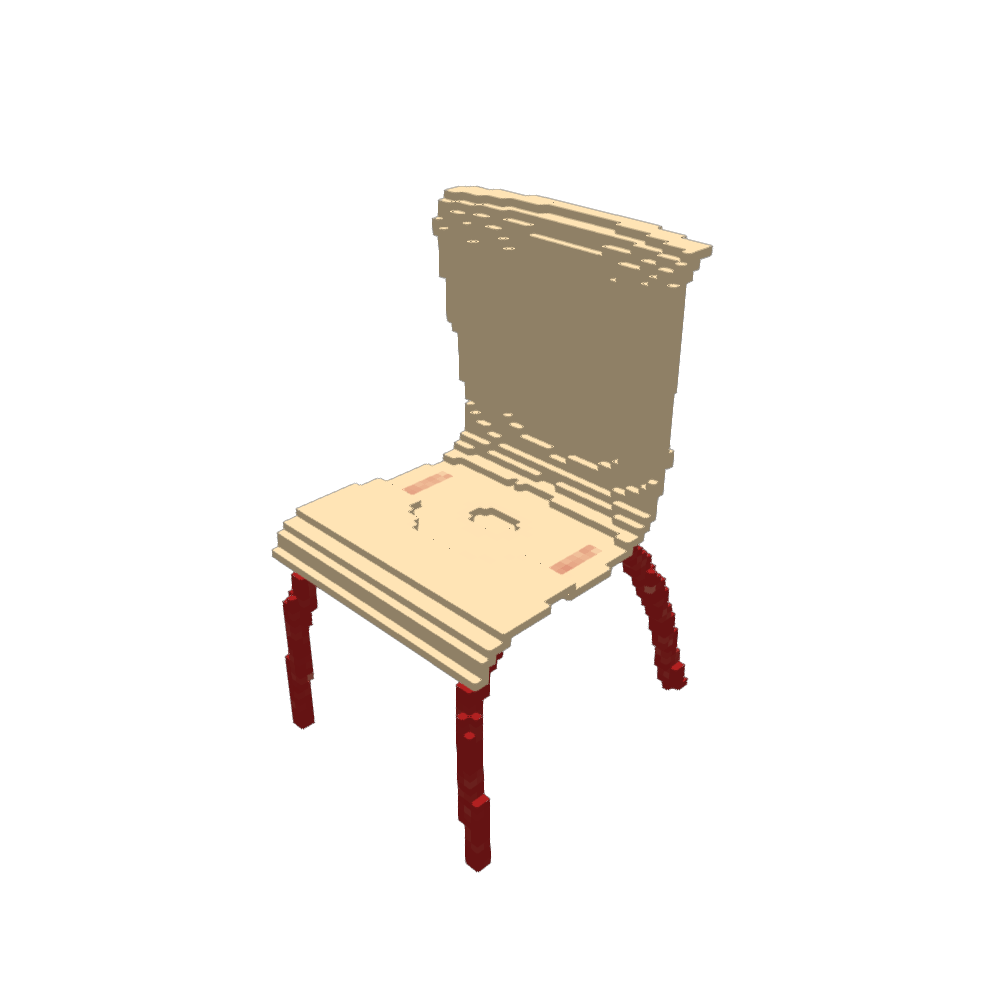} \\
\bottomrule
\end{tabularx}
\caption{
Retrieved shapes from test set using \bimodi, \bimodv, and \trimodiv for custom sentences.
Note all models retrieve shapes that match the color (\textit{dark brown}), materials (\textit{wooden}, \textit{glass}), shape (\textit{circular}), and the presence and absence of chair arms.
} 
\label{fig:retrieval-custom-visualization}
\end{figure}

\subsection{Retrieval on ShapeNet 13 categories}
\label{sec:supp:exp-c13}

\begin{figure}
\centering 
\includegraphics[trim=100 82 100 75,clip,width=\linewidth]{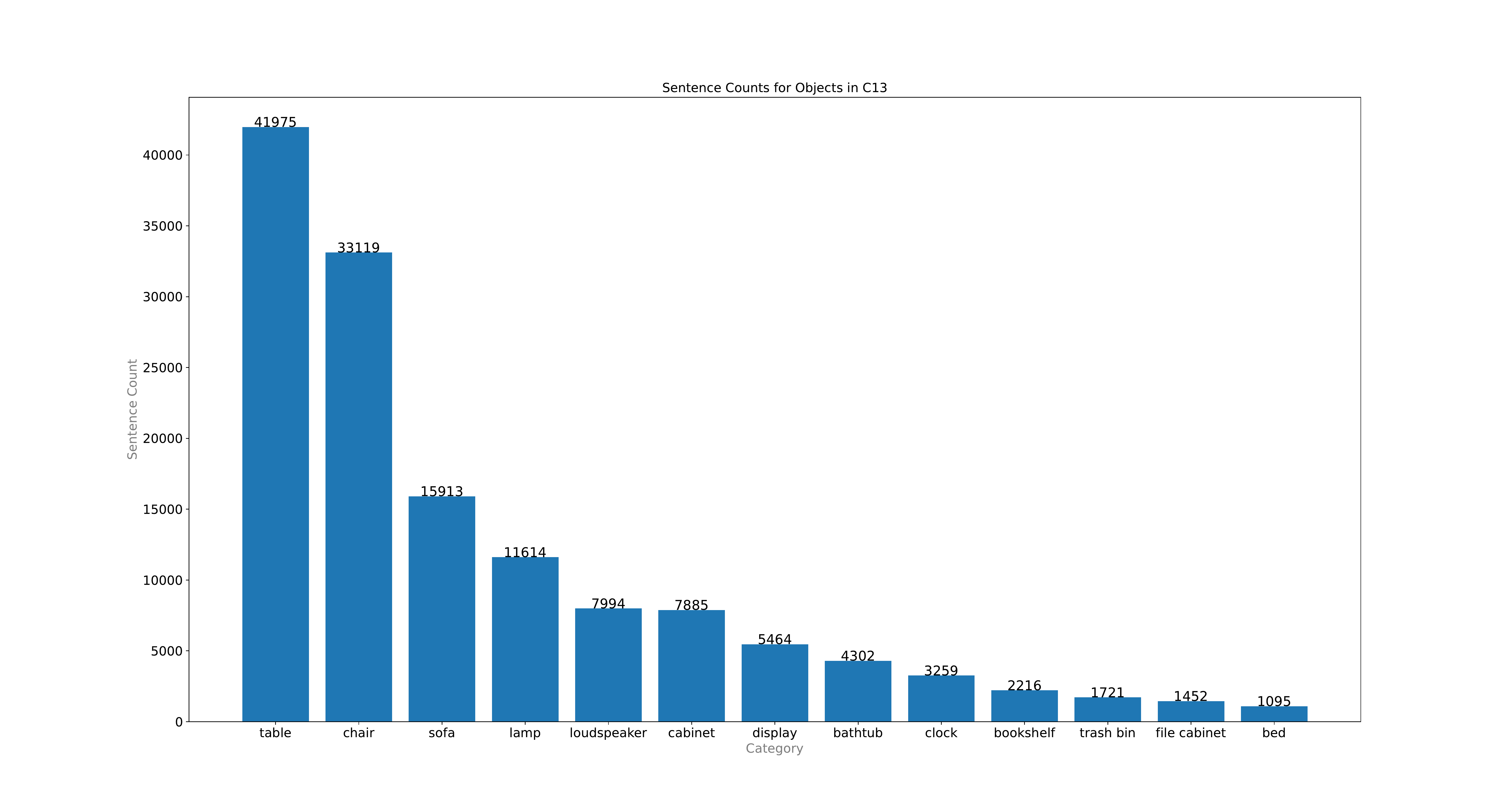}
\vspace{-8pt}
\caption{Histogram of number of sentences for the 13 different object categories ShapeNetCore.  We collected additional sentences for shapes other than `chairs and table' to investigate the performance of our models on a more diverse object dataset.}
\label{fig:c13-hist} 
\end{figure}
To show the effectiveness of our method for retrieval beyond `chairs and tables', we also collected a set of descriptions for $11$ additional categories of objects from ShapeNet~\cite{chang2015shapenet}.  Using a similar setting as \citet{chen2018text2shape}, we asked Amazon Mechanical Turk workers to provide descriptions for a random set of $5$ objects.  We exclusively engaged high-quality workers (with acceptance rate of $>95\%$ on more than 200 HITs) from countries that have English as a native language (US, Canada, Great Britain and Australia).  We collected up to $5$ descriptions per object.  This together with the original `chairs and tables' dataset, results in a dataset with over 138K descriptions for $27,510$ objects across $13$ object categories (see \Cref{fig:c13-hist} for distribution of sentences for each object category).  

\begin{table}
\centering
\resizebox{\linewidth}{!}
{
\begin{tabular}{@{}rrrrrrr@{}}
\toprule
& Text & Image & Voxels & RR@1 & RR@5 & NDCG@5\\
\midrule
ZS$^*$ & CLIP & CLIP & - & 6.81 & 16.97 & 11.97 \\
\midrule
\bimodi & CLIP & CLIP & - & 5.78 & 20.17 & 13.03   \\
\midrule
\bimodi & GRU & MVCNN & - & 8.44 & 25.64 & 17.15 \\
\bimodv & GRU & - & 3D-CNN  & 8.87 & 27.91 & 18.52 \\
\trimodiv & GRU & MVCNN & 3D-CNN & \textbf{10.63} & \textbf{31.01} & \textbf{21.03} \\
\bottomrule
\end{tabular}
}
\vspace{-6pt}
\caption{Shape retrieval results on ShapeNet c13 val set.  We observe a similar trend as on the `chairs and tables' dataset.}
\label{tab:retrieval-c13}
\end{table}

We use the same standard settings as in our Text2Shape experiments with with 6 multi-view images, resolution $128^2$ and $64^3$ for images and voxels respectively and a batch size of 128. 
Experiments on this ShapeNet C13 dataset show a similar trend as for the `chairs and table' dataset, with \trimodiv outperforming \bimodi and \bimodv (see \Cref{tab:retrieval-c13}).  Comparison with zero-shot CLIP (ZS$^*$) shows that the pretrained CLIP model is able to retrieve some relevant shapes in a zero-shot setting on this broader set of shapes, but has lower performance. Our \bimodi with GRU and MVCNN trained from scratch is able to outperform the bimodal models with CLIP text and image encoder trained on the data, but still underperforms \trimodiv, showing the value of trimodal embedding. With more categories and less training data for ShapeNet C13 compared to the `chairs and tables' dataset, the pretrained CLIP embeddings is more effective on ShapeNet C13.

\section{Qualitative results and discussion}
\label{sec:supp:qual}

We provide visualizations of the embedding space (\cref{sec:supp:tsne}), types of different errors(\cref{sec:supp:error-examples}), additional qualitative examples of text-to-shape retrieval results (\cref{sec:supp:result-visuals}) and a discussion of the shape similarity metrics (\cref{sec:supp:shape-sim-discussion}).

\begin{figure*}
\centering 
\includegraphics[width=\textwidth]{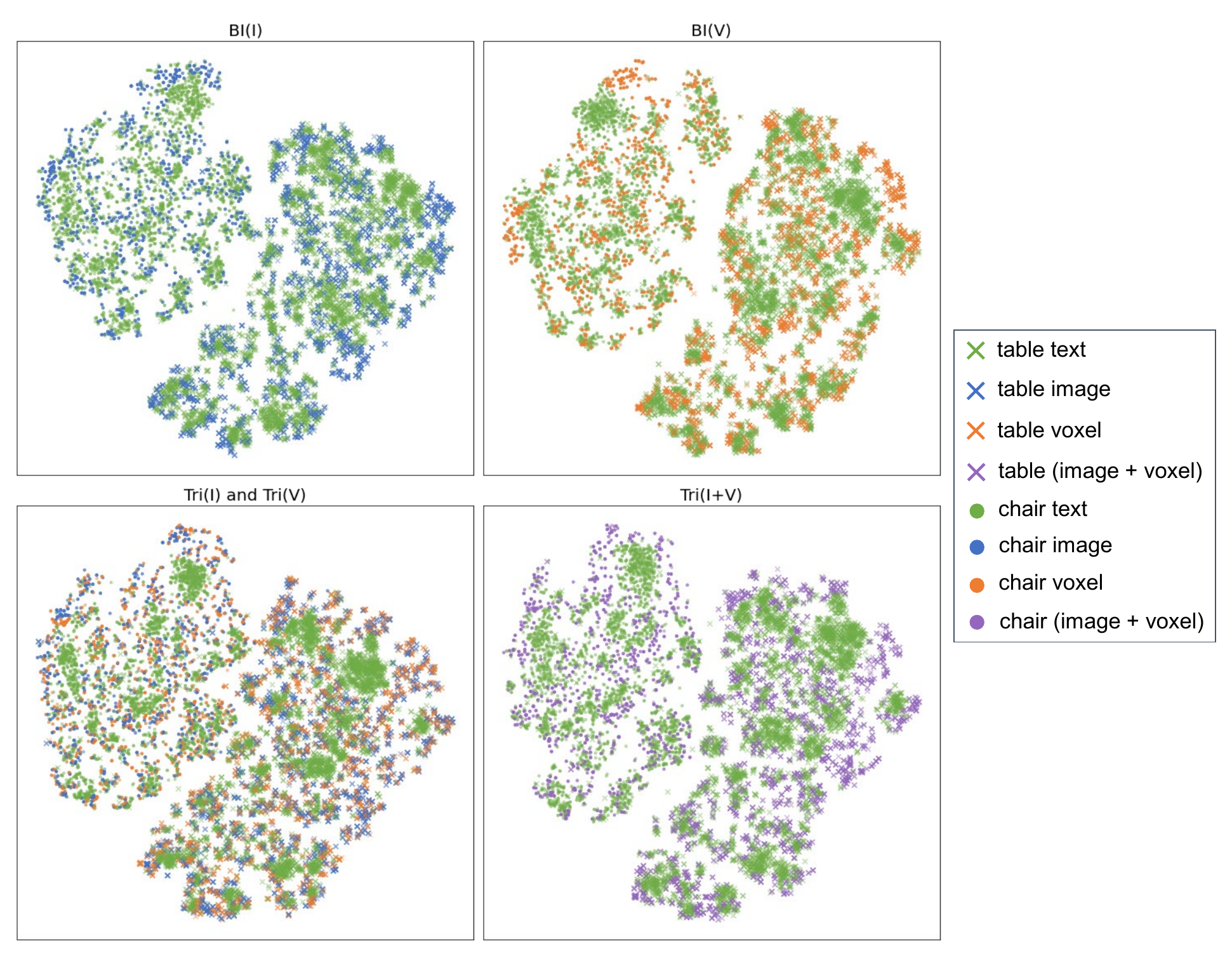}
\vspace{-6pt}
\caption{Joint embedding spaces for different models with t-SNE.  The modalities are indicated by color (\textcolor{text}{green} for text, \textcolor{image}{blue} for image, \textcolor{voxel}{orange} for voxel, and \textcolor{imagevoxel}{purple} for image + voxel). Tables are indicated by $\times$ and chairs by {\large{$\bullet$}}.  By training with the NT-XEnt contrastive loss, we are able to push the different modalities together and separate embeddings for tables (lower right) and chairs (upper left).}
\label{fig:tsne-embedding-space-for-models} 
\end{figure*}
\begin{table*}
\centering
\resizebox{\linewidth}{!}
{
\begin{tabular}{@{}ccccc|ccccc@{}}
\toprule
  & RR@1($\uparrow$) & RR@5($\uparrow$) & NDCG@5($\uparrow$) &  MRR($\uparrow$)& CD($\downarrow$) & NC($\uparrow$) & $F1^{0.1}$($\uparrow$) & $F1^{0.3}$($\uparrow$) & $F1^{0.5}$($\uparrow$) \\
\midrule
\bimodi & 11.61 $\pm$ 0.20 & 30.65 $\pm$ 0.19 & 21.36 $\pm$ 0.23 & 21.46 $\pm$ 0.25 & 2.01 $\pm$ 0.02 & 0.62 $\pm$ 0.002 & 11.97 $\pm$ 0.20 & 34.37 $\pm$ 0.31 & 48.89 $\pm$ 0.36 \\
\trimodi & 12.19 $\pm$ 0.45 & 32.33 $\pm$ 0.60 & 22.54 $\pm$ 0.54 & 22.62 $\pm$ 0.49 & 1.91 $\pm$ 0.02 & 0.63 $\pm$ 0.002 & 12.49 $\pm$ 0.21 & 35.56 $\pm$ 0.28 & 50.28 $\pm$ 0.29 \\
\bimodv & 9.59 $\pm$ 0.27 & 27.14 $\pm$ 0.48 & 18.54 $\pm$ 0.13 & 19.03 $\pm$ 0.08 & 1.96 $\pm$ 0.02 & 0.62 $\pm$ 0.002 & 12.21 $\pm$ 0.09 & 35.01 $\pm$ 0.18 & 49.60 $\pm$ 0.24  \\
\trimodv & 9.83 $\pm$ 0.21 & 27.75 $\pm$ 0.35 & 18.97 $\pm$ 0.21 & 19.32 $\pm$ 0.20 & 1.89 $\pm$ 0.03 & 0.63 $\pm$ 0.002 & 12.64 $\pm$ 0.13 & 35.75 $\pm$ 0.30 & 50.44 $\pm$ 0.37 \\
\trimodiv & \textbf{12.52 $\pm$ 0.28} & \textbf{32.67 $\pm$ 0.61} & \textbf{22.87 $\pm$ 0.46} & \textbf{22.68 $\pm$ 0.32} & \textbf{1.88 $\pm$ 0.02} & \textbf{0.63 $\pm$ 0.001} & \textbf{12.85 $\pm$ 0.17} & \textbf{36.02 $\pm$ 0.32} & \textbf{50.70 $\pm$ 0.35} \\
\bottomrule
\end{tabular}
}
\vspace{-6pt}
\caption{Comparison of bimodal and trimodal models for text-to-shape retrieval on the validation set. Models are trained with a batch size of 128, solid color voxels at a resolution of $64^3$, 6 multi-view images at a resolution of $128^2$ each. Having a trimodal embedding (\trimodi,\trimodv) gives better performance than the bimodal embeddings (\bimodi,\bimodv).  By summing the image and voxel representations from the trimodal embeddings (\trimodiv), we can further improve retrieval performance.}
\label{tab:retrieval-modality-shapenet-full}
\end{table*}

\subsection{Visualization for the embedding space}
\label{sec:supp:tsne}

We visualize the joint embedding spaces for different models by projecting our embeddings to 2D using t-distributed stochastic neighbor embedding (t-SNE)~\cite{van2008visualizing}.
The embedding spaces for the bimodal and trimodal models are shown in  \cref{fig:tsne-embedding-space-for-models}. We also take a closer look at the \trimodiv embedding space in \cref{fig:tsne-with-images} and show that the embeddings of similar shapes cluster together.

\subsection{Examples of error types}
\label{sec:supp:error-examples}

\cref{fig:retrieval-error-example} shows examples of the types of errors we analyzed in the main paper.
These failure cases demonstrate potential directions for future work.

\subsection{Additional visualizations of retrieval results}
\label{sec:supp:result-visuals}

\begin{figure*}%
\centering
\setkeys{Gin}{width=0.8\linewidth}
\setlength\tabcolsep{1.5pt}
\begin{tabularx}{\linewidth}{@{}p{1.5cm}YYYYY@{}}
\toprule
4 &\multicolumn{4}{p{12.5cm}}{Four black legged blue coloured rectangle shaped tennis table} &
\includegraphics[trim=50 200 50 285,clip]{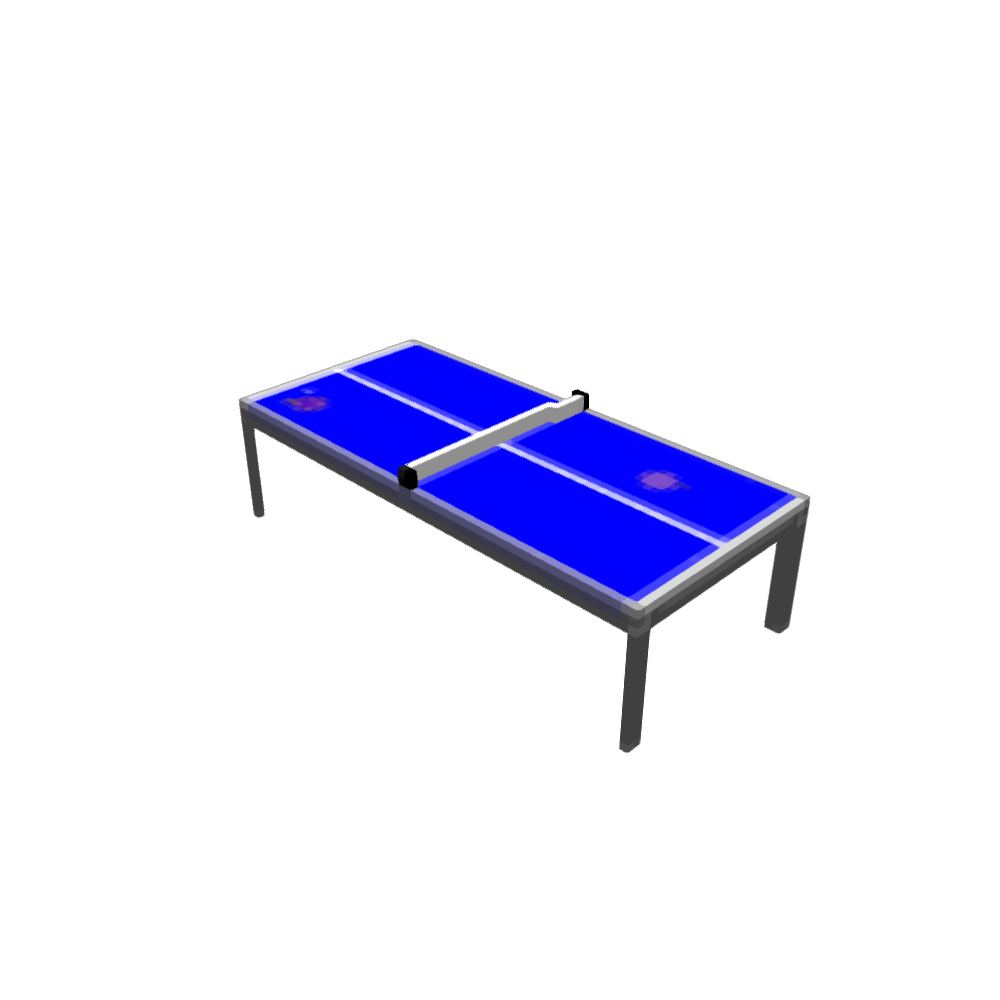} \\
[-0.3cm]
\bimodi & 
\includegraphics[trim=50 200 50 250,clip]{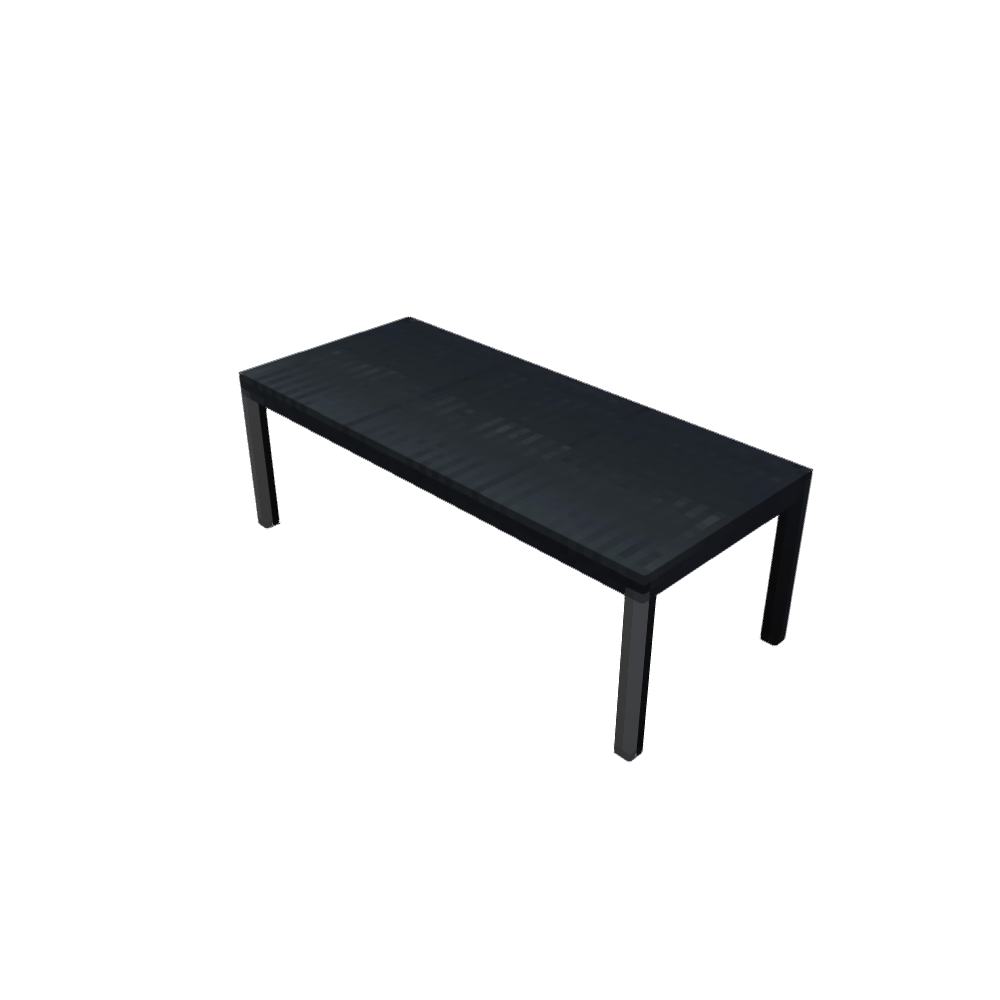} & 
\includegraphics[trim=50 200 50 250,clip]{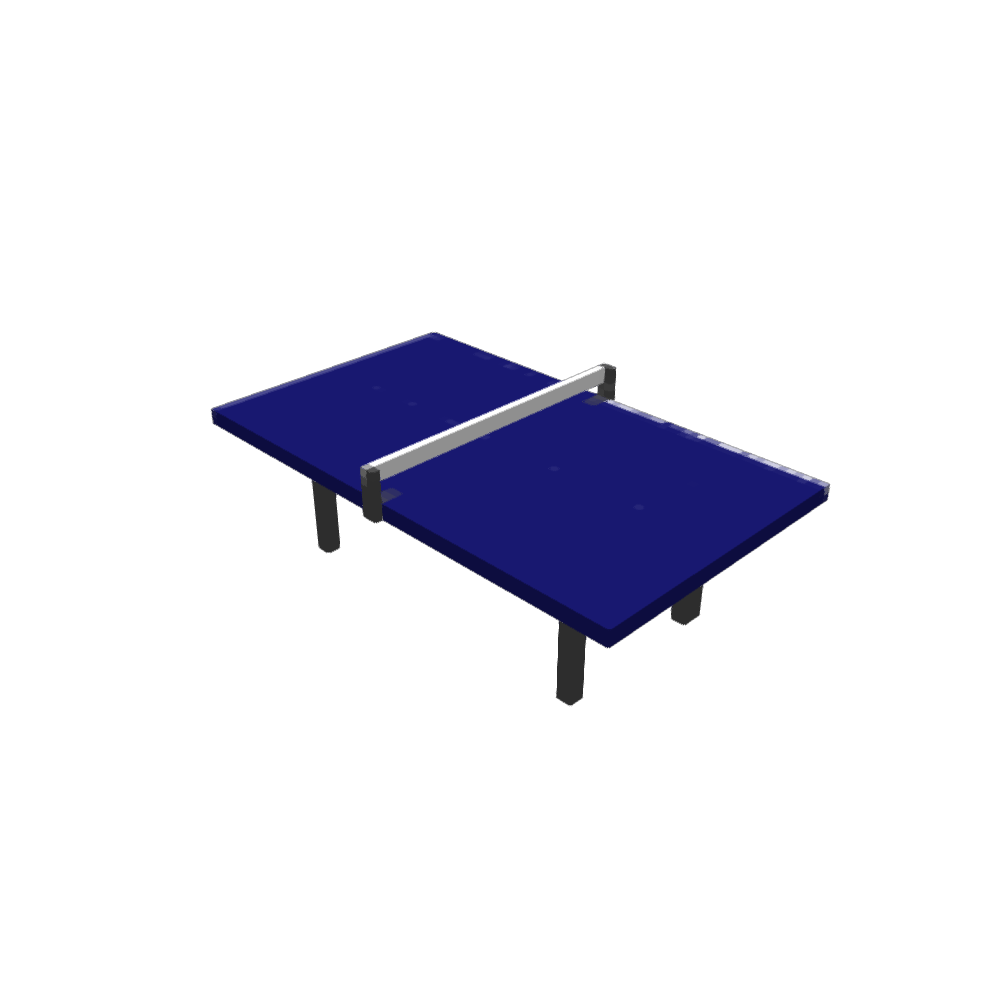} &
\includegraphics[trim=50 200 50 250,clip]{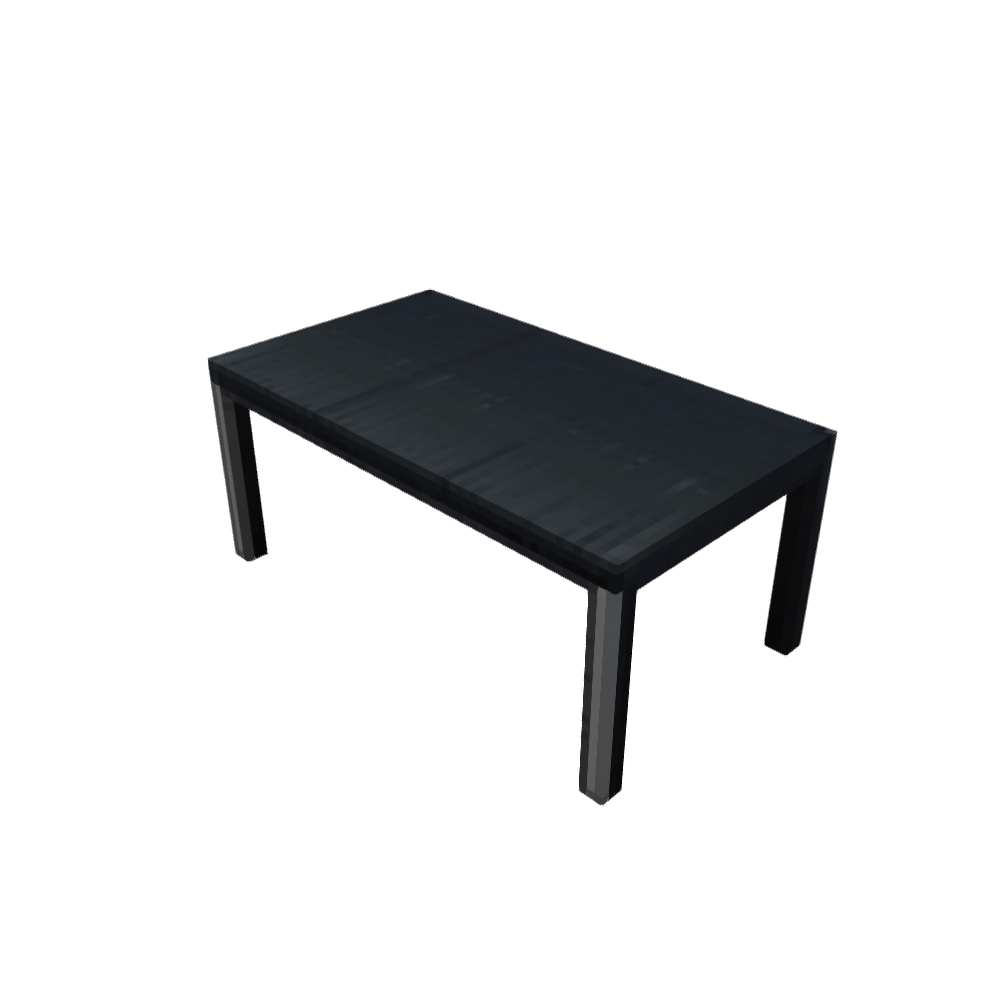} &
\includegraphics[trim=50 200 50 250,clip]{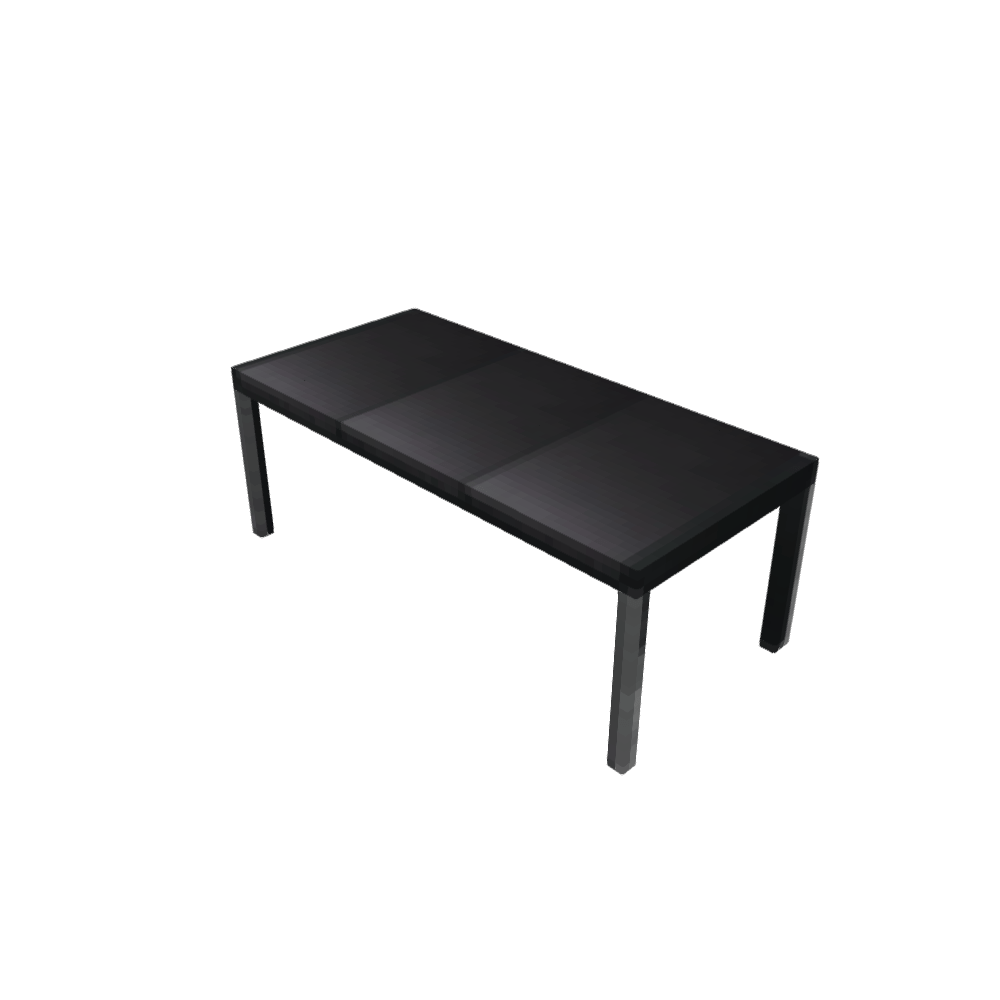} &
\includegraphics[trim=50 200 50 250,clip]{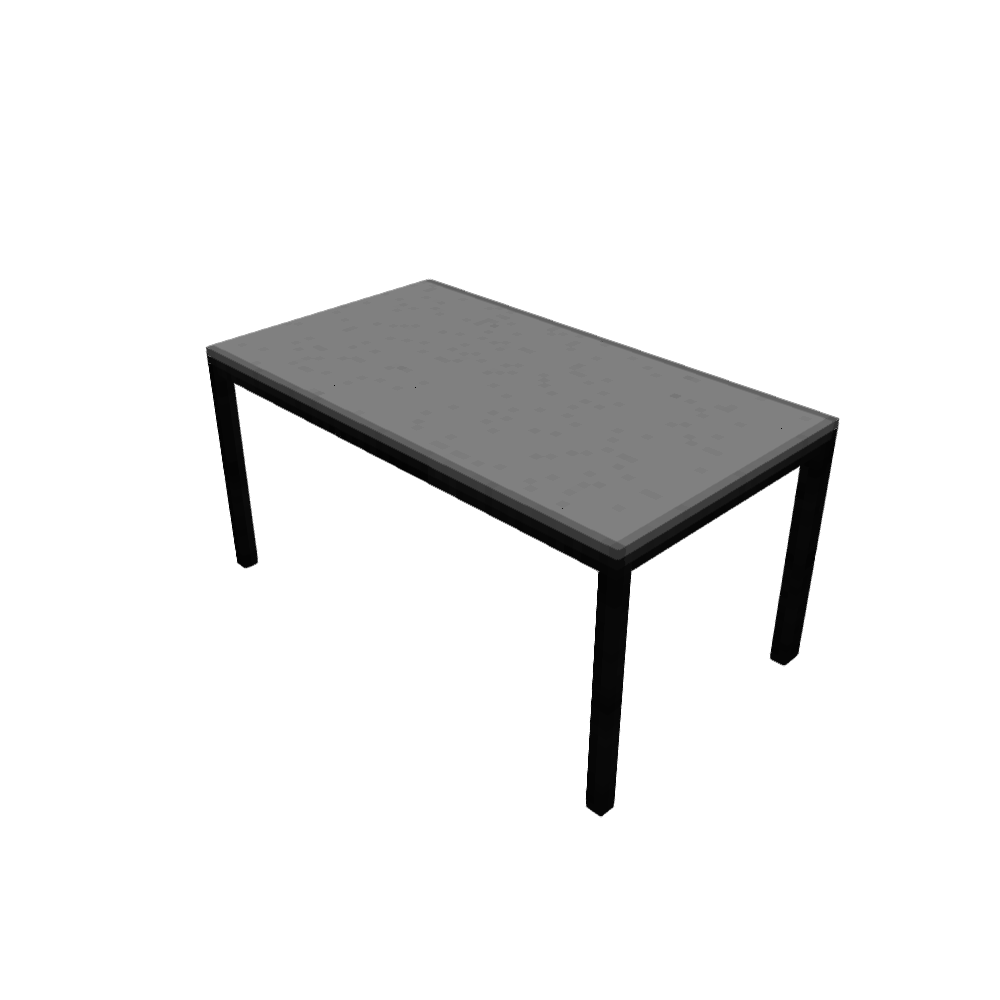} \\
[-0.3cm]
\bimodv & 
\includegraphics[trim=50 200 50 250,clip]{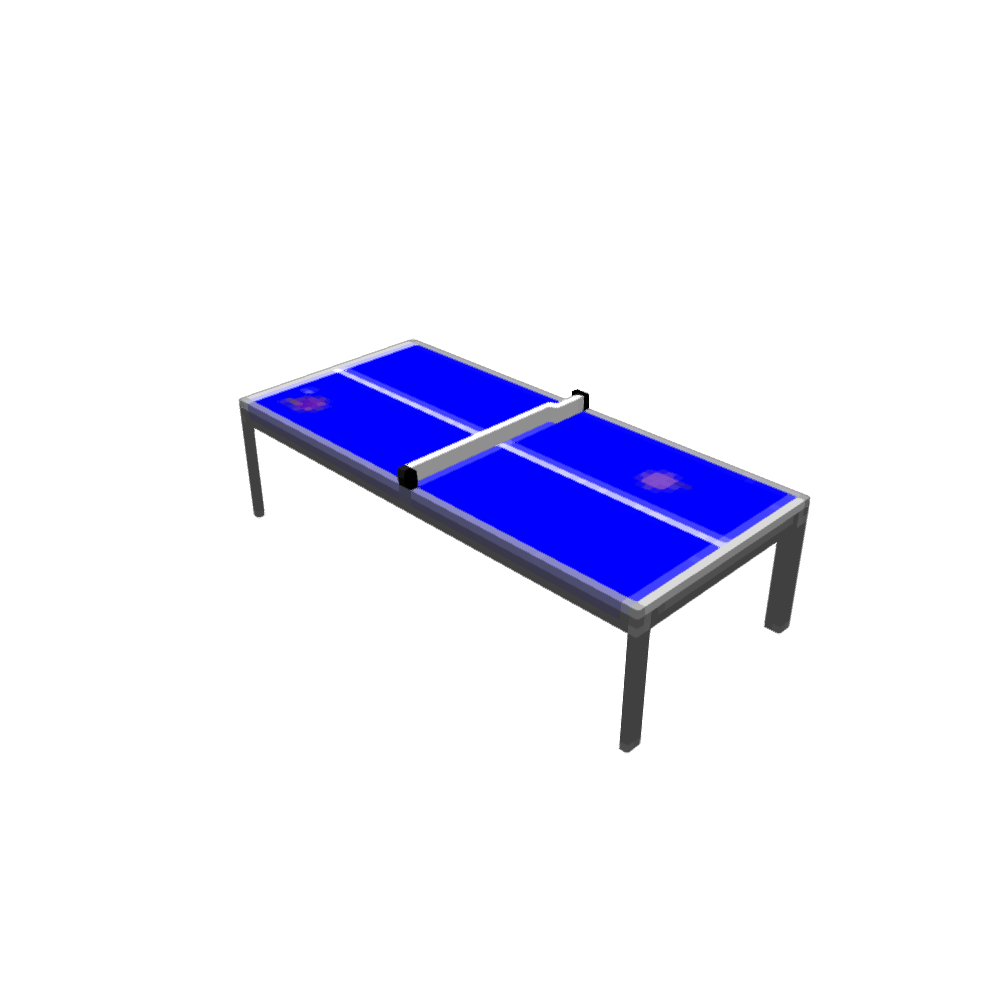} & 
\includegraphics[trim=50 200 50 250,clip]{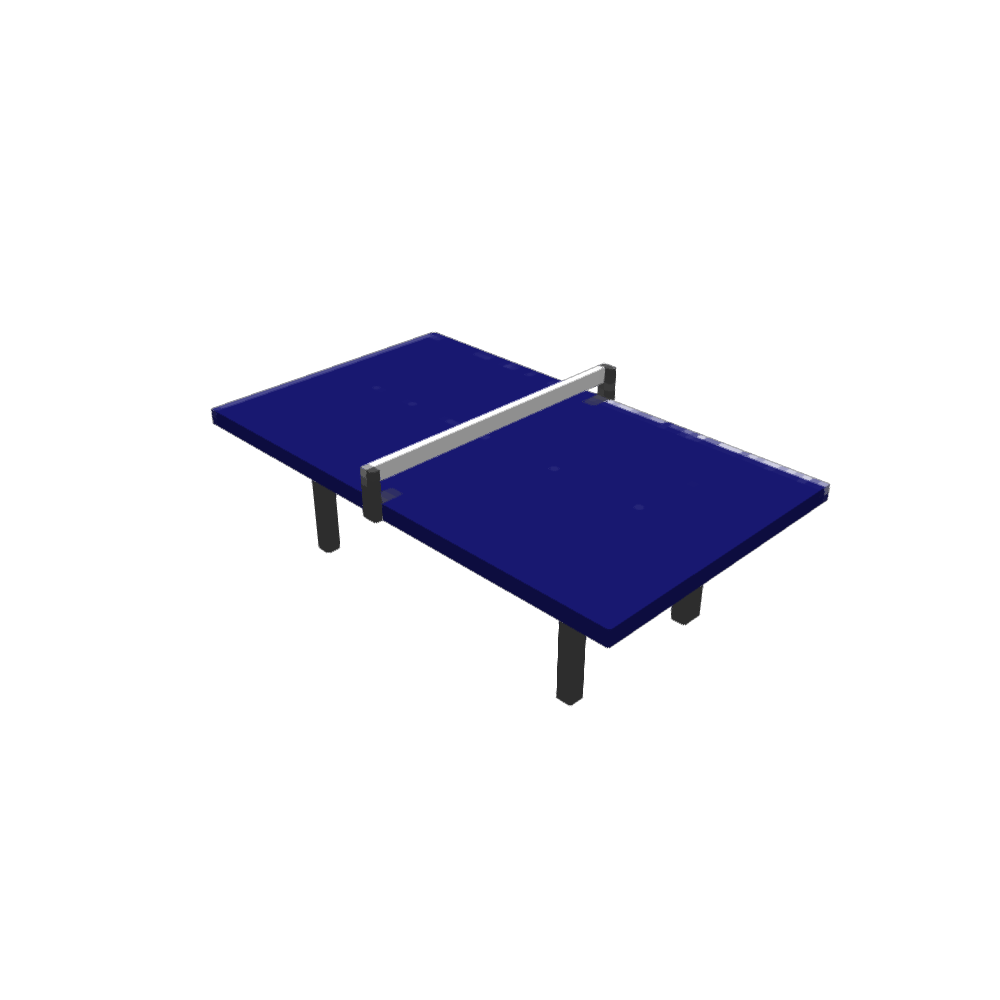} & 
\includegraphics[trim=50 200 50 250,clip]{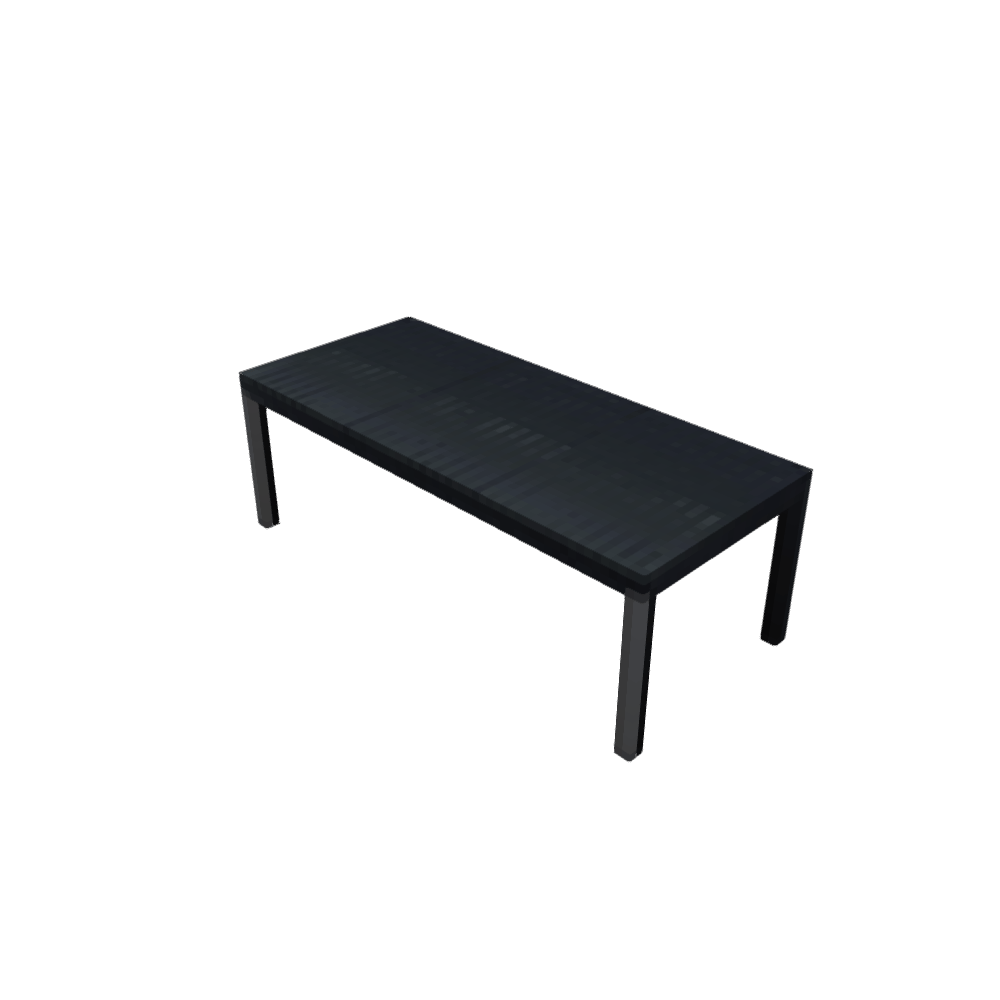} &
\includegraphics[trim=50 200 50 250,clip]{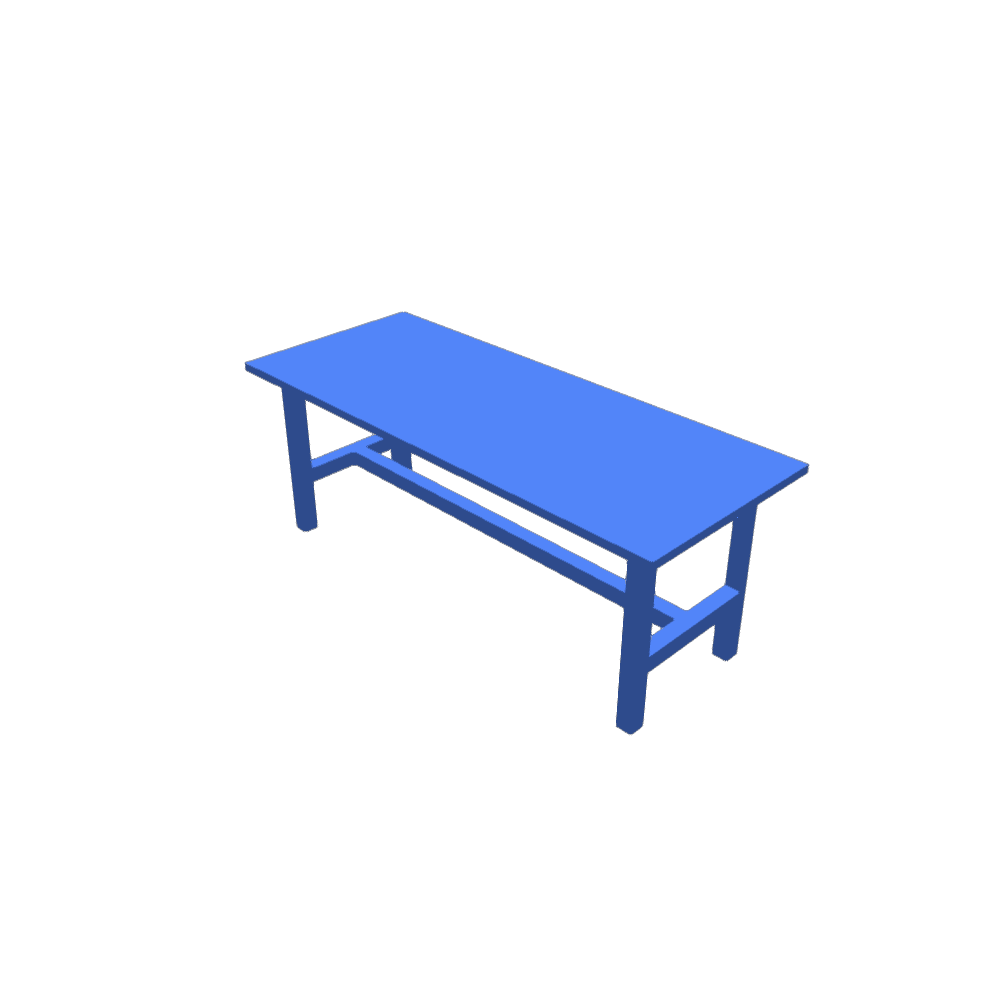} & 
\includegraphics[trim=50 200 50 250,clip]{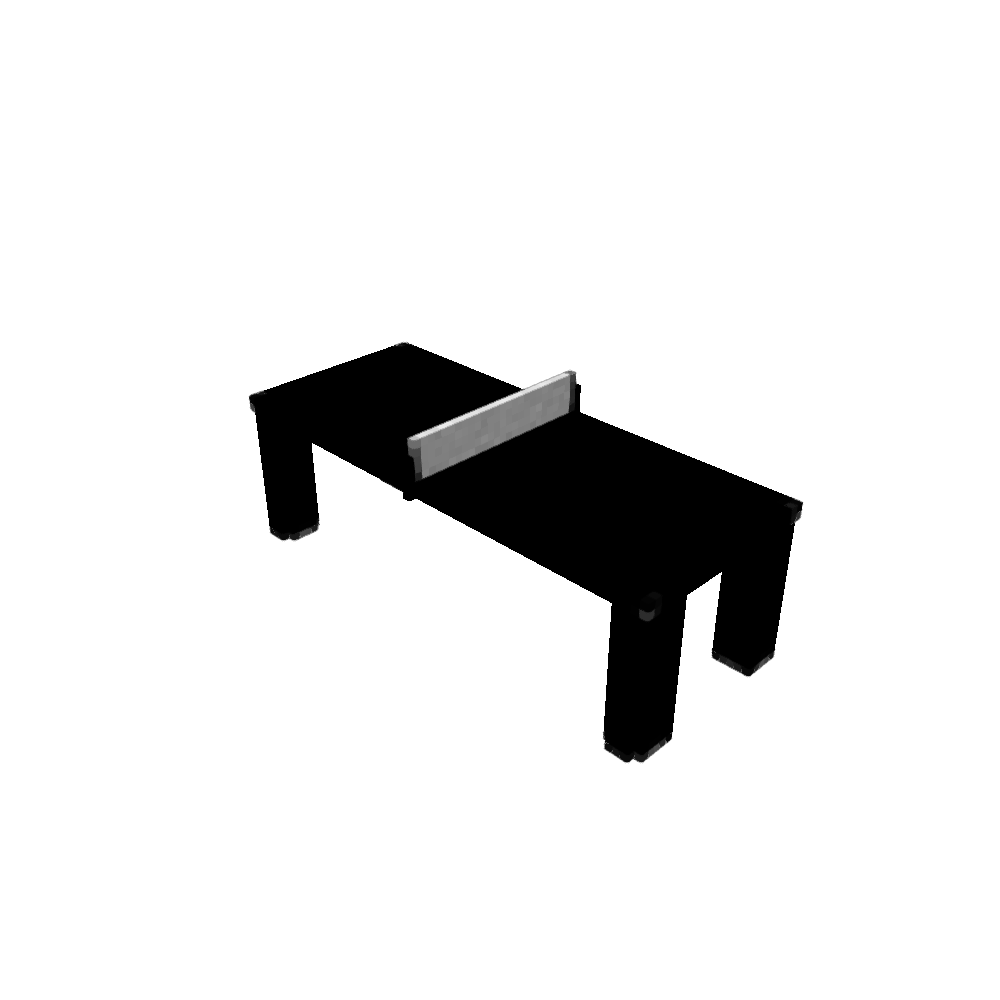} \\
[-0.3cm]
\trimodiv & 
\includegraphics[trim=50 200 50 250,clip]{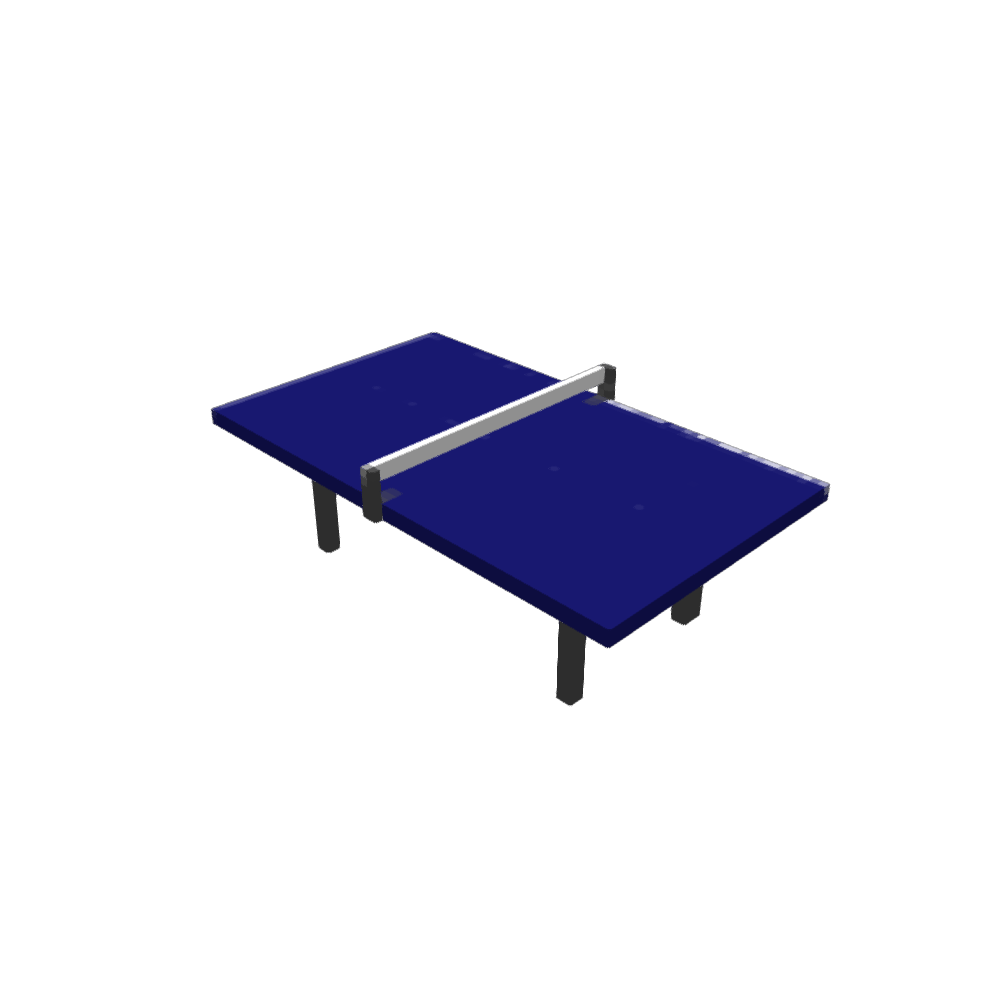} &
\includegraphics[trim=50 200 50 250,clip]{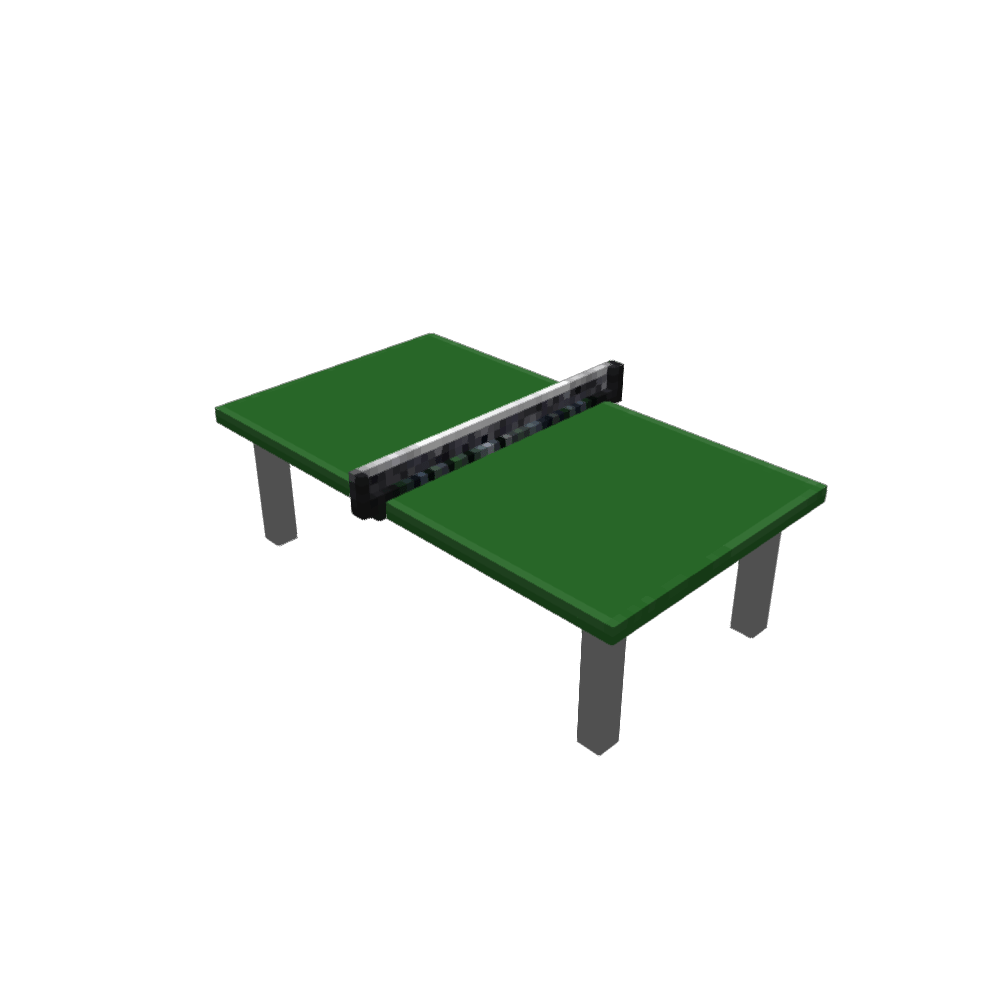} & 
\includegraphics[trim=50 200 50 250,clip]{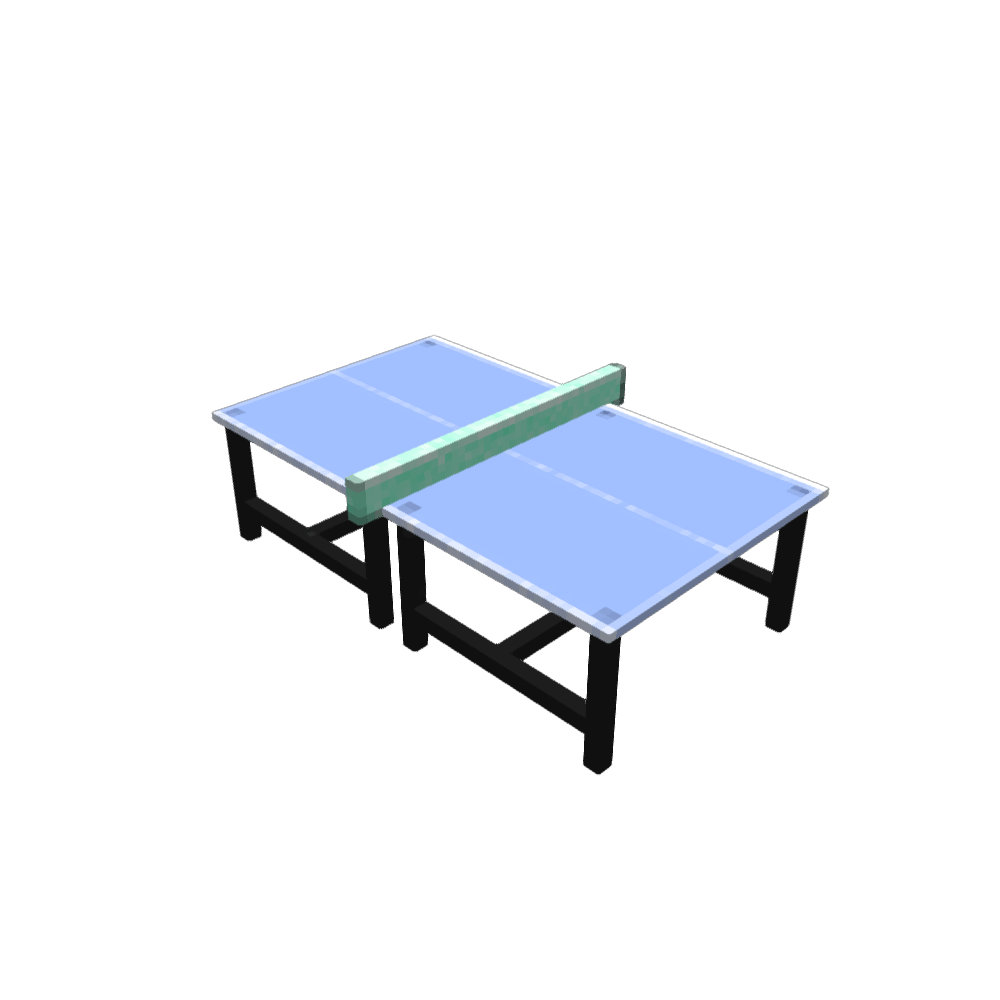} & 
\includegraphics[trim=50 200 50 250,clip]{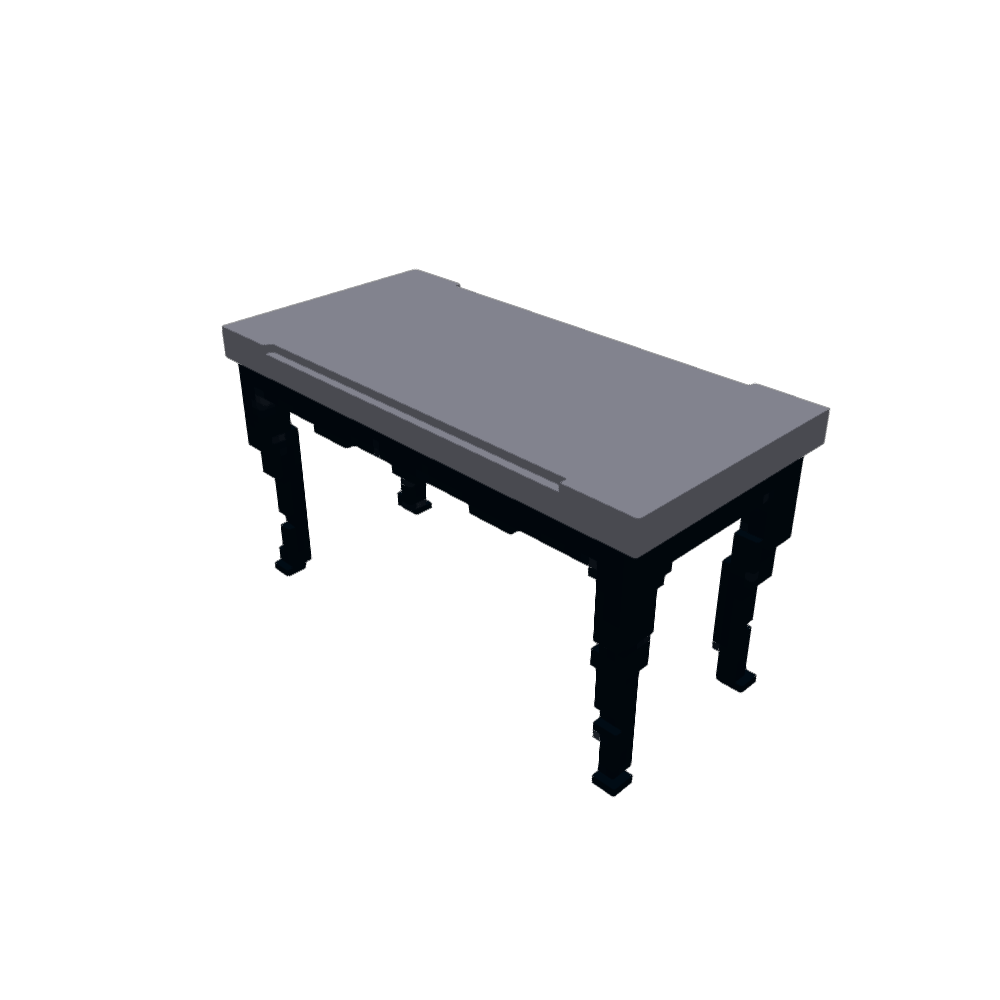} &
\includegraphics[trim=50 200 50 250,clip]{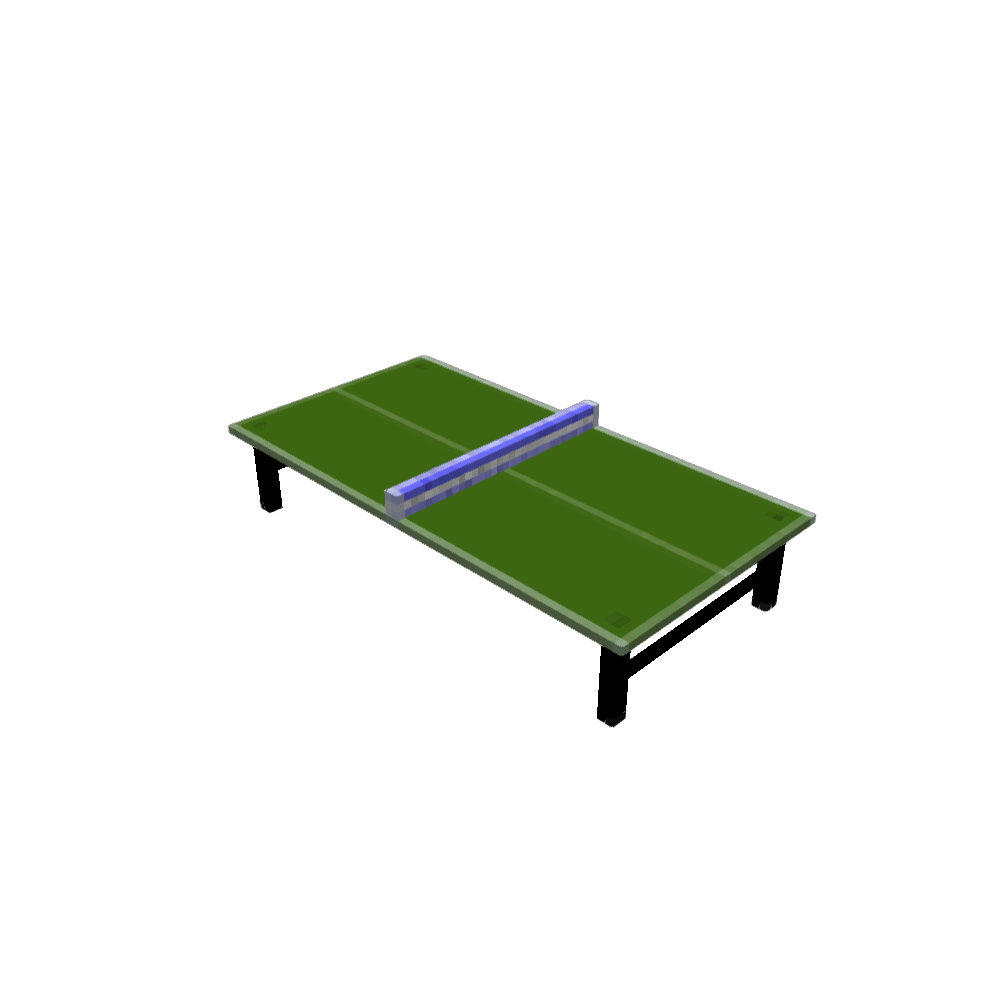} \\
[-0.1cm]
\midrule
5 &\multicolumn{4}{p{12.5cm}}{It looks like one you would use at a picnic.  It is wooden and has bench seating.} &
\includegraphics[trim=50 200 50 260,clip]{figures/comparison/groundtruth/e0eb9f5421ef2c4584904c716bc3b619.png} \\
[-0.2cm]
\bimodi & 
\includegraphics[trim=50 40 50 150,clip]{figures/comparison/i128b128/picnic-table/668aa5d430fd6d4e8f7d9678498f2295.png} & 
\includegraphics[trim=50 40 50 150,clip]{figures/comparison/i128b128/picnic-table/bffe3e68857faf7f4d1242a685303c47.png} &
\includegraphics[trim=50 40 50 150,clip]{figures/comparison/i128b128/picnic-table/68b26c9353e65419c3e46f55b34610cf.png} & 
\includegraphics[trim=50 40 50 150,clip]{figures/comparison/i128b128/picnic-table/8914307dbc15a20387f0bafef1e13471.png} & 
\includegraphics[trim=50 40 50 150,clip]{figures/comparison/i128b128/picnic-table/81628a0b5f7f9ad7ba94feecf6f7a200.png}  \\
[-0.35cm]
\bimodv & 
\includegraphics[trim=50 200 50 250,clip]{figures/comparison/v64b128/picnic-table/93078952823dddaa5e56625f6688e473.png} & 
\includegraphics[trim=50 200 50 250,clip]{figures/comparison/v64b128/picnic-table/e0eb9f5421ef2c4584904c716bc3b619.png} &
\includegraphics[trim=50 200 50 250,clip]{figures/comparison/v64b128/picnic-table/8b7f2caf571342398b8e4fade0702996.png} & 
\includegraphics[trim=50 200 50 250,clip]{figures/comparison/v64b128/picnic-table/a5d5fc6b0bb7881419fb4103277a6b93.png} &
\includegraphics[trim=50 200 50 250,clip]{figures/comparison/v64b128/picnic-table/3838913e27df8fe5287005440c82669a.png} \\
[-0.35cm]
\trimodiv & 
\includegraphics[trim=50 200 50 250,clip]{figures/comparison/v64i128b128/picnic-table/62f75f68a559cd9c5edbe4a62f5ee3a4.png} & 
\includegraphics[trim=50 200 50 250,clip]{figures/comparison/v64i128b128/picnic-table/e0eb9f5421ef2c4584904c716bc3b619.png} & 
\includegraphics[trim=50 200 50 250,clip]{figures/comparison/v64i128b128/picnic-table/3c514c4f53e3f1ed4b3c42e318f3affc.png} & 
\includegraphics[trim=50 200 50 250,clip]{figures/comparison/v64i128b128/picnic-table/f078a5940280c0a22c6c98851414a9d8.png} &
\includegraphics[trim=50 200 50 250,clip]{figures/comparison/v64i128b128/picnic-table/a24af284041888bd5f05ba8053abf6cf.png} \\
[-0.1cm]
\midrule
6 &\multicolumn{4}{p{12.5cm}}{Standard wooden small side table with a single drawer and bottom shelf} &
\includegraphics[trim=15 100 15 130,clip]{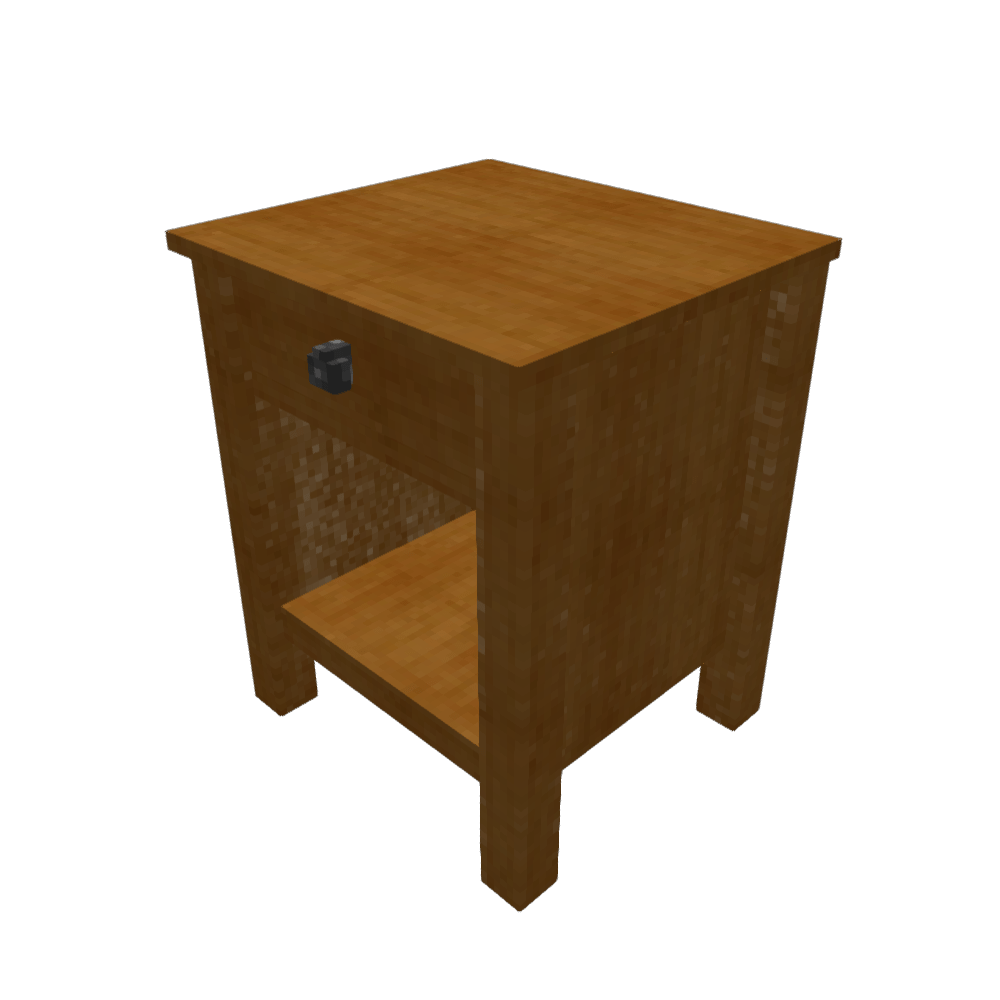} \\
[-0.1cm]
\bimodi & 
\includegraphics[trim=15 100 15 120,clip]{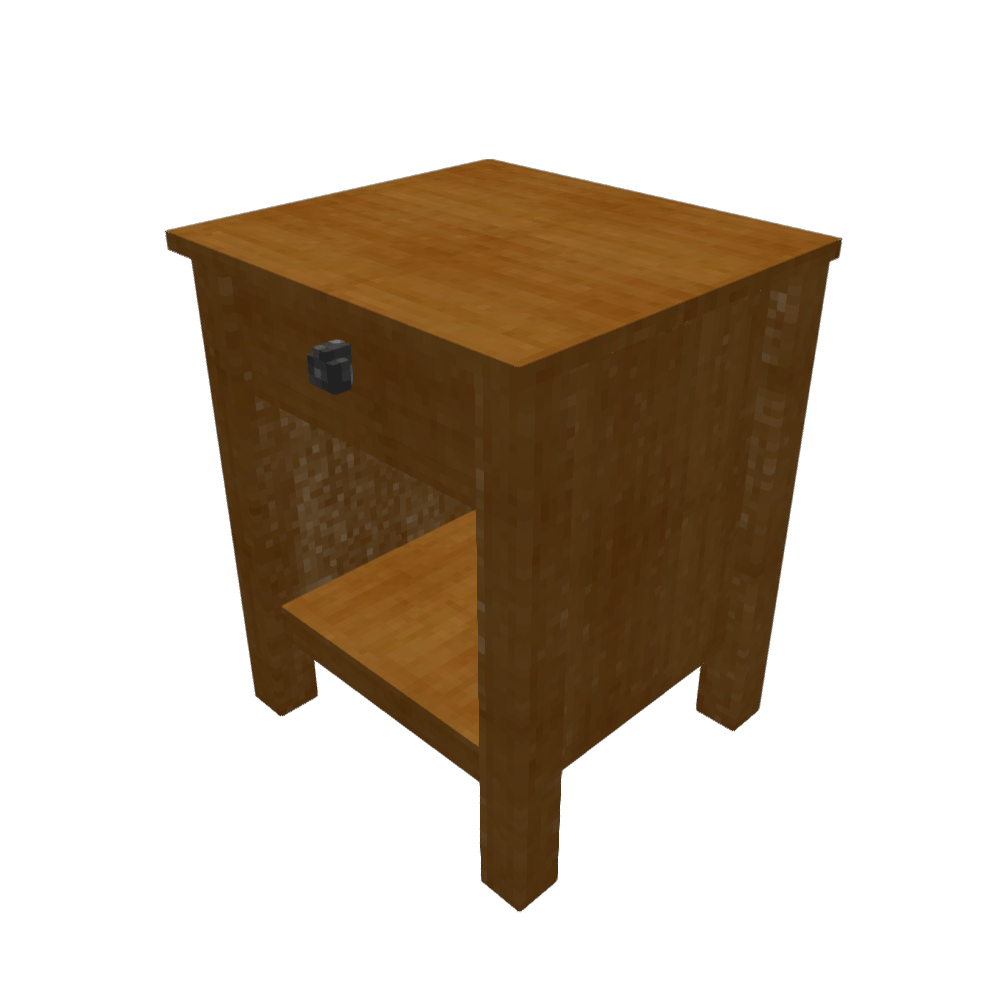} & 
\includegraphics[trim=15 100 15 120,clip]{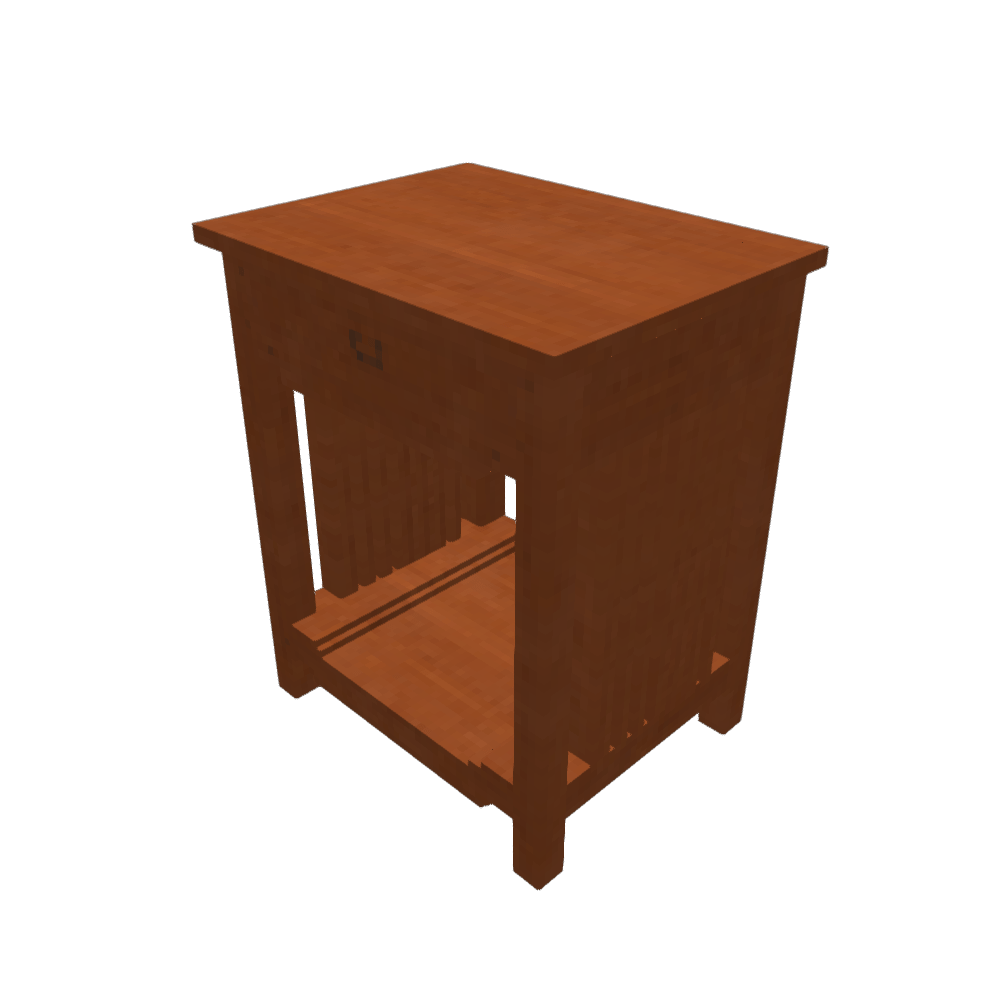} &
\includegraphics[trim=15 100 15 120,clip]{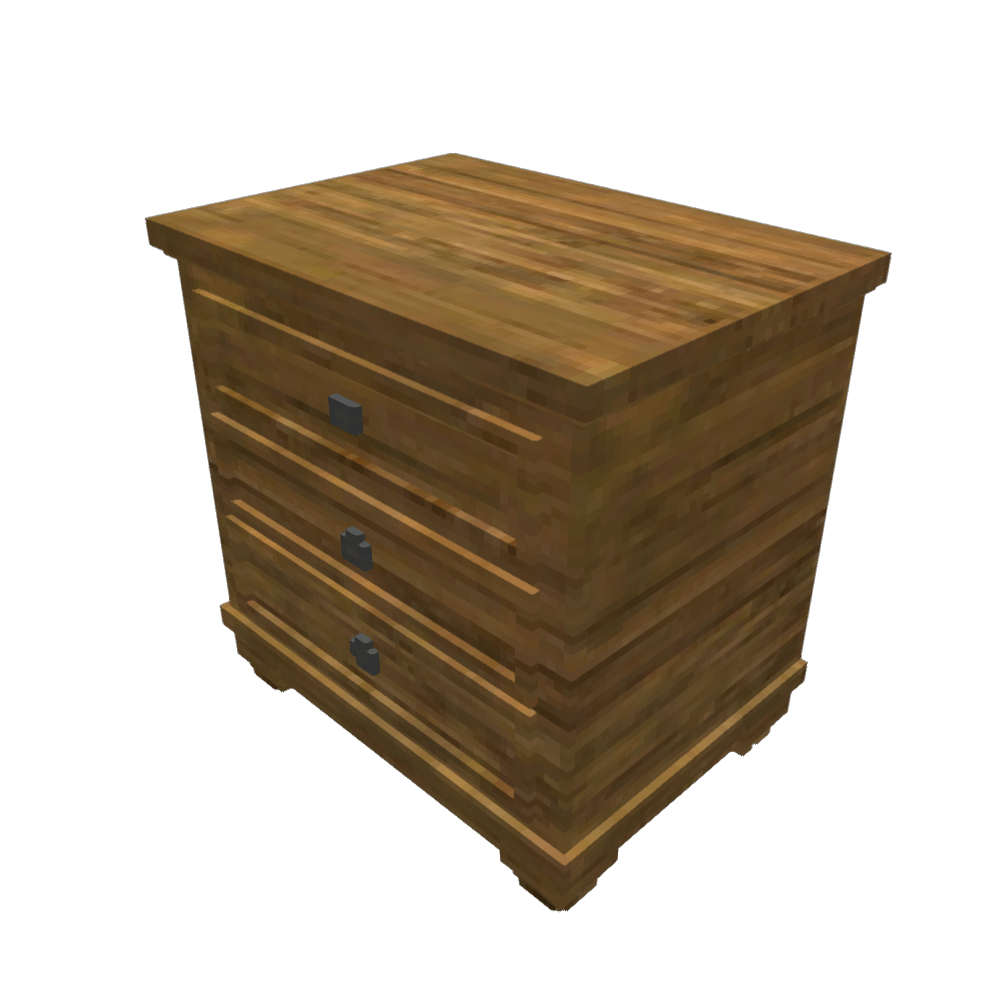} &
\includegraphics[trim=15 100 15 120,clip]{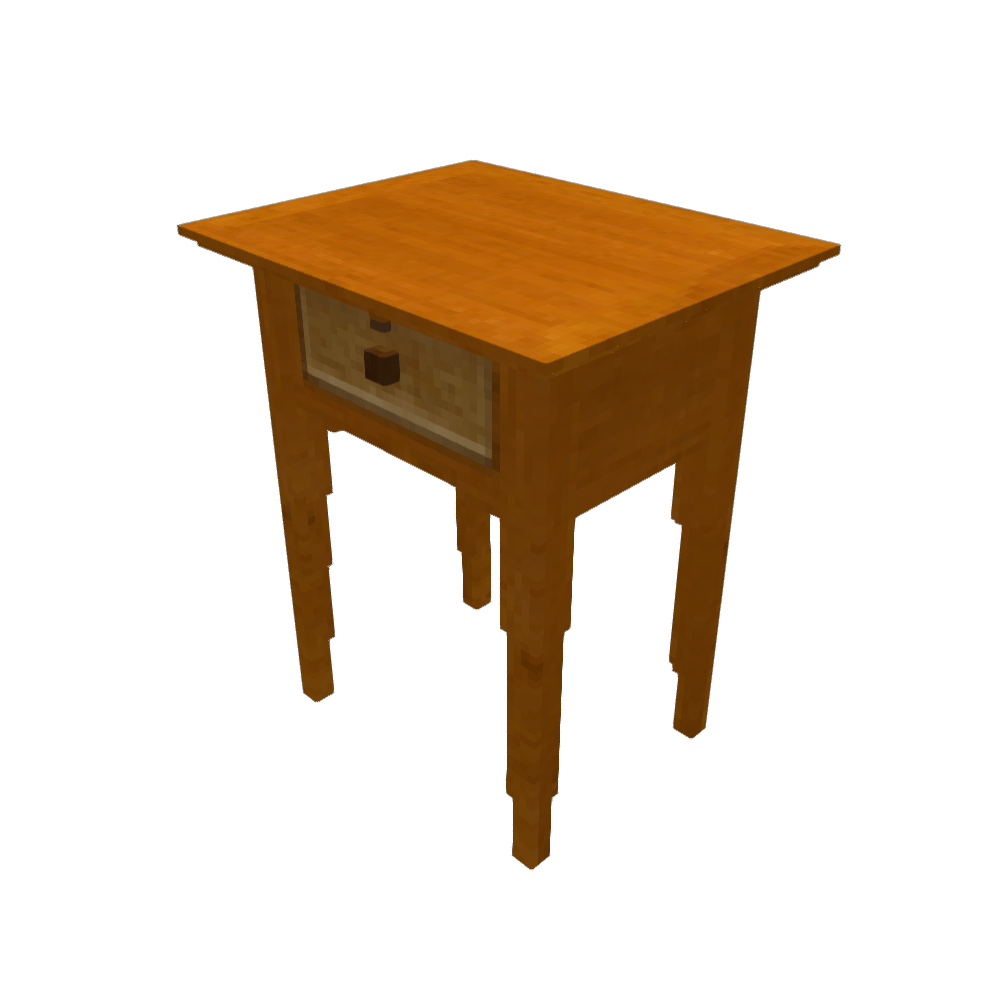} & 
\includegraphics[trim=15 100 15 120,clip]{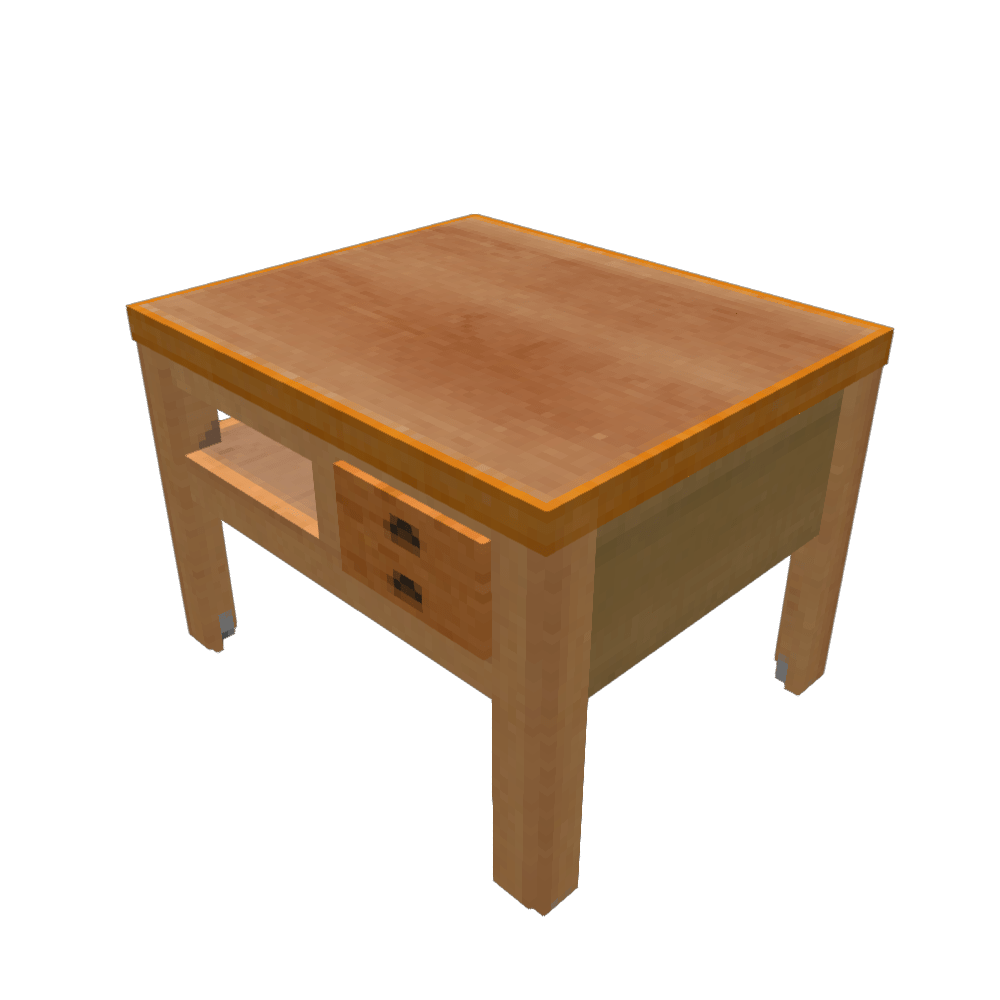} \\
[-0.1cm]
\bimodv & 
\includegraphics[trim=15 100 15 120,clip]{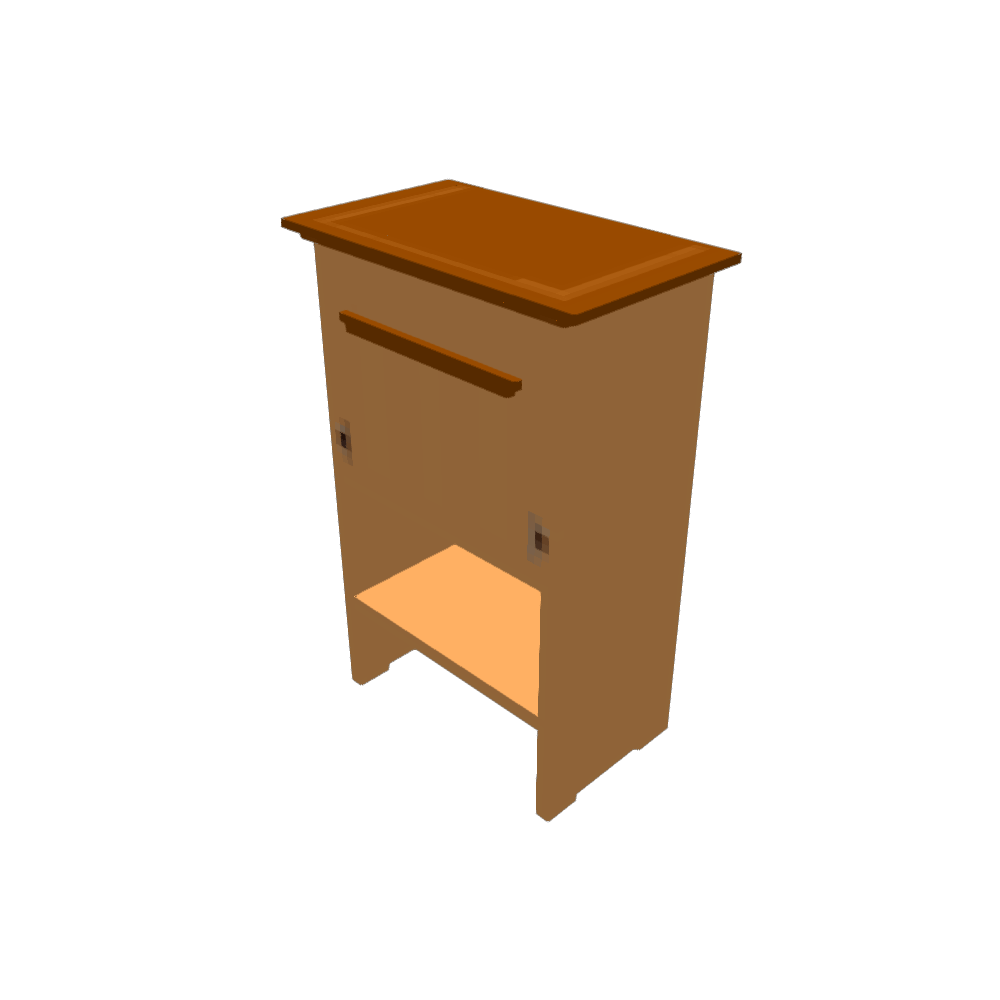} &
\includegraphics[trim=15 100 15 120,clip]{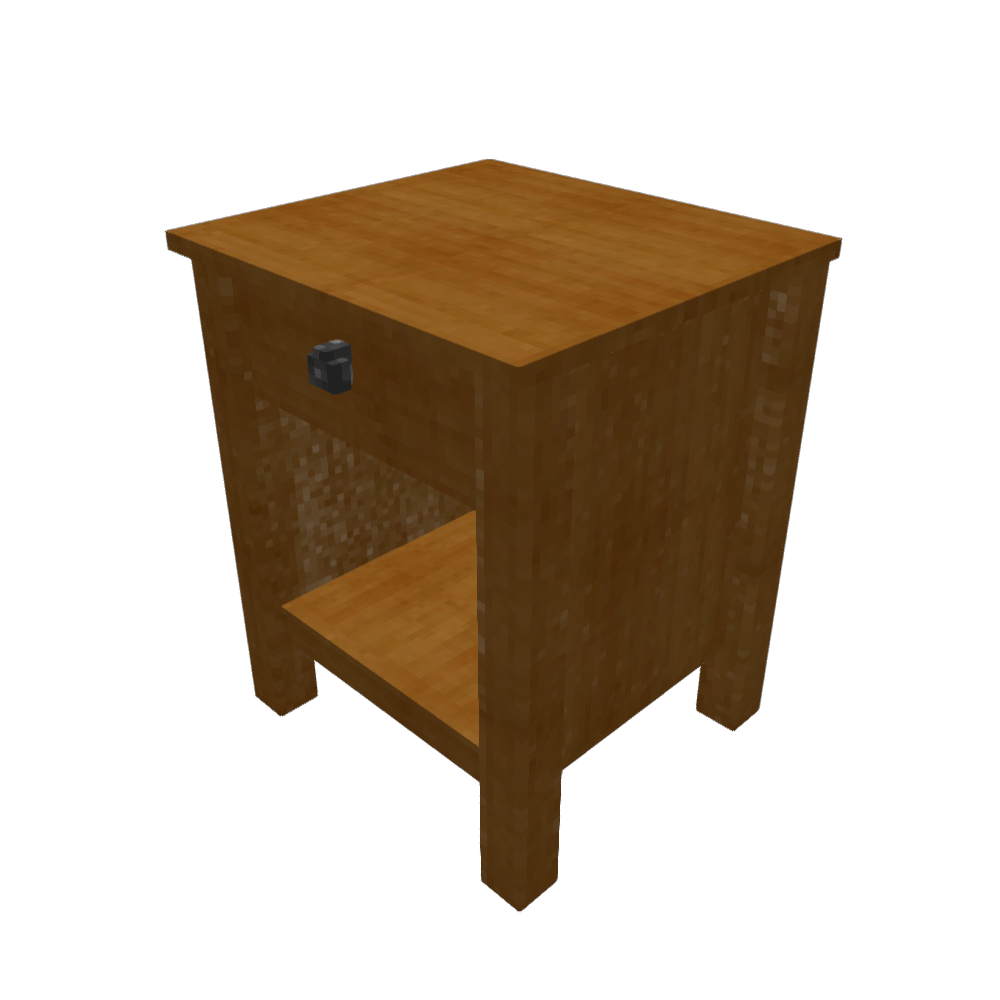} &
\includegraphics[trim=15 100 15 120,clip]{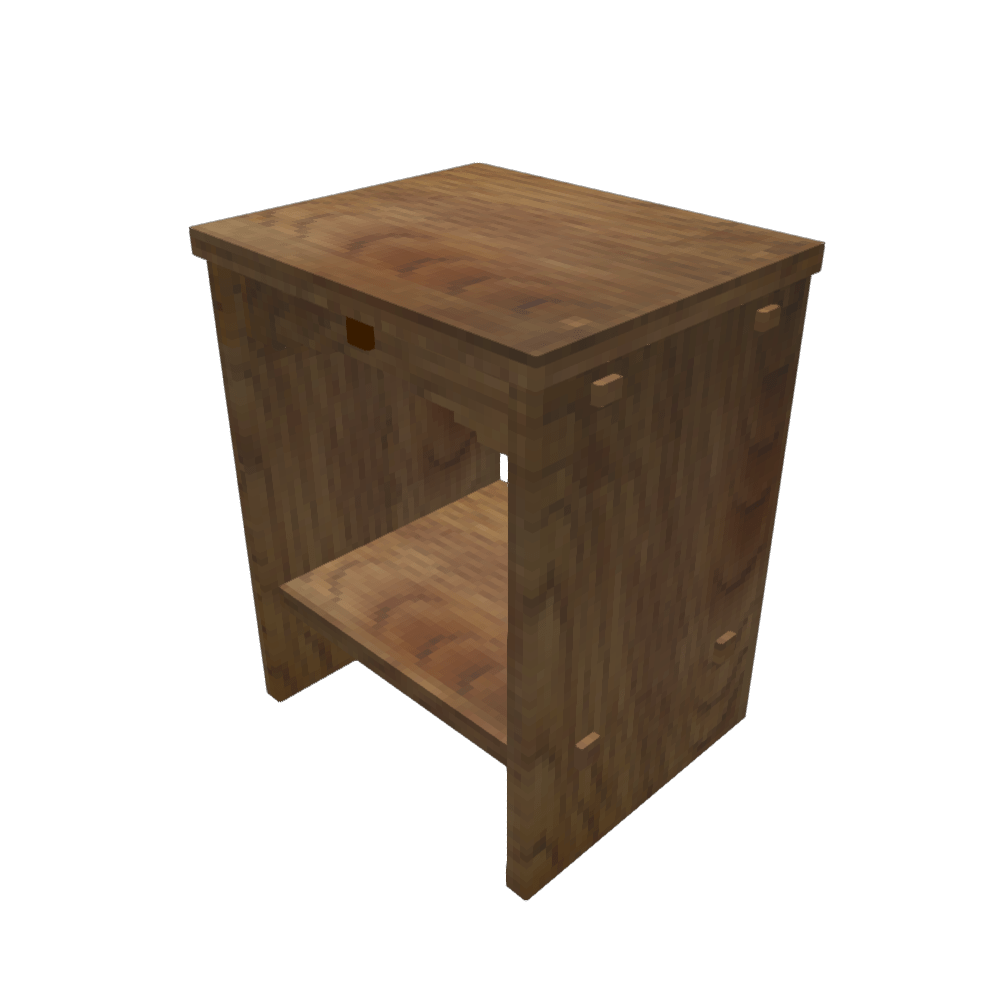} & 
\includegraphics[trim=15 100 15 120,clip]{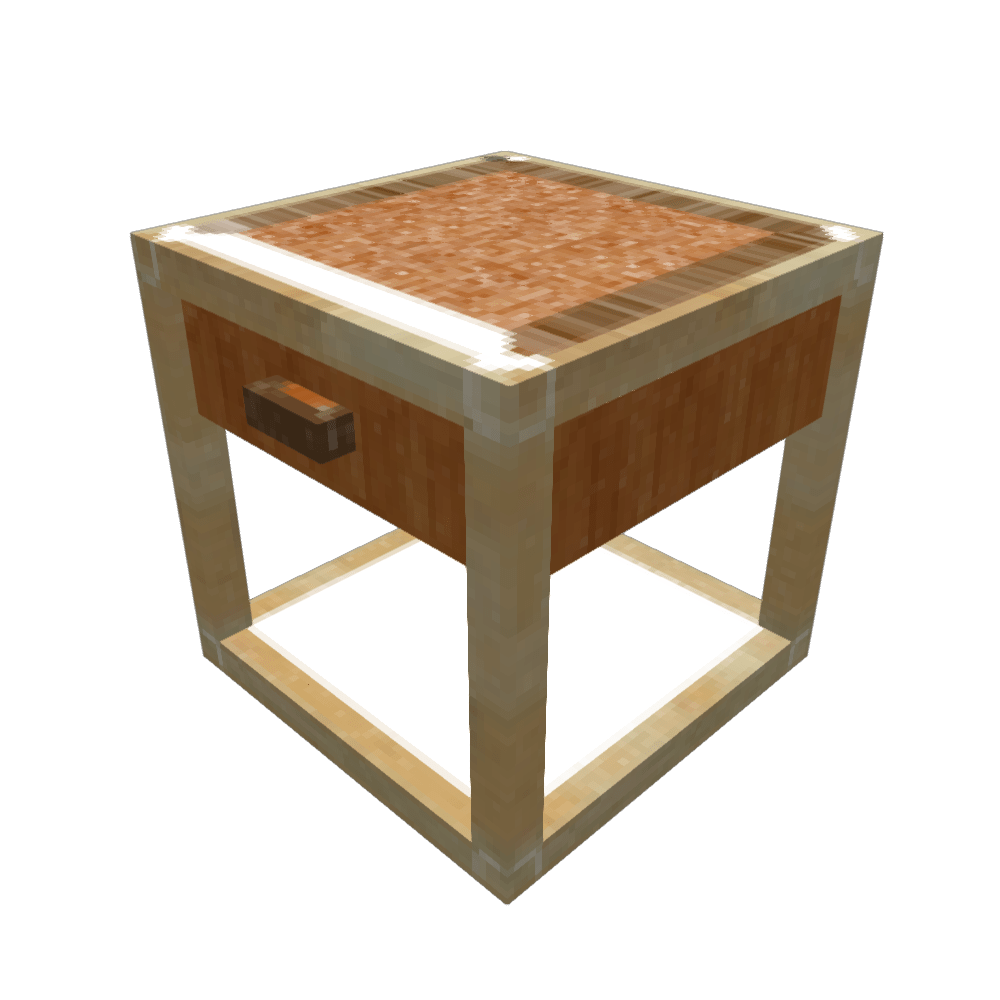} & 
\includegraphics[trim=15 100 15 120,clip]{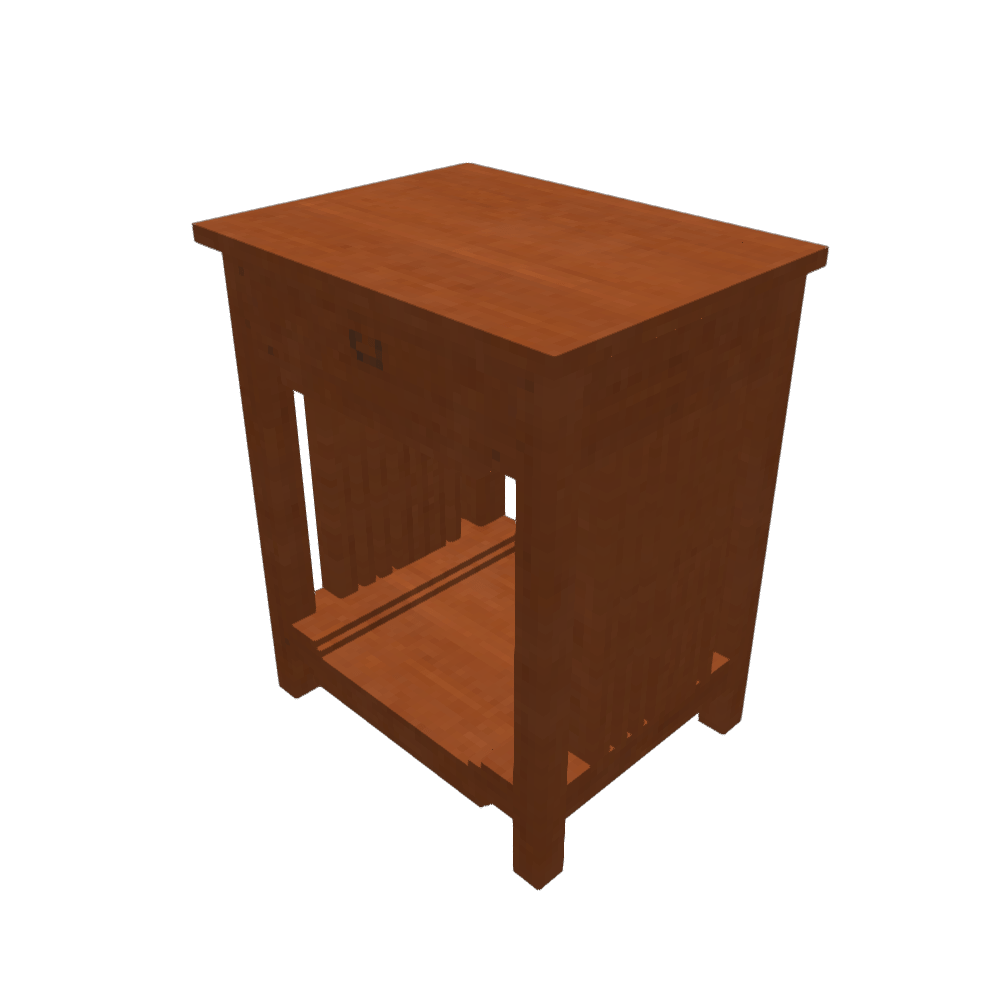}\\
[-0.1cm]
\trimodiv & 
\includegraphics[trim=15 100 15 120,clip]{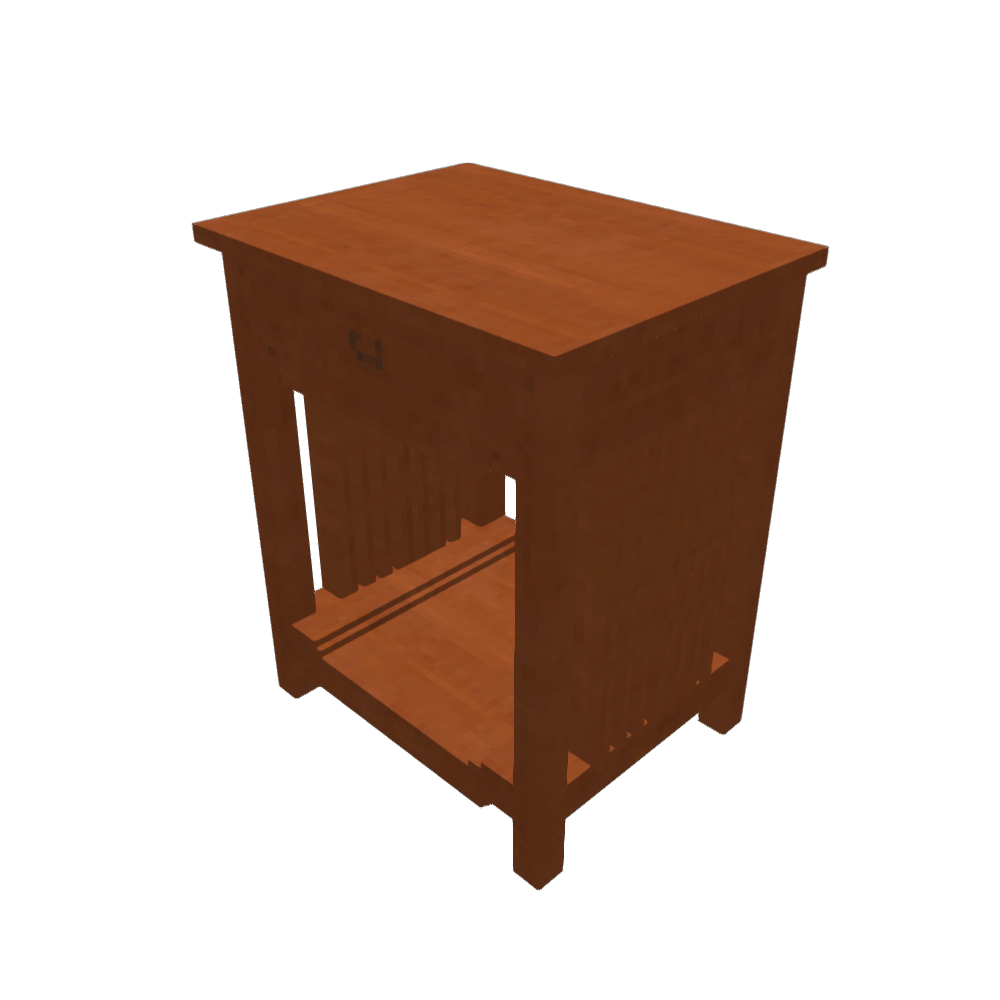} & 
\includegraphics[trim=15 100 15 120,clip]{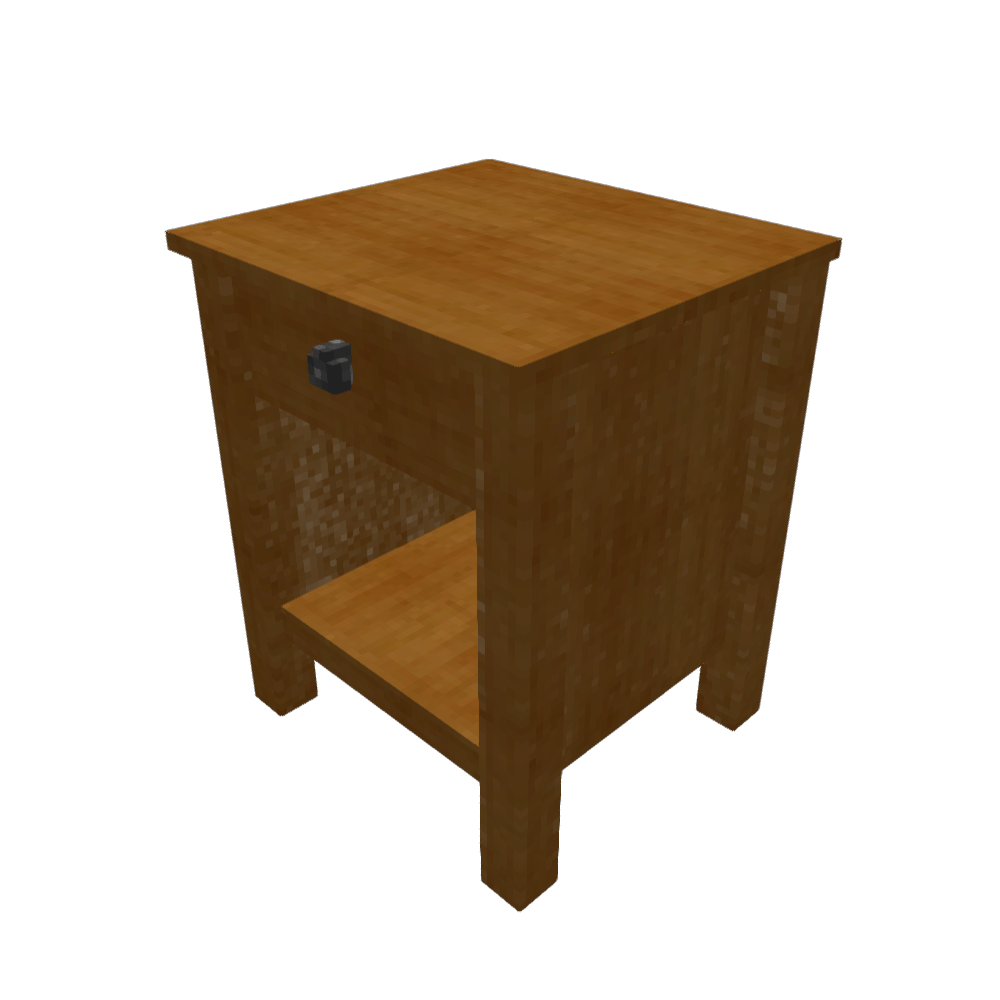} &
\includegraphics[trim=15 100 15 120,clip]{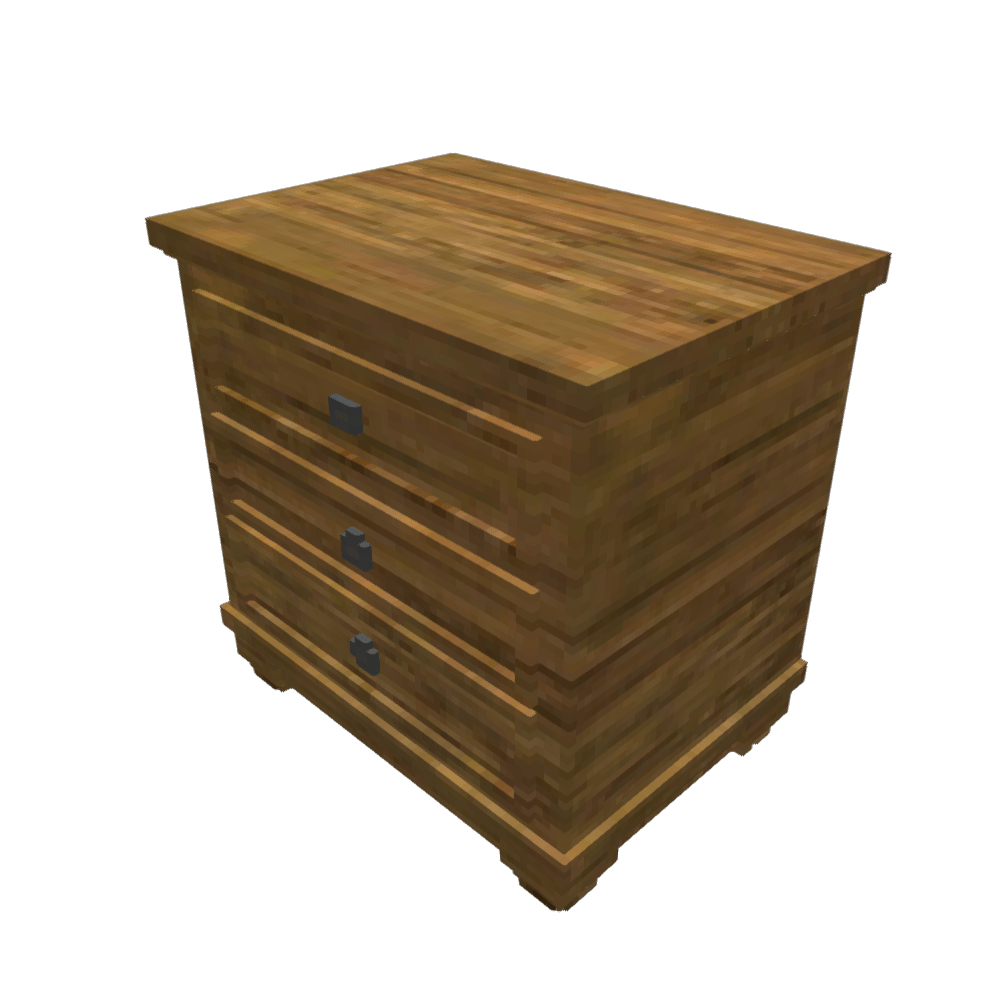} &
\includegraphics[trim=15 100 15 120,clip]{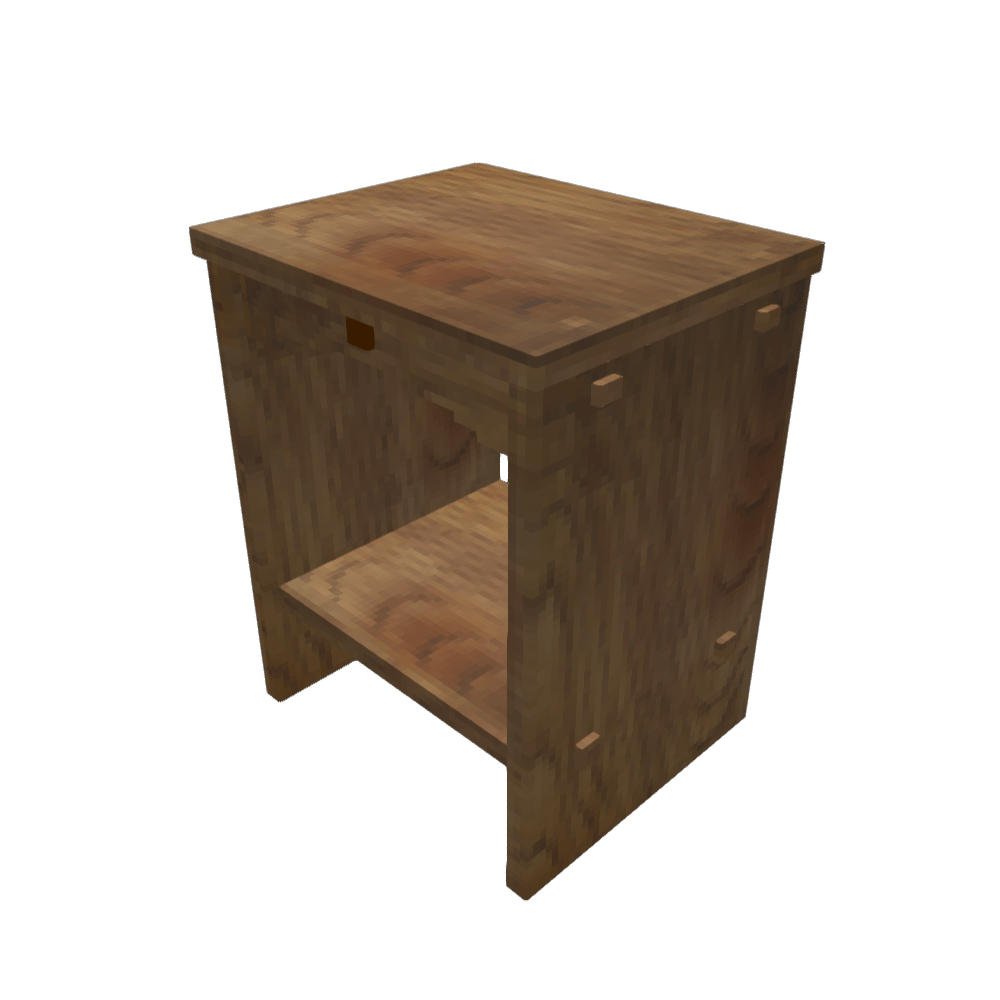} & 
\includegraphics[trim=15 100 15 120,clip]{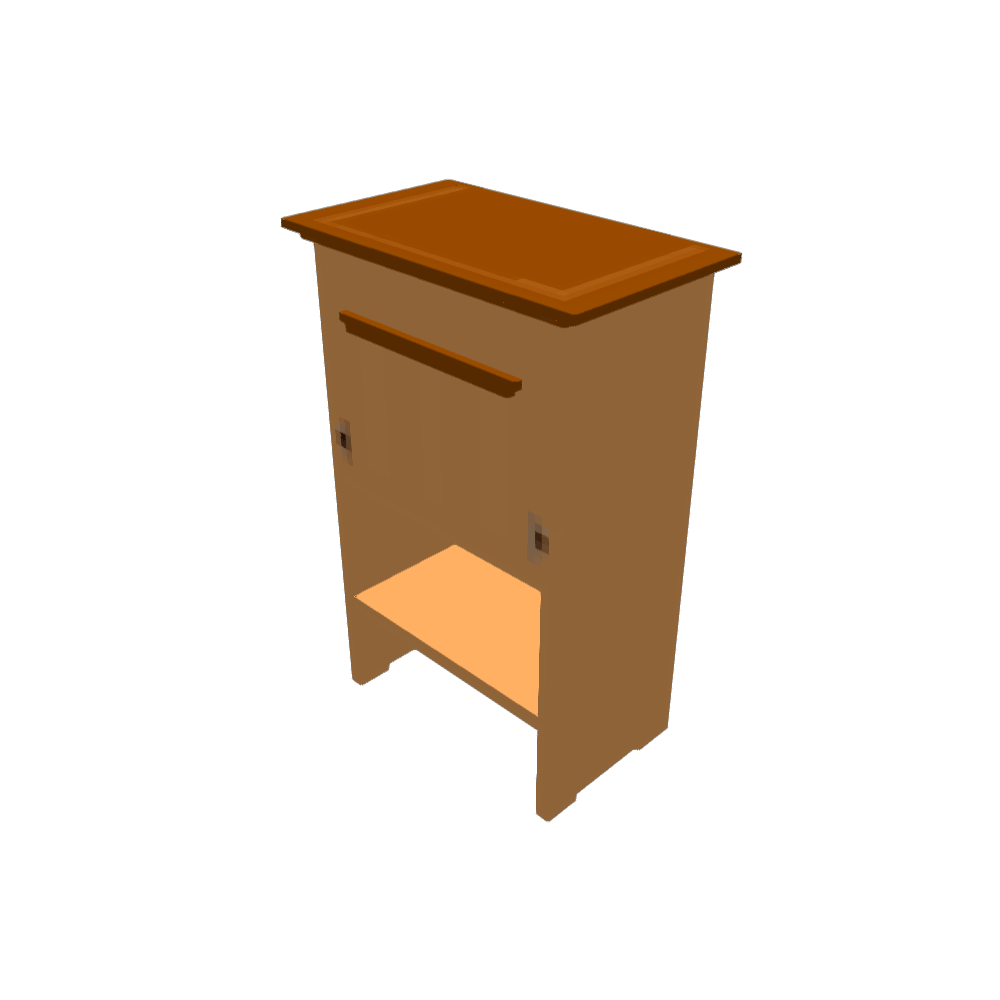} \\
[-0.1cm]
\bottomrule
\end{tabularx}
\vspace{-0.25cm}
\caption{Examples of top 5 retrieved shapes from the validation set using \bimodi, \bimodv, and \trimodiv. We can see that \bimodi understands abstract concepts such as \textit{stretched}, \textit{tennis} and \textit{picnic} poorly (examples 1,4,5). It is also challenging to pick up on small details (\textit{notch} in example 3).  Although the retrieval results of \trimodiv are not always the best among the three models (\bimodv results are better for example 6), \trimodiv  is the most stable overall and retrieves more results that are consistent with the description (examples 2,4,5).
} 
\label{fig:retrieval-comparison-visualization}
\end{figure*}

\cref{fig:retrieval-custom-visualization} shows the best matching shapes each model retrieves for novel queries.
These results show our network can be used for easy and rapid search through large 3D collections.  We show  qualitative comparisons of the models in \cref{fig:retrieval-comparison-visualization}. From examples 1,4 and 5, we see that \bimodi is unable to match high-level words such as \textit{stretched}, \textit{tennis} and \textit{picnic} as well as \bimodv and \trimodiv. Examples 2 and 3 show that sometimes \bimodv also proposes poor retrieval results. Since \trimodiv considers both images and voxels, it is more robust and retrieves fewer incorrect results than \bimodi and \bimodv (examples 2,4,5).  It is also challenging for the models to match small details (\textit{notch} in example 3) and the correct number of parts (\textit{single drawer} in example 6).

\subsection{Shape similarity metrics}
\label{sec:supp:shape-sim-discussion}

In addition to measuring $F1^{\tau}$ for $\tau=0.1$ we also follow prior shape retrieval work~\cite{kuo2020mask2cad} and measure $F1^{\tau}$ for $\tau=0.3, 0.5$, as well as the Chamfer Distance (CD), and (Abstract) Normal Consistency (NC).
We note that these are all point-wise metrics and we sample 10K points uniformly on the mesh surface of GT and retrieved shapes for computing these metrics.
Note that both CD and $F1^{\tau}$ depend on the absolute scale of meshes.
To compute them, we follow \citet{fan2016point} who define 1 unit as 1/10 of the largest length of the ground truth's bounding box and rescale the ground truth and retrieved meshes individually.
See \cref{tab:retrieval-modality-shapenet-full} (which expands on Tab. 3 from the main paper).
While our results show that the average of these metrics increases overall for our best model, \trimodiv, we find that these shape similarity metrics are not very informative in measuring fine shape details and do not always capture whether two shapes are semantically similar.
Nevertheless, these metrics are popular in 3D shape generation and retrieval literature and we report them here for completeness.

\begin{figure*}
\centering
\setkeys{Gin}{width=0.8\linewidth}
\setlength\tabcolsep{1.5pt}
\begin{tabularx}{\linewidth}{@{}p{1.5cm}YYYYY@{}}
\toprule
1 &\multicolumn{4}{p{12.0cm}}{\small{A wooden designer chair, which is good for a stretched sitting.}} &
\includegraphics[trim=50 150 50 275,clip]{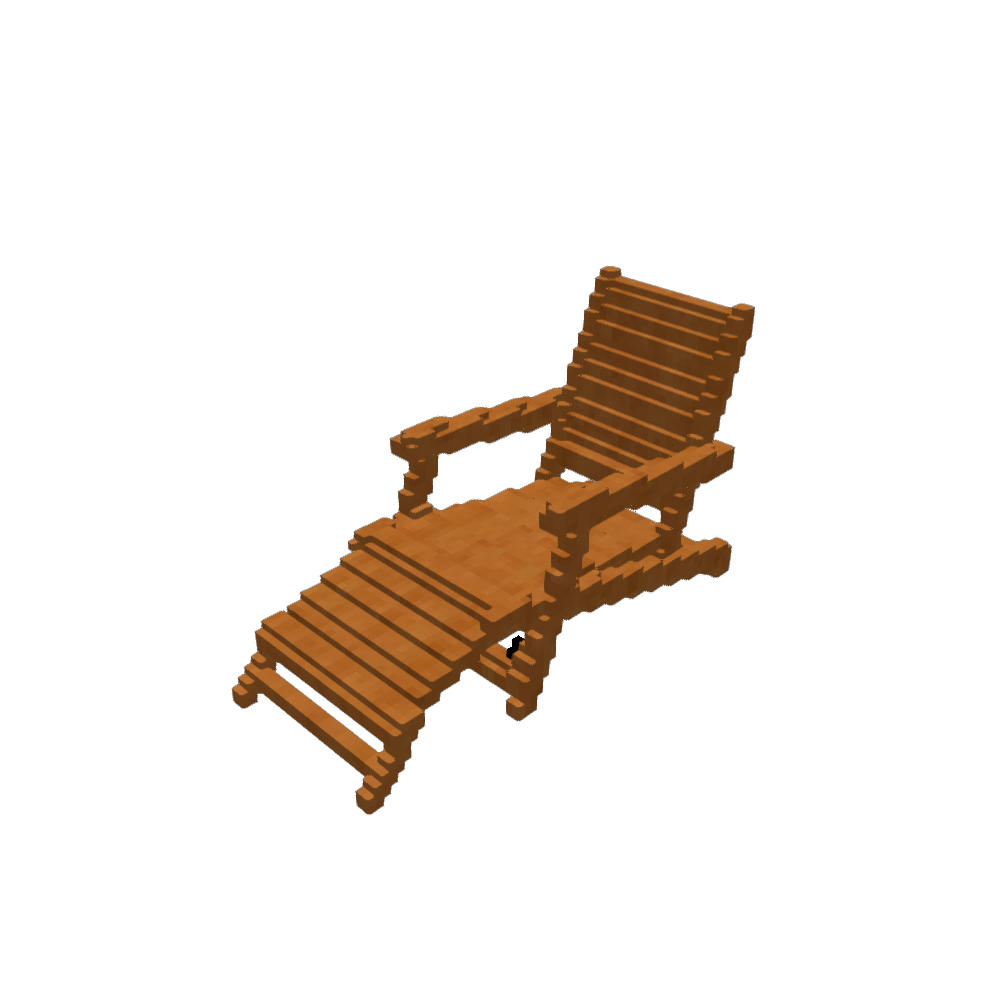} \\
[-0.17cm]
\bimodi & 
\includegraphics[trim=50 180 50 200,clip]{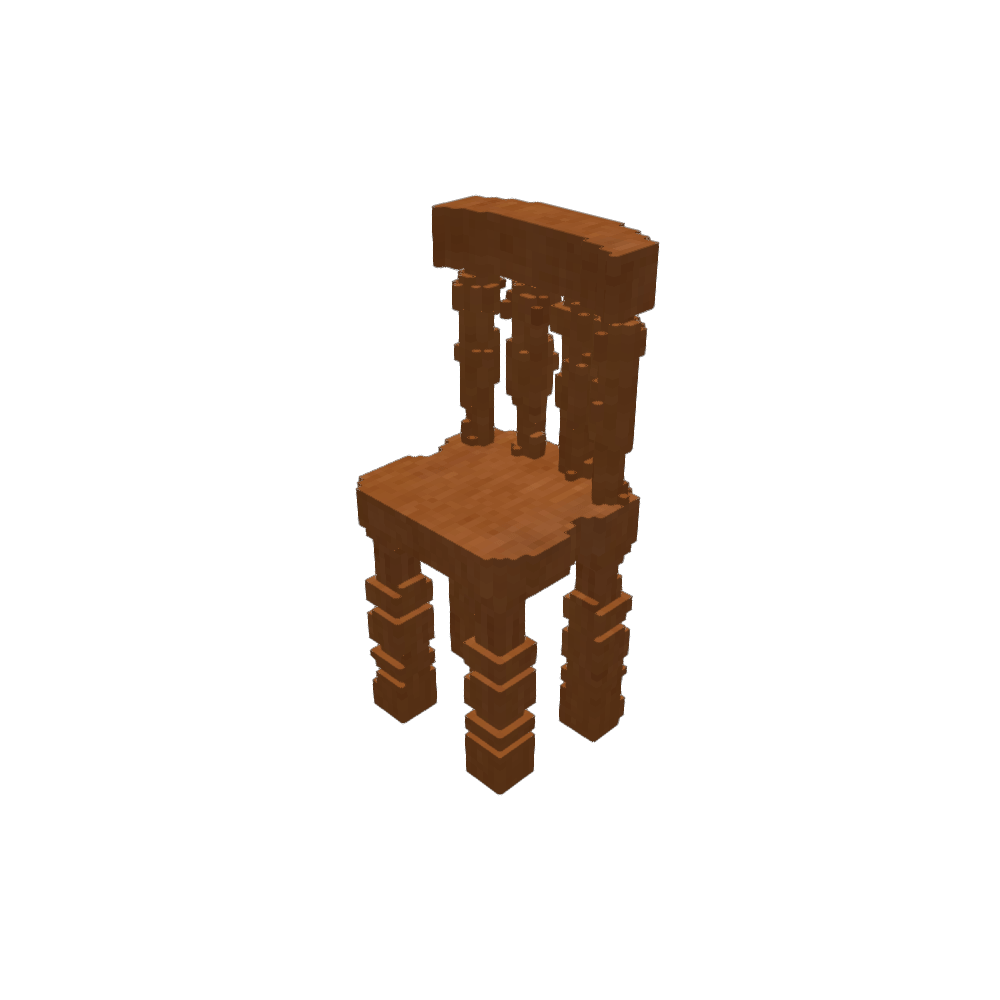} & 
\includegraphics[trim=50 180 50 200,clip]{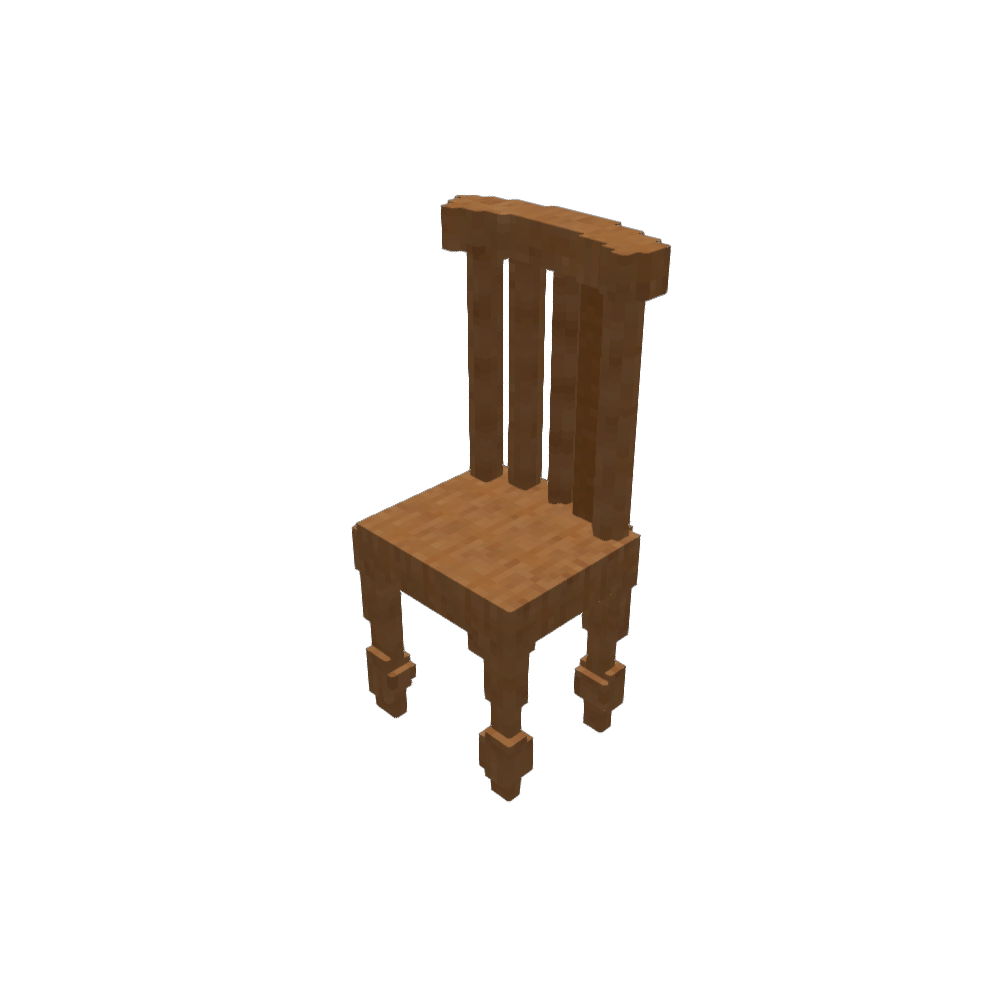} & 
\includegraphics[trim=50 180 50 200,clip]{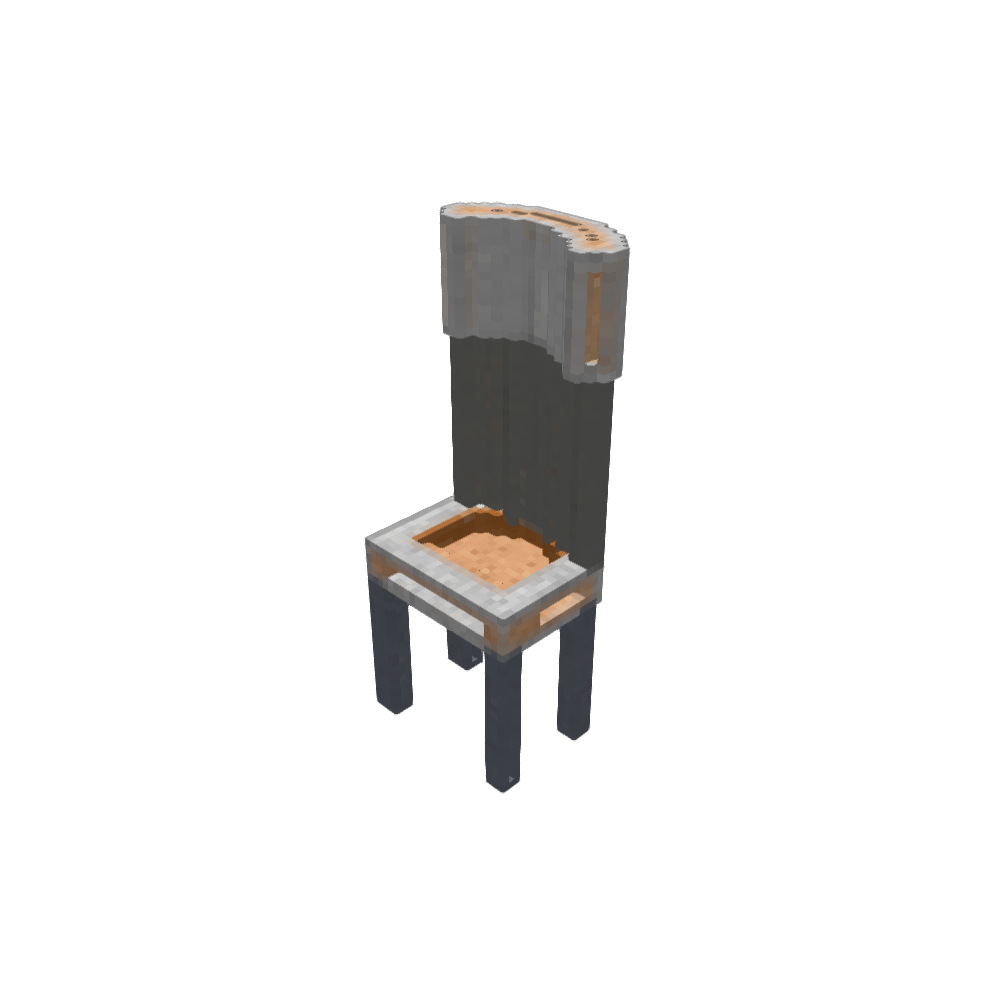} & 
\includegraphics[trim=50 180 50 200,clip]{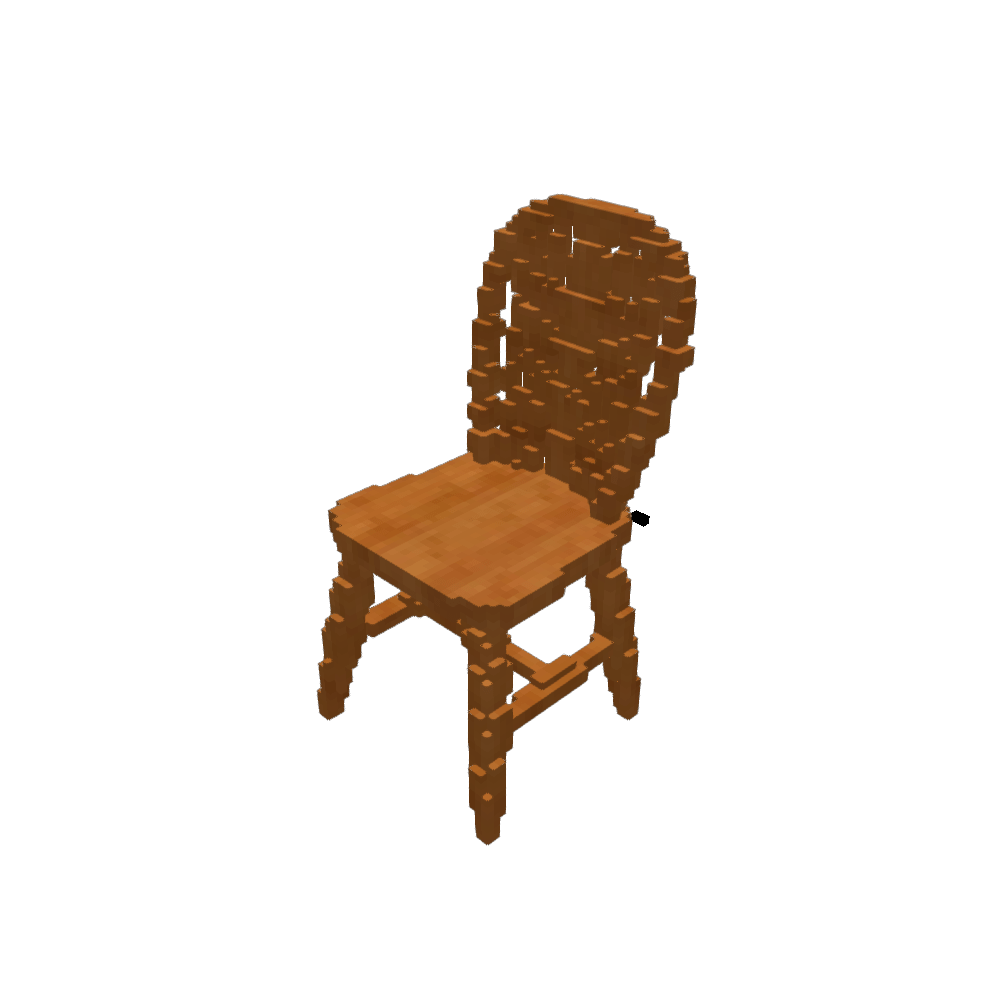} & 
\includegraphics[trim=50 180 50 200,clip]{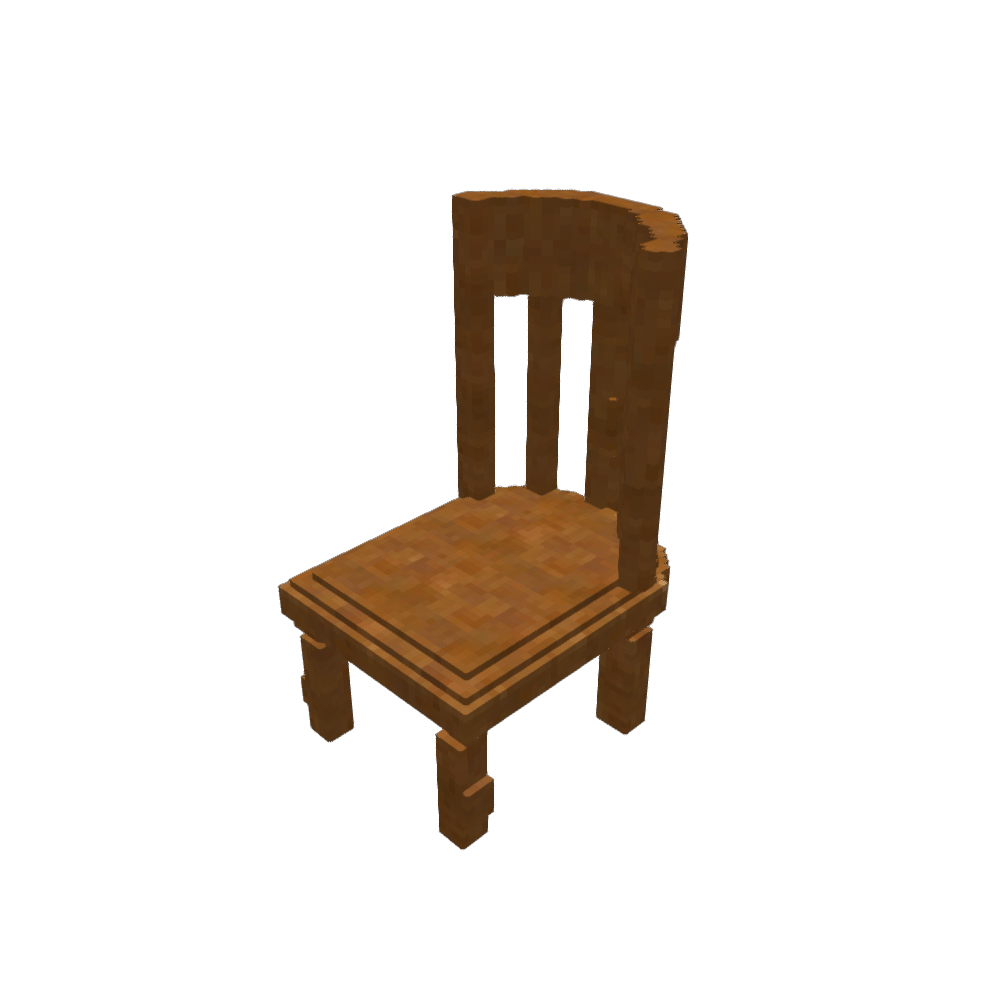} \\
[-0.12cm]
\bimodv & 
\includegraphics[trim=50 100 50 150,clip]{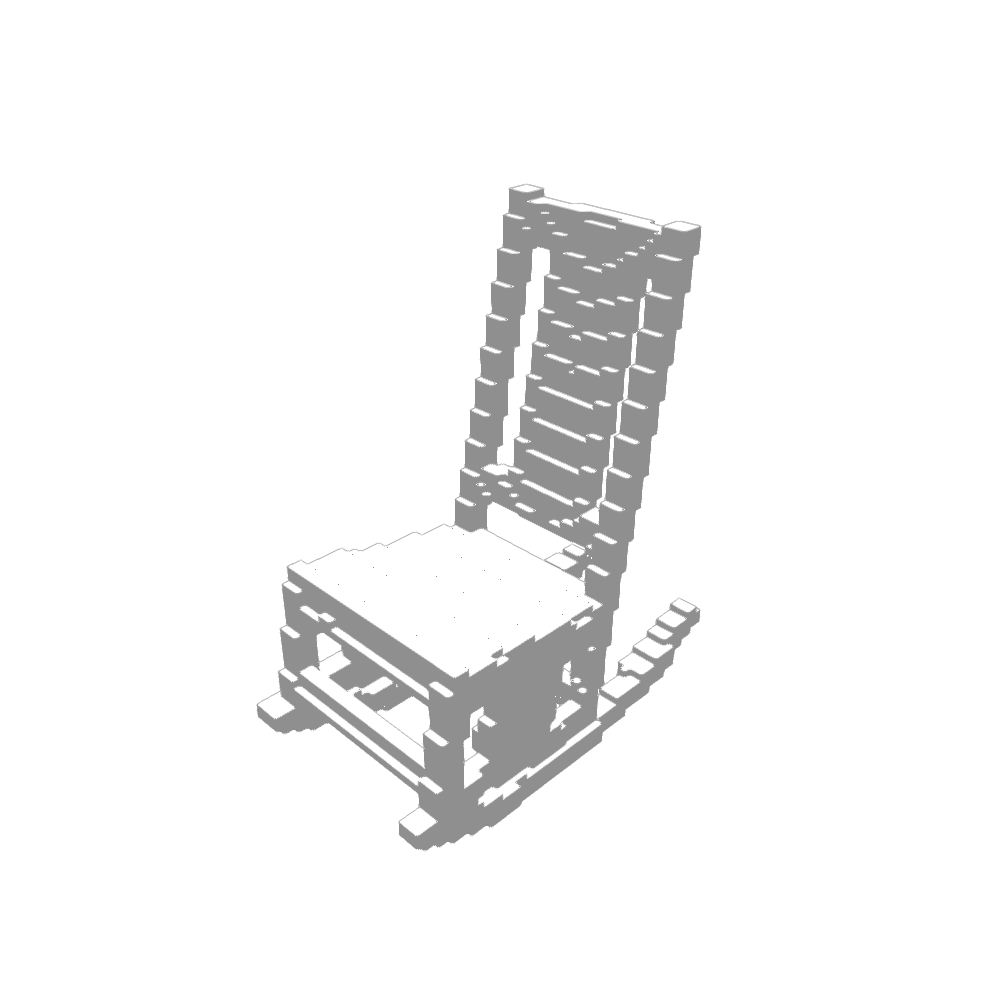} & 
\includegraphics[trim=50 100 50 150,clip]{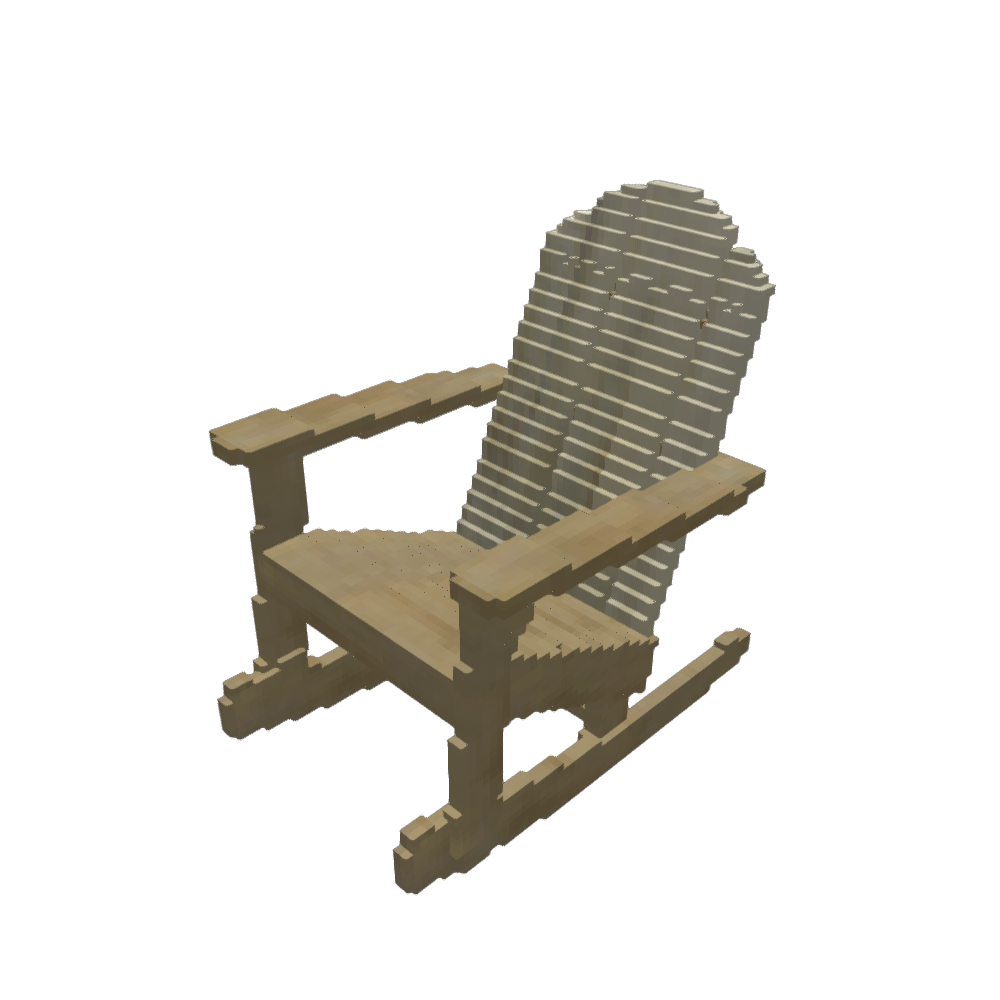} & 
\includegraphics[trim=50 100 50 150,clip]{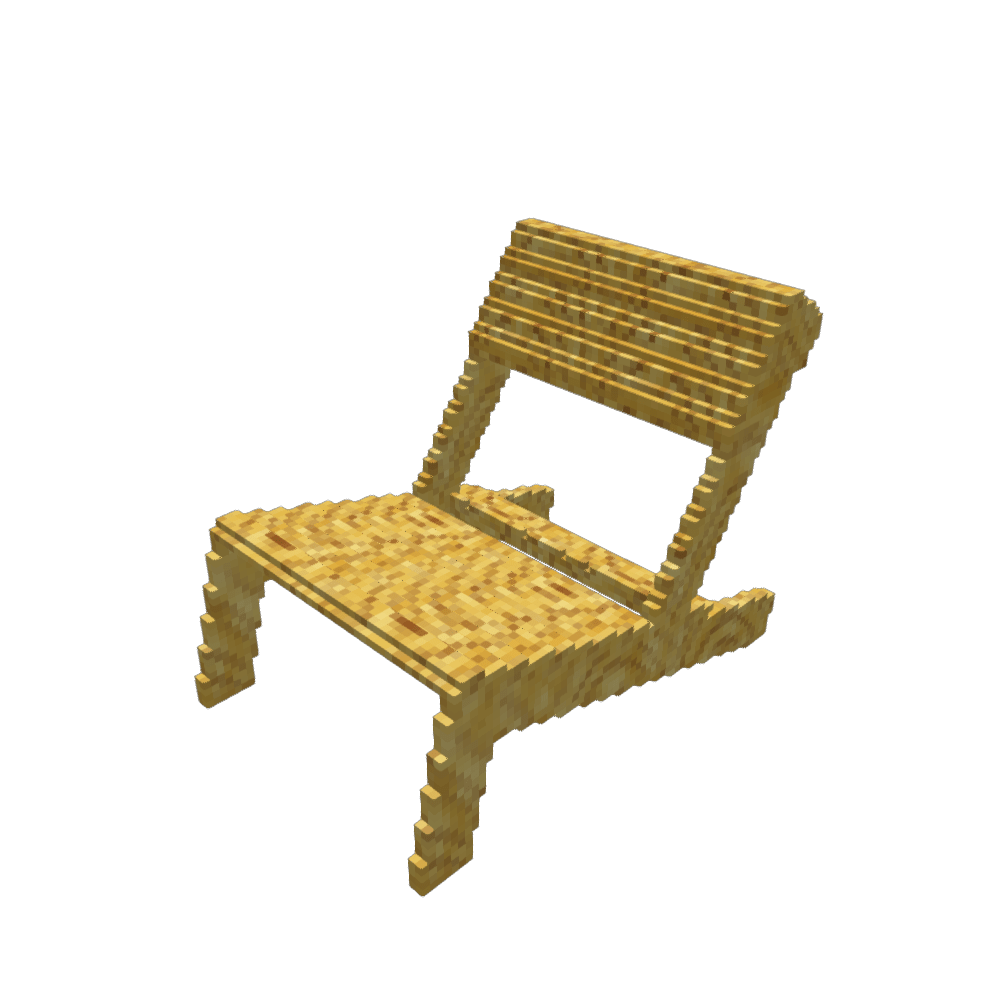} & 
\includegraphics[trim=50 100 50 150,clip]{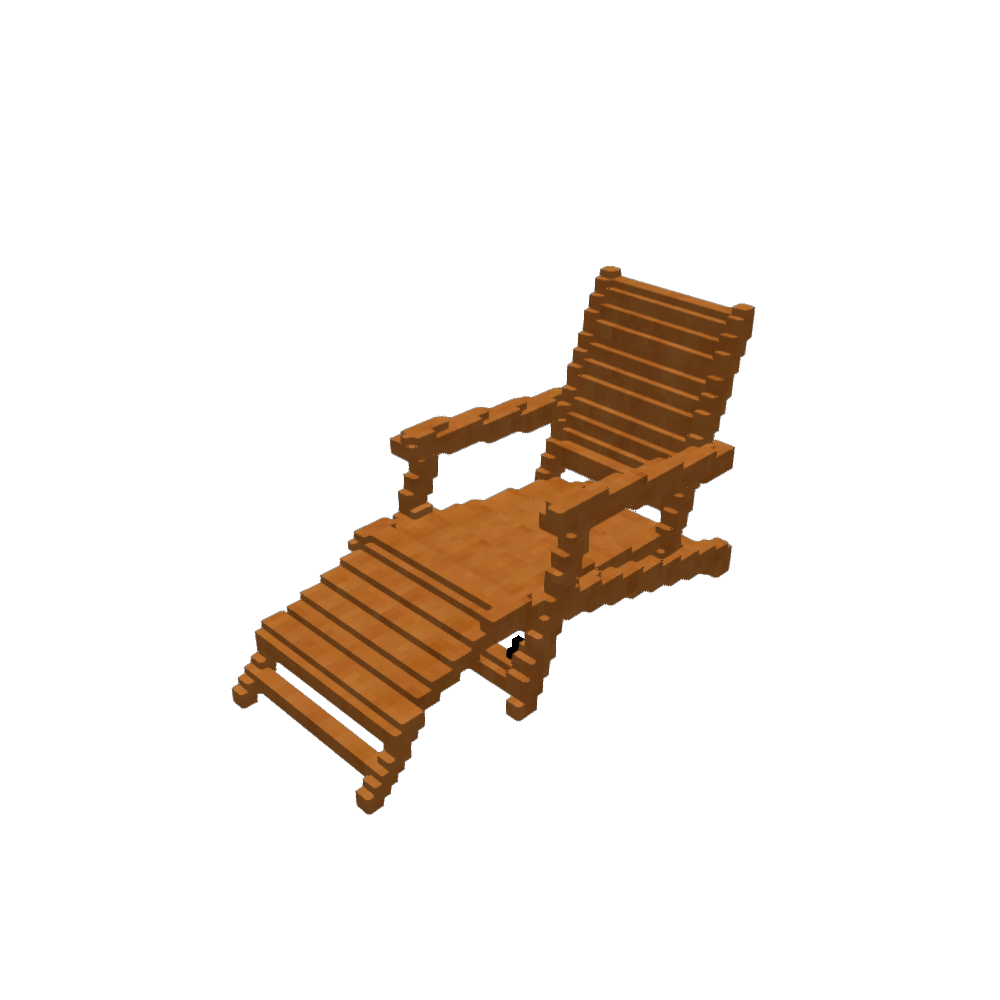} & 
\includegraphics[trim=50 100 50 150,clip]{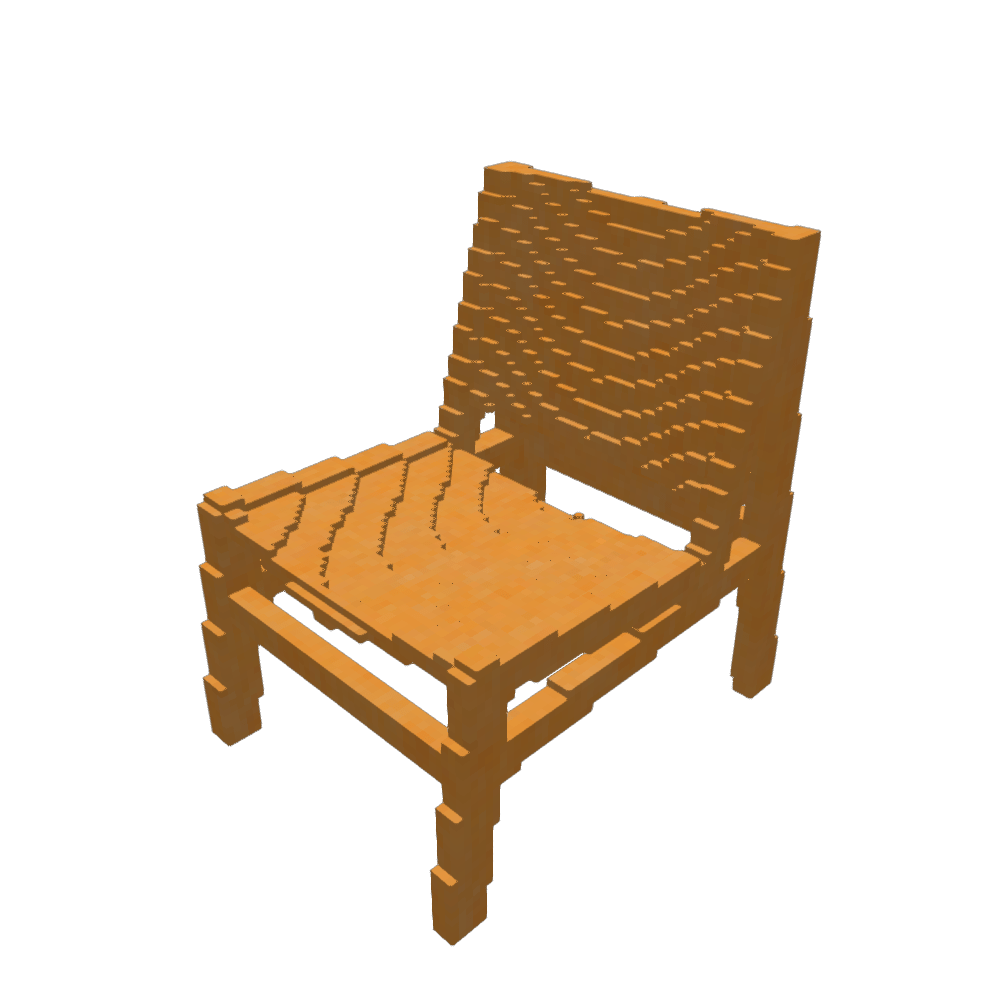} \\
[-0.12cm]
\trimodiv & 
\includegraphics[trim=50 80 50 150,clip]{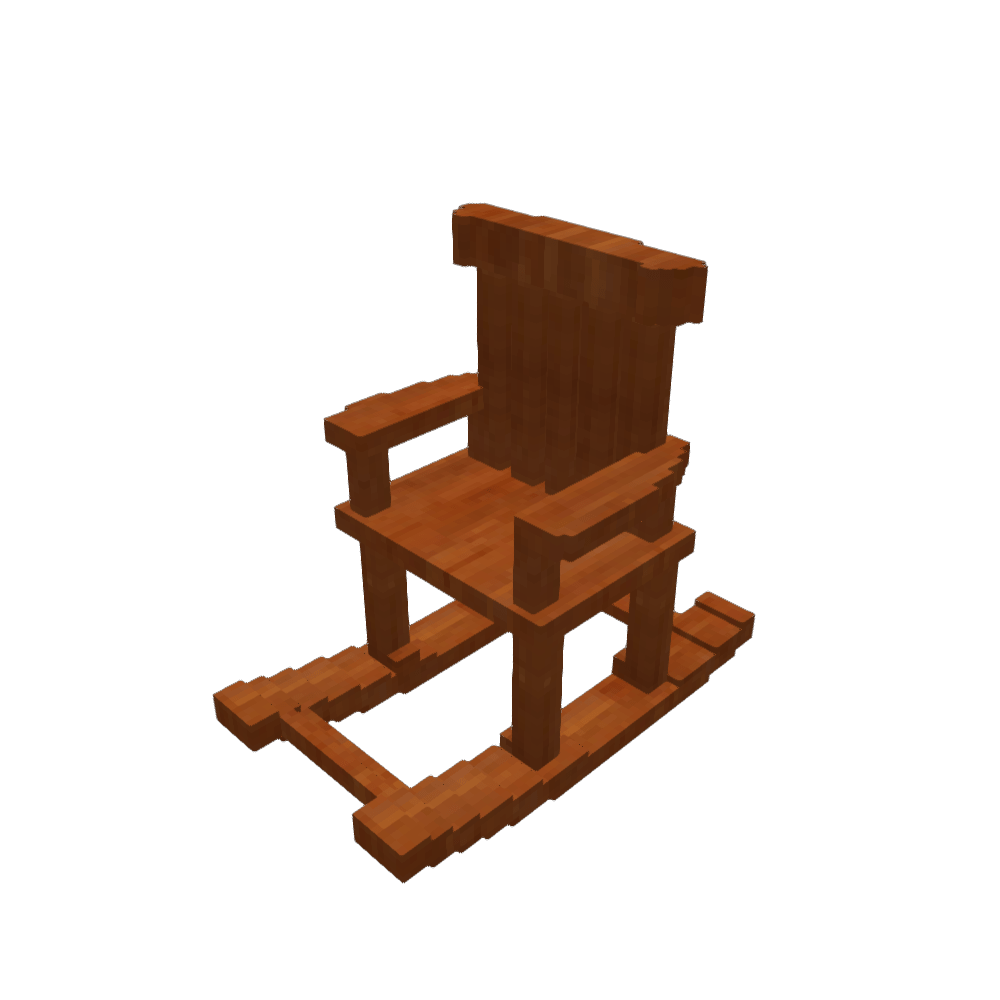} & 
\includegraphics[trim=50 80 50 150,clip]{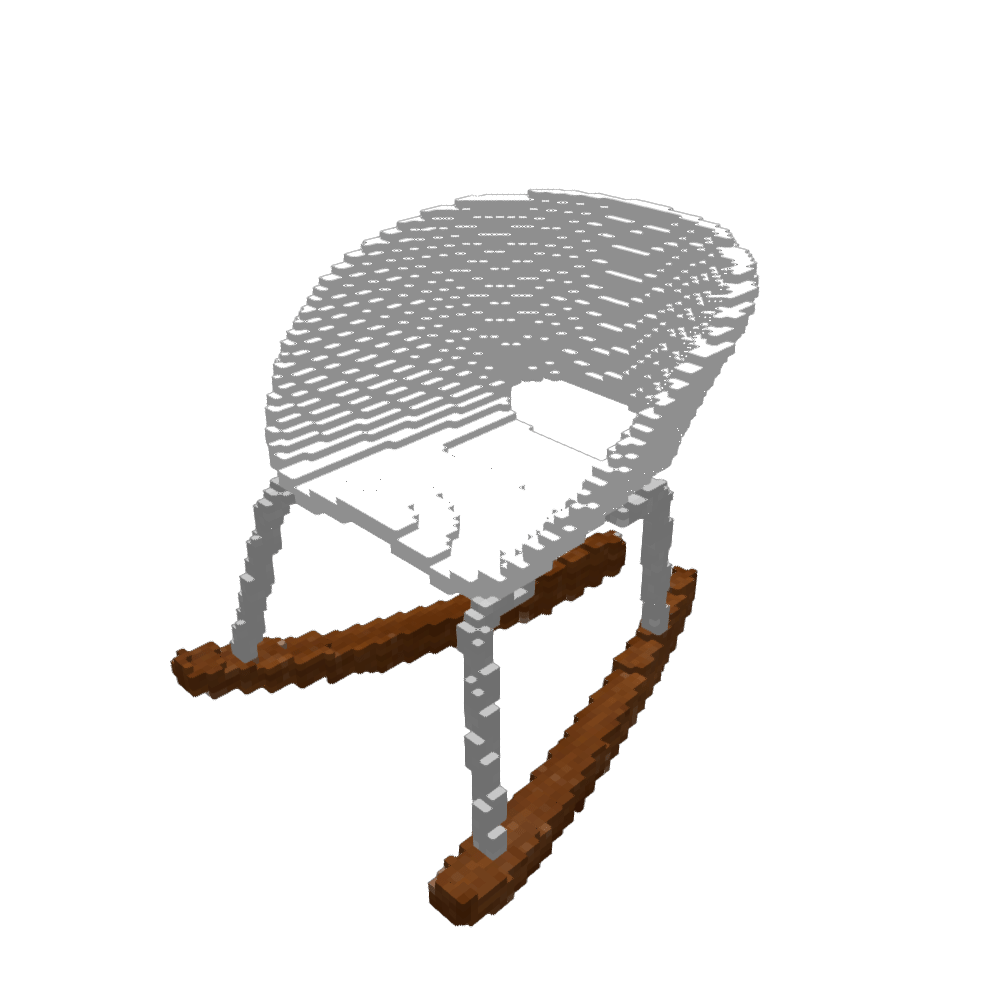} &
\includegraphics[trim=50 80 50 150,clip]{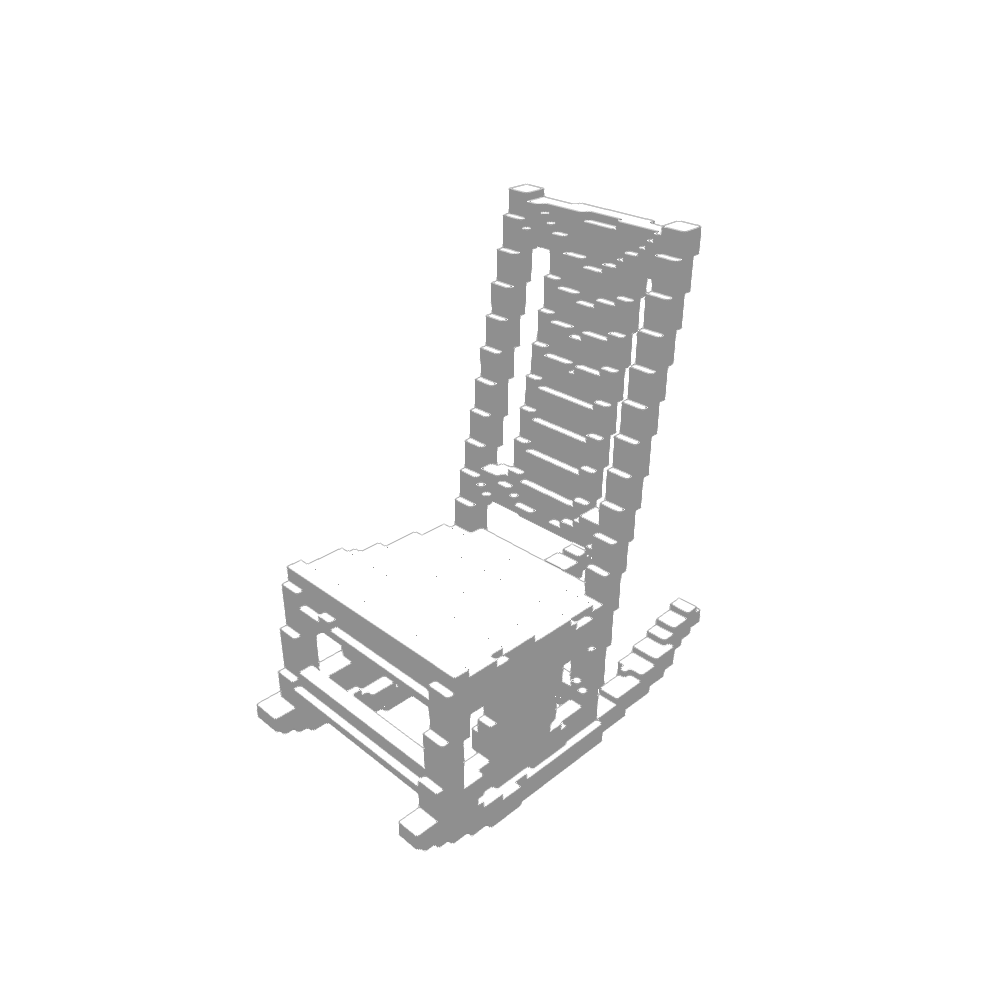} & 
\includegraphics[trim=50 80 50 150,clip]{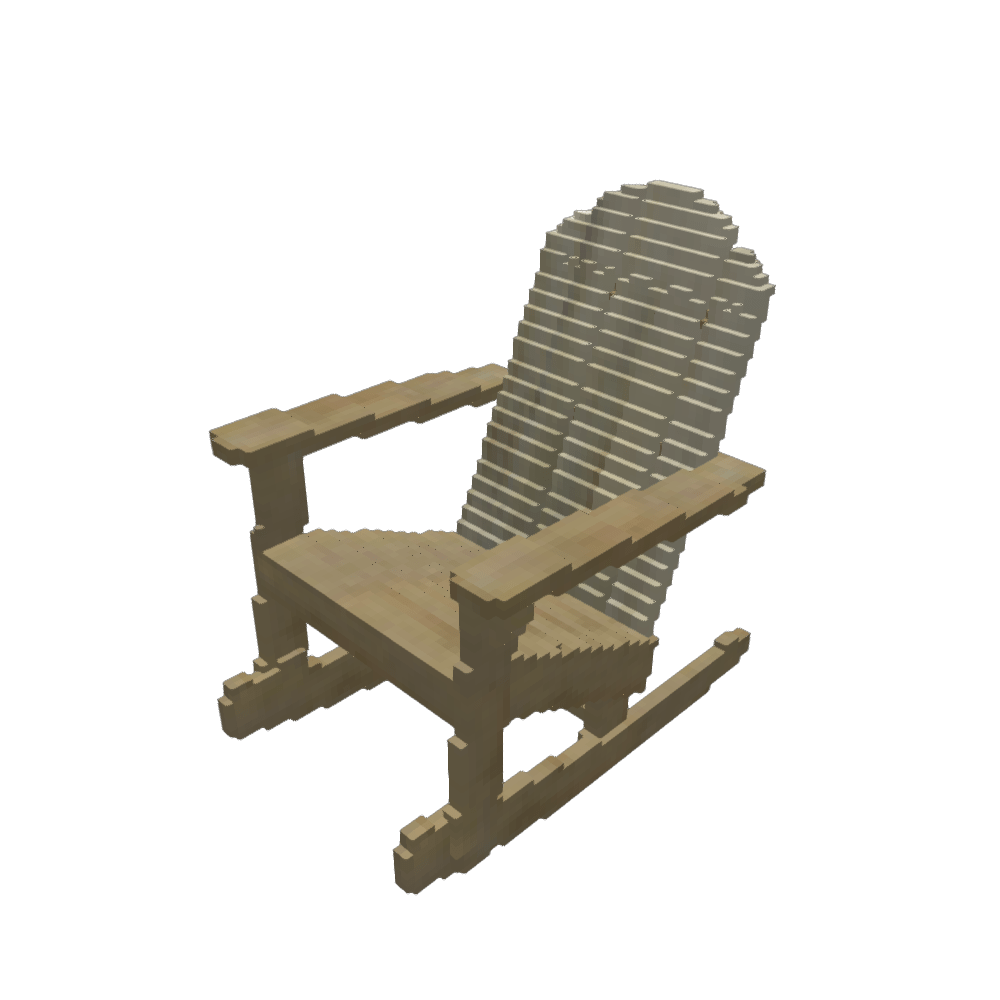} & 
\includegraphics[trim=50 80 50 150,clip]{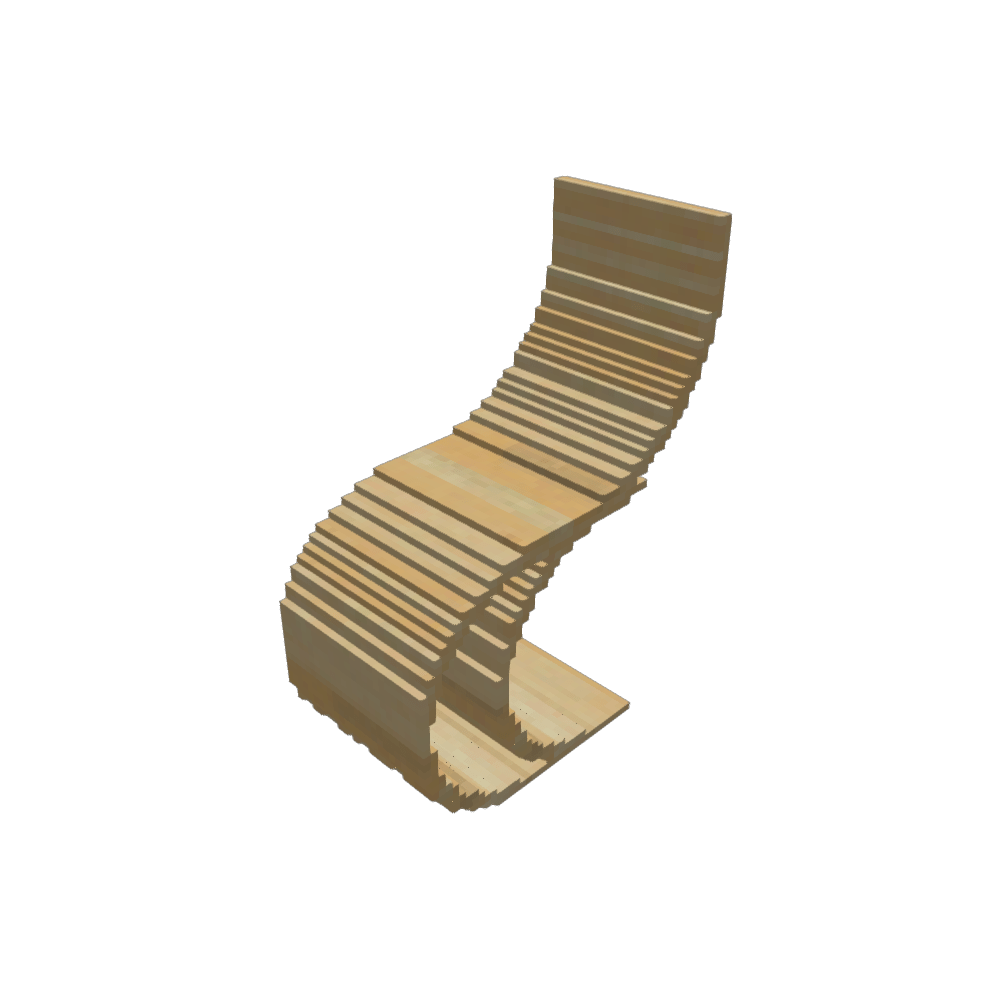} \\
[-0.1cm]
\midrule
2 &\multicolumn{4}{p{12.0cm}}{This short bar stool has a curved metal back in gray. The round cushion is blue and appears to be vinyl.}  &
\includegraphics[trim=50 100 50 200,clip]{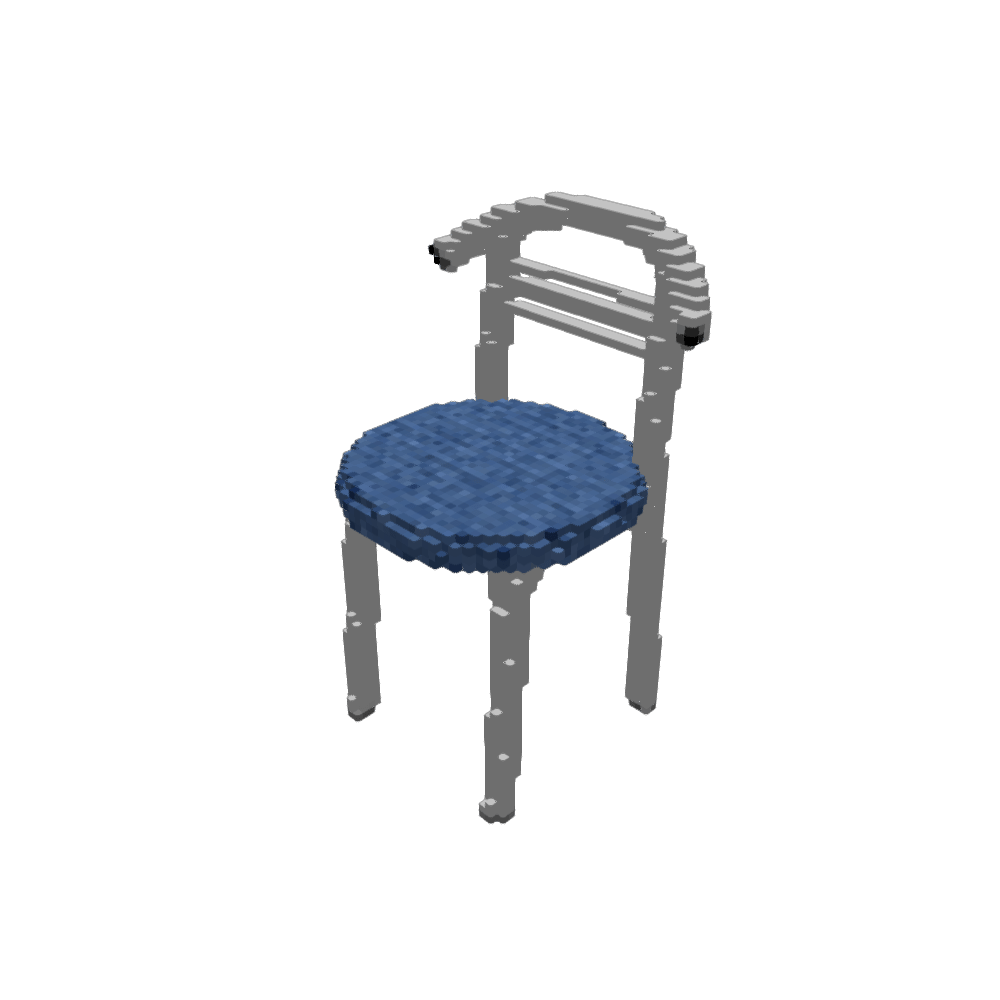} \\
[-0.35cm]
\bimodi & 
\includegraphics[trim=50 80 50 170,clip]{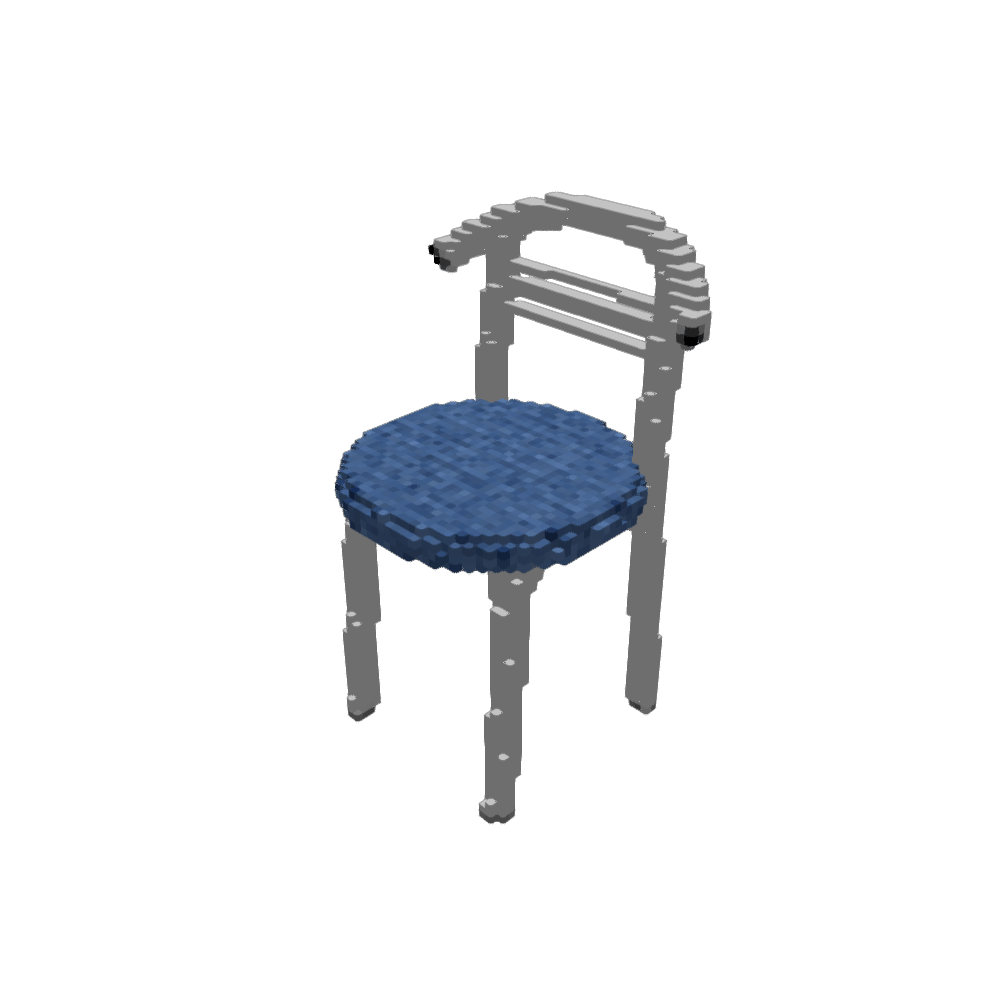} & 
\includegraphics[trim=50 80 50 170,clip]{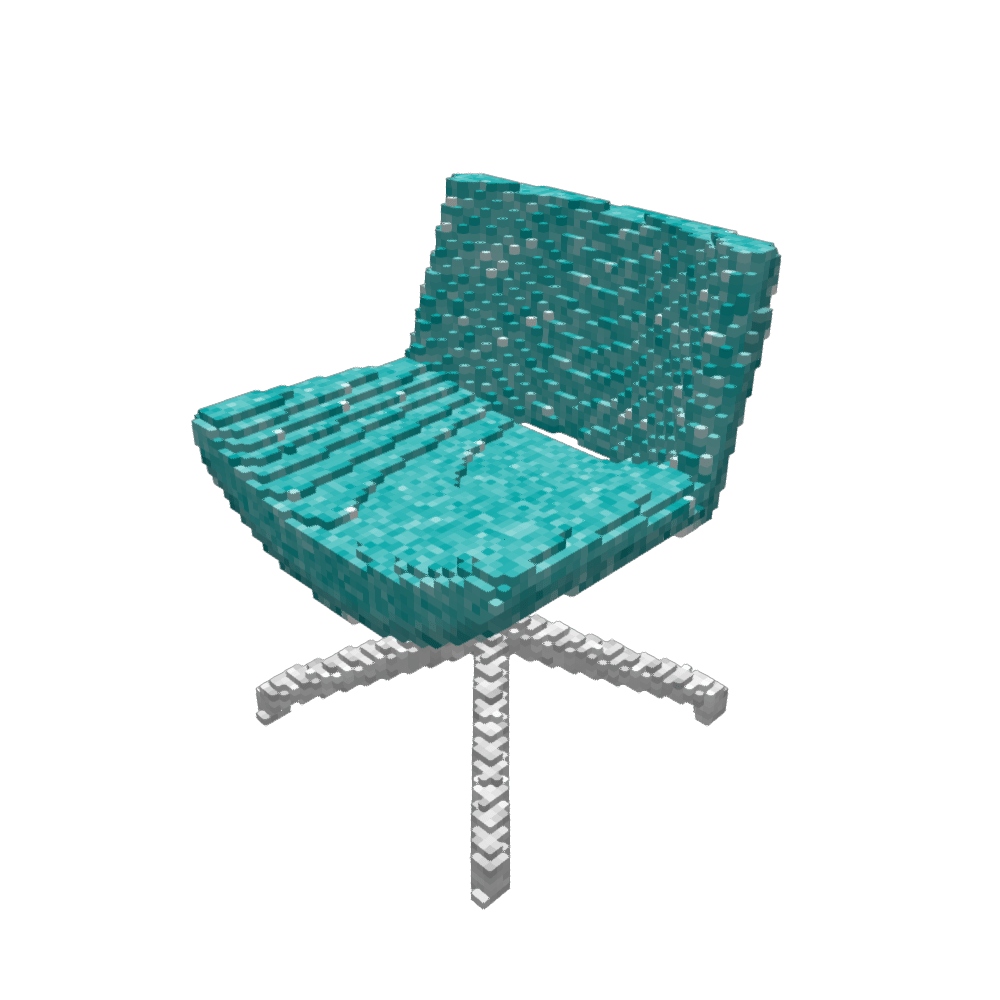} &
\includegraphics[trim=50 80 50 170,clip]{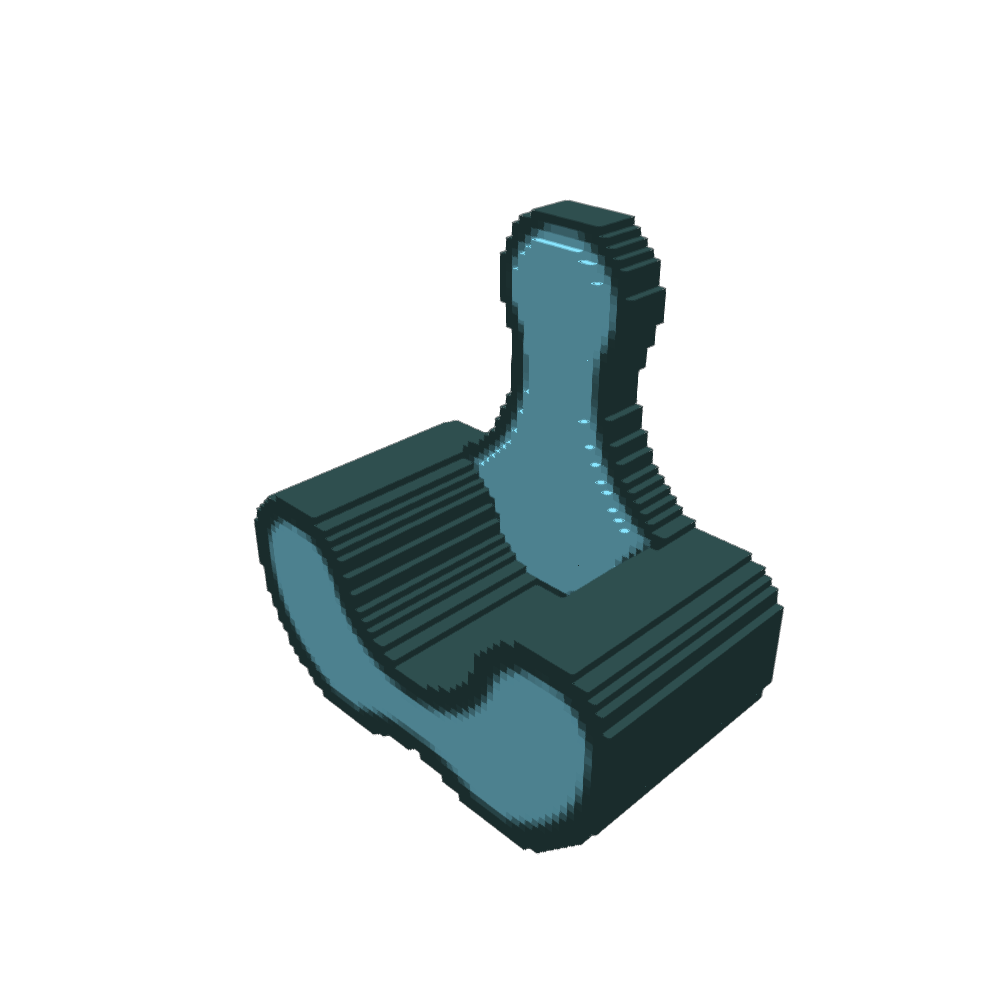} &
\includegraphics[trim=50 80 50 170,clip]{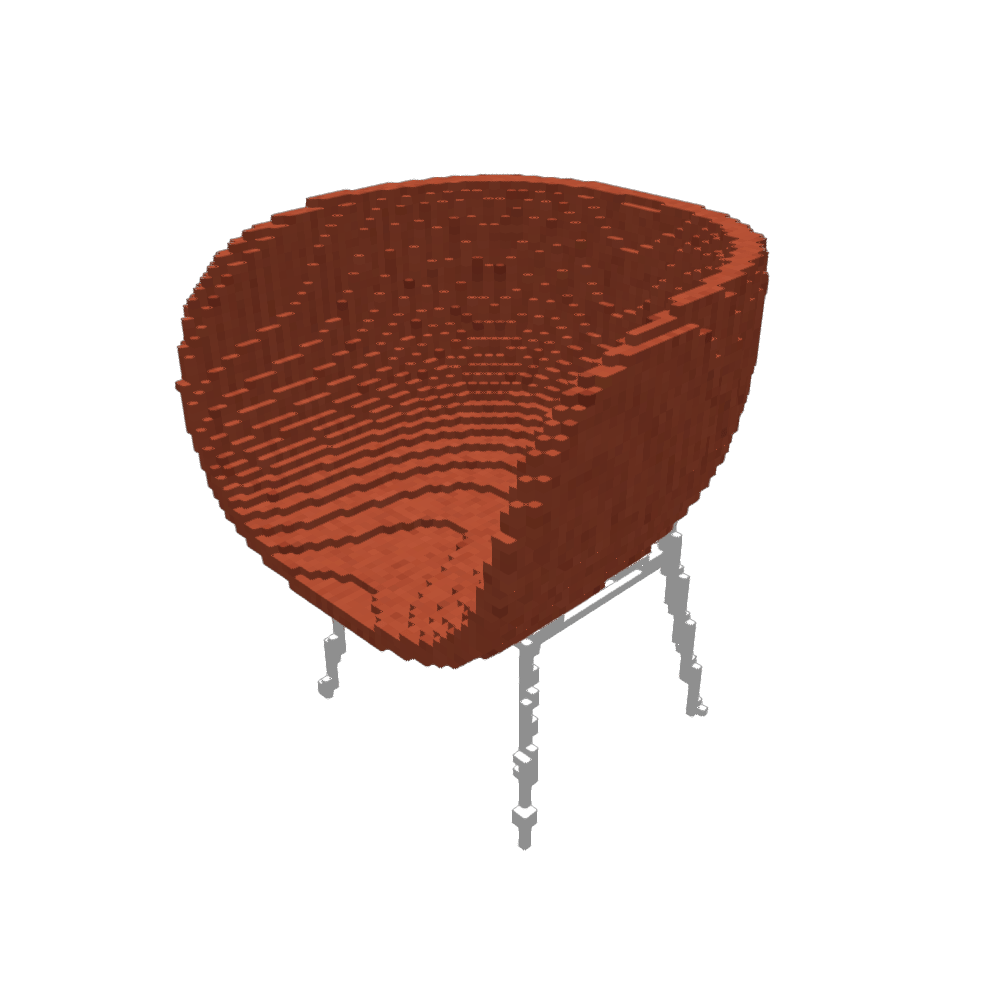} &
\includegraphics[trim=50 80 50 170,clip]{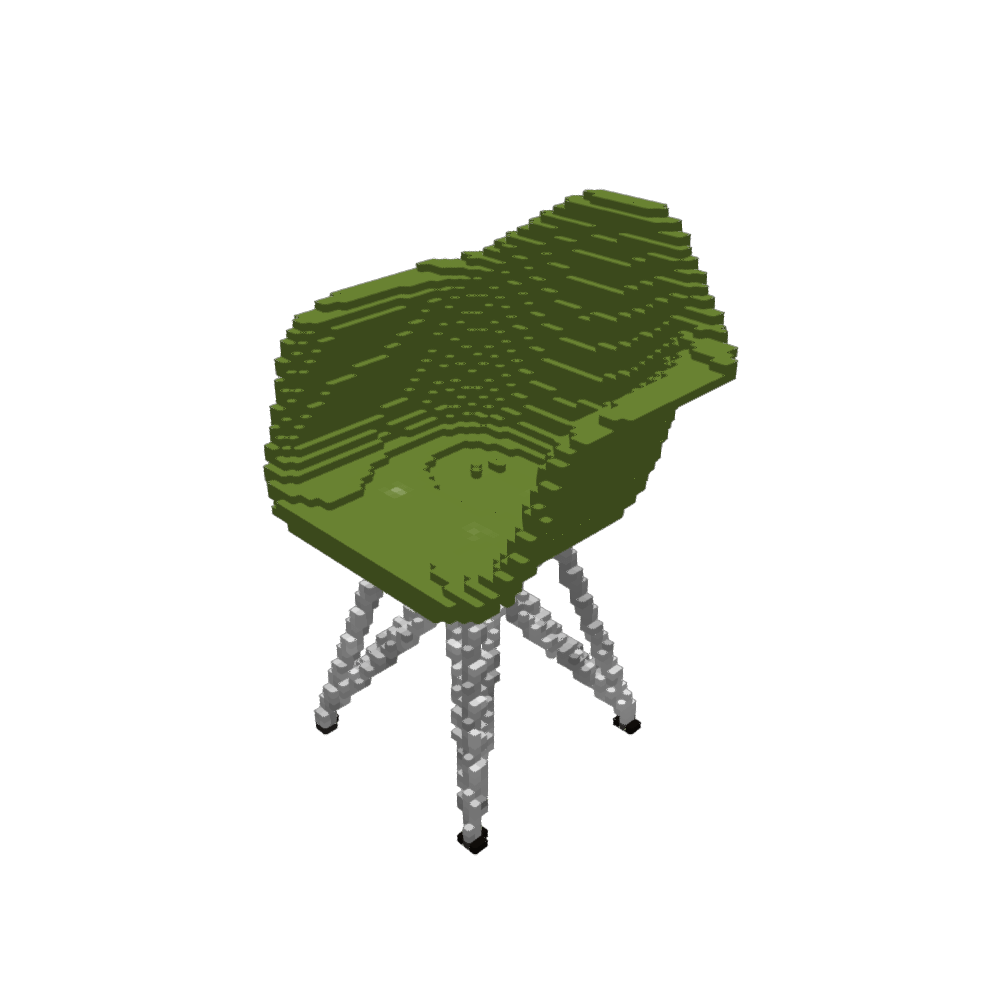} \\
[-0.15cm]
\bimodv & 
\includegraphics[trim=50 80 50 170,clip]{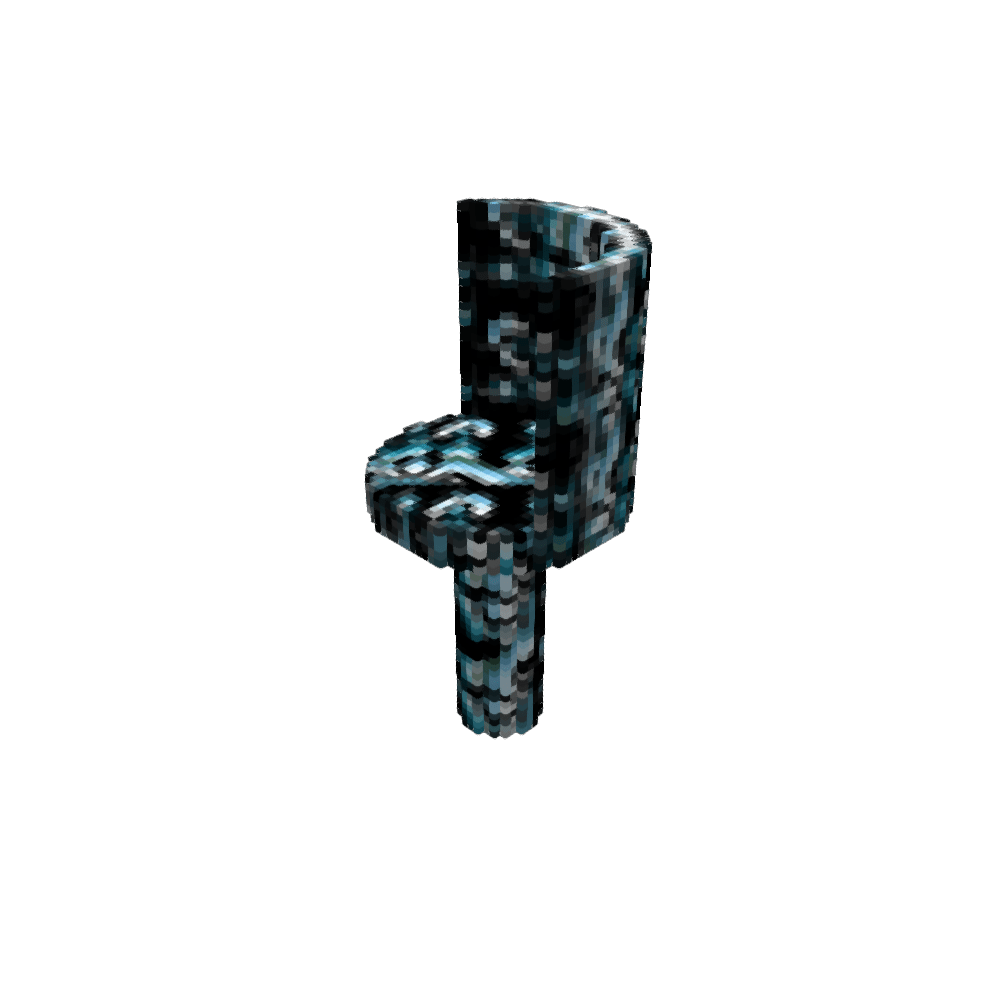} &
\includegraphics[trim=50 80 50 170,clip]{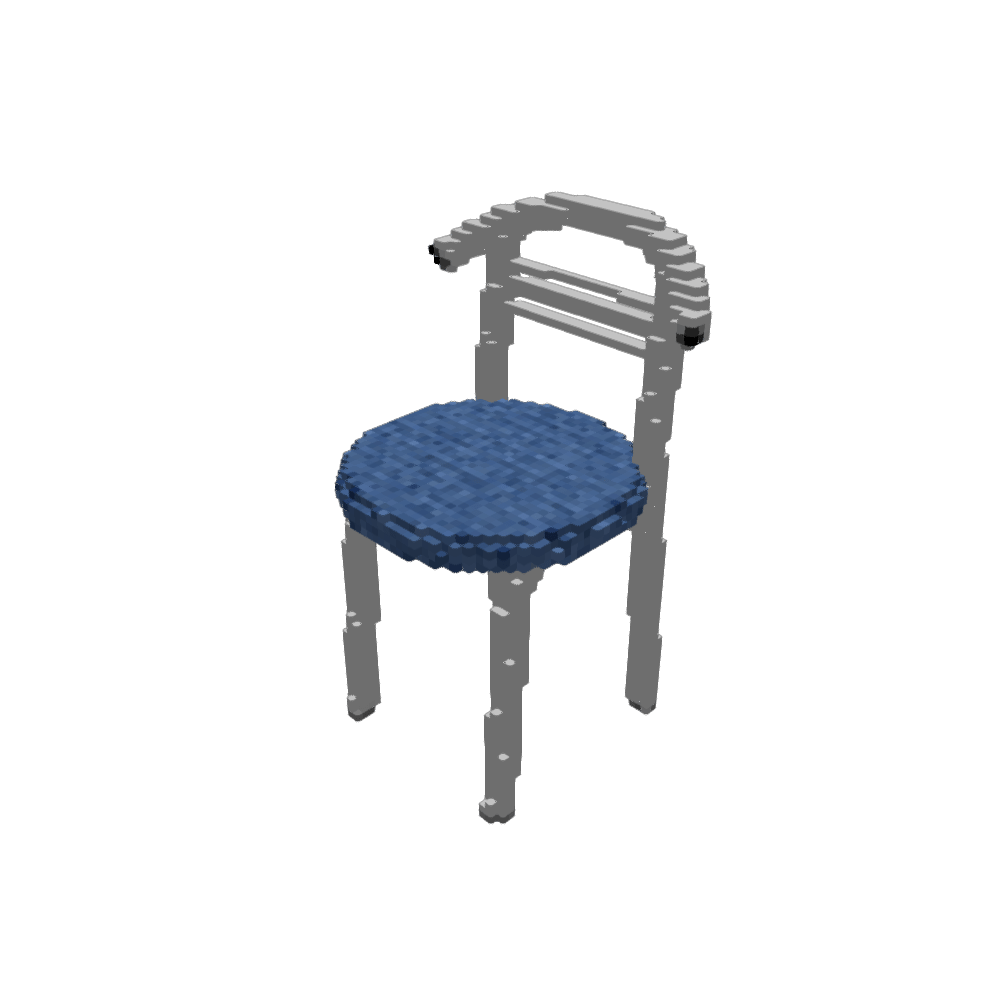} & 
\includegraphics[trim=50 80 50 170,clip]{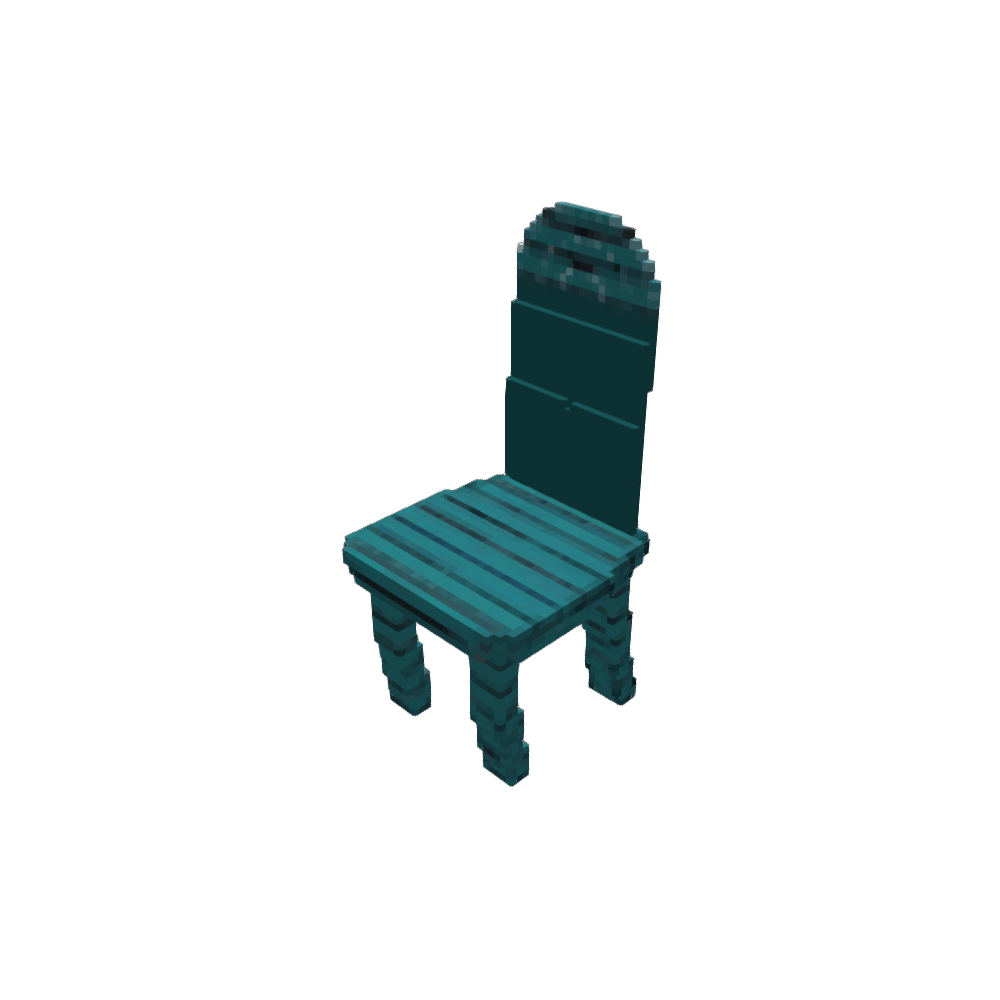} & 
\includegraphics[trim=50 80 50 170,clip]{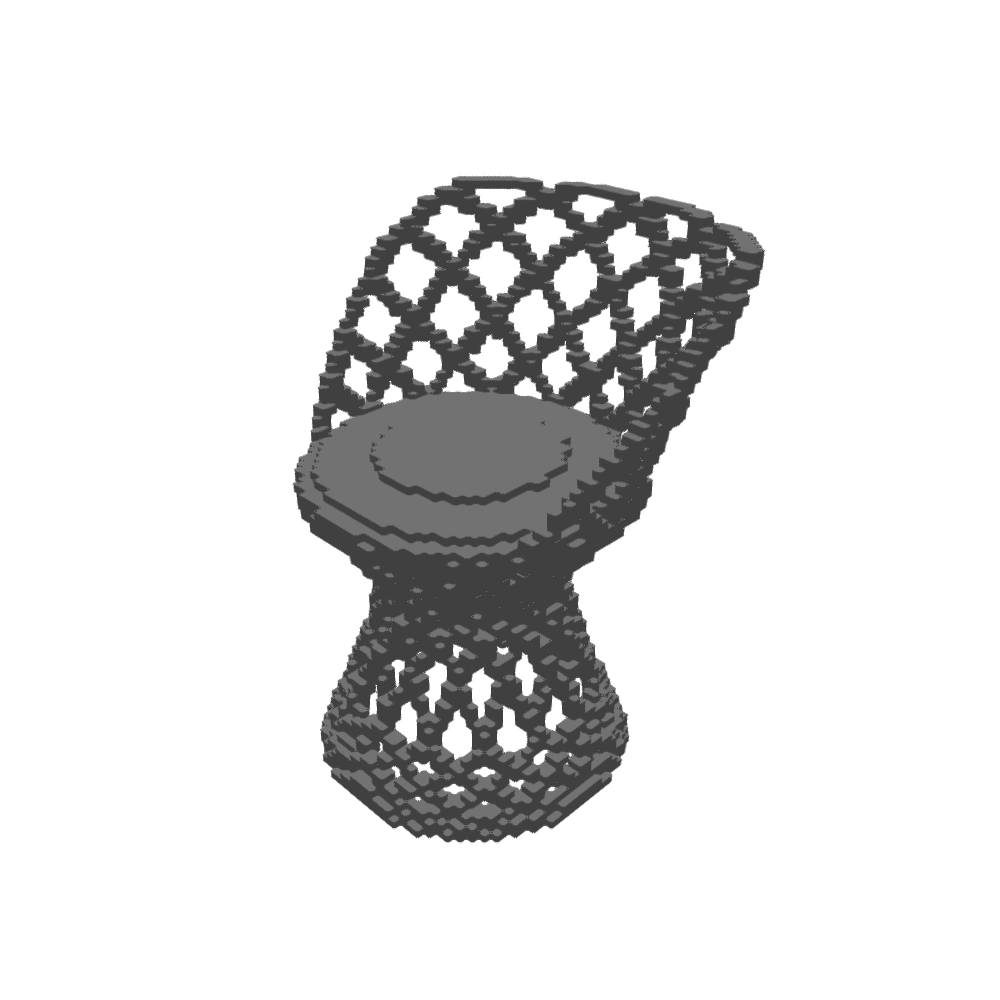} & 
\includegraphics[trim=50 80 50 170,clip]{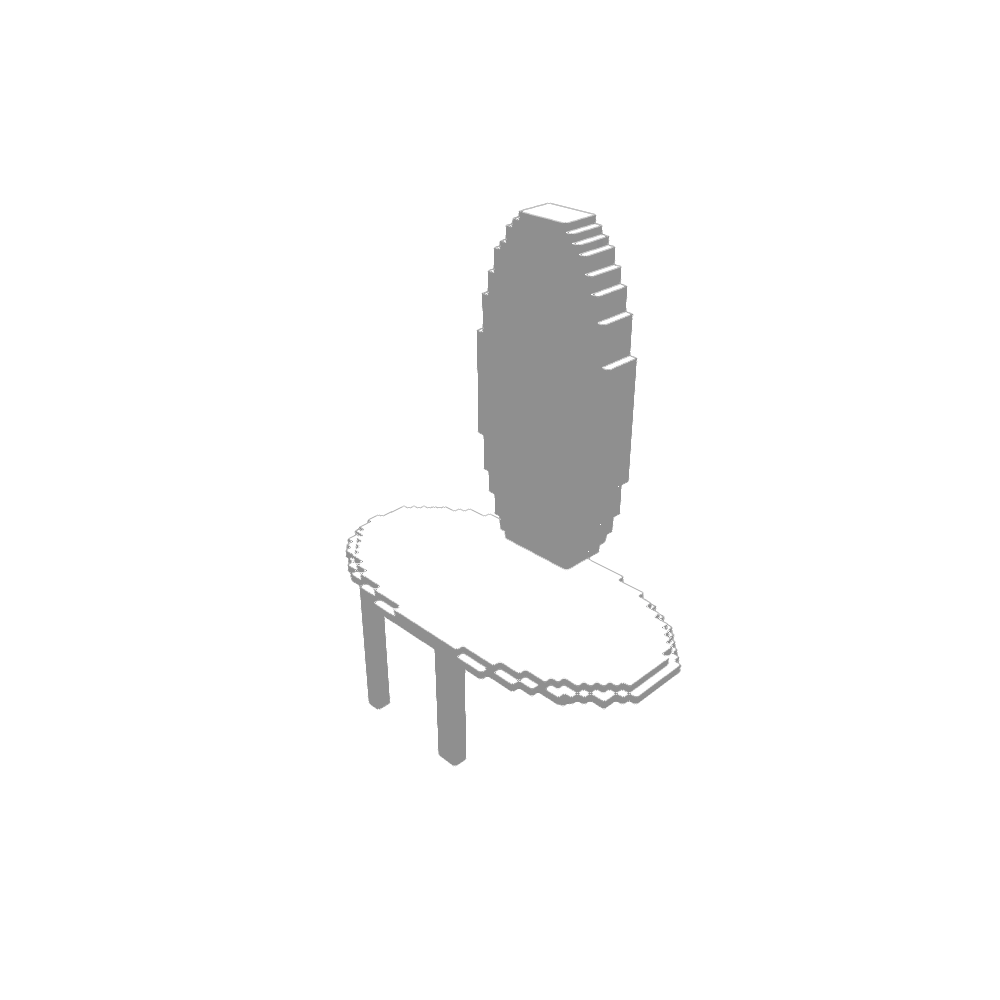} \\
[-0.25cm]
\trimodiv & 
\includegraphics[trim=50 100 50 170,clip]{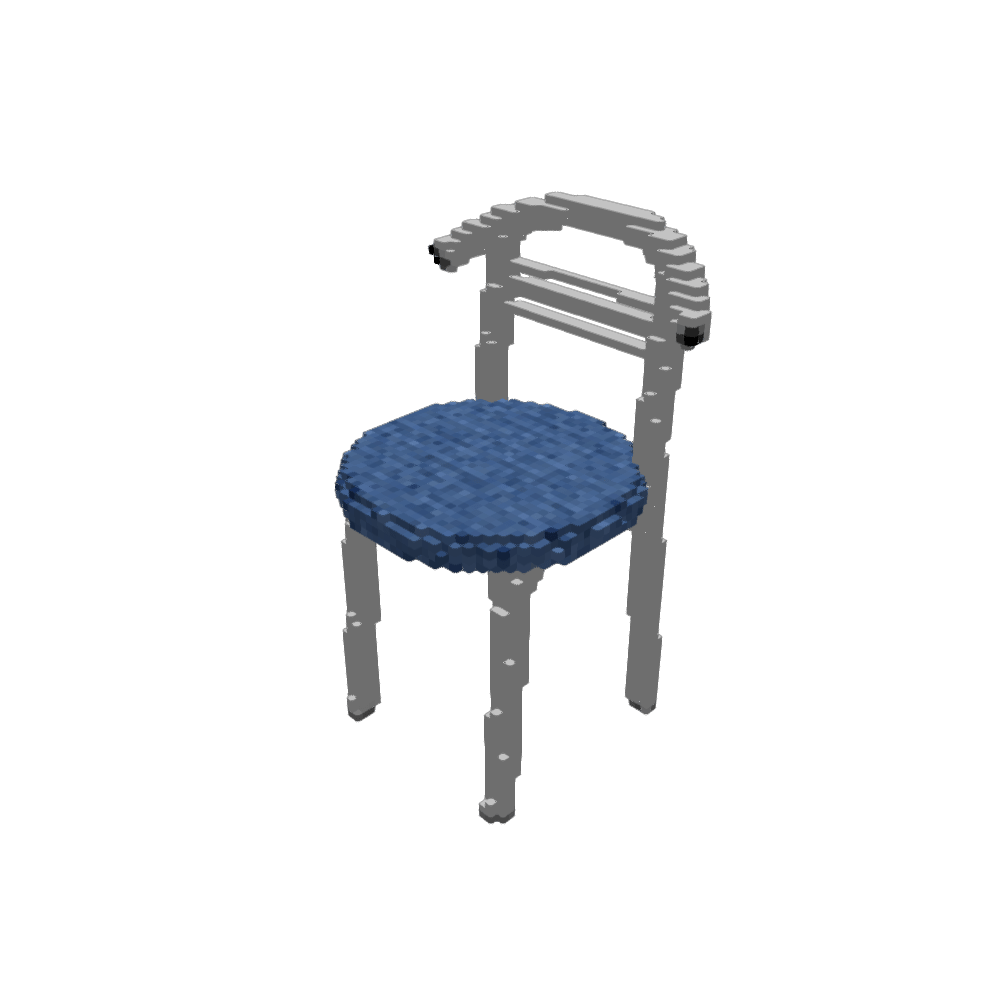} & 
\includegraphics[trim=50 100 50 170,clip]{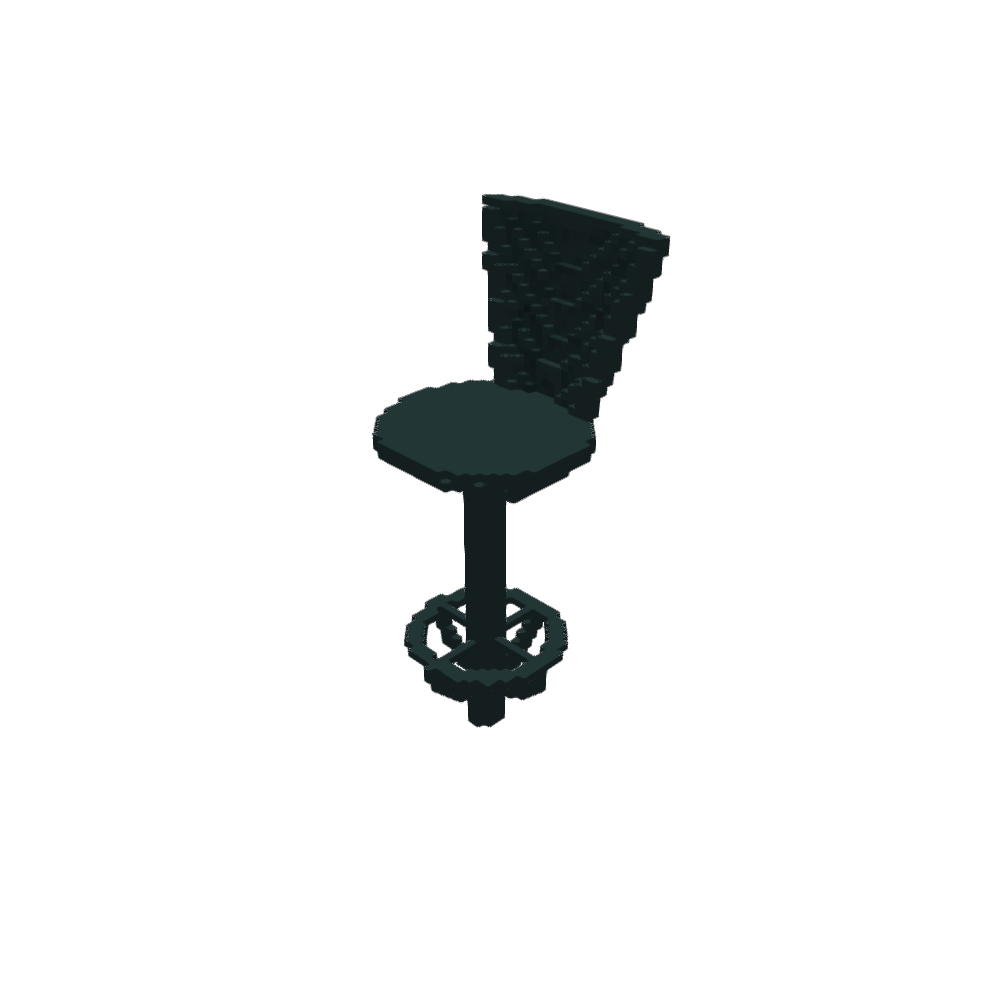} & 
\includegraphics[trim=50 100 50 170,clip]{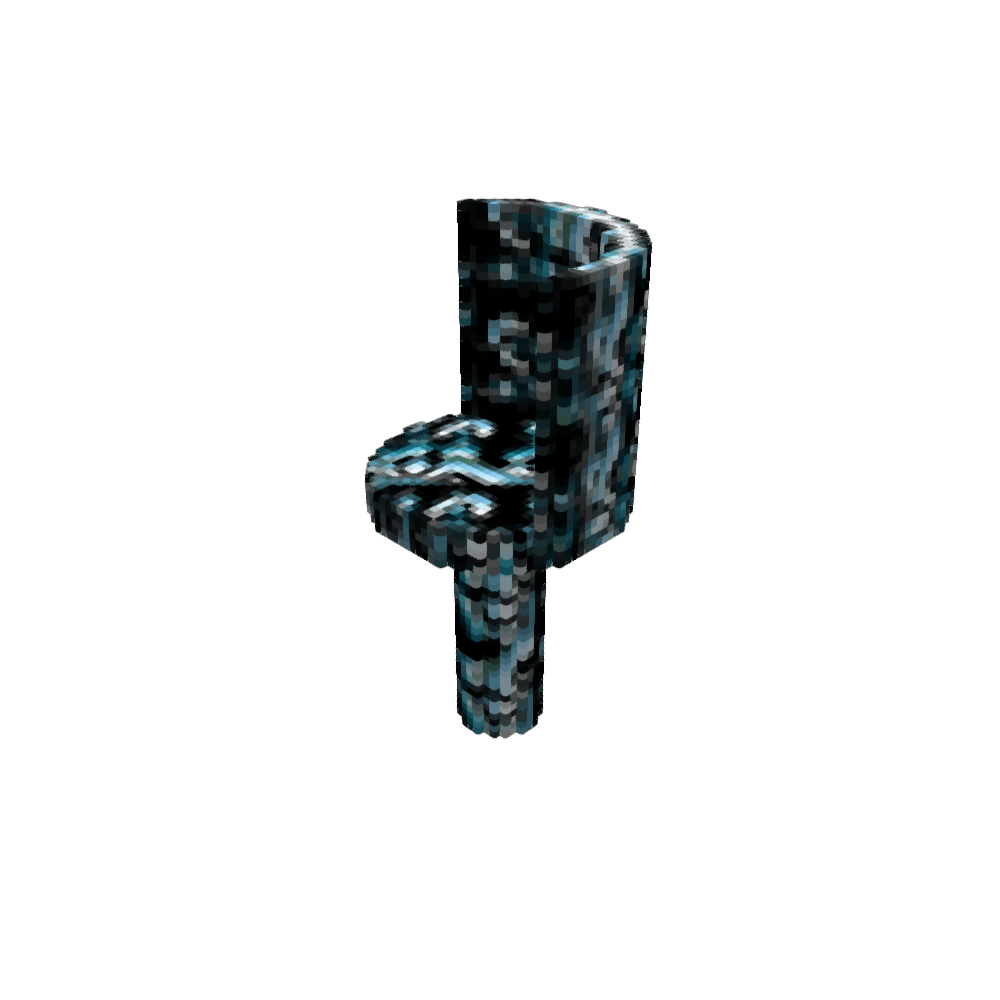} &
\includegraphics[trim=50 100 50 170,clip]{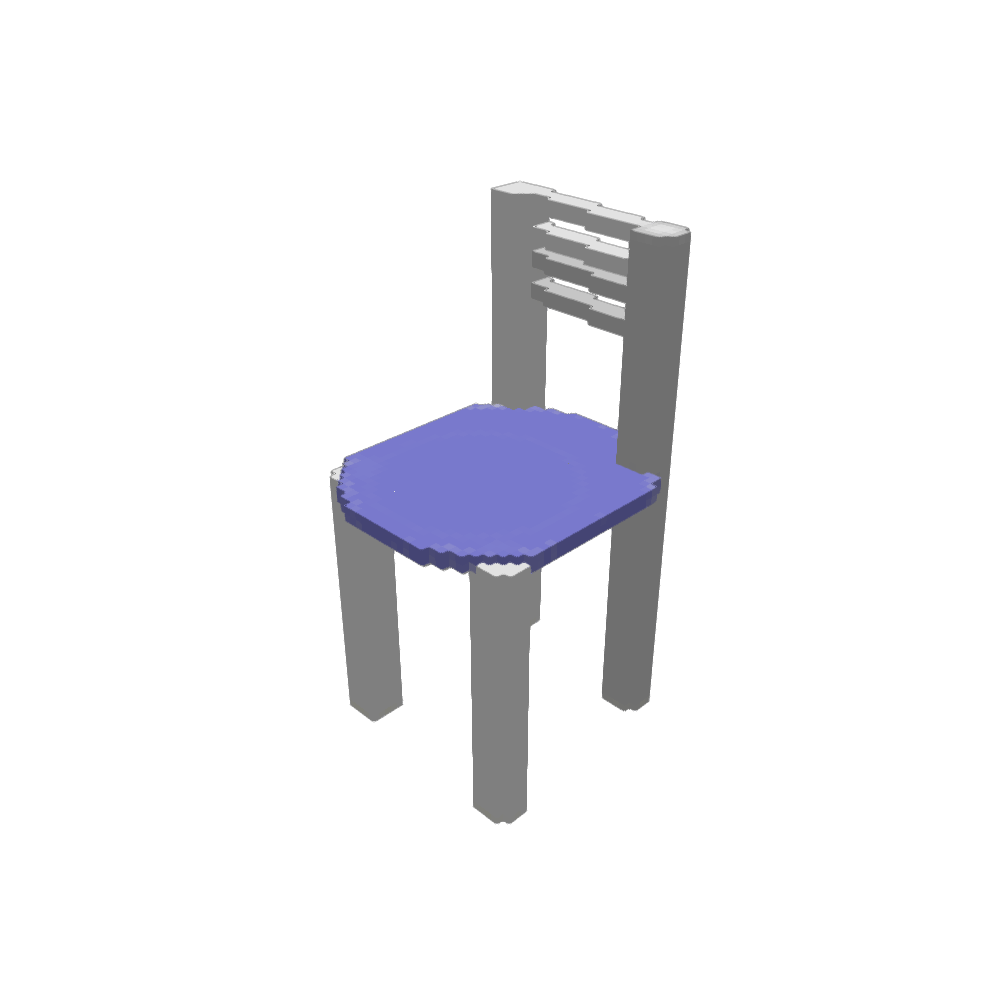} &
\includegraphics[trim=50 100 50 170,clip]{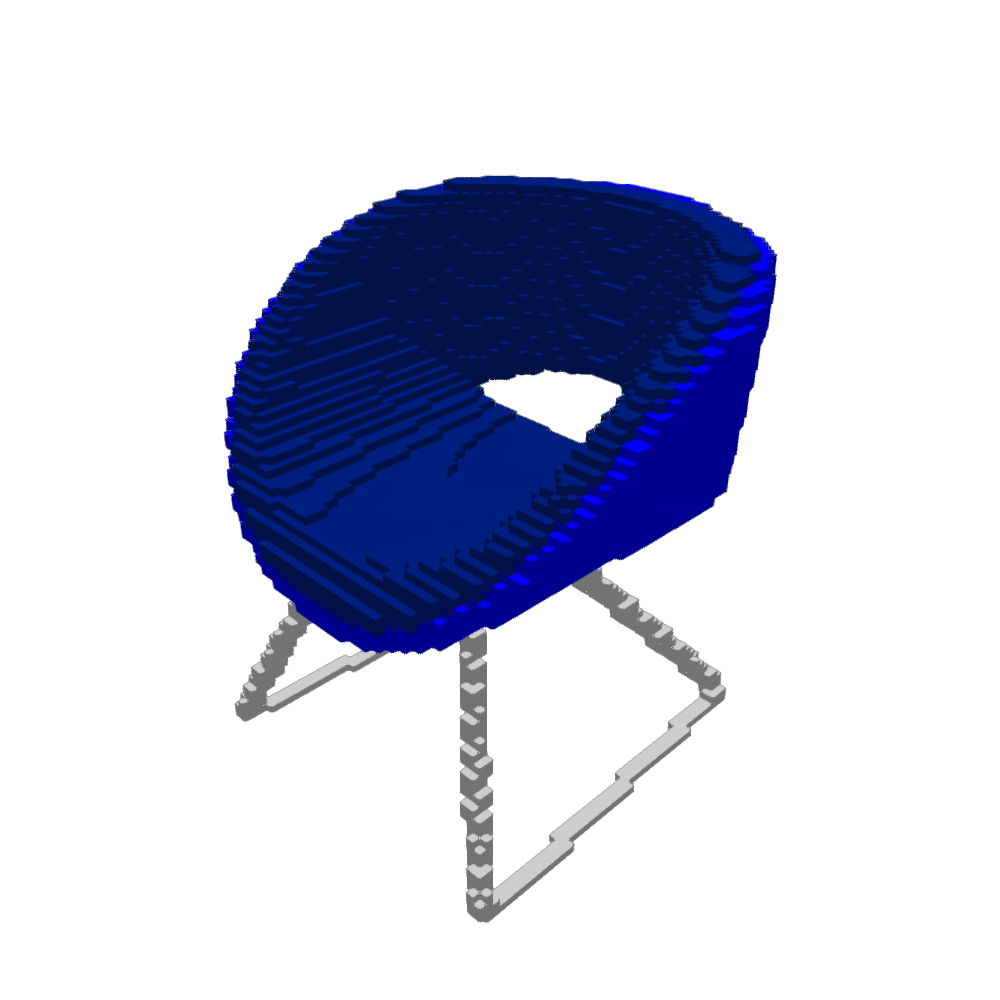} \\
[-0.1cm]
\midrule
3 &\multicolumn{4}{p{12.0cm}}{a low seat with olive exterior and grey-ish interior. There is a notch where one's neck might rest. It looks to be raised on a very low pedestal, maybe brown in color} &
\includegraphics[trim=25 120 25 160,clip]{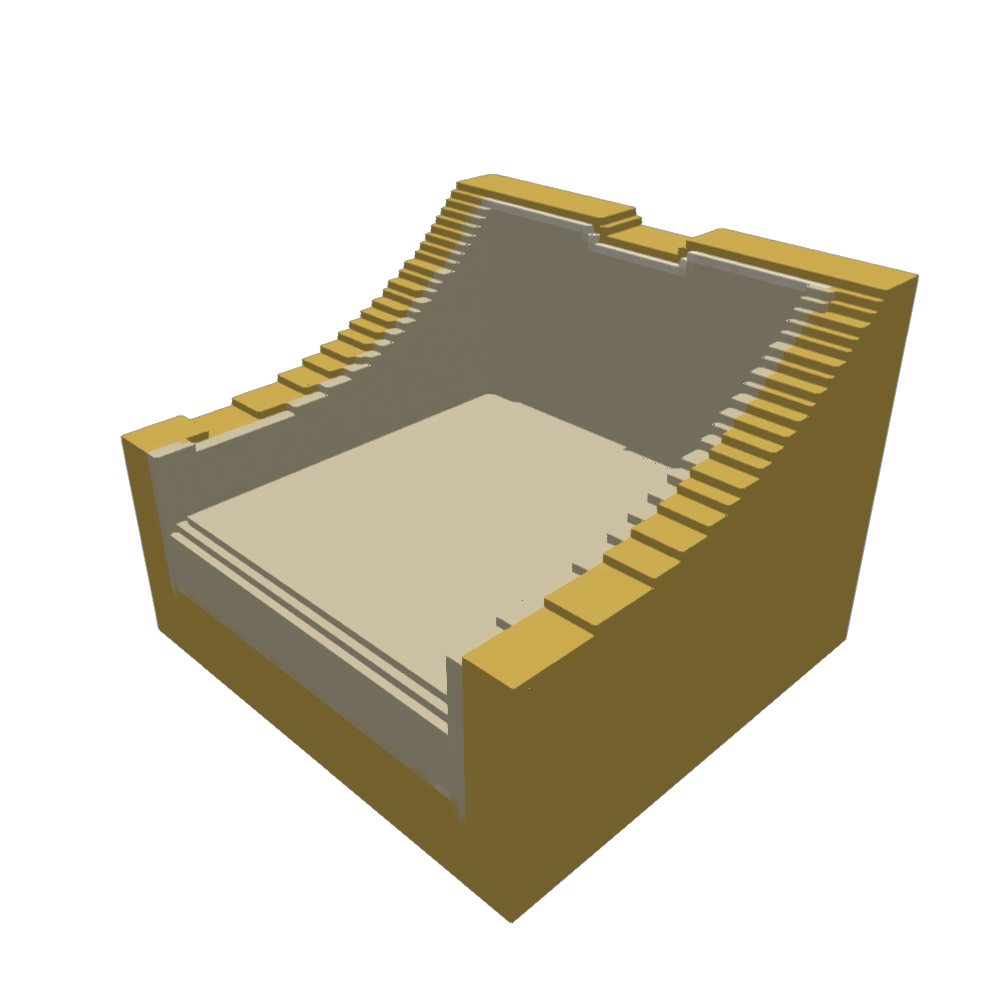} \\
[-0.4cm]
\bimodi & 
\includegraphics[trim=25 40 25 120,clip]{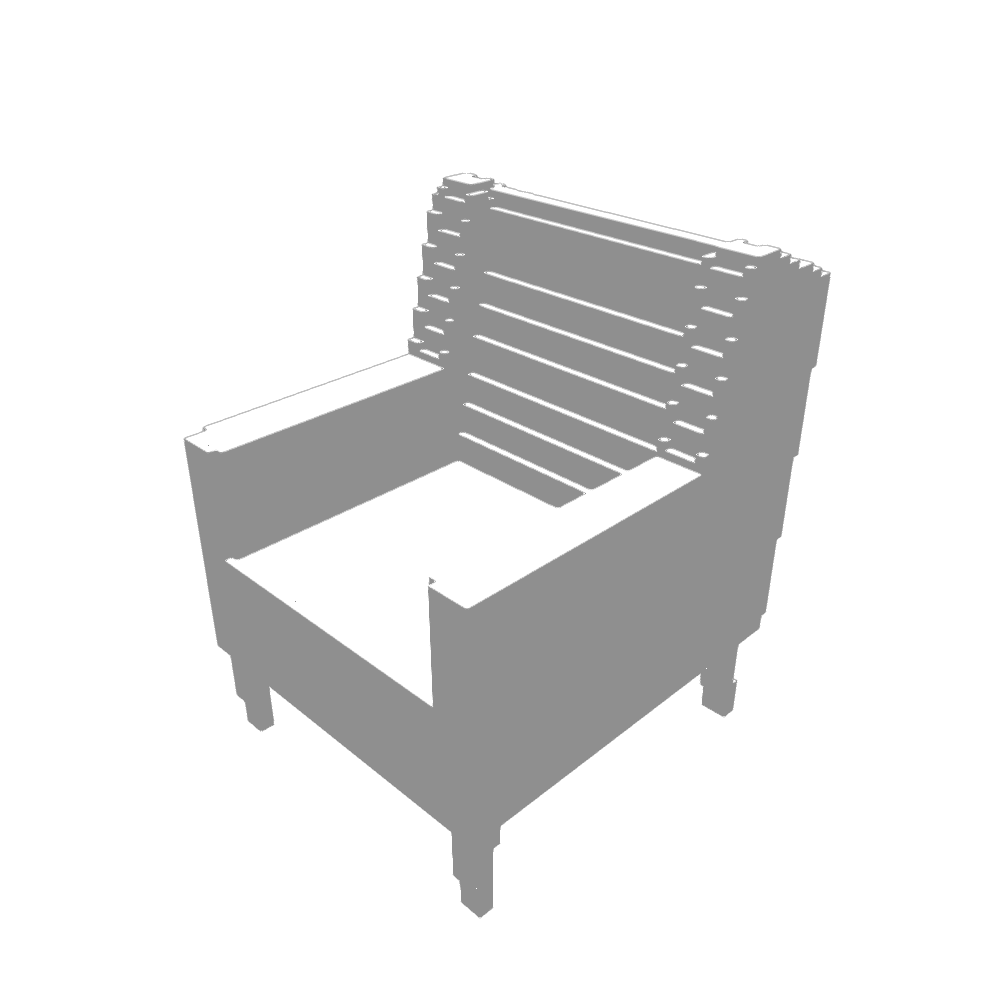} & 
\includegraphics[trim=25 40 25 120,clip]{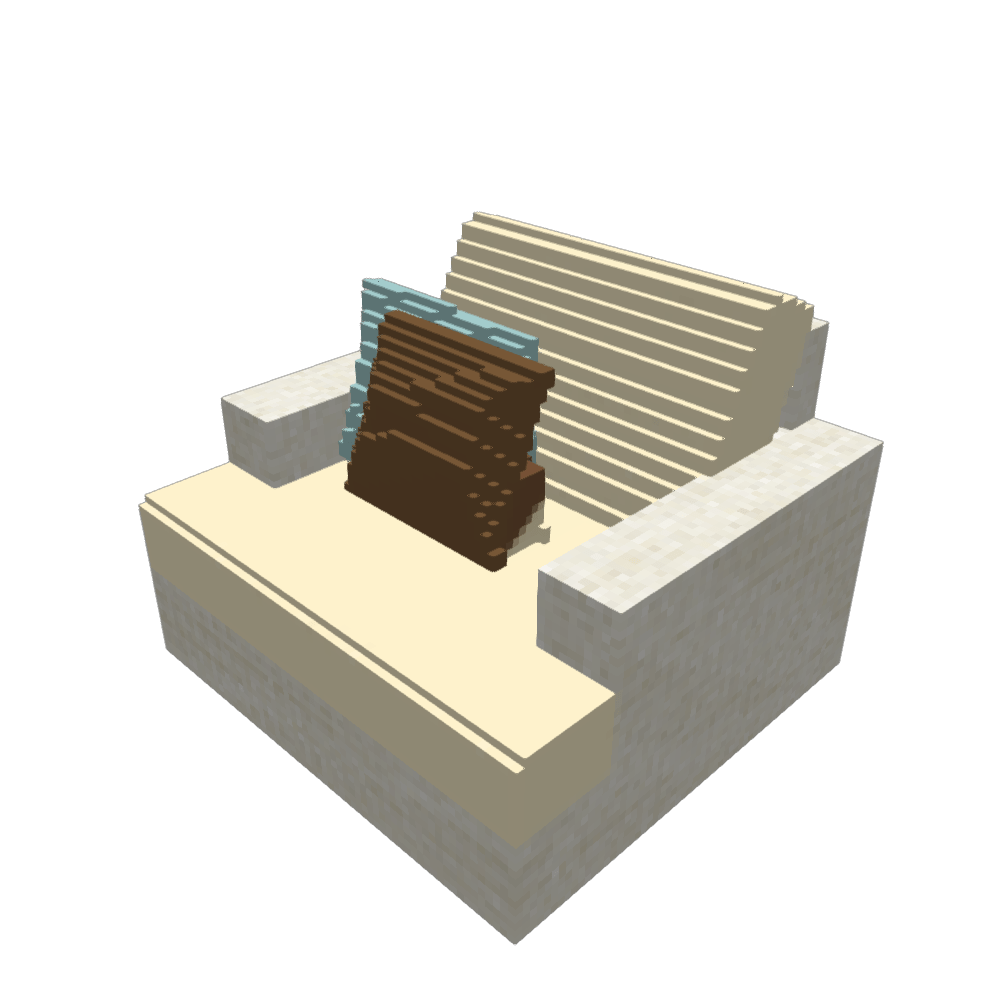} & 
\includegraphics[trim=25 40 25 120,clip]{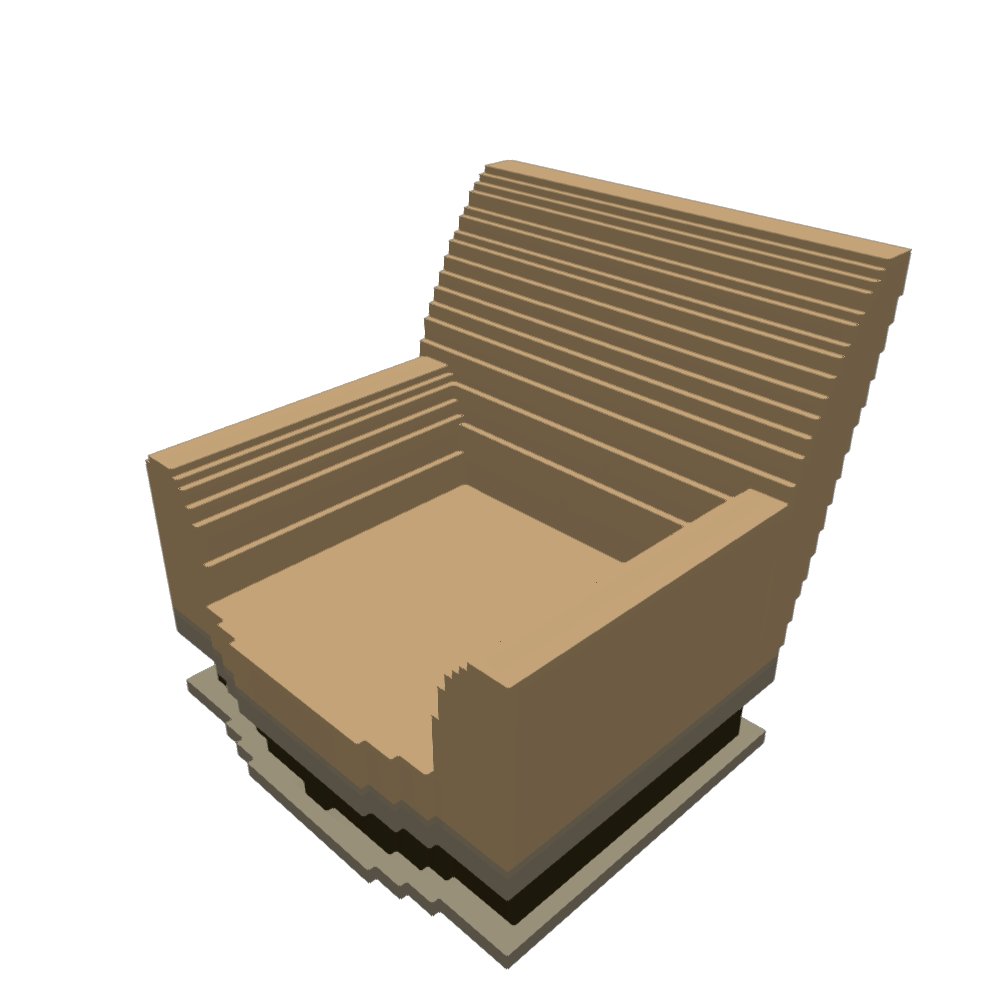} & 
\includegraphics[trim=25 40 25 120,clip]{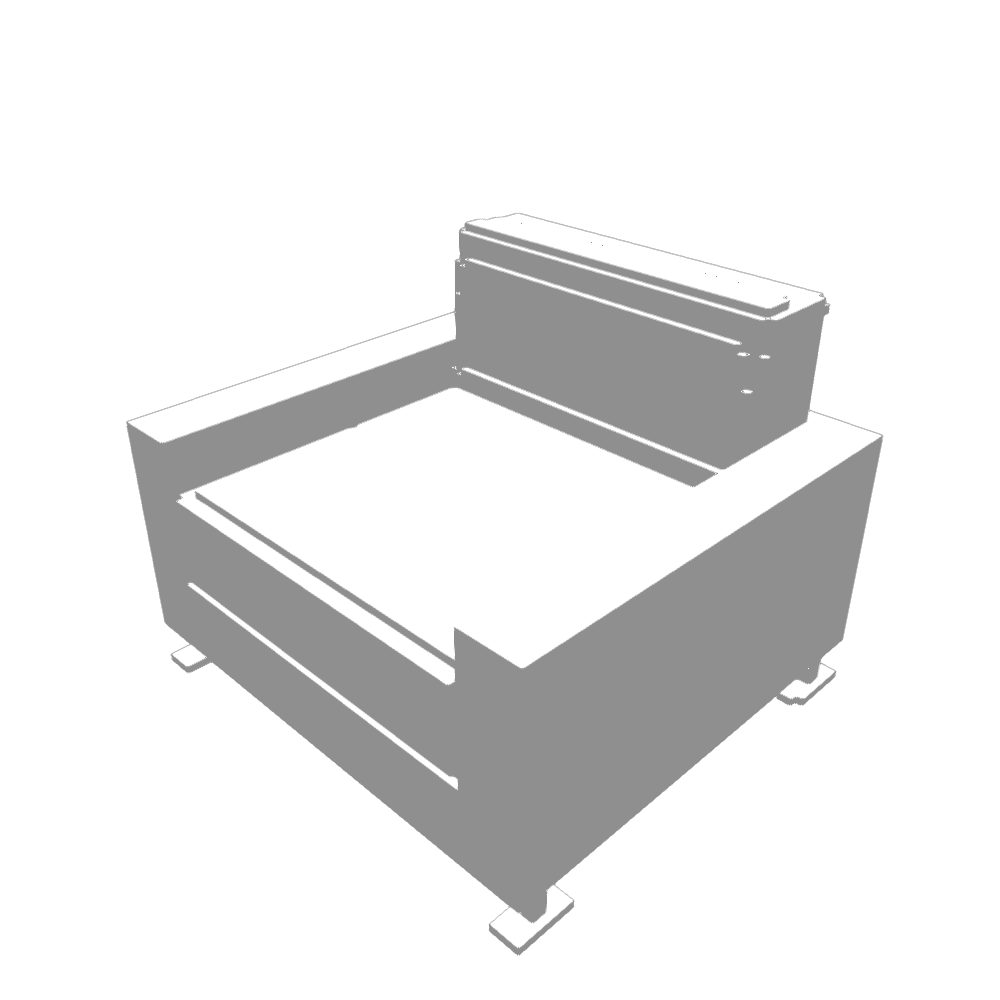} &
\includegraphics[trim=25 40 25 120,clip]{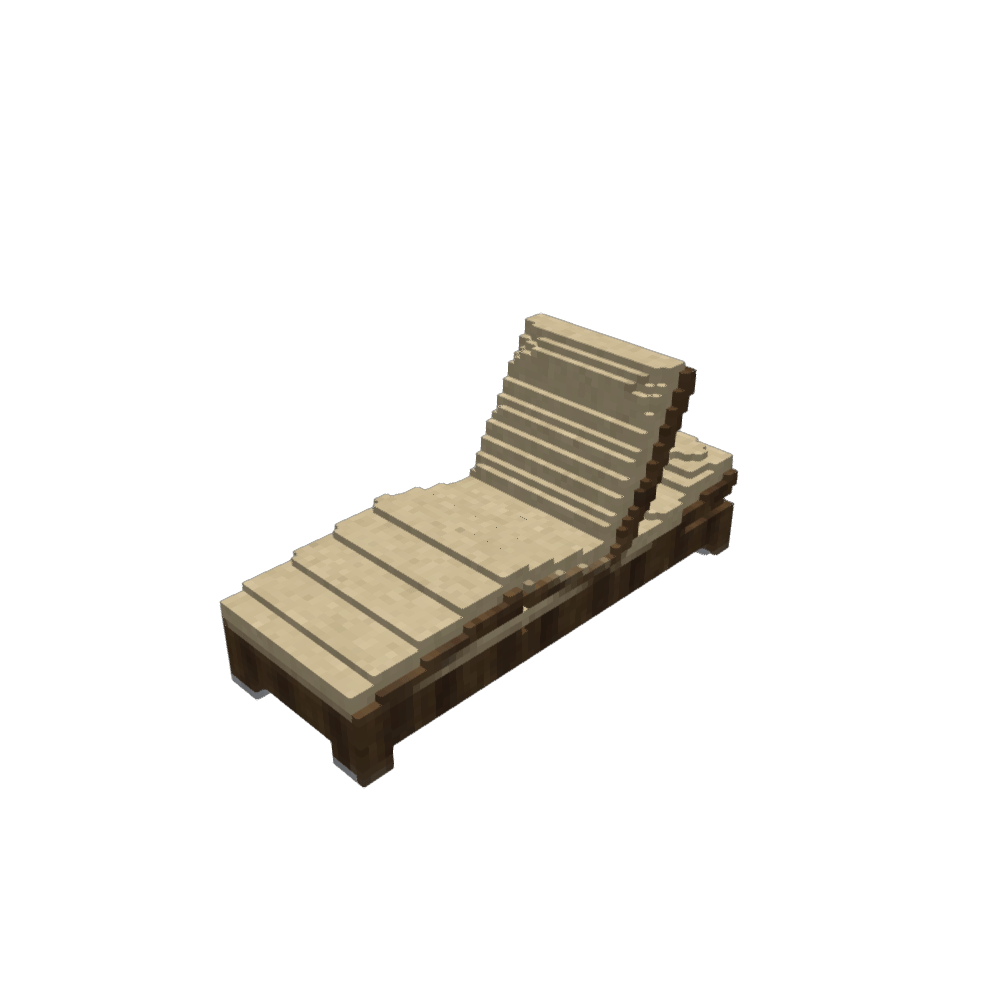} \\
[-0.15cm]
\bimodv & 
\includegraphics[trim=50 260 50 220,clip]{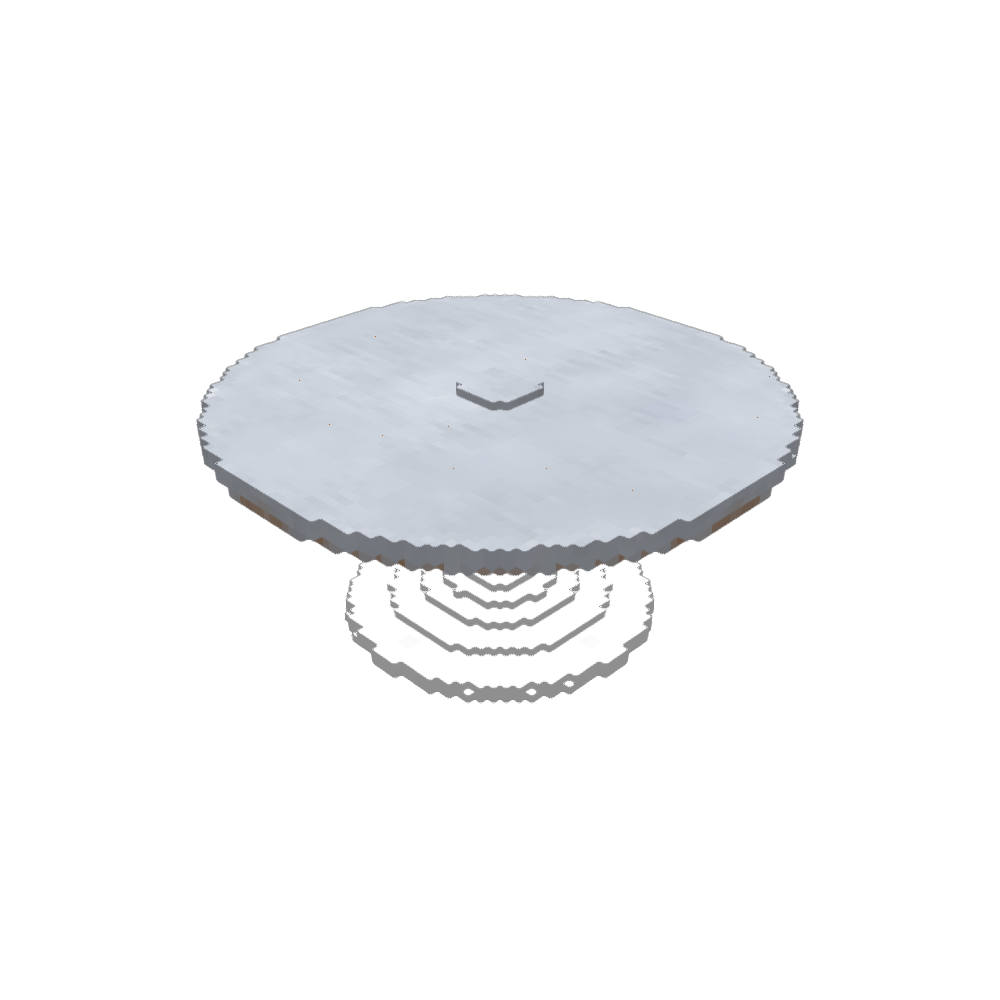} & 
\includegraphics[trim=50 320 50 260,clip]{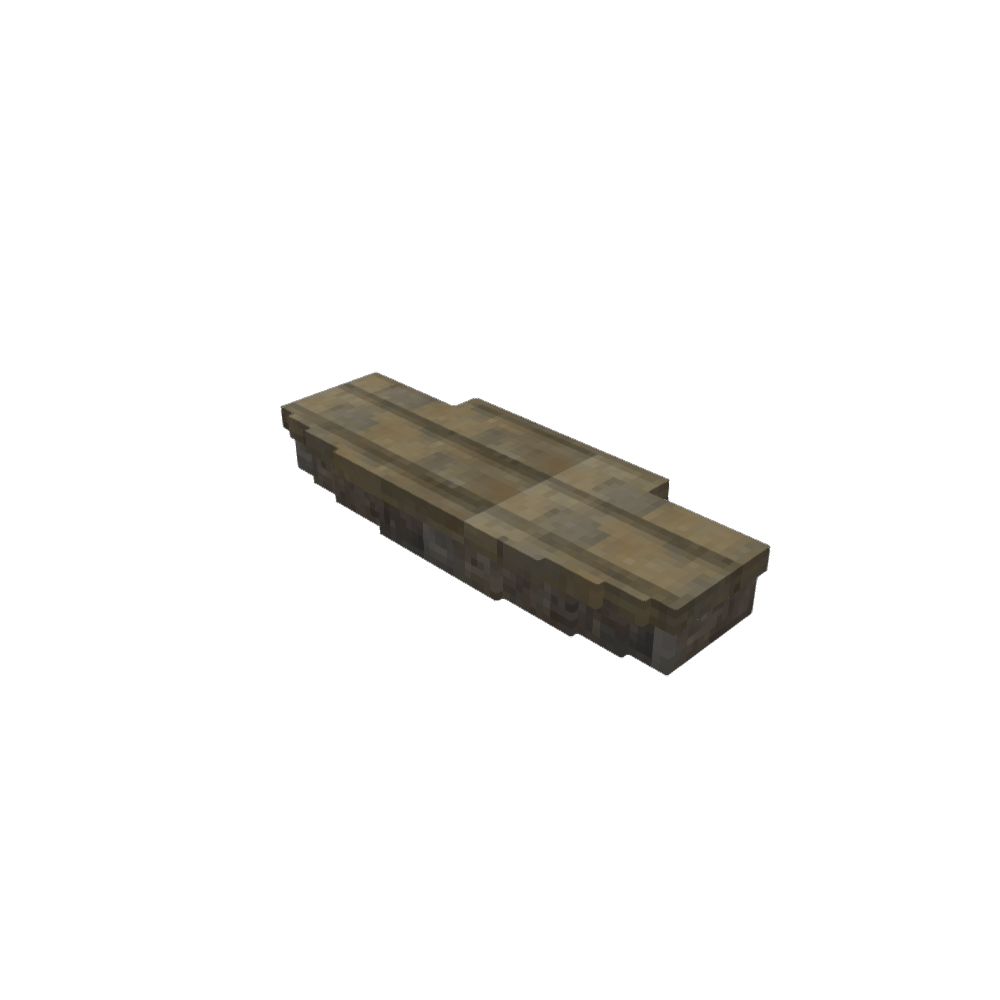} & 
\includegraphics[trim=50 280 50 200,clip]{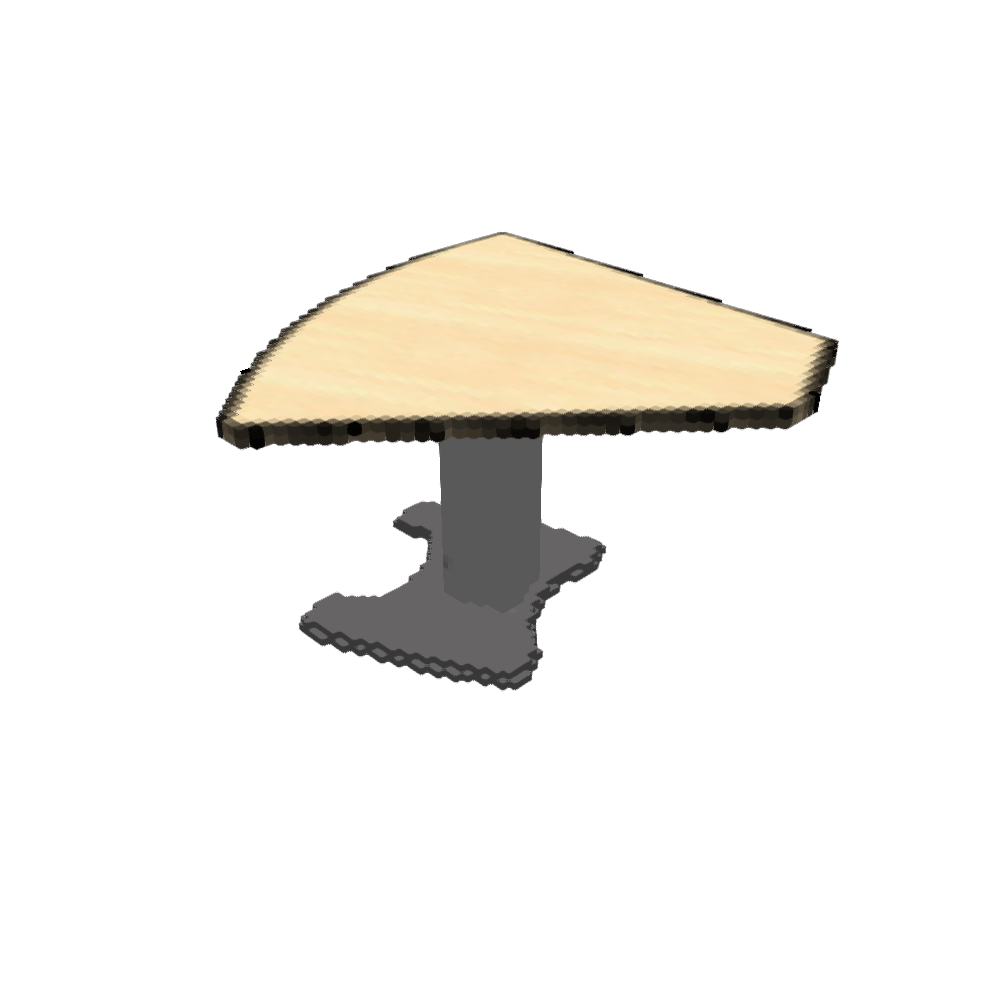} & 
\includegraphics[trim=50 260 50 220,clip]{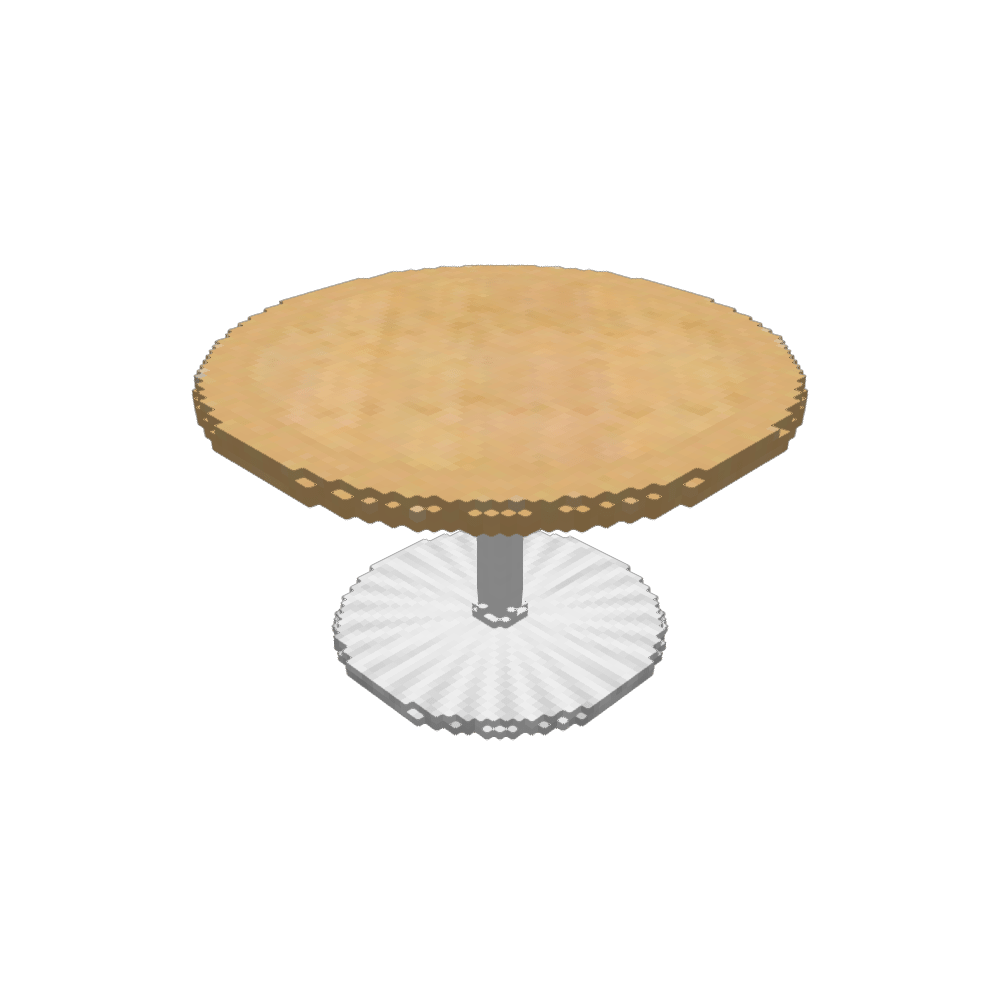} & 
\includegraphics[trim=50 280 50 200,clip]{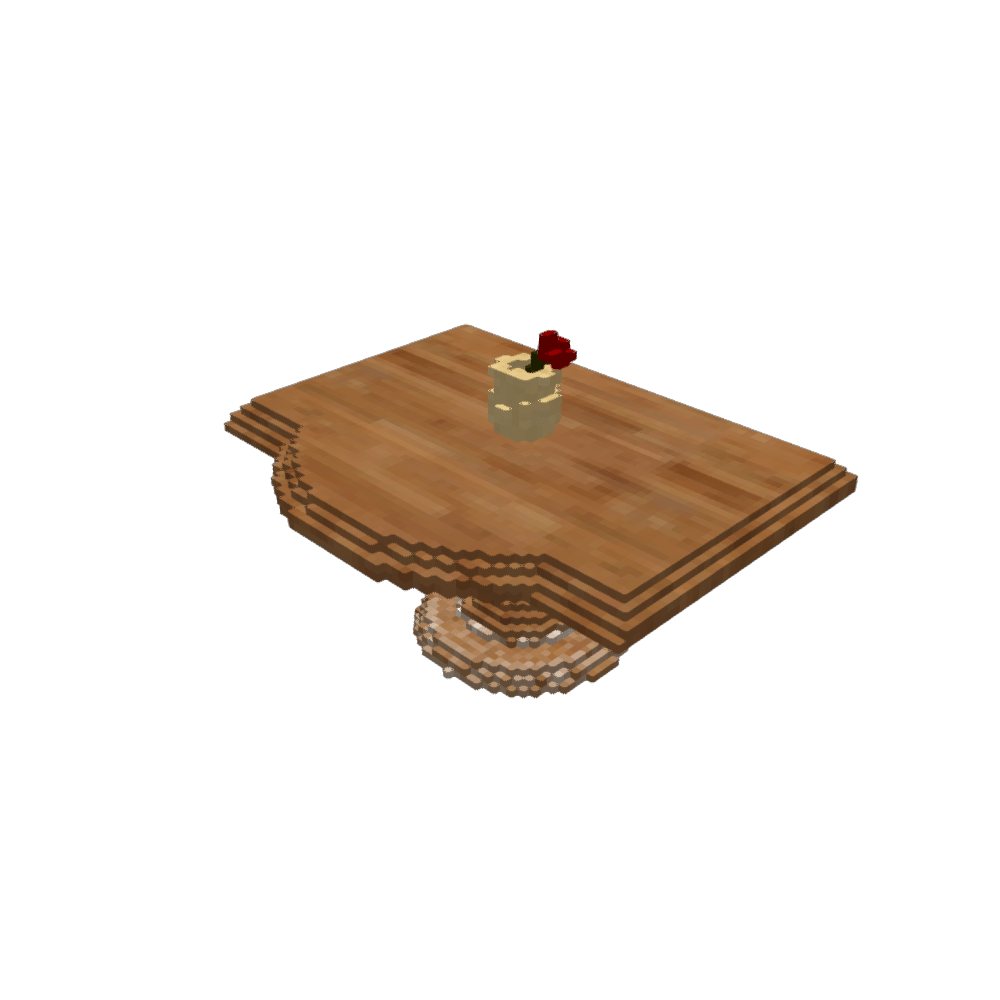} \\
[-0.15cm]
\trimodiv & 
\includegraphics[trim=50 180 50 200,clip]{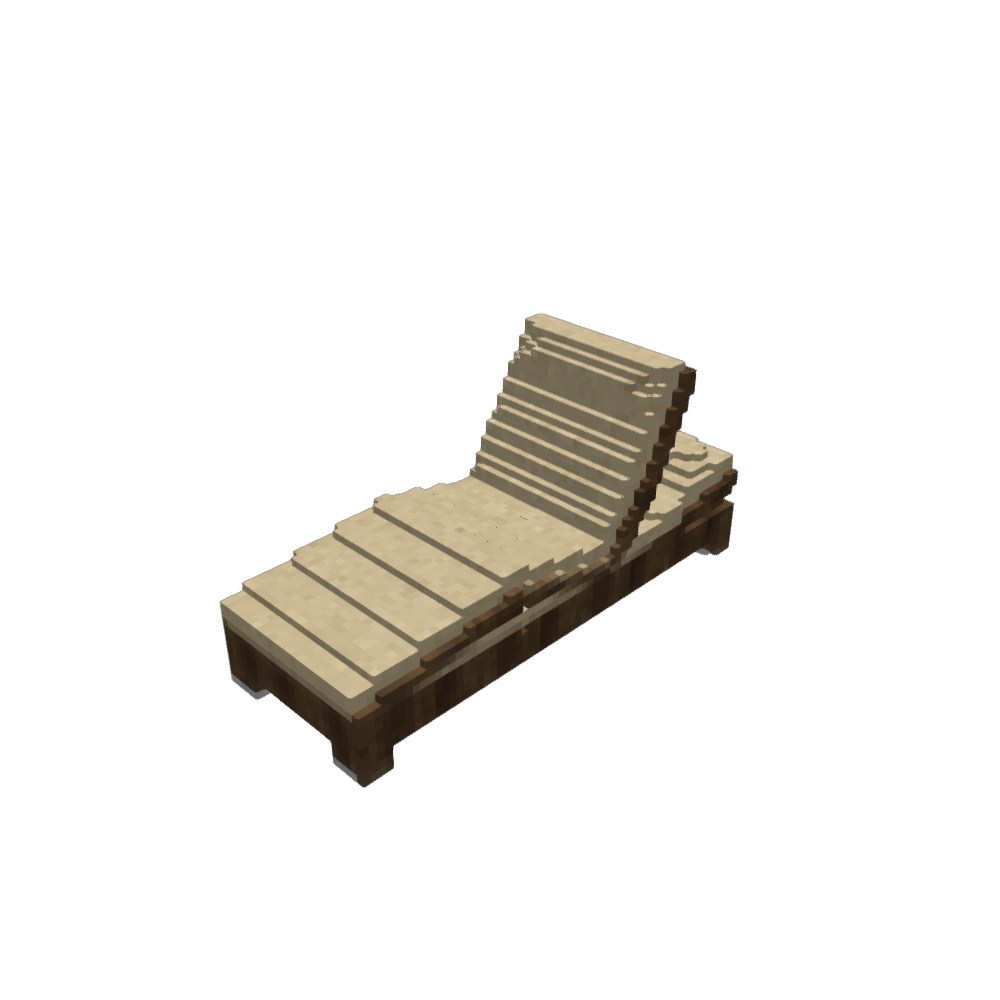} & 
\includegraphics[trim=50 180 50 200,clip]{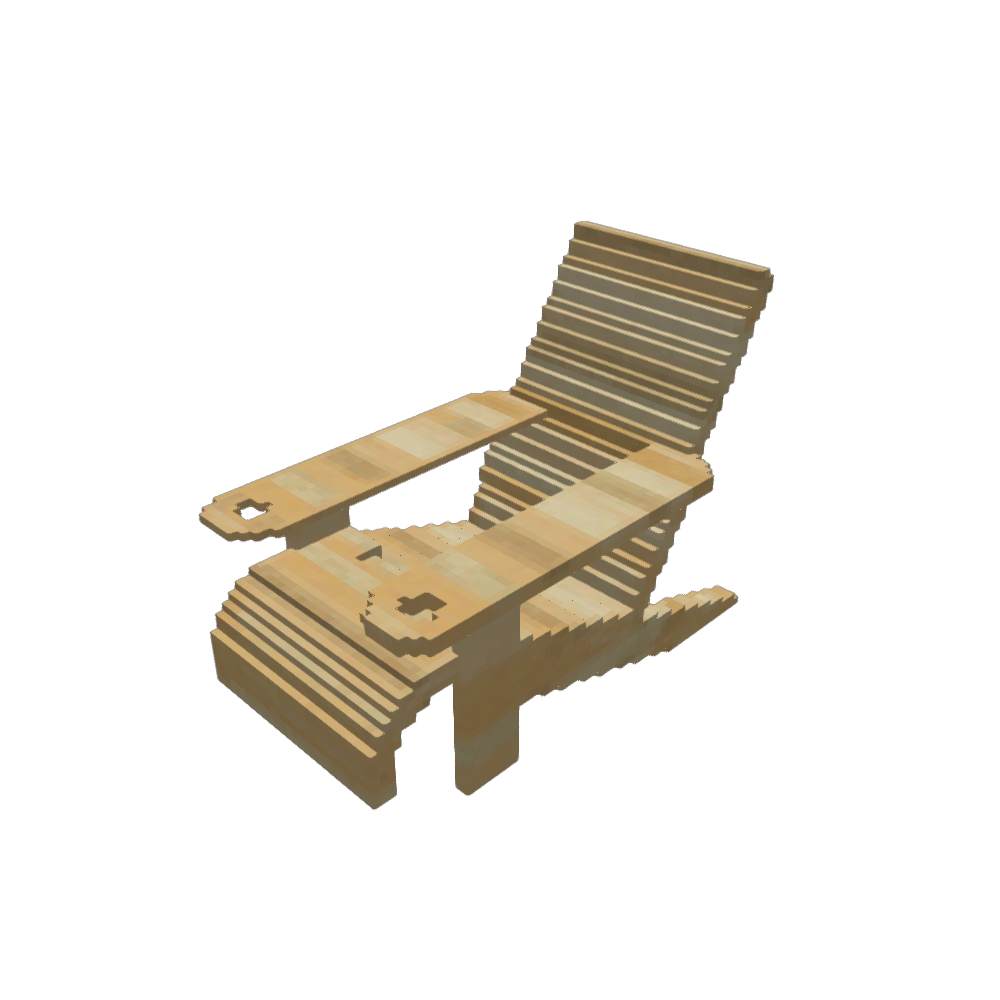} & 
\includegraphics[trim=50 150 50 230,clip]{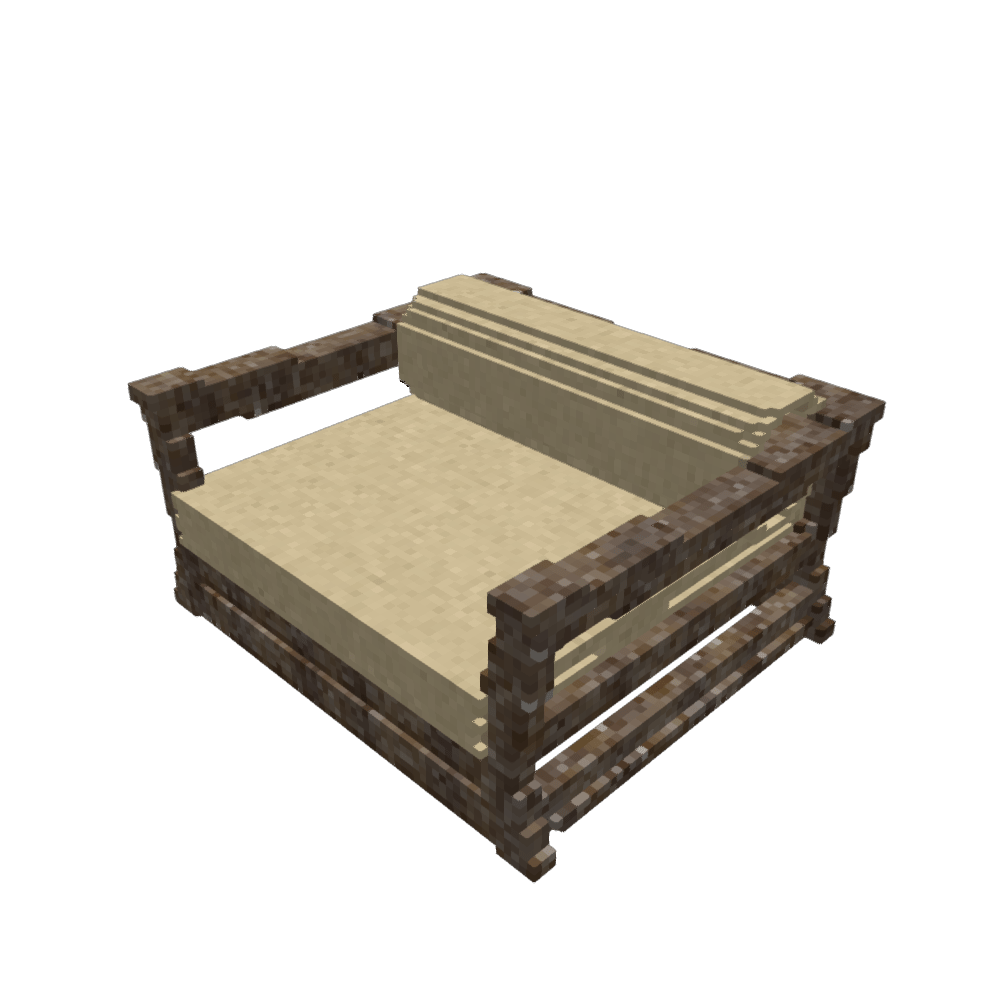} & 
\includegraphics[trim=50 220 50 260,clip]{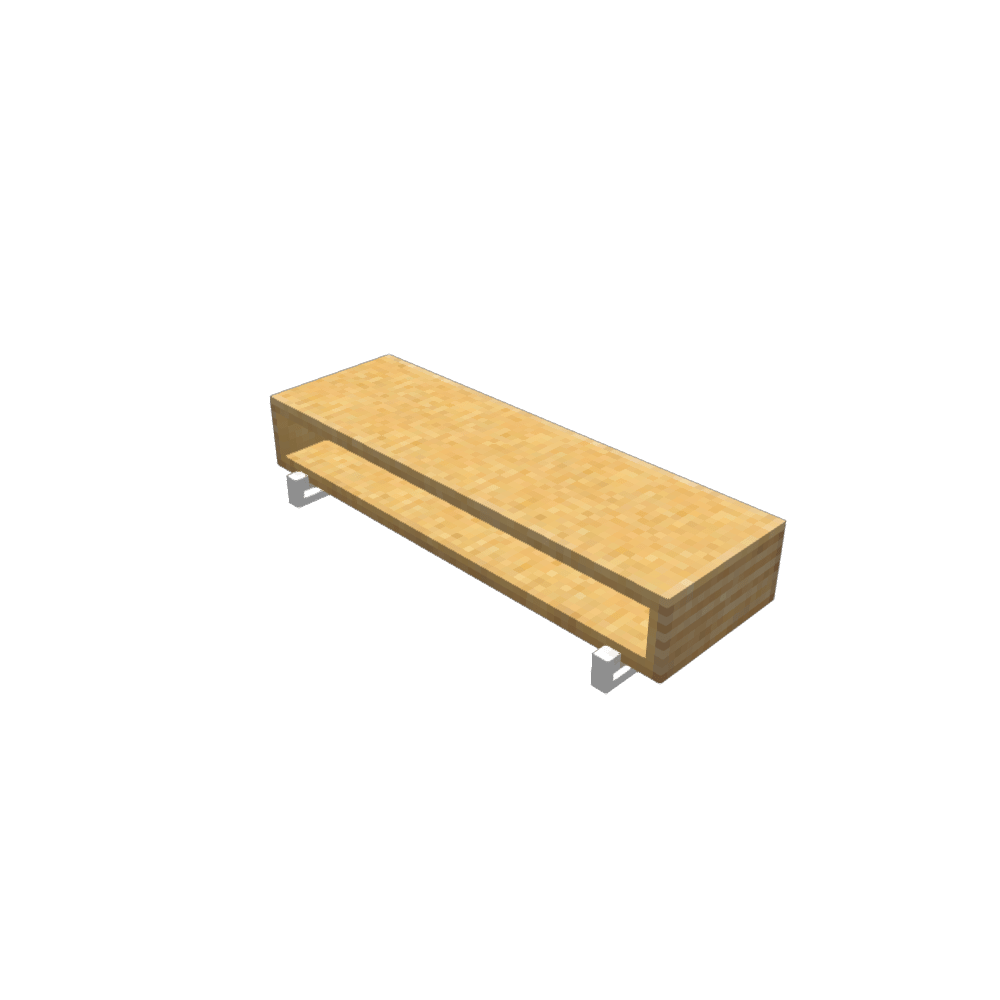} &
\includegraphics[trim=50 220 50 260,clip]{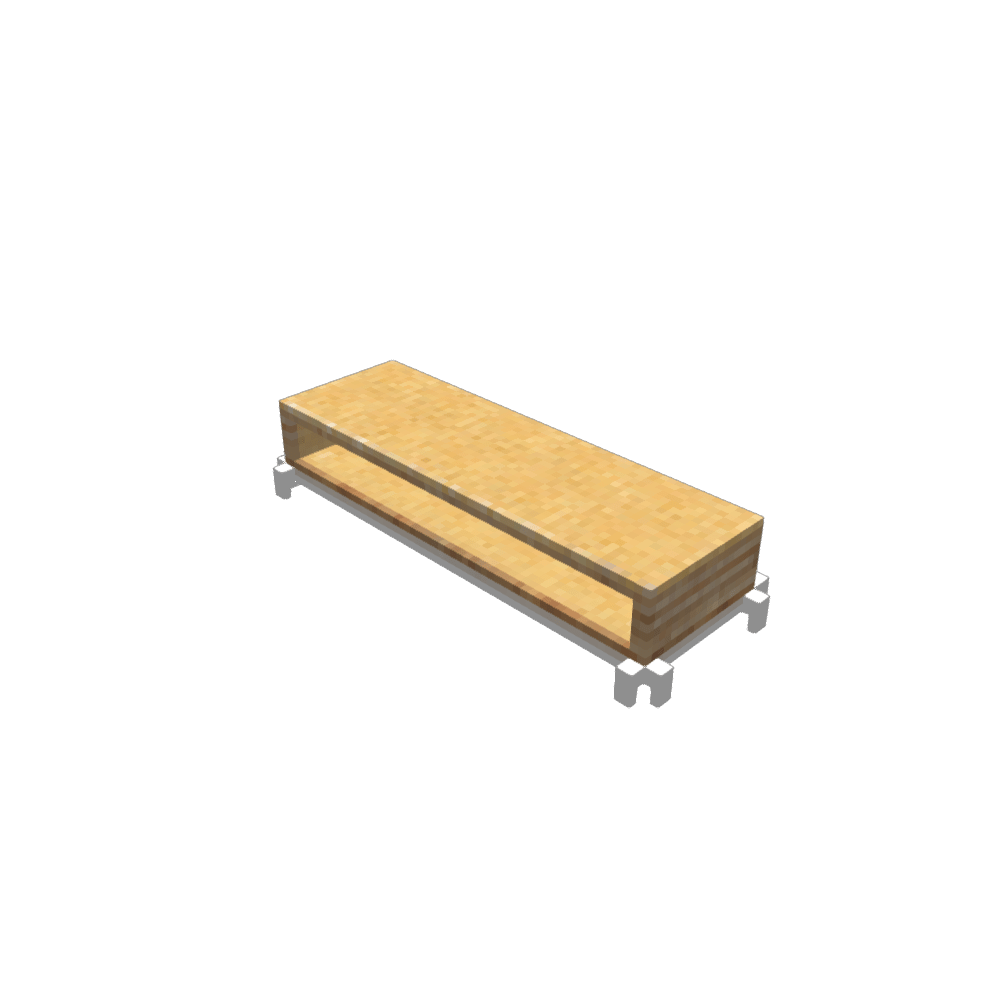} \\
[-0.1cm]
\bottomrule
\end{tabularx}
\end{figure*}

\end{document}